\documentclass{article} % For LaTeX2e
\usepackage{iclr2015,times}
\usepackage{hyperref}
\usepackage{url}

\usepackage[export]{adjustbox}
\usepackage{graphicx}
\usepackage{subcaption}

\usepackage{enumitem}
\usepackage{multirow}
\usepackage{amsmath,amssymb} % define this before the line numbering.
\usepackage{amsthm}
\usepackage{color}
\usepackage{xspace}
\usepackage{mathtools}
\usepackage{algorithm}

\usepackage{algolyx}
\makeatother

\newtheorem{corollary}{Corollary}
\newtheorem{thrm}{Theorem}

\newcommand{\eg}{\emph{e.g\@.}\xspace}

\DeclareMathOperator*{\loss}{loss}
\DeclareMathOperator*{\argmin}{arg\,min}
\DeclareMathOperator*{\argmax}{arg\,max}

\newcommand{\includepart}[1]{%
\includegraphics[height=0.40in,frame={0.5pt} {-0.5pt}]%
{images/HOG/part_filters_neg/jpg/part-filter_#1}}

\newcommand{\lolt}{\ell 1 / \ell 2}
\newcommand{\plc}{z}

\newcommand{\sq}[2][0]{% "sq" for "squeeze"
  \mbox{$\medmuskip=#1mu\displaystyle#2$}%
}

\title{
\begin{flushleft}
  Automatic Discovery and Optimization of Parts for Image Classification
\end{flushleft}
}

\author{
\begin{tabular}[t]{ c c c c } 
\textbf{Sobhan Naderi Parizi} & \textbf{Andrea Vedaldi} & \textbf{Andrew Zisserman} & \textbf{Pedro Felzenszwalb} \\
\textnormal{Brown University} & \multicolumn{2}{c}{\textnormal{University of Oxford}} & \textnormal{Brown University} \\
\textnormal{sobhan@brown.edu} & \multicolumn{2}{c}{\textnormal{\{vedaldi, az\}@robots.ox.ac.uk}} & \textnormal{pff@brown.edu}
\end{tabular}
}

\iclrfinalcopy % Uncomment for camera-ready version

\iclrconference % Uncomment if submitted as conference paper instead of workshop

\begin{document}

\maketitle

\begin{abstract}
Part-based representations have been shown to be very useful for image
classification. Learning part-based models is often viewed as a
two-stage problem. First, a collection of informative parts is
discovered, using heuristics that promote part distinctiveness and
diversity, and then classifiers are trained on the vector of part
responses. In this paper we unify the two stages and learn the
image classifiers and a set of shared parts jointly. We generate 
an initial pool of parts by randomly sampling part candidates and
 selecting a good subset using $\lolt$ regularization. 
All steps are driven \emph{directly} by the same objective namely 
the classification loss on a training set. This lets us do away with
engineered heuristics.
We also introduce the notion of \emph{negative parts}, 
intended as parts that are negatively correlated with one or 
more classes. Negative parts are complementary to the parts 
discovered by other methods, which look only for positive
correlations.
 \end{abstract}

\section{Introduction}
\begin{figure}[b]
\centering
\includegraphics[height=0.38in, width=0.38in,frame=0.5pt -0.5pt]{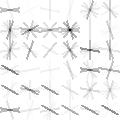}\hfill
\includegraphics[height=0.38in, width=0.38in]{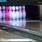}\hfill
\includegraphics[height=0.38in, width=0.38in]{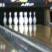}\hfill
\includegraphics[height=0.38in, width=0.38in]{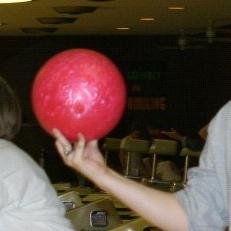}\hfill
\includegraphics[height=0.38in, width=0.38in]{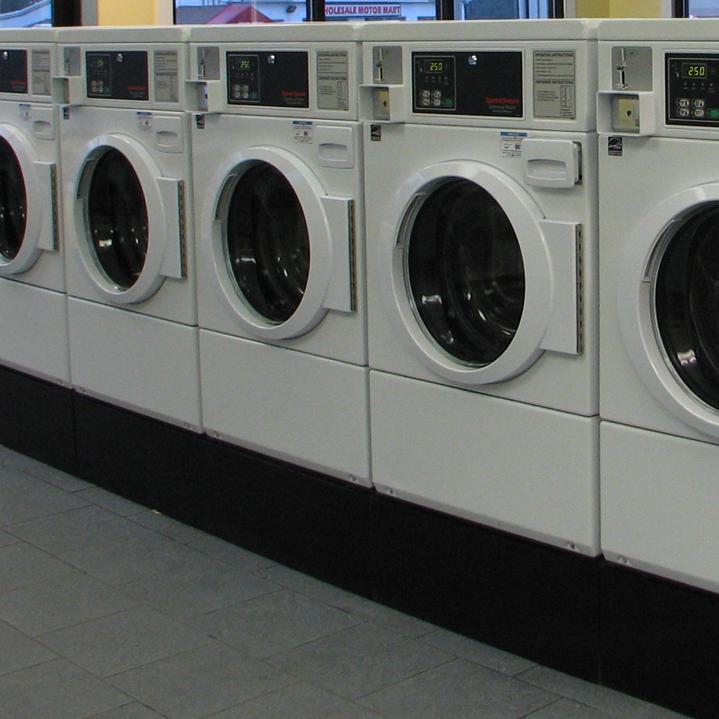}\hfill
\includegraphics[height=0.38in, width=0.38in]{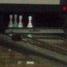} \hspace{0.4cm}
\includegraphics[height=0.38in, width=0.38in,frame=0.5pt -0.5pt]{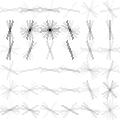}\hfill
\includegraphics[height=0.38in, width=0.38in]{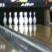}\hfill
\includegraphics[height=0.38in, width=0.38in]{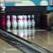}\hfill
\includegraphics[height=0.38in, width=0.38in]{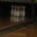}\hfill
\includegraphics[height=0.38in, width=0.38in]{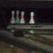}\hfill
\includegraphics[height=0.38in, width=0.38in]{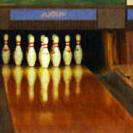}\newline
\includegraphics[height=0.38in, width=0.38in,frame=0.5pt -0.5pt]{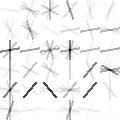}\hfill
\includegraphics[height=0.38in, width=0.38in]{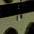}\hfill
\includegraphics[height=0.38in, width=0.38in]{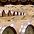}\hfill
\includegraphics[height=0.38in, width=0.38in]{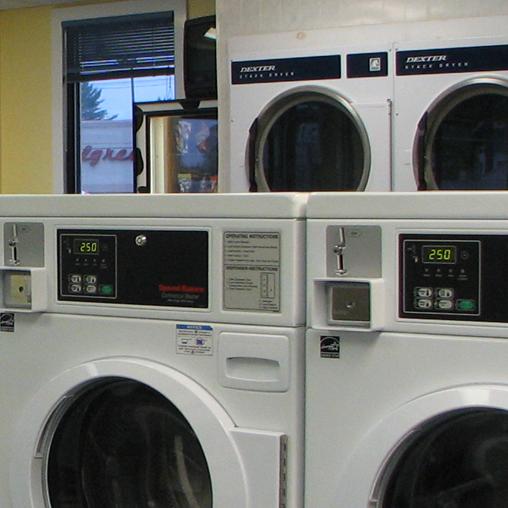}\hfill
\includegraphics[height=0.38in, width=0.38in]{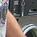}\hfill
\includegraphics[height=0.38in, width=0.38in]{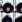} \hspace{0.4cm}
\includegraphics[height=0.38in, width=0.38in,frame=0.5pt -0.5pt]{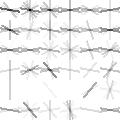}\hfill
\includegraphics[height=0.38in, width=0.38in]{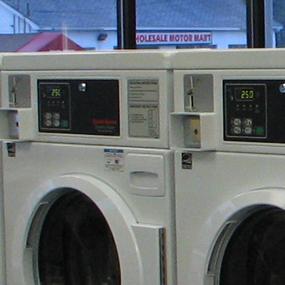}\hfill
\includegraphics[height=0.38in, width=0.38in]{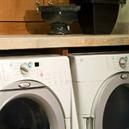}\hfill
\includegraphics[height=0.38in, width=0.38in]{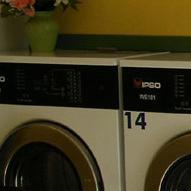}\hfill
\includegraphics[height=0.38in, width=0.38in]{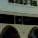}\hfill
\includegraphics[height=0.38in, width=0.38in]{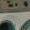}\newline
\includegraphics[height=0.38in, width=0.38in,frame=0.5pt -0.5pt]{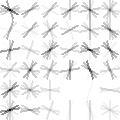}\hfill
\includegraphics[height=0.38in, width=0.38in]{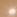}\hfill
\includegraphics[height=0.38in, width=0.38in]{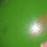}\hfill
\includegraphics[height=0.38in, width=0.38in]{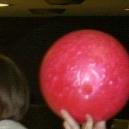}\hfill
\includegraphics[height=0.38in, width=0.38in]{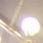}\hfill
\includegraphics[height=0.38in, width=0.38in]{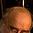} \hspace{0.4cm}
\includegraphics[height=0.38in, width=0.38in,frame=0.5pt -0.5pt]{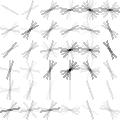}\hfill
\includegraphics[height=0.38in, width=0.38in]{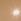}\hfill
\includegraphics[height=0.38in, width=0.38in]{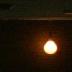}\hfill
\includegraphics[height=0.38in, width=0.38in]{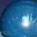}\hfill
\includegraphics[height=0.38in, width=0.38in]{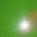}\hfill
\includegraphics[height=0.38in, width=0.38in]{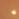} \\
\vspace{-0.75em}
\caption{Part filters before (left) and after joint training (right) and top scoring detections for each.}\label{fig:filters_before_and_after}
\vspace{-1.25em}
\end{figure}
Computer vision makes abundant use of the concept of ``part''. There
are at least three key reasons why parts are useful for representing objects
or scenes. One reason is the existence of non-linear and non-invertible nuisance
factors in the generation of images, including occlusions. By
breaking an object or image into parts, at least some of these may be visible
and recognizable. A second reason is that parts can be recombined 
in a model to express a combinatorial number of variants of an object or scene. 
For example parts corresponding to objects (\eg a laundromat and a desk)
can be rearranged in a scene, and parts of objects (\eg the face and
the clothes of a person) can be replaced by other parts.  A third
reason is that parts are often distinctive of a particular
(sub)category of objects (\eg cat faces usually belong to
cats).

Discovering good parts is a difficult problem that has recently raised
considerable interest (\cite{juneja13,doersch13,sun13}).  The quality
of a part can be defined in different ways.  Methods such as
(\cite{juneja13,doersch13}) decouple learning parts and image
classifiers by optimizing an intermediate objective that is only
heuristically related to classification.  Our first contribution is to
learn instead a \emph{system of discriminative parts jointly with the
  image classifiers}, optimizing the overall classification
performance on a training set.  We propose a unified framework for
training all of the model parameters jointly
(Section~\ref{sec:joint_training}). We show that joint training can
substantially improve the quality of the models
(Section~\ref{sec:experiments}).

A fundamental challenge in part learning is a classical chicken-and-egg problem: 
without an appearance model, examples of a part cannot be found, 
and without having examples an appearance model cannot be learned. 
To address this methods such as (\cite{juneja13,HOEIMCVPR13}) start from a single random example to initialize a part model, and alternate between finding more examples and retraining the part model. As the quality of the learned part depends on the initial random seed, thousands of parts are generated and a distinctive and diverse subset is extracted by means of some heuristic. Our second contribution is to propose a simple and effective alternative (Section~\ref{sec:init}). We still initialize a large pool of parts from random examples; we use these initial part models, each trained from a single example, to train image classifiers using $\lolt$ regularization as in (\cite{sun13}). This removes uninformative and redundant parts through group sparsity. This simple method produces better parts than more elaborate alternatives.  Joint training (Section~\ref{sec:experiments}) improve the quality of the parts further. 

Our pipeline, comprising random part initialization, part selection, 
and joint training is summarized in Figure~\ref{fig:pipeline}. 
In Section~\ref{sec:experiments} we show empirically that,
although our part detectors have the same form as the models 
in (\cite{juneja13,sun13}), they can reach a higher level of 
performance using a fraction of the number of parts. 
This translates directly to test time speedup.
%We also demonstrate importance of flip-invariant representation
%for image classification obtained by averaging features extracted 
%from an image and its right-to-left flip as done in~\cite{doersch13}.
We present experiments with both HOG~(\cite{dalal05}) and
CNN~(\cite{cnn}) features and improve the state-of-the-art results on
the MIT-indoor dataset (\cite{mit_indoor}) using CNN features.

A final contribution of our paper is the introduction of the concept
of negative parts, i.e.\ parts that are negatively correlated with respect to a class (Section~\ref{sec:negative_parts}). These parts are still informative as ``counter-evidence'' for the class. In certain formulations, negative parts
are associated to negative weights in the model and in others with negative 
weight differences. 

\begin{figure}
\hfill
\fbox{\begin{minipage}[c][0.9in][t]{1.7in}
\centering
{\bf Random Part Initialization}
\small
\begin{enumerate}[leftmargin=.5cm]
\itemsep2pt \parskip0pt \parsep0pt
\raggedright
\item Extract feature from a patch at random image and location.
\item Whiten the feature.
\item Repeat to construct a pool of candidate parts.
\end{enumerate}
\end{minipage}}
\fbox{\begin{minipage}[c][0.9in][t]{1.75in}
\centering
{\bf Part Selection}
\small
\begin{enumerate}[leftmargin=.5cm]
\itemsep2pt \parskip0pt \parsep0pt
\raggedright
\item Train part weights $u$ with $\lolt$ regularization.
\item Discard parts that are not used according to $u$.
\end{enumerate}
\end{minipage}}
\fbox{\begin{minipage}[c][0.9in][t]{1.65in}
\centering
{\bf Joint Training}
\small
\begin{enumerate}[leftmargin=.5cm]
\raggedright
\itemsep2pt \parskip0pt \parsep0pt
\item Train part weights $u$ keeping part filters $w$ fixed.
\item Train part filters $w$ keeping part weights $u$ fixed.
\item Repeat until convergence.
\end{enumerate}
\end{minipage}}
\hfill
\vspace{-0.8em}
\caption{Our pipeline.  Part selection and joint training are
  driven by classification loss.  Part selection is important because
  joint training is computationally demanding.}
\label{fig:pipeline}
\vspace{-1.25em}
\end{figure}

%\paragraph{Related Work.}

\subsection{Related Work}
Related ideas in part learning have been recently explored in 
(\cite{singh2012,juneja13,sun13,doersch13}). 
The general pipeline in all of these approaches is a two-stage
procedure that involves pre-training a set of discriminative parts
followed by training a classifier on top of the vector of the part
responses.  The differences in these methods lay in the details of how
parts are discovered. Each approach uses a different heuristic to find
a collection of parts such that each part scores high on a subset of
categories (and therefore is discriminative) and, collectively, they
cover a large area of an image after max-pooling (and therefore are
descriptive).  Our goal is similar, but we achieve part diversity, 
distinctiveness, and coverage as natural byproducts of optimizing 
the ``correct'' objective function, i.e.\ the final image 
classification performance. 

Reconfigurable Bag of Words (RBoW) model~\cite{naderi12} is another
part-based model used for image classification. 
RBoW uses latent variables to define a mapping from image regions to 
part models. In contrast, the latent variables in our model 
define a mapping from parts to image regions.

% pff: I think this is redundant
%These methods also introduced heuristics to select subsets of good parts 
%from an initial pool; we also perform part selection in our randomized 
%part initialization, but even in this case we are guided by the minimization 
%of classification error, resulting in more informative parts.

It has been shown before (\cite{girshick13}) that joint training is 
important for the success of part-based models in object detection. 
Differently from them, however, we share parts among multiple classes 
and define a joint optimization in which multiple classifiers are 
learned concurrently. In particular, the same part can vote strongly 
for a subset of the classes and against another subset.
   
%The work most closely related to ours is the recent paper by Lobel et al.~\cite{lobel13}. 
The most closely related work to ours is (\cite{lobel13}). Their model 
has two sets of parameters; a dictionary of visual words $\theta$ and 
a set of weights $u$ that specifies the importance the visual words in 
each category.  
%% The score of category $y$ on image $x$ is computed by
%% max-pooling features in a fixed set of regions in the image.  Let
%% $\Phi(x, \theta)_{r, j}$ be the maximum response of visual word
%% $\theta_j$ on the pooling region $r$.  The score of category $y$ on
%% image $x$ is a dot-product $u_y \cdot \Phi(x, \theta)$.  The
%% max-pooling in $\Phi$ makes this a latent variable model.  
Similar to what we do here,~\cite{lobel13} trains $u$ and $\theta$ 
jointly (visual words would be the equivalent of part filters in 
our terminology). However, they assume that $u$ is non-negative.  
This assumption does not allow for ``negative parts'' as we describe 
in Section \ref{sec:negative_parts}.

%Their method requires recomputing maximum-response of visual words in
%all regions multiple times which would take prohibitively long in our
%setting. We show that one can use a caching mechanism to considerably
%reduce the amount of this computation. We also prove that the caching
%mechanism leads to an exact solution.

The concept of negative parts and relative attributes (\cite{parikh11}) are related 
in that both quantify the relative strength of visual patterns.  Our parts are trained 
jointly using using image category labels as the only form of suppervision, whereas 
the relative attributes in (\cite{parikh11}) were trained independently using labeled 
information about the strength of hand picked attributes in training images. 
%More importantly, we train negative parts jointly with image classifiers whereas 
%the relative attributes in \cite{parikh11} are pre-trained.

\section{Part-Based Models and Negative Parts}
\label{sec:negative_parts}

%This section describes our part-based image classifiers and discusses the notion of negative parts, including in relationship to DPMs~\cite{dpm}.
We model an image class using a collection of parts. A part may capture the appearance of
an entire object (\eg bed in a \emph{bedroom} scene), a part of an
object (\eg drum in the \emph{laundromat} scene), a rigid composition
of multiple objects (\eg rack of clothes in a \emph{closet} scene), or a region type (\eg ocean in a \emph{beach} scene).

Let $x$ be an image.
We use $H(x)$ to denote the space of latent values for a part.  In our
experiments $H(x)$ is a set of positions and scales in a scale pyramid. 
To test if the image $x$ contains part $j$ at location $z_j \in H(x)$, we extract features $\psi(x,z_j)$ and take the dot product of this feature vector with a part filter $w_j$.  Let 
$s(x,z_j,w_j)$ denote the response of part $j$ at location $z_j$ in $x$.  Since the location of a part is unknown, it is
treated as a latent variable which is maximized over.  This defines the response $r(x,w_j)$ of a part in an image,
\begin{equation}\label{eq:part-scores}
s(x,z_j,w_j) = w_j \cdot \psi(x,z_j),
\qquad 
r(x,w_j) = \max_{z_j \in H(x)} s(x,z_j,w_j).
\end{equation}
%The maximization over the latent variable results in a response that reflects whether the part occurs anywhere in the image or not. 
Given a collection of $m$ parts $w=(w_1, \ldots, w_m)$, their responses are collected in an $m$-dimensional vector of part responses $r(x,w) = (r(x,w_1);\dots;r(x,w_m))$. In practice, filter responses are pooled within several distinct \emph{spatial subdivisions} (\cite{Lazebnik06}) to encode weak geometry.  In this case we have $R$ pooling regions and $r(x,w)$ is an $mR$-dimensional vector maximizing part responses in each pooling region.  For the rest of the paper we assume a single pooling region to simplify notation.

Part responses can be used to predict the class of an image. For example, high response for ``bed'' and ``lamp'' would suggest the image is of a ``bedroom'' scene.
%In binary classification the goal is to predict if an image belongs to a foreground class  $y=+1$ (``bedroom'') or a background class $y=-1$ (``everything else'').
Binary classifiers are often used for multi-class classification with a one-vs-all setup.
DPMs (\cite{dpm}) also use binary classifiers to detect objects of each class.  For a binary classifier we can define a score function $f_\beta(x)$ for the foreground hypothesis.  The score combines part responses using a vector of part weights $u$,
\begin{align}\label{eq:part-model}
 f_\beta(x) = u \cdot r(x,w), \quad \beta = (u,w).
\end{align}
The binary classifier predicts $y=+1$ if $f_\beta(x) \ge 0$, and $y=-1$ otherwise.

{\bf Negative parts in a binary classifier:} If $u_j > 0$ we say part
$j$ is a \emph{positive part} for the foreground class and if $u_j <
0$ we say part $j$ is a \emph{negative part} for the foreground class.

Intuitively, a negative part provides counter-evidence for the
foureground class; i.e. $r(x, w_j)$ is negatively correlated with 
$f_\beta(x)$.  For example, since cows are not usually in bedrooms 
a high response from a \emph{cow} filter should penalize the 
score of a \emph{bedroom} classifier.

Let $\beta = (u,w)$ be the parameters of a binary classifier.  We can
multiply $w_j$ and divide $u_j$ by a positive value $\alpha$ to obtain
an equivalent model.  If we use $\alpha=|u_j|$ we obtain a model where
$u\in\{-1,+1\}^m$.  However, in general it is not possible to
transform a model where $u_j$ is negative into a model where $u_j$ is
positive because of the $\max$ in~\eqref{eq:part-scores}.

We note that, when $u$ is non-negative the score function $f_\beta(x)$ is convex
in $w$.  On the other hand, if there are negative parts, $f_\beta(x)$
is no longer convex in $w$.  If $u$ is non-negative then \eqref{eq:part-model} reduces to the scoring function of a latent SVM, and a special case of a DPM. By the argument above when $u$ is non-negative we can assume $u=\mathbf{1}$ and \eqref{eq:part-model} reduces to
\begin{equation}
 f_\beta(x) = \sum_{j=1}^m \max_{z_j \in H(x)} w_j \cdot \psi(x,z_j)
 = \max_{z\in Z(x)} w \cdot \Psi(x,z),
 \label{eq:CPBM_score_function1}
\end{equation}
where $Z(x) = H(x)^m$, and $\Psi(x,z) = (\psi(x, z_1); \dots; \psi(x, z_m))$. In the case of a DPM, the feature vector $\Psi(x,z)$ and the model parameters contain additional terms capturing spatial relationships between parts.  In a DPM all part responses are positively correlated with the score
of a detection.  Therefore DPMs do not use negative parts.

\subsection{Negative Parts in Multi-Class Setting}
\label{sec:negative_parts_multi_class}

In the previous section we showed that certain one-vs-all part-based classifiers, including 
DPMs, cannot capture counter-evidence from
negative parts.  This limitation can be
addressed by using more general models with two sets of parameters
$\beta=(u,w)$ and a scoring function $f_\beta(x) = u \cdot r(x,w)$, as long
as we allow $u$ to have negative entries.

\iffalse
Another example where negative parts may be useful is the case of challenging false alarms when detecting similar animals.~\cite{hoiem12} argues that similarity between objects such as \emph{cow} and \emph{horse} is a major factor that contributes to the high false positive rate in detecting these objects. However, despite the strong similarity between their body-shape, their heads are very different. A \emph{cow} detector with a negative part (that is trained to fire on \emph{horse}-head) can be the key to resolving such sources of ambiguity. In that case, the \emph{cow} detector fires only when there is strong evidence for \emph{cow}-body \emph{and} a \emph{horse}-head is not present. And similarly for the \emph{horse} detector.
\fi

%Now we consider the case of a multi-class classifier where different
%image categories share a set of part filters. 
Now we consider the case of a multi-class classifier where part responses are 
weighted differently for each category but all categories share the same set of part filters.
A natural consequence of part sharing is that a positive part for one class can 
be used as a negative part for another class. 
% In the multi-class case we have a
%single model with shared parts and part weights for
%each class.  

Let $\mathcal{Y} = \{1, \dots, n\}$ be a set of $n$ categories.  A
multi-class part-based model $\beta=(w,u)$ is defined by $m$ part
filters $w=(w_1,\ldots,w_m)$ and $n$ vectors of part weights
$u=(u_1,\ldots,u_n)$ with $u_y \in \mathbb{R}^m$.  The shared filters
$w$ and the weight vector $u_y$ define parameters
$\beta_y=(w,u_y)$ for a scoring function for class $y$.  For an input
$x$ the multi-class classifier selects the class with highest score
\begin{equation}
\hat{y}_{\beta}(x) = \argmax_{y \in \mathcal{Y}} f_{\beta_y}(x) = \argmax_{y \in \mathcal{Y}} u_{y} \cdot r(x, w) 
\label{eq:classifier}
\end{equation}
We can see $u$ as an $n \times m$ matrix.  Adding a
constant to a column of $u$ does not change the differences between
scores of two classes $f_{\beta_a}(x) - f_{\beta_b}(x)$.  This implies
 the function $\hat{y}$ is invariant to such transformations.  We
can use a series of such transformations to make all entries in $u$
non-negative, without changing the classifier. 
Thus, in a multi-class part-based model, unlike the binary case,
it is not a significant restriction to require the entries in $u$ to
be non-negative.  In particular the sign of an entry in $u_y$ does not
determine the type of a part (positive or negative) for class $y$.

{\bf Negative parts in a multi-class classifier:} 
If $u_{a,j} > u_{b,j}$ we say part $j$ is a \emph{positive part} for class $a$
\emph{relative} to $b$.  If $u_{a,j} < u_{b,j}$ we say part $j$ is
a \emph{negative part} for class $a$ \emph{relative} to $b$.

Although adding a constant to a column of $u$ does not affect
$\hat{y}$, it does impact the norms of the part weight vectors $u_y$.
For an $\ell_2$ regularized model the columns of $u$ will sum to zero.
Otherwise we can subtract the column sums from each column of $u$ to
decrease the $\ell_2$ regularization cost without changing $\hat{y}$
and therefore the classification loss.  We see that in the multi-class
part-based model constrainig $u$ to have non-negative entries only
affects the regularization of the model.  

%In this paper, we propose a framework for training a multi-class part-based classifier that uses a shared pool of part filters as well as a class-specific part weight matrix. The part weight matrix can have real values and, besides the relative importance of each part for each category, determines whether response of a part should be added or subtracted from the classification score of an image with respect to each category. 

\section{Joint Training}
\label{sec:joint_training}
In this section we propose an approach for joint training of all
parameters $\beta=(w,u)$ of a multi-class part-based model.  Training
is driven directly by classification loss.  Note that a
classification loss objective is sufficient to encourage diversity of
parts.  In particular joint training encourages part filters to
complement each other.  We have found that joint training leads to a
substantial improvement in performance (see
Section~\ref{sec:experiments}).  The use of classification loss to
train all model parameters also leads to a simple framework that
does not rely on multiple heuristics.

Let $\mathcal{D} = \{(x_i, y_i)\}_{i=1}^k$ denote a training set of
labeled examples.  We train $\beta$ using $\ell_2$ regularization for both the part
filters $w$ and the part weights $u$ (we think of each as a single vector) and the multi-class hinge
loss, resulting in the objective function:
\begin{align}
O(u,w) & = \lambda_w ||w||^2 + \lambda_u ||u||^2 + \sum_{i=1}^k \max \left\{0, 1+(\max_{y \neq y_i} u_y \cdot r(x_i, w)) - u_{y_i} \cdot r(x_i,w)\right\}\label{eq:np_objective0} \\
       & = \lambda_w ||w||^2 + \lambda_u ||u||^2 + \sum_{i=1}^k \max \left\{0,\ %
 1+\max_{y \neq y_i} (u_y - u_{y_i}) \cdot r(x_i, w)\right\}\label{eq:np_objective_function}
\end{align}
We use block coordinate descent for training, as summarized in Algorithm \ref{alg:np_optimization}. This alternates between
(Step 1) optimizing $u$ while $w$ is fixed and (Step 2)
optimizing $w$ while $u$ is fixed. The first step reduces to a convex
structural SVM problem (line~3 of Algorithm~\ref{alg:np_optimization}). 
If $u$ is non-negative the second step could be reduced to a
latent structural SVM problem defined by (\ref{eq:np_objective0}).
We use a novel approach that
allows $u$ to be negative (lines~4-7 of Algorithm~\ref{alg:np_optimization}) described below.

%That said, we observed that constraining $u$ to be non-negative
%does not affect classification performance in our experiments. The following paragraphs describe the two steps in further detail.

%The loss function $L_i$ is not convex. If $w$'s are fixed, however,
%it becomes convex (in fact $O$ becomes a Structural-SVM
%objective). If $u$'s are fixed, on the other hand, the loss function
%becomes a sum-of-convex-concave functions and, as such, can be
%minimized to a reasonable local minimum using the CCCP method
%~\cite{yuille03}.

%Algorithm \ref{alg:np_optimization} outlines the training procedure.
%Line 3 minimizes $O$ over $u$ for fixed $w$.  This is a convex
%problem.  Lines 4-7 minimize $O$ over $w$ for fixed $u$ using CCCP.
%In the rest of this section we explain how the two major steps
%(\ref{ln:alg:update_u} \& \ref{ln:alg:update_w}) in this iterative
%procedure are carried out.

\begin{algorithm}[t]
\begin{algor}[1]
	\item [{*}] initialize the part filters $w=(w_1, \dots, w_m)$
	\item [repeat] %\{joint-training loop\}
		\item [{*}] $u := \argmin_{u'} O(u', w)$ (defined in Equation \ref{eq:np_objective_function}) \label{ln:alg:update_u}
%		\item [{*}] optimize $O$ over $\boldsymbol{\beta}$ keeping $w$'s fixed \label{ln:alg:update_u}
%		\item [{*}] $t := 0$
%		\item [{*}] $w^{(t)} := w$
		\item [repeat] %\{latent-variable-update loop\}
			\item[{*}] $w^{\text{old}} := w$
			\item [{*}] $w := \argmin_{w'} B_u(w', w^{\text{old}})$ (defined in Equation \ref{eq:np_objective_function_bound}) \label{ln:alg:update_w}
%			\item [{*}] optimize $O$ over $\boldsymbol{\beta}$ keeping $u$'s fixed \label{ln:alg:update_w}
%			\item [{*}] use $w^{(t)}$ to construct $O_u^{(t)}$, a convex upper bound of $O$ (see Eq. \ref{eq:np_objective_function_bound})
%			\item [{*}] $w^{(t+1)} := \argmin_w O_u^{(t)}(w)$ \label{ln:alg:update_w}
%			\item [{*}] $t := t+1$ \label{ln:alg:proceed}
%			\item [{*}] $w := \argmin_{w'} O_u(w', w)$ (see Equation \ref{eq:np_objective_function_bound}) \label{ln:alg:update_w}
		\item [until] convergence
%		\item [{*}] $w := w^{(t)}$
	\item [until] convergence
	\item [{*}] output $\beta=(u, w)$
\end{algor}
\caption{\label{alg:np_optimization} Joint training of model parameters by optimizing $O(u,w)$ in Equation
  \ref{eq:np_objective_function}.}
\end{algorithm}
%\vspace{-1.25em}

\subsubsection*{Step 1: Learning Part Weights (line 3 of Algorithm \ref{alg:np_optimization})}

%\label{sec:learning_u}
This involves computing $\argmin_{u'} O(u', w)$.  Since $w$ is fixed
$\lambda_w ||w||^2$ and $r(x_i, w)$ are constant. This makes the
optimization problem equivalent to training a multi-class SVM where
the $i$-th training example is represented by an $m$-dimensional
vector of part responses $r(x_i, w)$.  This is a convex problem that can be
solved using standard methods.

\subsubsection*{Step 2: Learning Part Filters (lines 4-7 of Algorithm \ref{alg:np_optimization})}

%\label{sec:learning_w}
This involves computing $\argmin_{w'} O(u, w')$.  Since $u$ is fixed
$\lambda_u ||u||^2$ is constant. 
While $r(x_i, w_j)$ is convex in $w$ (it is a maximum of linear functions) 
the coefficients $u_{y, j} - u_{y_i, j}$ may be negative.  This makes the 
objective function (\ref{eq:np_objective_function}) non-convex.
Lines 4-7 of Algorithm \ref{alg:np_optimization} implement the CCCP algorithm (\cite{yuille03}).
In each iteration we construct a convex bound using the previous estimate of $w$
and update $w$ to be the minimizer of the bound.

%The key to convergence of this iterative procedure is that the minimum value of the bound is non-increasing over time.
%decreases as we update $w$ in line 6 of the algorithm; that is 
%$\forall t$ if $w^{t+1} = \argmax_{w'} B_u(w', w^t)$ and $w^{t+2} = \argmax_{w'} B_u(w', w^{t+1})$ then $B_u(w^{t+2}, w^{t+1}) \leq B_u(w^{t+1}, w^t)$. 

Let $s(x, z, w) =
(s(x, z_1, w_1); \dots; s(x, z_m, w_m))$ to be the vector of part
responses in image $x$ when the latent variables are fixed to $z=(z_1, \dots, z_m)$. We
construct a convex upper bound on $O$ by replacing $r(x_i, w_j)$ with
$s(x_i, z_{i,j}, w_j)$ in (\ref{eq:np_objective_function}) when
$u_{y, j} - u_{y_i, j} < 0$.  We make the bound tight for the last estimate of the part filters $w^{\text{old}}$
by selecting $z_{i,j} = \argmax_{z_j \in H(x_i)} s(x_i, z_j,
w^{\text{old}}_j)$. Then a convex upper bound that touches $O$ at $w^{\text{old}}$ is
given by $\lambda_u ||u||^2 + B_u(w, w^{\text{old}})$ with
\begin{align}
	\hspace{-0.2cm}
	\sq{B_u(w, w^{\text{old}}) = \lambda_w ||w||^2 
	+ \sum_{i=1}^k \max\left\{0,\ %
	1 + \max_{y \neq y_i} (u_y - u_{y_i}) \cdot
	\left[ 
	S_{y, y_i} r(x_i, w) + 
	\bar{S}_{y, y_i} s(x_i, z_i, w)
	\right]
	\right\}} \label{eq:np_objective_function_bound}
\end{align}
Here, for a pair of categories $(y, y')$, $S_{y, y'}$ and $\bar{S}_{y, y'}$ are $m \times m$ diagonal 0-1 matrices such that $\bar{S}_{y, y'}(j, j)=1-S_{y, y'}(j, j)$ and $S_{y, y'}(j, j) = 1$ if and only if $u_{y, j} - u_{y', j} \geq 0$.
The matrices $S$ and $\bar{S}$ select 
$r(x_i, w_j)$ when $u_{y, j} - u_{y_i, j} \geq 0$ and 
$s(x_i, z_{i,j}, w_j)$ when
$u_{y, j} - u_{y_i, j} < 0$.  This implements the convex upper-bound outlined above.

Line \ref{ln:alg:update_w} of the algorithm updates the part filters by minimizing $B_u(w, w^\text{old})$.
%  i.e.\ $w^{\text{new}}=\argmin_{w'} B_u(w', w^{\text{old}})$.
Optimizing this function requires significant computational and memory
resources.  
In the supplementary material (Section~\ref{sec:notes_on_optimization_supplement}) we give details of how our optimization
method works in practice.

\section{Part Generation and Selection}
\label{sec:init}
The joint training objective in (\ref{eq:np_objective_function}) is non-convex making Algorithm~\ref{alg:np_optimization} sensitive to initialization. Thus, the choice of initial parts can be crucial in training models that perform well in practice. We devote the first two steps of our pipeline to finding good initial parts (Figure~\ref{fig:pipeline}). We then use those parts to initialize the joint training procedure of Section \ref{sec:joint_training}.

In the first step of our pipeline we randomly generate a large pool of initial parts. Generating a part involves picking a random training image (regardless of the image category labels) and extracting features from a random subwindow of the image followed by whitening (\cite{LDA}). To whiten a feature vector $f$ we use $\Sigma^{-1}(f - \mu)$ where $\mu$ and $\Sigma$ are the mean and covariance of all patches in all training images. We estimate $\mu$ and $\Sigma$ from 300,000 random patches. We use the norm of the whitened features to estimate discriminability of a patch. Patches with large whitened feature norm are farther from the mean of the background distribution in the whitened space and, hence, are expected to be more discriminative.  Similar to (\cite{Aubry13}) we discard the $50\%$ least discriminant patches from each image prior to generating random parts.

Our experimental results with HOG features (Figure~\ref{fig:performance_HOG}) show that
randomly generated parts using the procedure described here perform
better than or comparable to previous methods that are much more involved (\cite{juneja13,doersch13,sun13}).   When using CNN features we get
very good results using random parts alone, even before part-selection
and training of the part filters (Figure~\ref{fig:performance_CNN}).

Random part generation may produce redundant or useless
parts. In the second step of our pipeline we train image
classifiers $u$ using $\lolt$ regularization (a.k.a. group lasso) to
select a subset of parts from the initial random pool.
We group entries in each column of $u$. Let $\rho_j$ 
denote the $\ell 2$-norm of the $j$-th column of $u$. 
The $\lolt$ regularization is defined by 
$R_g(u) = \lambda \sum_{j=1}^m \rho_j$.

If part $j$ is not uninformative or redundant $\rho_j$ (and therefore all entries in the $j$-th column of $u$)
will be driven to zero by the regularizer.  We train models using different values for $\lambda$
to generate a target number of parts. The number of selected parts decreases monotonically as $\lambda$ increases. 
Figure~\ref{fig:L1L2_norms} in the supplement shows this. We found it important to retrain $u$ after part selection using $\ell 2$ regularization to obtain good classification performance.

\section{Experiments}
\label{sec:experiments}
We evaluate our methods on the MIT-indoor dataset (\cite{mit_indoor}). 
We compare performance of models with randomly generated parts, selected parts, 
and jointly trained parts. We also compare performance of HOG and CNN features. The dataset has 67 indoor scene classes. There are about 80 training and 20 test images per class.  Recent part-based methods that do well on this dataset (\cite{juneja13,doersch13,sun13}) 
use a large number of parts (between 3350 and 13400).

\textbf{HOG features:}
We resize images (maintaining aspect ratio) to have about $2.5M$ pixels. 
We extract 32-dimensional HOG features (\cite{dalal05,dpm}) at multiple scales. 
Our HOG pyramid has 3 scales per octave. This yields about 11,000 patches per image. Each part filter $w_j$ models a $\sq{6 \times 6}$ grid of HOG features, so $w_j$ and $\psi(x, z_j)$ are both $1152$-dimensional.

\textbf{CNN features:}
We extract CNN features at multiple scales from overlapping patches of fixed size $\sq{256 \times 256}$ and with stride value $256/3=85$. We resize images (maintaining aspect ratio) to have about $5M$ pixels in the largest scale. We use a scale pyramid with 2 scales per octave. This yields about 1200 patches per image. 
We extract CNN features using Caffe (\cite{jia2014caffe}) and the hybrid neural network from (\cite{places}).
The hybrid network is pre-trained on images from ImageNet (\cite{imagenet}) and PLACES (\cite{places}) datasets. 
We use the output of the 4096 units in the penultimate fully connected layer of the network (fc7). We denote these features by \emph{HP} in our plots. 

\textbf{Part-based representation:}
Our final image representation is an $mR$-dimensional vector of part responses where 
$m$ is the number of shared parts and $R$ is the number of spatial pooling regions. 
We use $R=5$ pooling regions arranged in a $\sq{1 \times 1} + \sq{2 \times 2}$ grid.  
To make the final representation invariant to horizontal image flips we average 
the $mR$-dimensional vector of part responses for image $x$ and its right-to-left mirror image
$x'$ to get $\left[ r(x, w) +  r(x', w) \right] / 2$ as in (\cite{doersch13}). 

We first evaluate the performance of random parts. Given a pool of randomly initialized parts (Section~\ref{sec:init}), we train the part weights $u$ using a standard $\ell 2$-regularized linear SVM; we then repeat the experiment by selecting few parts from a large pool using $\lolt$ regularization (Section~\ref{sec:init}). Finally, we evaluate joint training (Section \ref{sec:joint_training}). While joint training significantly improves performance, it comes at a significantly increased computational cost.

\begin{figure}
        \centering
        \begin{subfigure}[b]{0.49\textwidth}
                \includegraphics[width=0.85\textwidth]{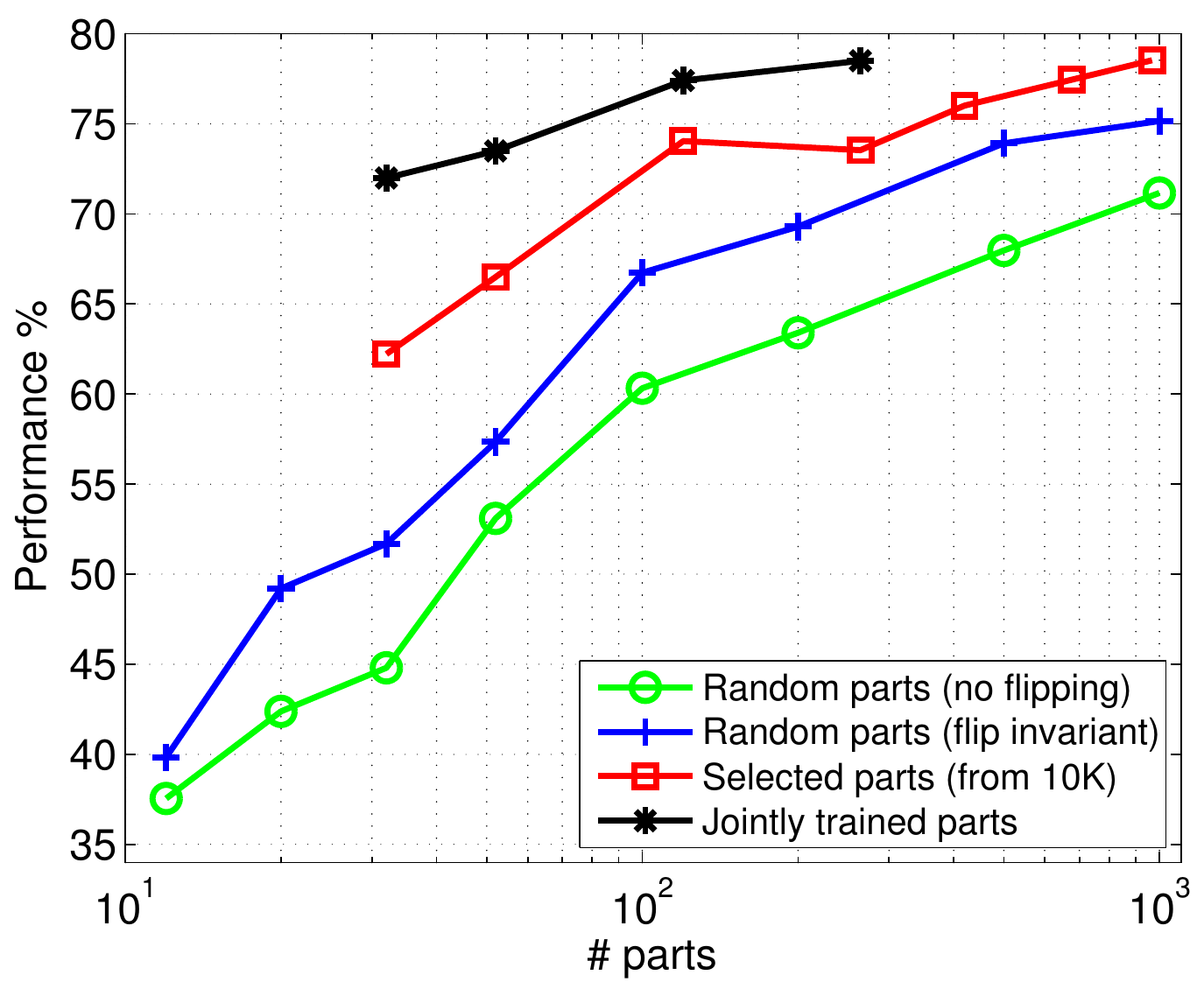}
%		\vspace{-0.3em}
%                \caption{10-Class Subset}
%                \label{fig:performance_HOG_subset}
        \end{subfigure}
	\hfill
        \begin{subfigure}[b]{0.49\textwidth}
                \includegraphics[width=0.85\textwidth]{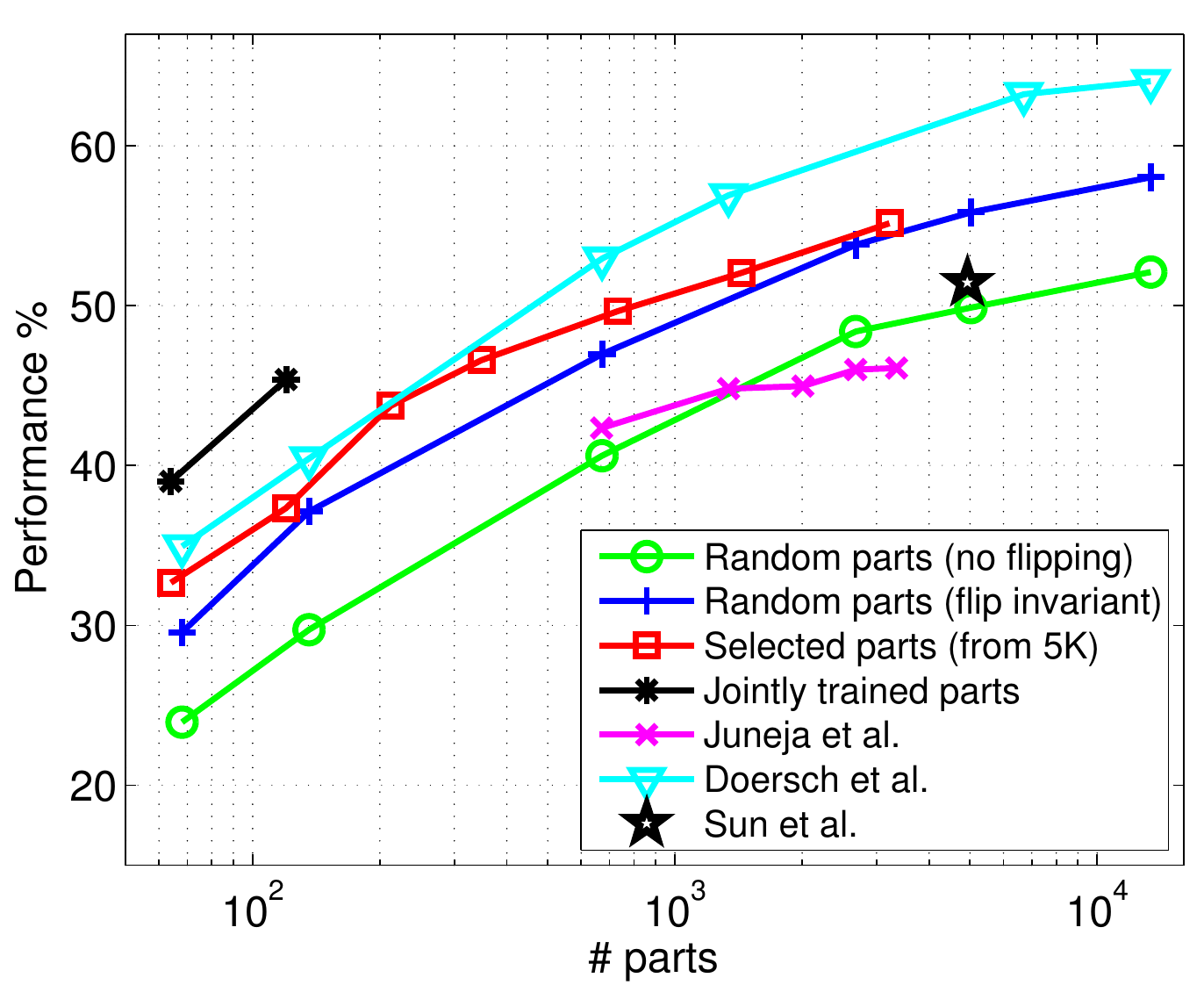}
%		\vspace{-0.3em}
%                \caption{Full Dataset}
%                \label{fig:performance_HOG_full}
        \end{subfigure}
	\vspace{-0.75em}
	\caption{Performance of HOG features on 10-class subset (left) and full MIT-indoor dataset (right).}
	\label{fig:performance_HOG}
	\vspace{-1.25em}
\end{figure}

Figure~\ref{fig:performance_HOG} shows performance of HOG features on the MIT-indoor dataset.  Because of the high dimensionality of the HOG features and the large space of potential placements in a HOG pyramid we consider a 10-class subset of the dataset for experiments with a large number of parts using HOG features. The subset comprises \emph{bookstore}, \emph{bowling}, \emph{closet}, \emph{corridor}, \emph{laundromat}, \emph{library}, \emph{nursery}, \emph{shoeshop}, \emph{staircase}, and \emph{winecellar}. 
Performance of random parts increases as we use more parts. Flip invariance and part selection consistently improve results. Joint training improves the performance even further by a large margin achieving the same level of performance as the selected parts using much fewer parts. On the full dataset, random parts already outperform the results from~\cite{juneja13}, flip invariance boosts the performance beyond~\cite{sun13}. Joint training dominates other methods. However, we could not directly compare with the best performance of~\cite{doersch13} due to the very large number of parts they use.

\begin{figure}
        \centering
        \begin{subfigure}[b]{0.49\textwidth}
                \includegraphics[width=0.85\textwidth]{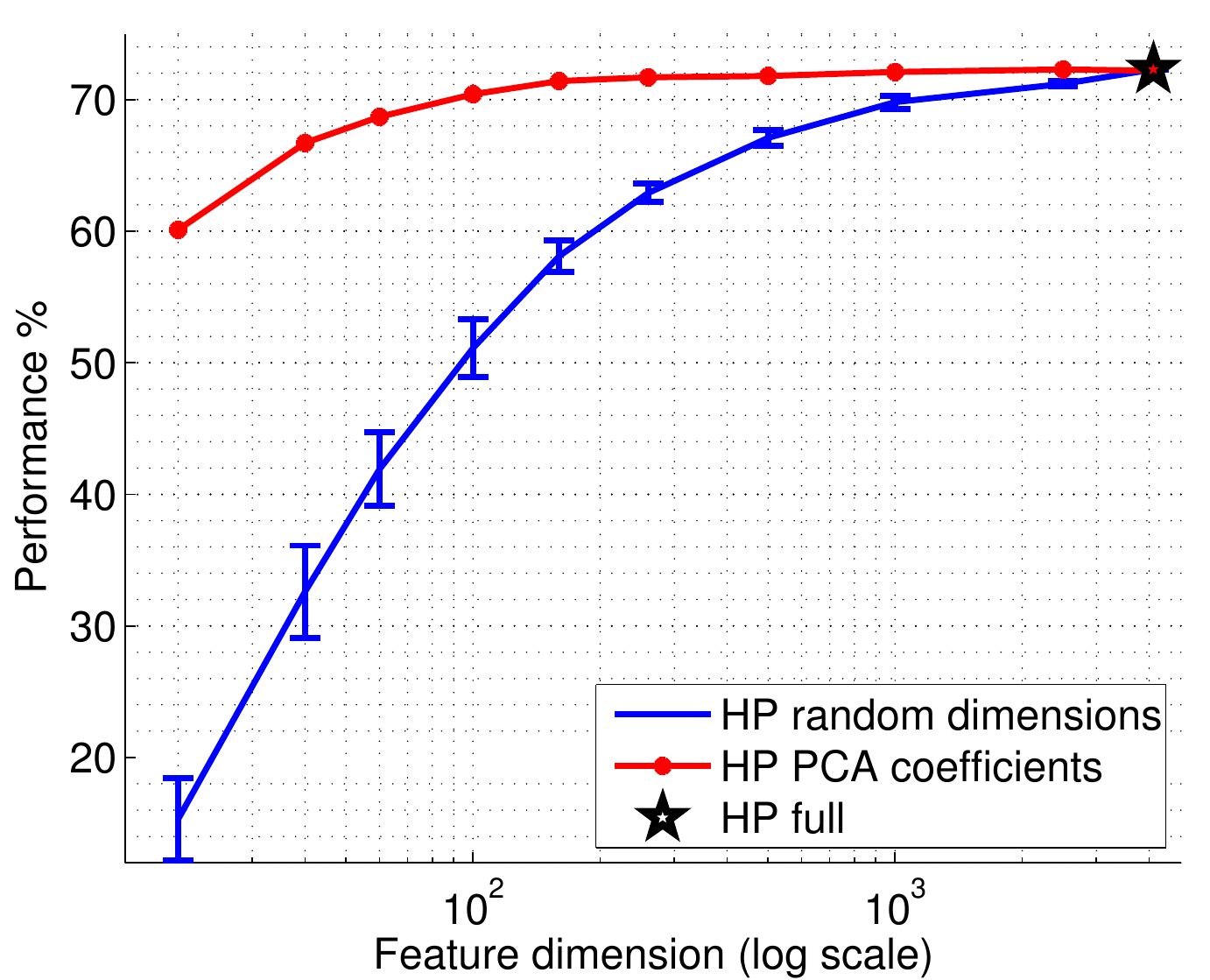}
%		\vspace{-0.3em}
%                \caption{Effect of feature dimensionality on performance}
%                \label{fig:performance_CNN_dimensions}
        \end{subfigure}
	\hfill
        \begin{subfigure}[b]{0.49\textwidth}
                \includegraphics[width=0.85\textwidth]{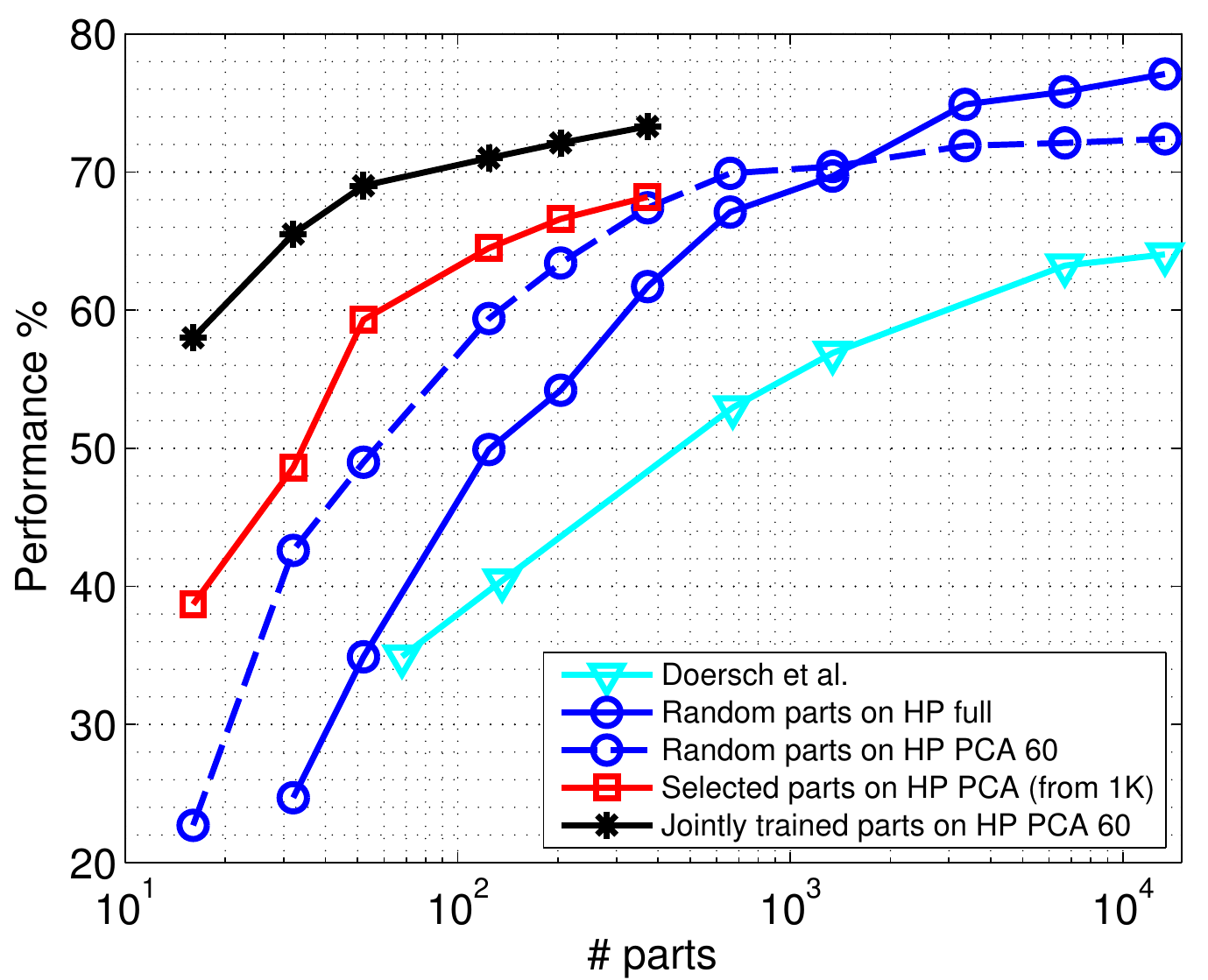}
%		\vspace{-0.3em}
%                \caption{Performance of part-based models using CNN}
%                \label{fig:performance_CNN_joint}
        \end{subfigure}
	\vspace{-0.75em}
	\caption{Performance of CNN features on the full MIT-indoor dataset. \emph{HP} denotes the hybrid features from~\cite{places}. \textbf{Left:} the effect of dimensionality reduction on performance of the CNN features extracted from the entire image. Two approaches are compared; random selection over 5 trials (blue curve) and PCA (red curve). \textbf{Right:} part-based models with random parts (blue curves), selected parts from 1K random parts (red curve), and jointly trained parts (black curve).}
	\label{fig:performance_CNN}
	\vspace{-1.25em}
\end{figure}

Figure~\ref{fig:performance_CNN} shows performance of CNN features on MIT-indoor dataset. As a baseline we extract CNN features from the entire image (after resizing to $\sq{256 \times 256}$ pixels) and train a multi-class linear SVM. This obtains $72.3\%$ average performance. This is a strong baseline. \cite{sharif14} get $58.4\%$ using CNN trained on ImageNet. They improve the result to $69\%$ after data augmentation.

We applied PCA on the 4096 dimensional features to make them more compact. This is essential for making the joint training tractable both in terms of running time and memory footprint. Figure~\ref{fig:performance_CNN}-left shows the effect of PCA dimensionality reduction. It is surprising that we lose only $1\%$ in accuracy with 160 PCA coefficients and only $3.5\%$ with 60 PCA coefficients. We also show how performance changes when a random subset of dimensions is used. For joint training we use 60 PCA coefficients.

Figure~\ref{fig:performance_CNN}-right shows performance of our part-based models using CNN features. For comparison with HOG features we also plot result of \cite{doersch13}. Note that part-based representation improves over the CNN extracted on the entire image. With 13400 random parts we get $77.1\%$ (vs $72.3\%$ for CNN on the entire image). The improvement is from $68.2\%$ to $72.4\%$ when we use 60 PCA coefficients. Interestingly, the 60 PCA coefficients perform better than the full CNN features when only a few parts are used (up to 1000). The gap increases as the number of parts decreases. 

We do part selection and joint training using 60 PCA coefficients of the CNN features. We select parts from an initial pool of 1000 random parts.  Part selection is most effective when a few parts are used. Joint training improves the quality of the selected parts. With only 372 jointly trained parts we obtain $73.3\%$ classification performance which is even better than 13400 random parts ($72.4\%$).

The significance of our results is two fold: 1) we demonstrate a very simple 
and fast to train pipeline for image classification using randomly generated parts; 
2) we show that using part selection and joint training we can obtain similar or 
higher performance using much fewer parts. The gain is largest for CNN features 
($13400/372 \approx 36$ times). This translates to 36x speed up in test time. 
See Section~\ref{sec:processing_time_supplement} of the supplement for 
detailed run-time analysis of our method.
%In Figure~\ref{fig:performance_HOG}-left random parts (the blue curve) reach 73.9\% using 500 parts while jointly trained parts (the black curve) reach 73.5\% using 52 parts. In Figure~\ref{fig:performance_HOG}-right random parts (the blue curve) reach 47.0\% using 672 parts while jointly trained parts (the black curve) reach 45.4\% using 120 parts.

\subsection{Visualization of the Model}
\label{sec:visualization}
\begin{figure*}[t]
	\centering
	\includegraphics[width=0.8\textwidth]{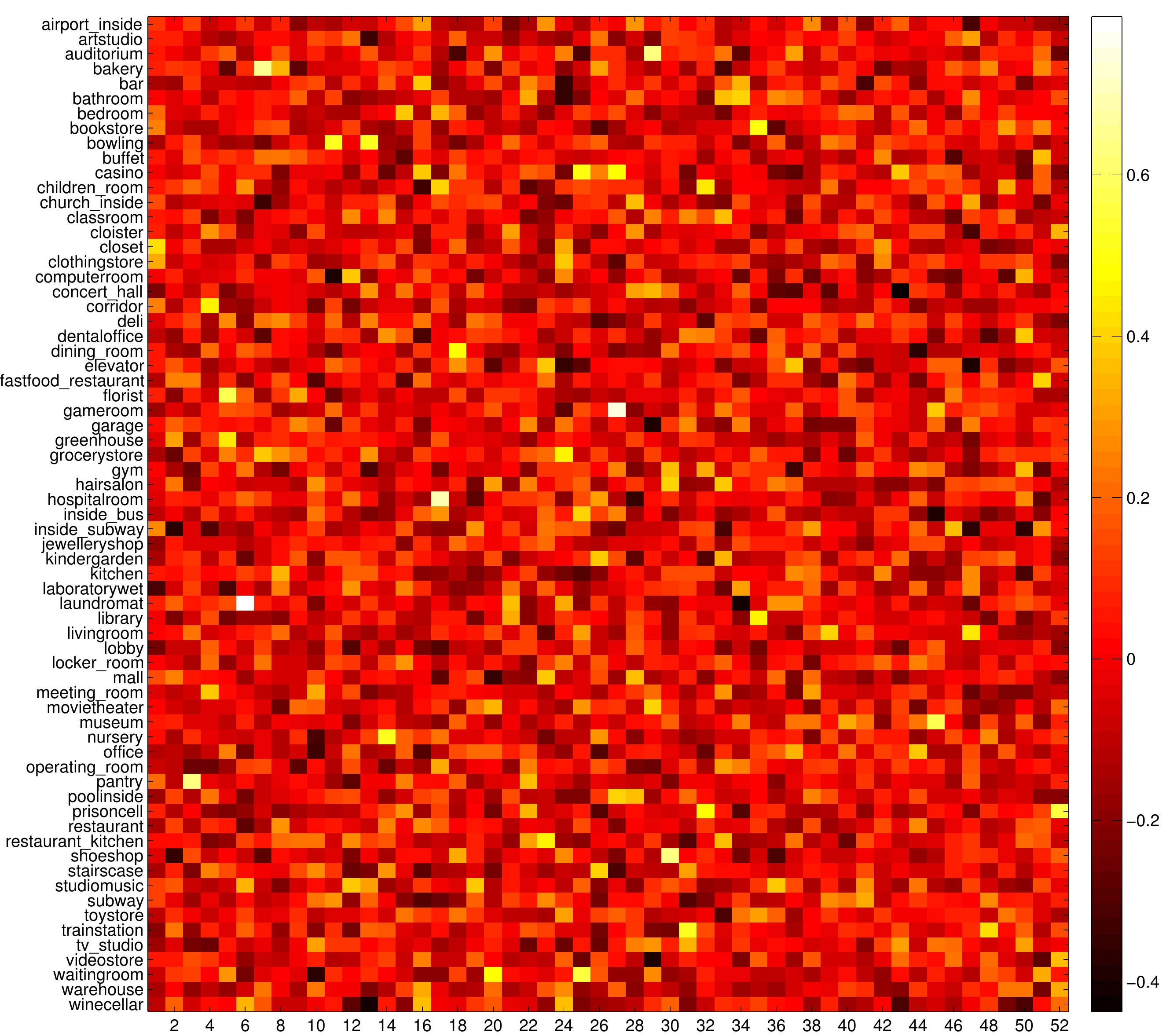}
	\vspace{-1em}
	\caption{Part weights after joint training a model with 52 parts on the full dataset. Here patches are represented using 60 PCA coefficients on CNN features. Although the model uses 5 pooling regions (corresponding to cells in $\sq{1 \times 1} + \sq{2 \times 2}$ grids) here we show the part weights only for the first pooling region corresponding to the entire image.}
	\label{fig:u_matrix_full}
	\vspace{-1.25em}
\end{figure*}
\begin{figure*}
	\centering
	\renewcommand{\tabcolsep}{0.75pt}
	\begin{tabular}{ c  c  c c c c c c  c c c c c c }
	 & & \multicolumn{12}{c}{Top scoring patches on test images (multiple patches per image)} \\
	\multirow{4}{*}{\rotatebox{90}{\hspace{-1.2cm} Part 25}$\;$} &
	\multirow{2}{*}{\rotatebox{90}{Before}$\;$} &
			\includegraphics[height=0.35in, width=0.35in]{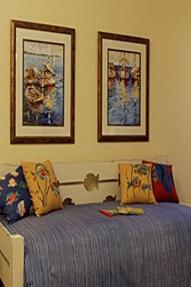} &
			\includegraphics[height=0.35in, width=0.35in]{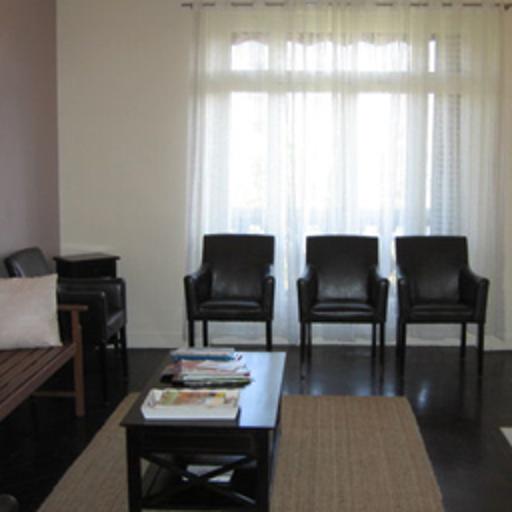} &
			\includegraphics[height=0.35in, width=0.35in]{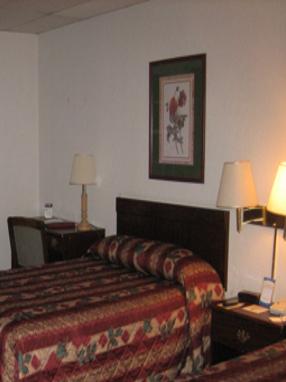} &
			\includegraphics[height=0.35in, width=0.35in]{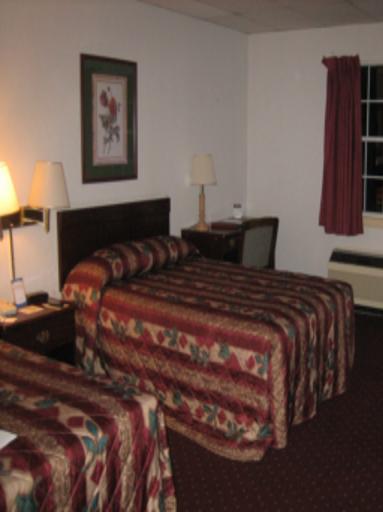} &
			\includegraphics[height=0.35in, width=0.35in]{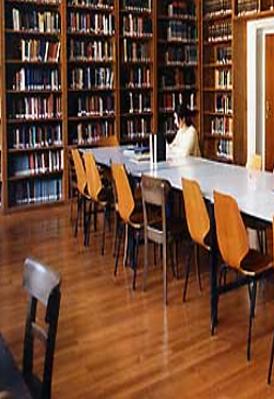} &
			\includegraphics[height=0.35in, width=0.35in]{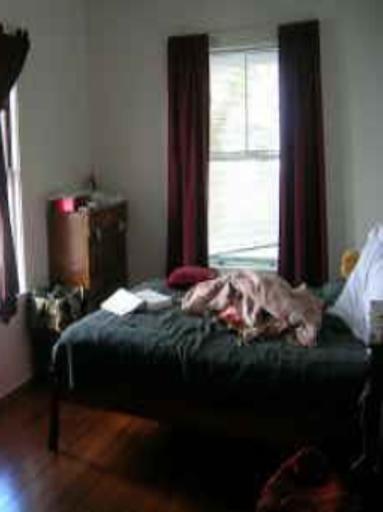} &
			\includegraphics[height=0.35in, width=0.35in]{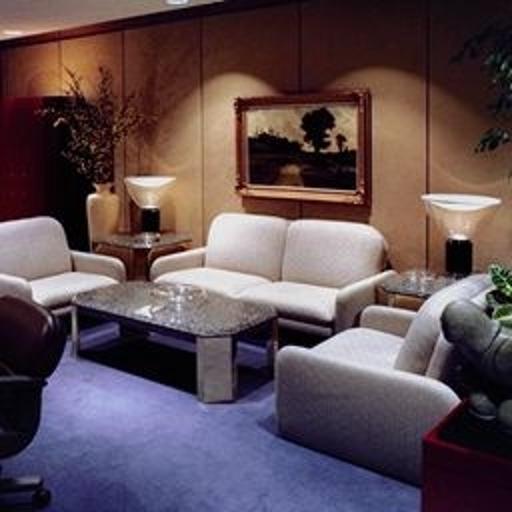} &
			\includegraphics[height=0.35in, width=0.35in]{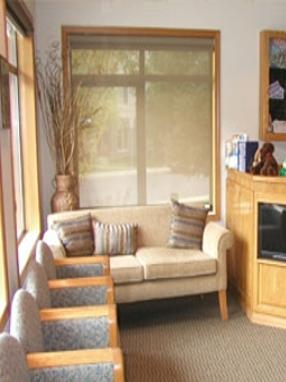} &
			\includegraphics[height=0.35in, width=0.35in]{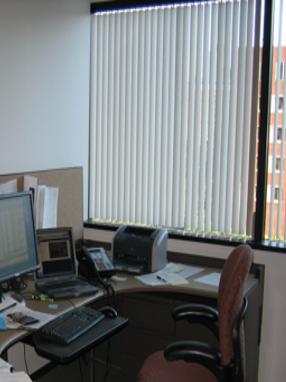} &
			\includegraphics[height=0.35in, width=0.35in]{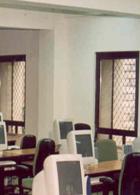} &
			\includegraphics[height=0.35in, width=0.35in]{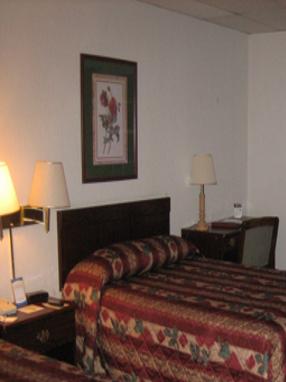} &
			\includegraphics[height=0.35in, width=0.35in]{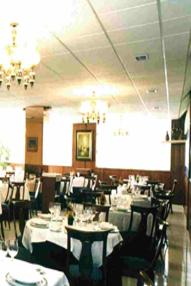} \\ [-0.05cm]
		& &	\includegraphics[height=0.35in, width=0.35in]{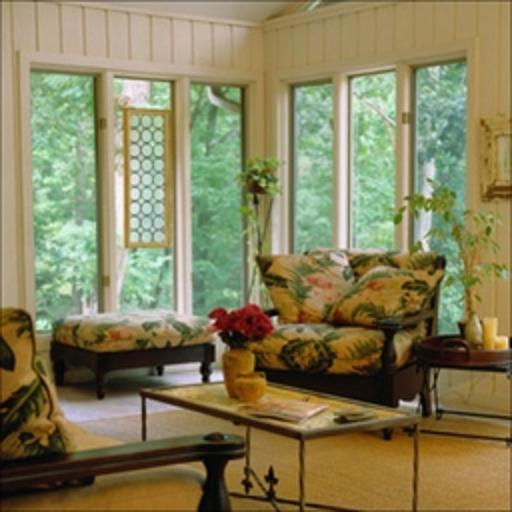} &
			\includegraphics[height=0.35in, width=0.35in]{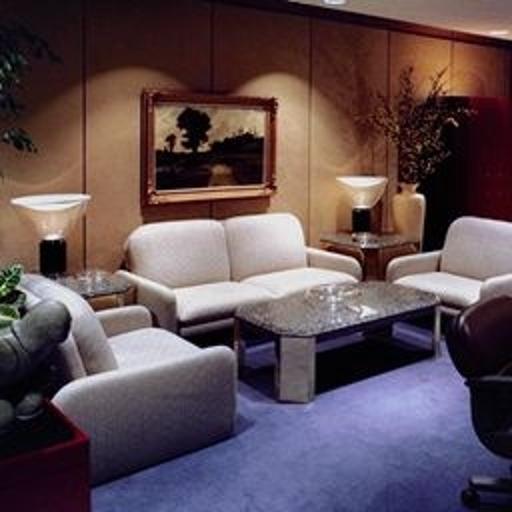} &
			\includegraphics[height=0.35in, width=0.35in]{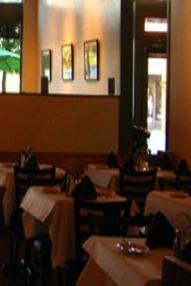} &
			\includegraphics[height=0.35in, width=0.35in]{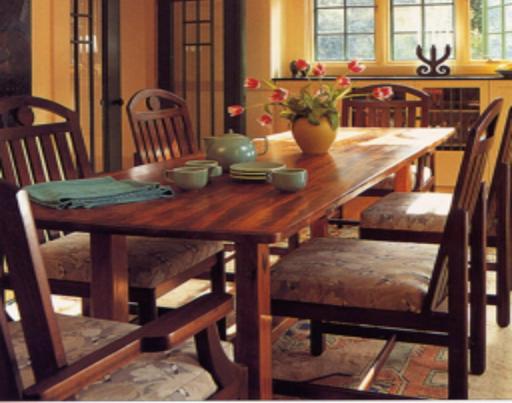} &
			\includegraphics[height=0.35in, width=0.35in]{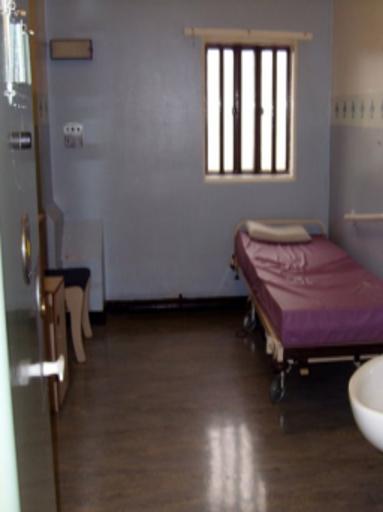} &
			\includegraphics[height=0.35in, width=0.35in]{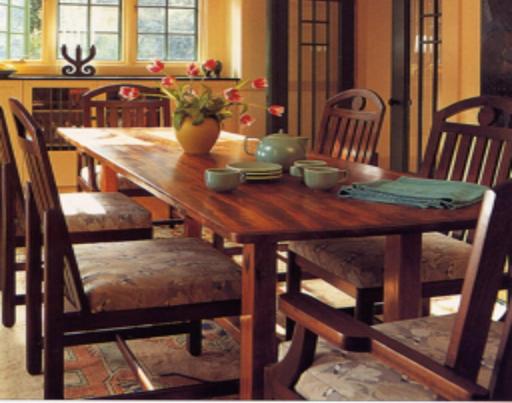} &
			\includegraphics[height=0.35in, width=0.35in]{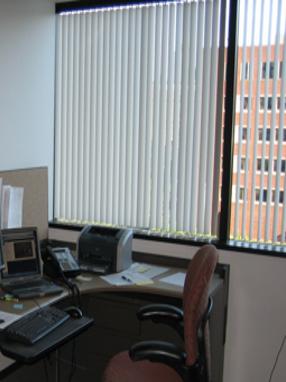} &
			\includegraphics[height=0.35in, width=0.35in]{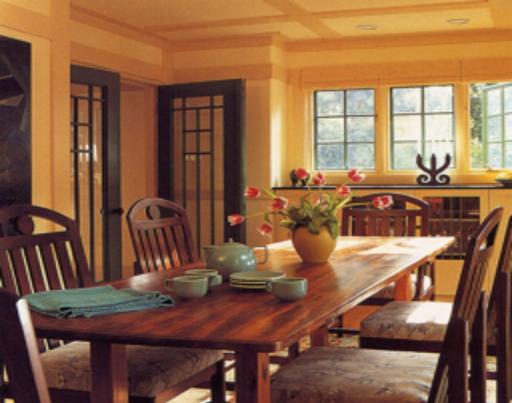} &
			\includegraphics[height=0.35in, width=0.35in]{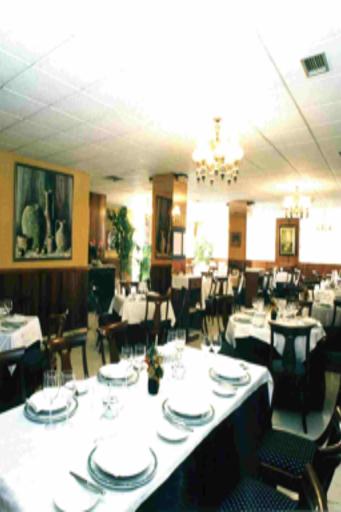} &
			\includegraphics[height=0.35in, width=0.35in]{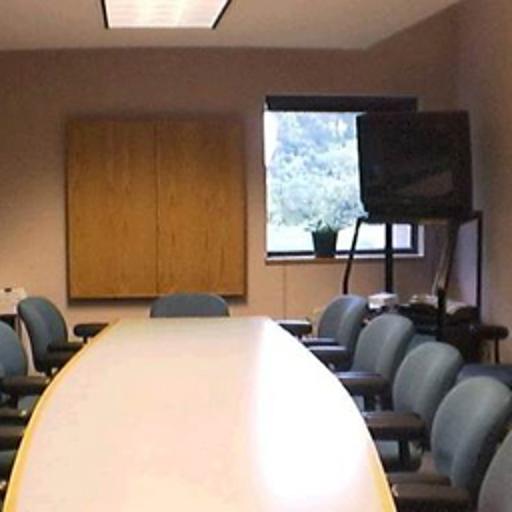} &
			\includegraphics[height=0.35in, width=0.35in]{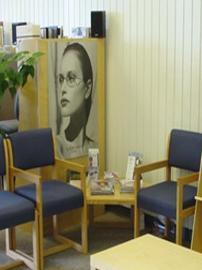} &
			\includegraphics[height=0.35in, width=0.35in]{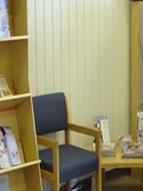}
			\\ [0.05cm]
	& \multirow{2}{*}{\rotatebox{90}{After}$\;$} &
			\includegraphics[height=0.35in, width=0.35in]{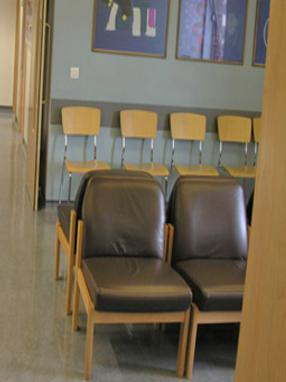} &
			\includegraphics[height=0.35in, width=0.35in]{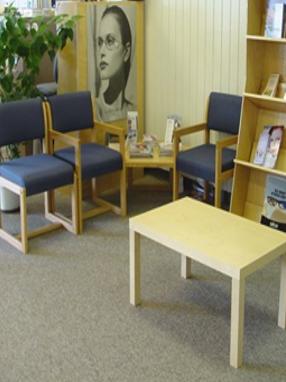} &
			\includegraphics[height=0.35in, width=0.35in]{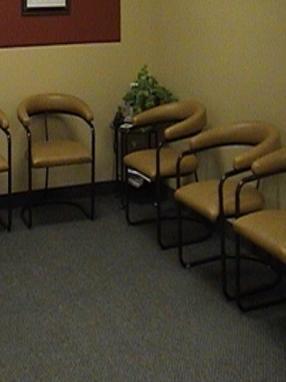} &
			\includegraphics[height=0.35in, width=0.35in]{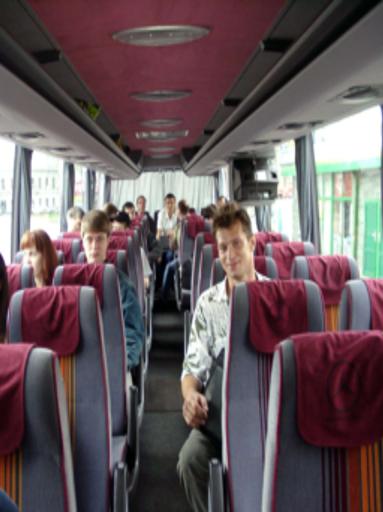} &
			\includegraphics[height=0.35in, width=0.35in]{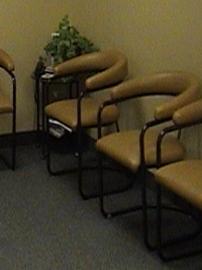} &
			\includegraphics[height=0.35in, width=0.35in]{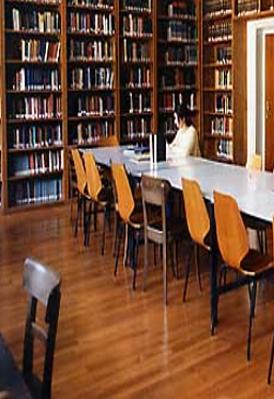} &
			\includegraphics[height=0.35in, width=0.35in]{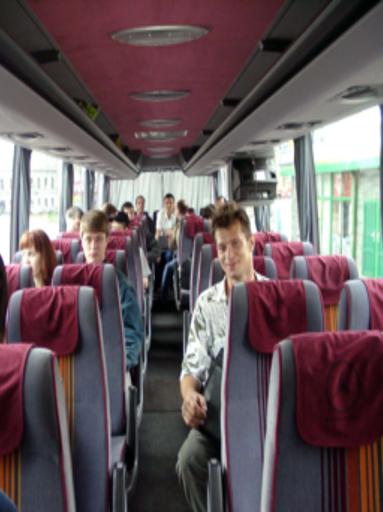} &
			\includegraphics[height=0.35in, width=0.35in]{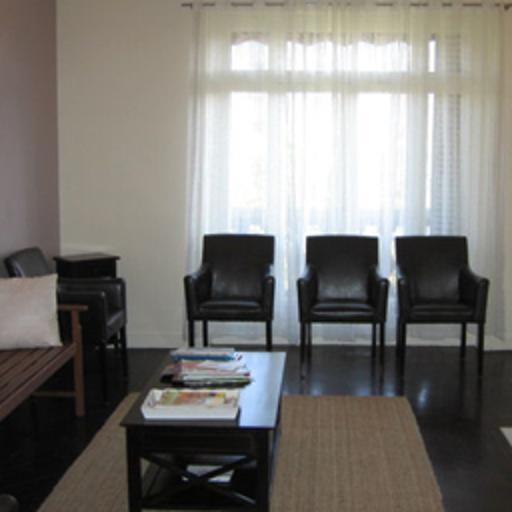} &
			\includegraphics[height=0.35in, width=0.35in]{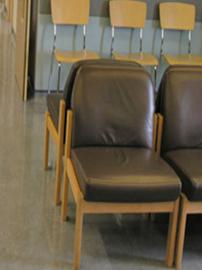} &
			\includegraphics[height=0.35in, width=0.35in]{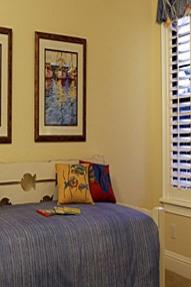} &
			\includegraphics[height=0.35in, width=0.35in]{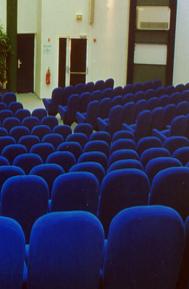} &
			\includegraphics[height=0.35in, width=0.35in]{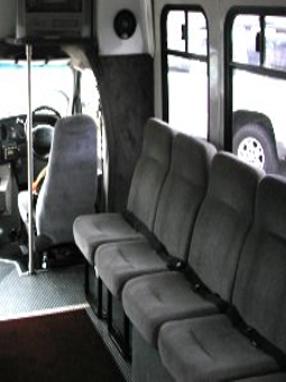} \\ [-0.05cm]
		& &	\includegraphics[height=0.35in, width=0.35in]{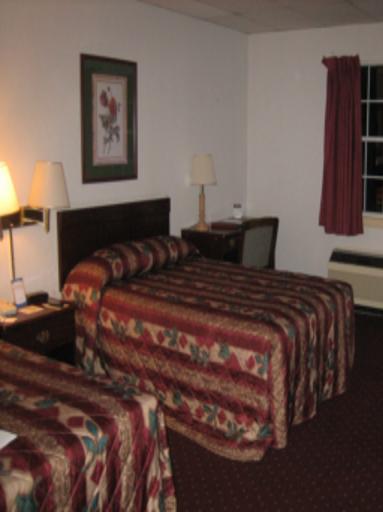} &
			\includegraphics[height=0.35in, width=0.35in]{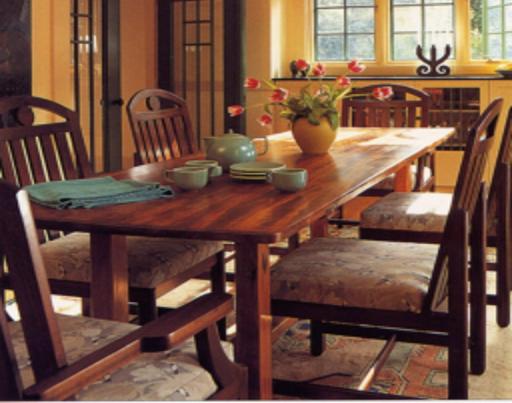} &
			\includegraphics[height=0.35in, width=0.35in]{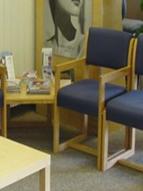} &
			\includegraphics[height=0.35in, width=0.35in]{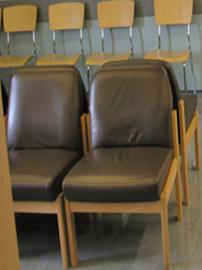} &
			\includegraphics[height=0.35in, width=0.35in]{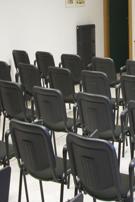} &
			\includegraphics[height=0.35in, width=0.35in]{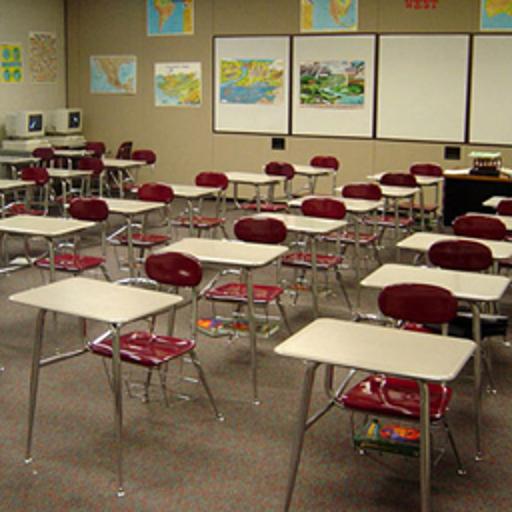} &
			\includegraphics[height=0.35in, width=0.35in]{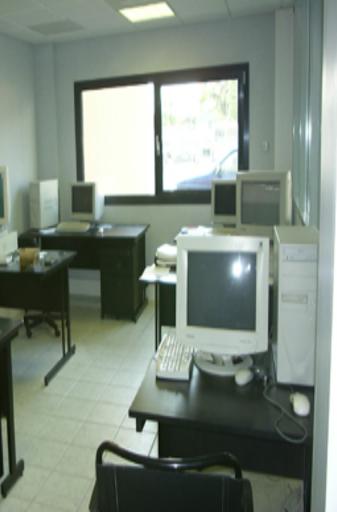} &
			\includegraphics[height=0.35in, width=0.35in]{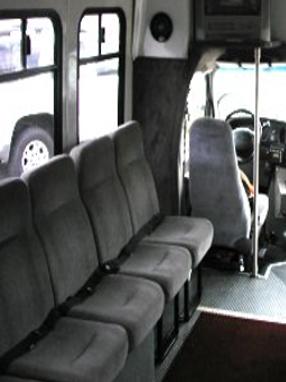} &
			\includegraphics[height=0.35in, width=0.35in]{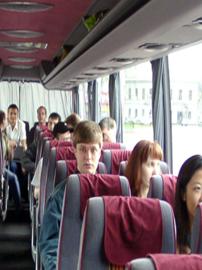} &
			\includegraphics[height=0.35in, width=0.35in]{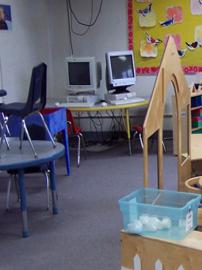} &
			\includegraphics[height=0.35in, width=0.35in]{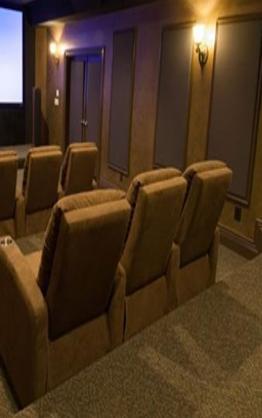} &
			\includegraphics[height=0.35in, width=0.35in]{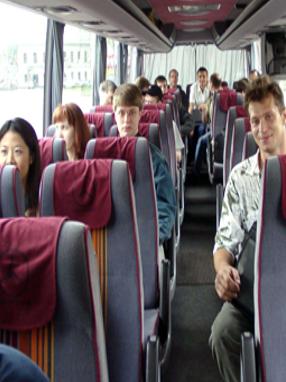}
			\\ [0.15cm]
	\multirow{4}{*}{\rotatebox{90}{\hspace{-1.2cm} Part 46}$\;$} &
	\multirow{2}{*}{\rotatebox{90}{Before}$\;$} &
			\includegraphics[height=0.35in, width=0.35in]{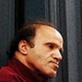} &
			\includegraphics[height=0.35in, width=0.35in]{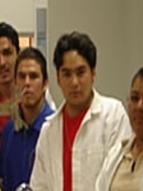} &
			\includegraphics[height=0.35in, width=0.35in]{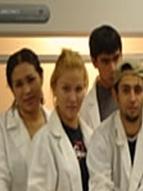} &
			\includegraphics[height=0.35in, width=0.35in]{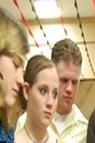} &
			\includegraphics[height=0.35in, width=0.35in]{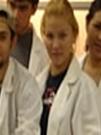} &
			\includegraphics[height=0.35in, width=0.35in]{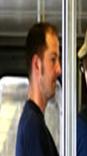} &
			\includegraphics[height=0.35in, width=0.35in]{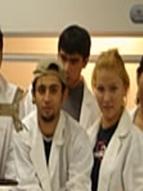} &
			\includegraphics[height=0.35in, width=0.35in]{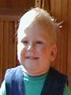} &
			\includegraphics[height=0.35in, width=0.35in]{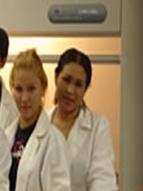} &
			\includegraphics[height=0.35in, width=0.35in]{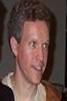} &
			\includegraphics[height=0.35in, width=0.35in]{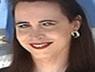} &
			\includegraphics[height=0.35in, width=0.35in]{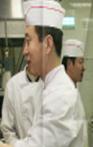} \\ [-0.05cm]
		& &	\includegraphics[height=0.35in, width=0.35in]{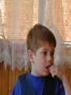} &
			\includegraphics[height=0.35in, width=0.35in]{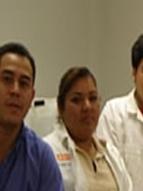} &
			\includegraphics[height=0.35in, width=0.35in]{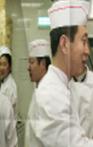} &
			\includegraphics[height=0.35in, width=0.35in]{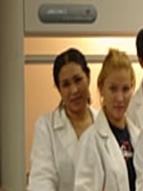} &
			\includegraphics[height=0.35in, width=0.35in]{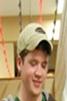} &
			\includegraphics[height=0.35in, width=0.35in]{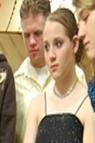} &
			\includegraphics[height=0.35in, width=0.35in]{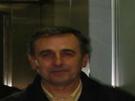} &
			\includegraphics[height=0.35in, width=0.35in]{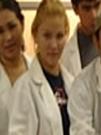} &
			\includegraphics[height=0.35in, width=0.35in]{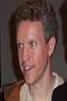} &
			\includegraphics[height=0.35in, width=0.35in]{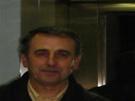} &
			\includegraphics[height=0.35in, width=0.35in]{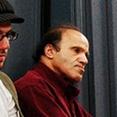} &
			\includegraphics[height=0.35in, width=0.35in]{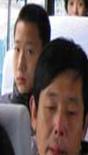}
			\\ [0.05cm]
	& \multirow{2}{*}{\rotatebox{90}{After}$\;$} &
			\includegraphics[height=0.35in, width=0.35in]{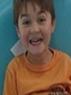} &
			\includegraphics[height=0.35in, width=0.35in]{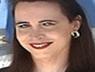} &
			\includegraphics[height=0.35in, width=0.35in]{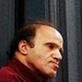} &
			\includegraphics[height=0.35in, width=0.35in]{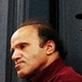} &
			\includegraphics[height=0.35in, width=0.35in]{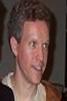} &
			\includegraphics[height=0.35in, width=0.35in]{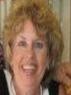} &
			\includegraphics[height=0.35in, width=0.35in]{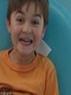} &
			\includegraphics[height=0.35in, width=0.35in]{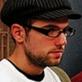} &
			\includegraphics[height=0.35in, width=0.35in]{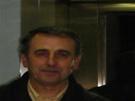} &
			\includegraphics[height=0.35in, width=0.35in]{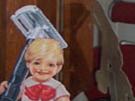} &
			\includegraphics[height=0.35in, width=0.35in]{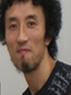} &
			\includegraphics[height=0.35in, width=0.35in]{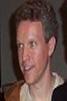} \\ [-0.05cm]
		& &	\includegraphics[height=0.35in, width=0.35in]{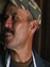} &
			\includegraphics[height=0.35in, width=0.35in]{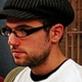} &
			\includegraphics[height=0.35in, width=0.35in]{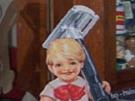} &
			\includegraphics[height=0.35in, width=0.35in]{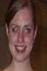} &
			\includegraphics[height=0.35in, width=0.35in]{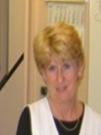} &
			\includegraphics[height=0.35in, width=0.35in]{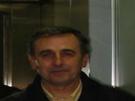} &
			\includegraphics[height=0.35in, width=0.35in]{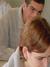} &
			\includegraphics[height=0.35in, width=0.35in]{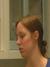} &
			\includegraphics[height=0.35in, width=0.35in]{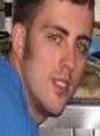} &
			\includegraphics[height=0.35in, width=0.35in]{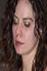} &
			\includegraphics[height=0.35in, width=0.35in]{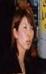} &
			\includegraphics[height=0.35in, width=0.35in]{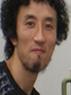}
	\end{tabular}
	\vspace{-1.2em}
	\caption{Top detections of two parts are shown before and after joint training on test images of the full MIT-indoor dataset. The numbers in the first column match the part indices in Figure~\ref{fig:u_matrix_full}.} 
	\label{fig:CNN_filters_before_and_after}
	\vspace{-1.25em}
\end{figure*}

Figure~\ref{fig:u_matrix_full} shows the part weight matrix after joint training a model with 52 parts on the full MIT-indoor dataset. This model uses 60 PCA coefficients from the \emph{HP} CNN features. Figure~\ref{fig:CNN_filters_before_and_after} shows top scoring patches for a few parts before and after joint training. The parts correspond to the model illustrated in Figure~\ref{fig:u_matrix_full}. The benefit of joint training is clear. The part detections are more consistent and ``clean'' after joint training.
%Part 17 seem to be detecting \emph{cribs} initially but after joint training it becomes a more accurate \emph{bed} detector while still firing on \emph{cribs}. 
The majority of the detections of part 25 before joint training are \emph{seats}. Joint training filters out most of the noise (mostly coming from \emph{bed} and \emph{sofa}) in this part. Part 46 consistently fires on \emph{faces} even before joint training. After joint training, however, the part becomes more selective to a single face and the detections become more localized.

\begin{figure*}[t]
	\vspace{-1cm}
	\renewcommand{\tabcolsep}{0.75pt}
	\begin{tabular}{ c   c c c c c c  }
	\rotatebox{90}{\hspace{0.27cm} Part 3}$\;$ &
				\includegraphics[height=0.63in, width=0.85in]{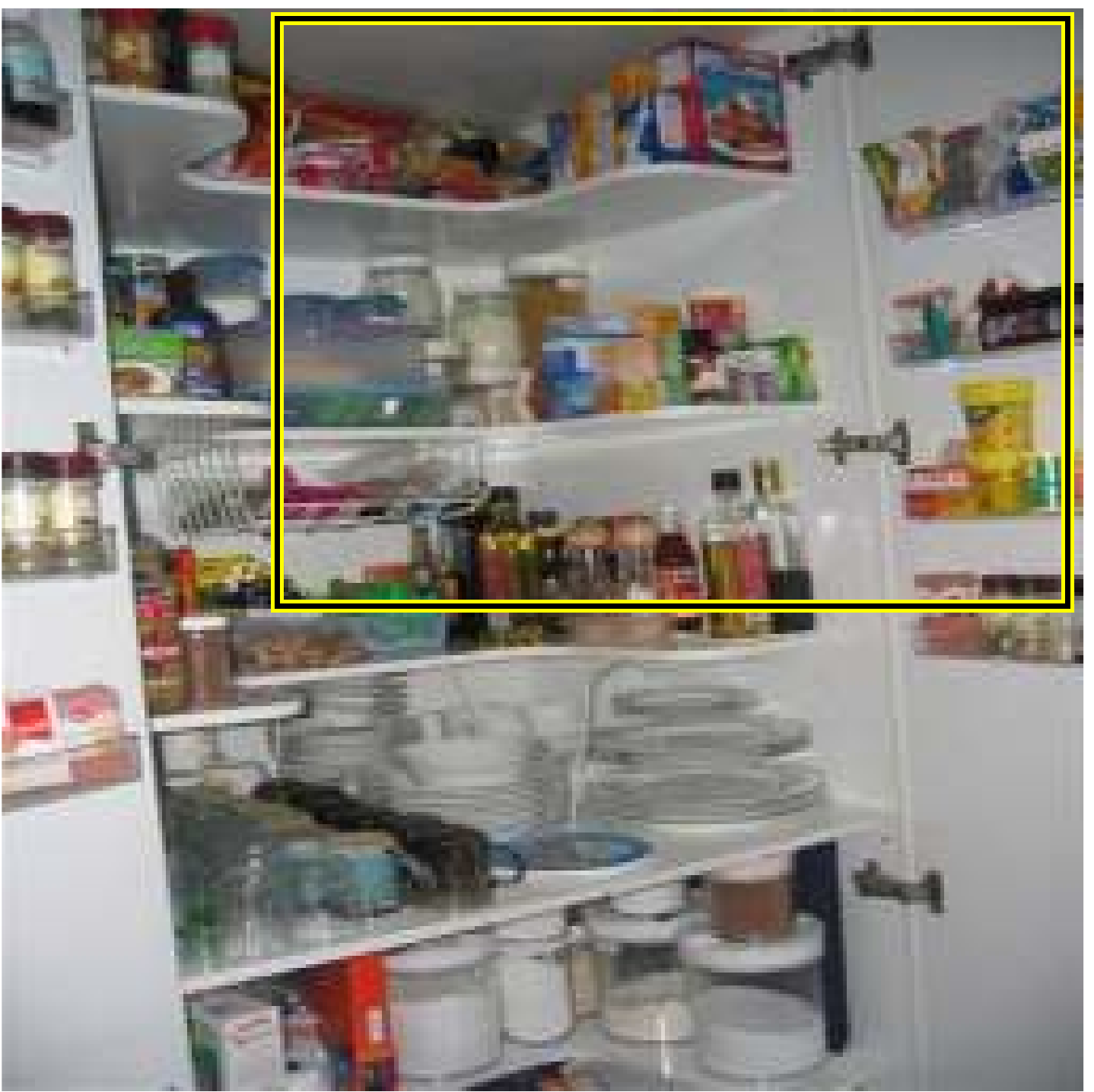} &
				\includegraphics[height=0.63in, width=0.85in]{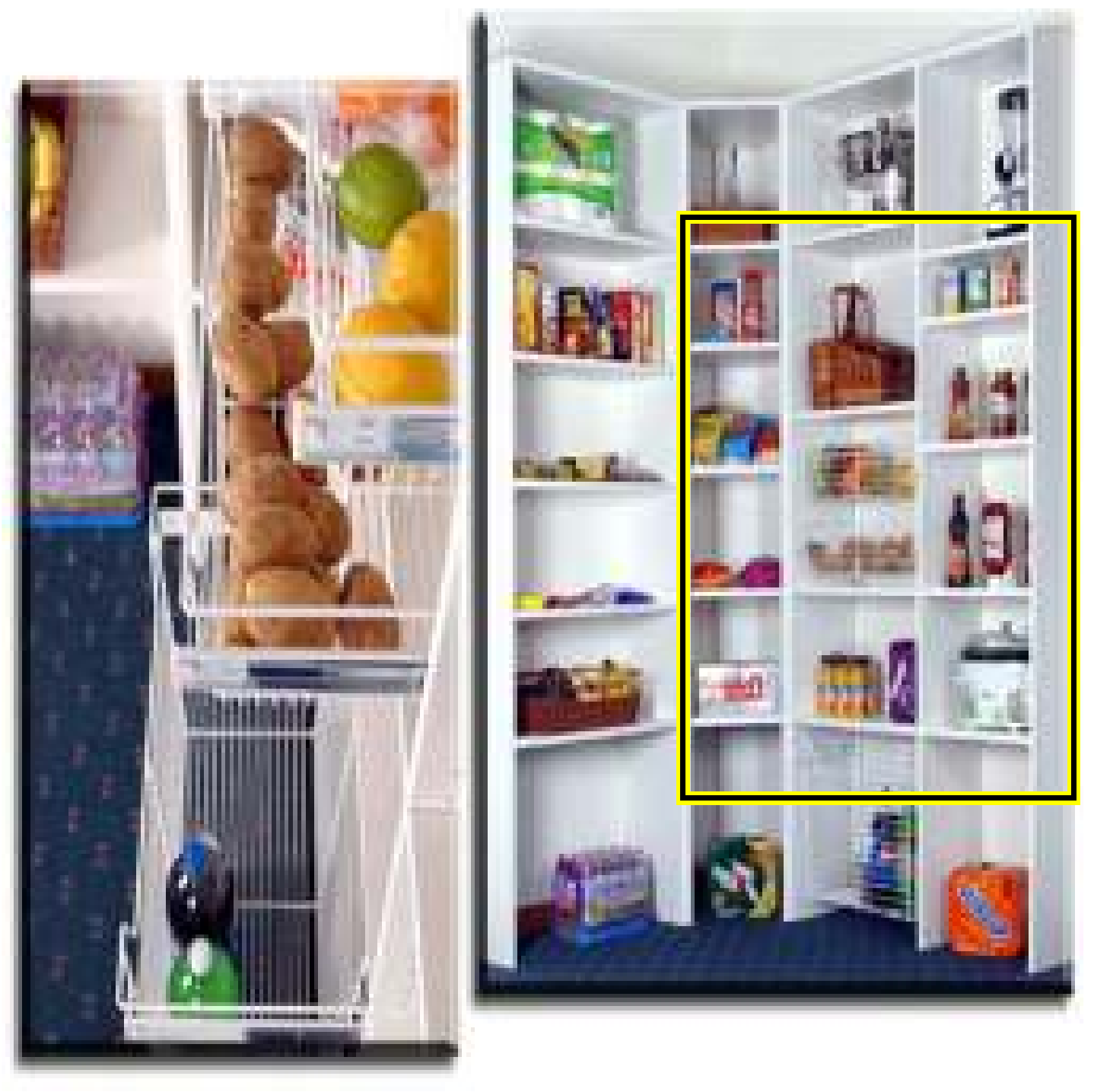} &
				\includegraphics[height=0.63in, width=0.85in]{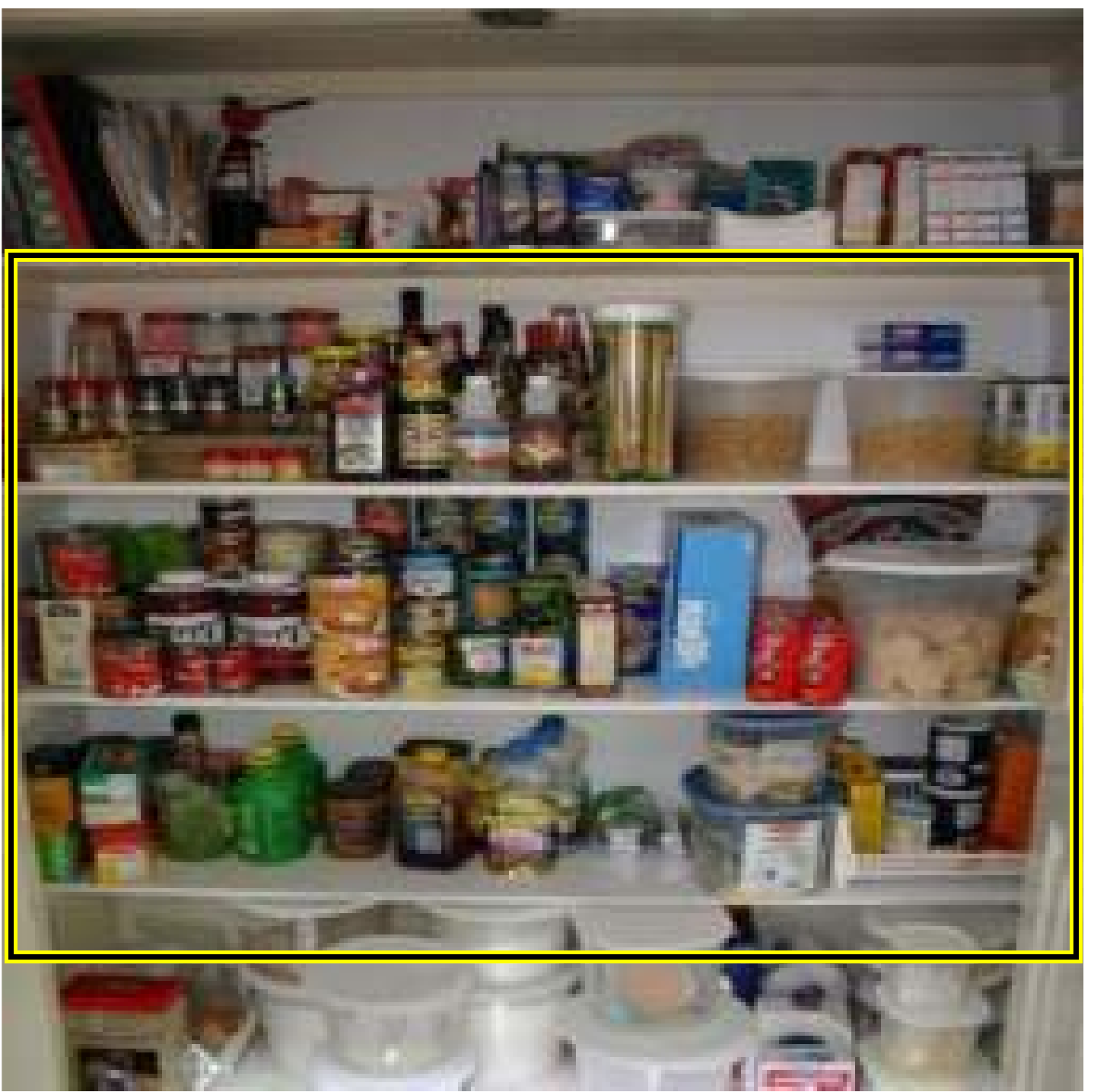} &
				\includegraphics[height=0.63in, width=0.85in]{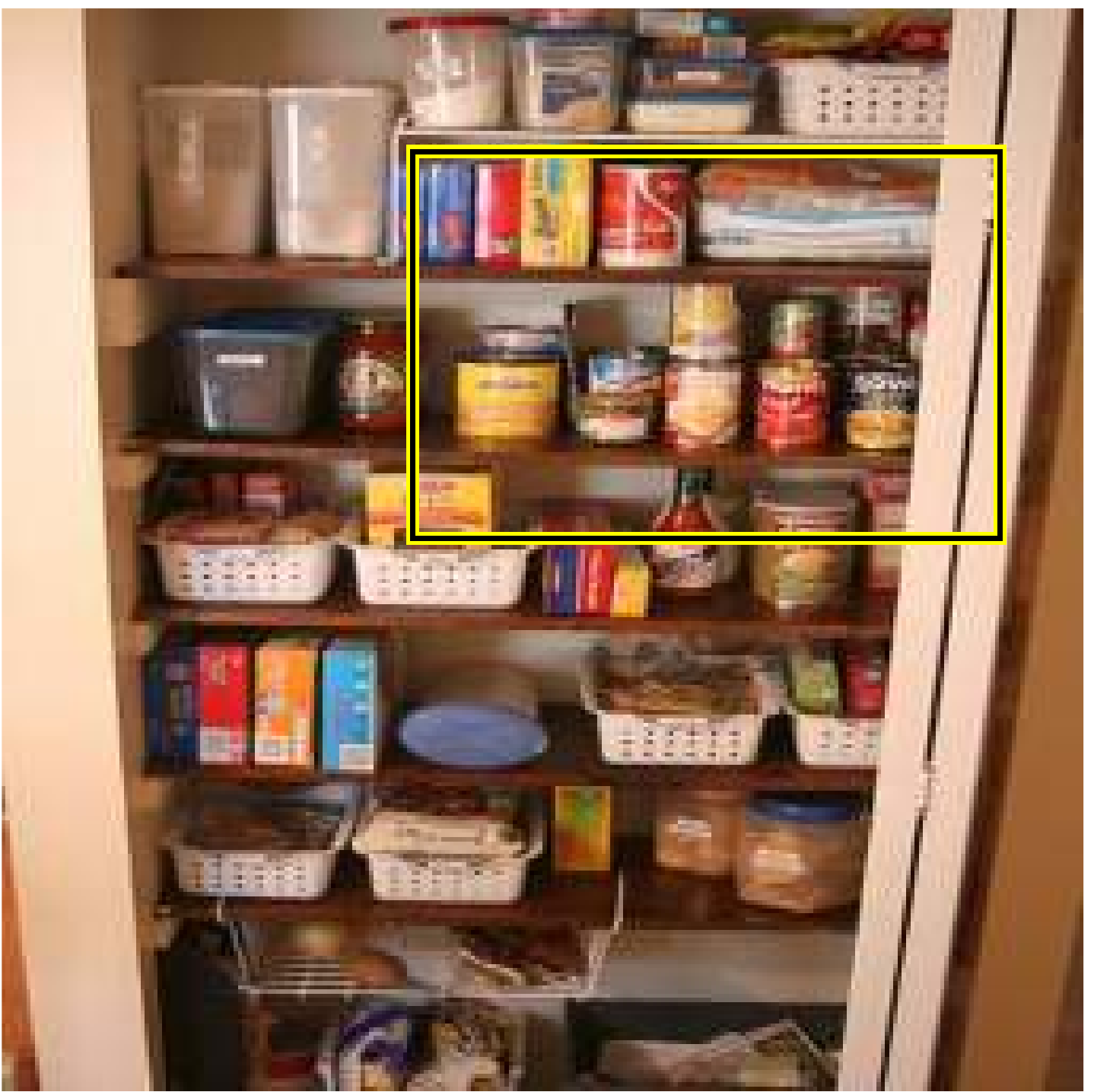} &
				\includegraphics[height=0.63in, width=0.85in]{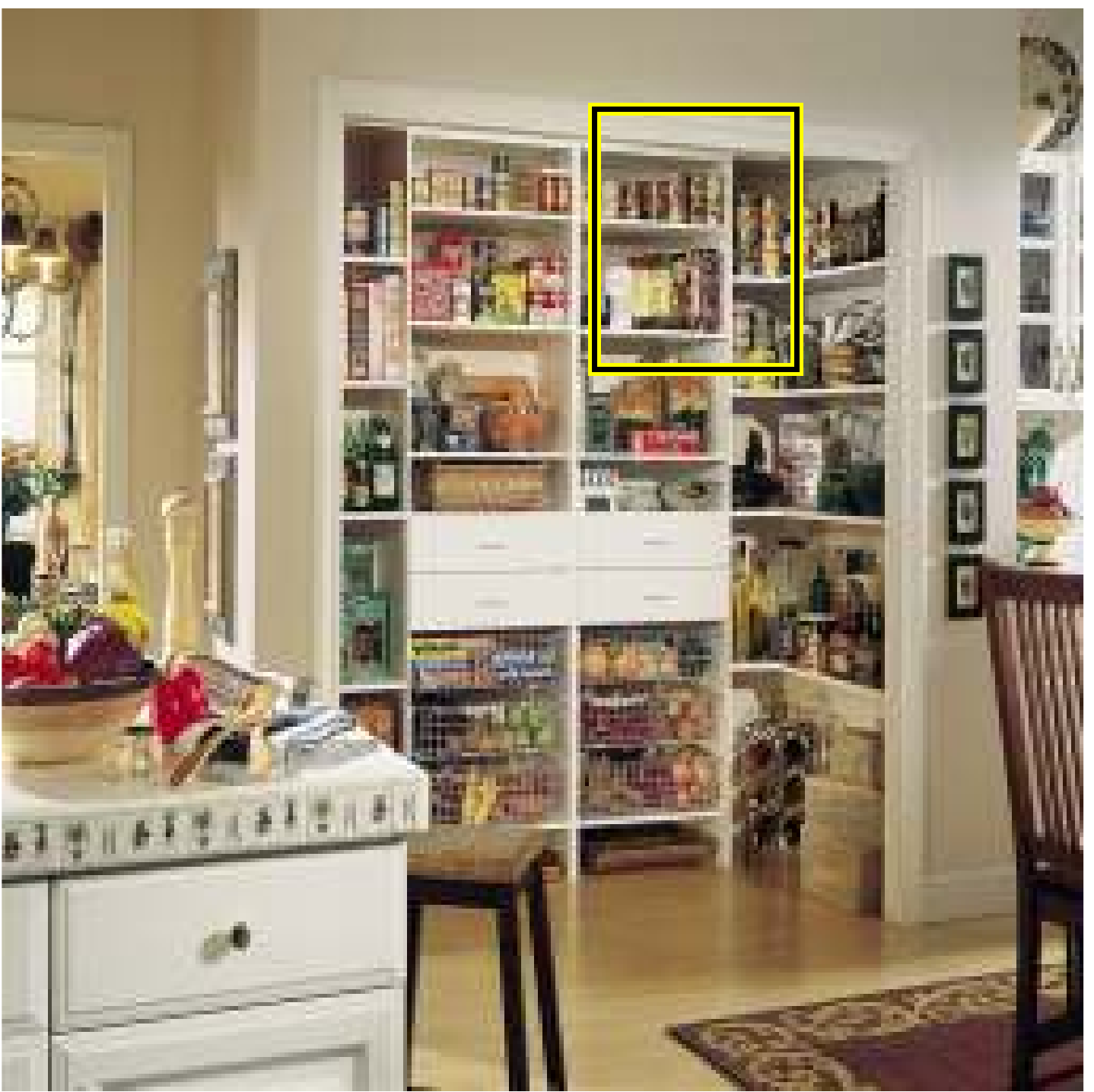} &
				\includegraphics[height=0.63in, width=0.85in]{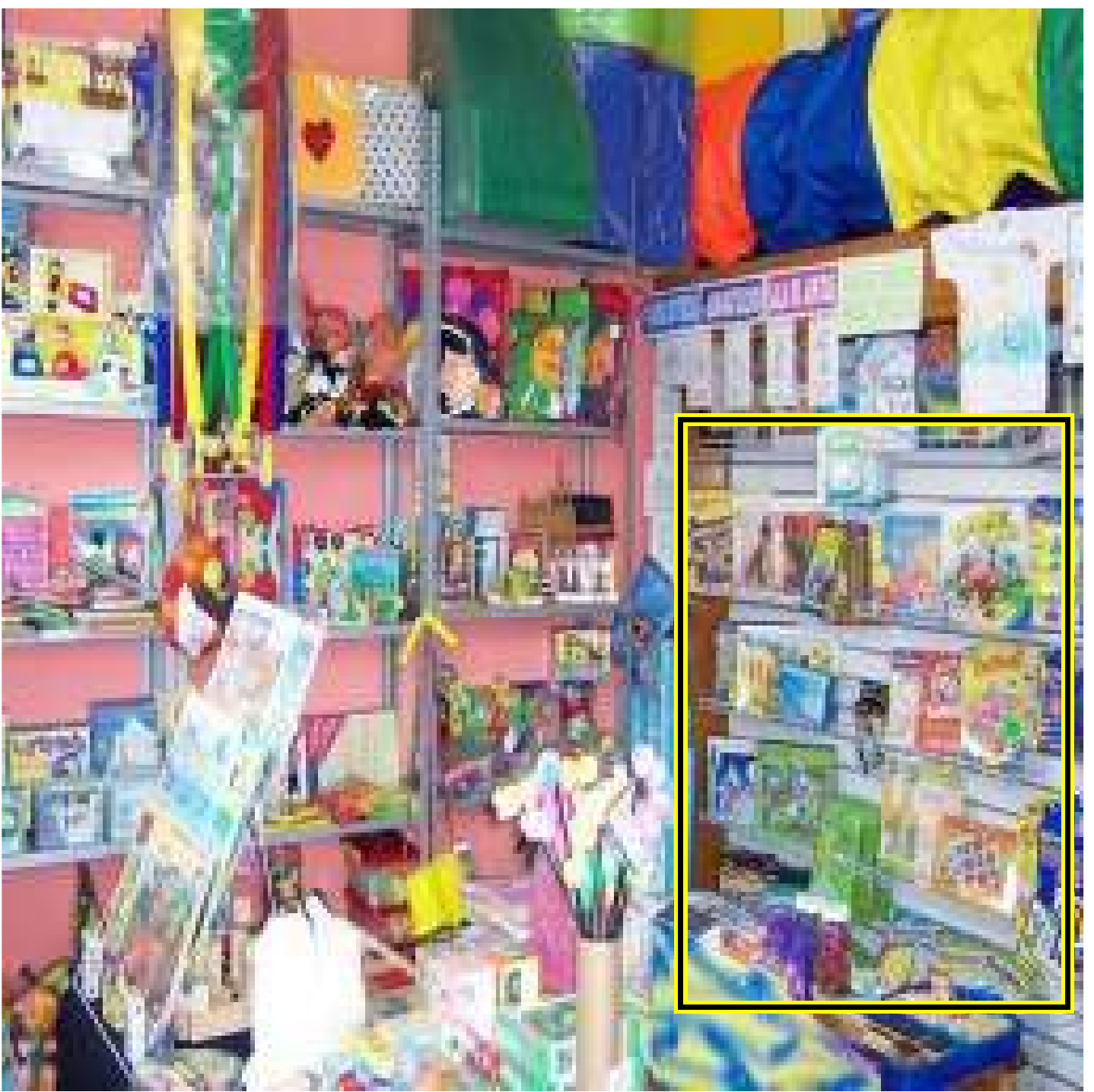} \\ [-0.05cm]
	\rotatebox{90}{\hspace{0.27cm} Part 6}$\;$ &
				\includegraphics[height=0.63in, width=0.85in]{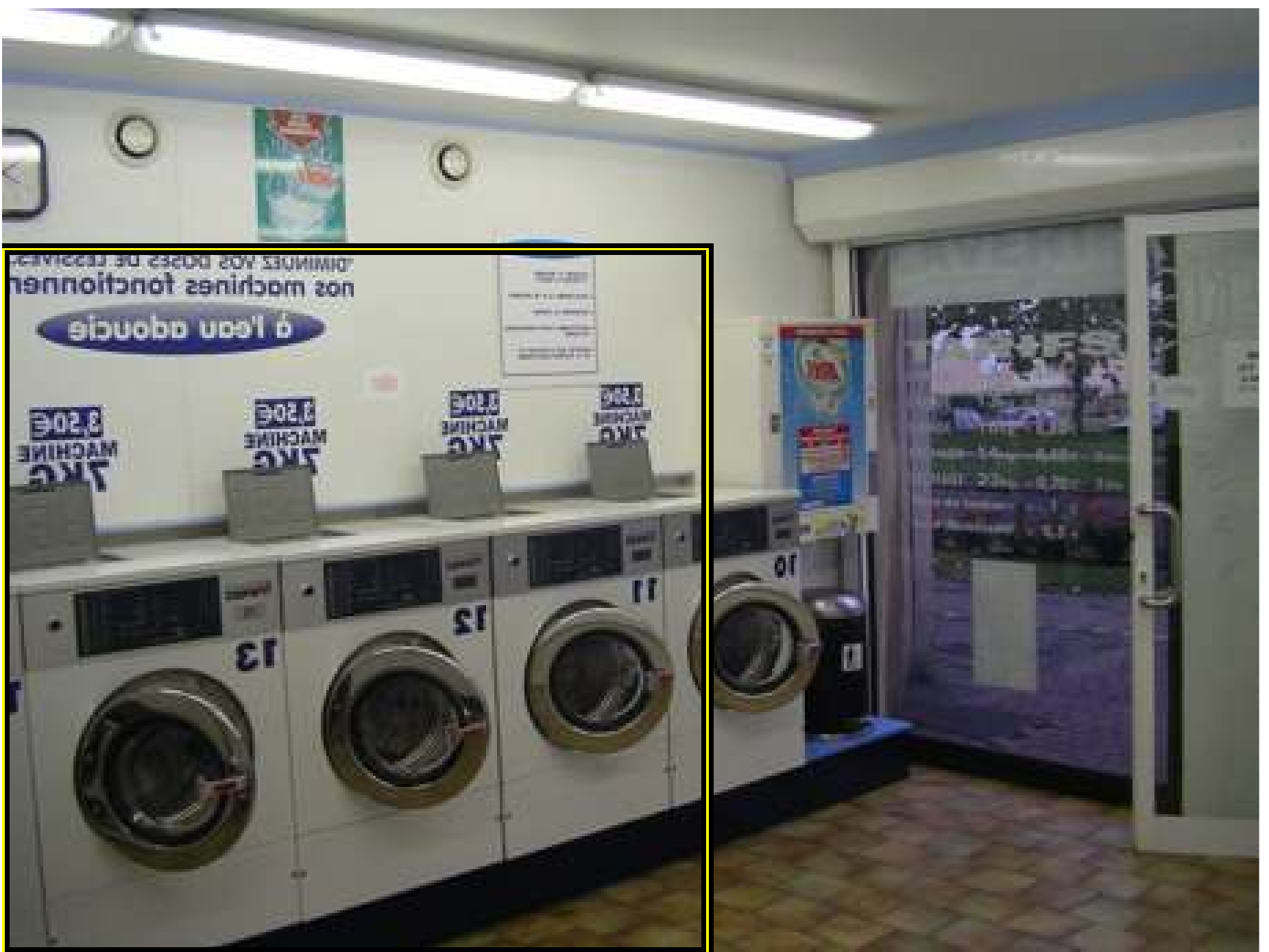} &
				\includegraphics[height=0.63in, width=0.85in]{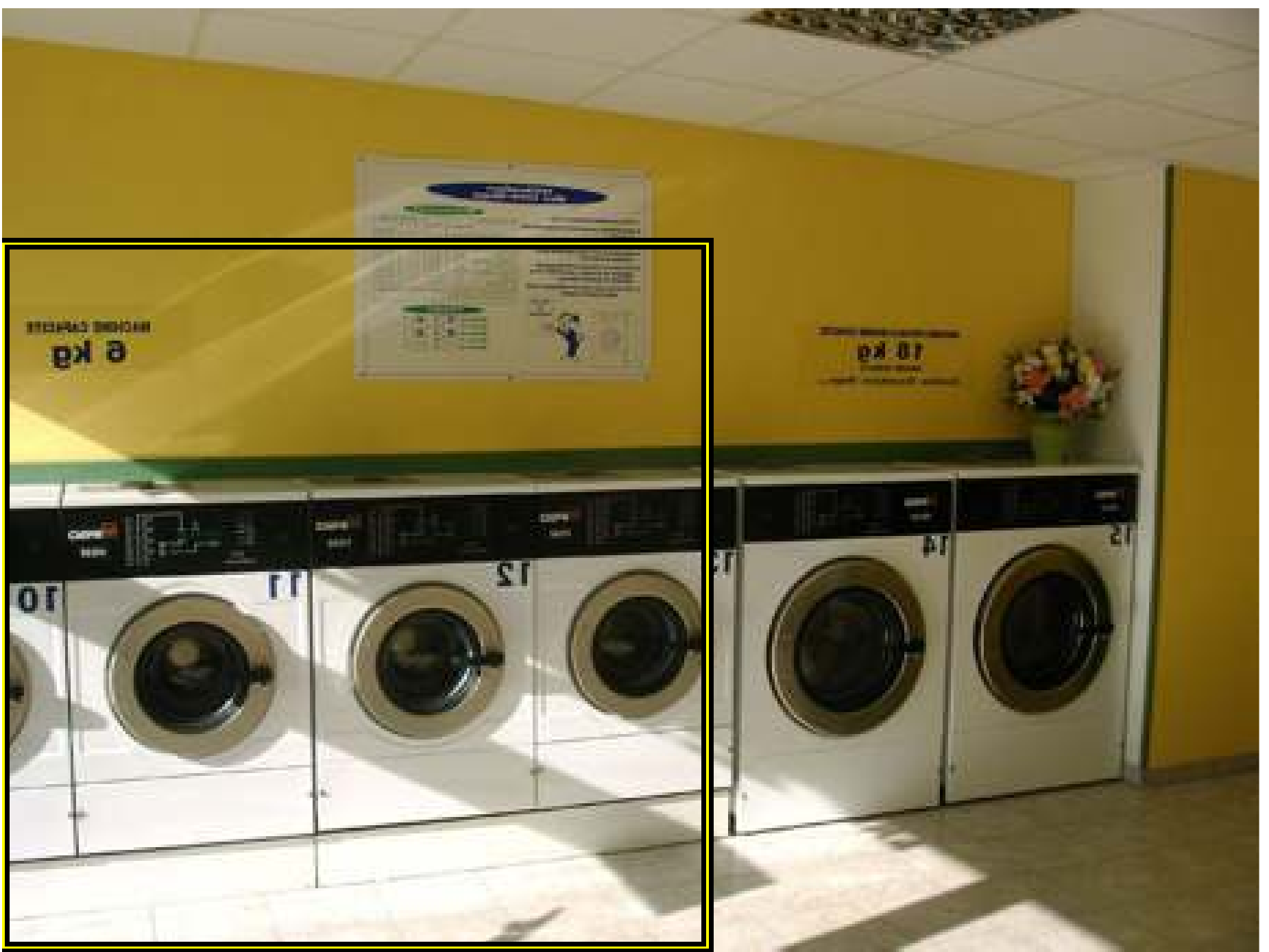} &
				\includegraphics[height=0.63in, width=0.85in]{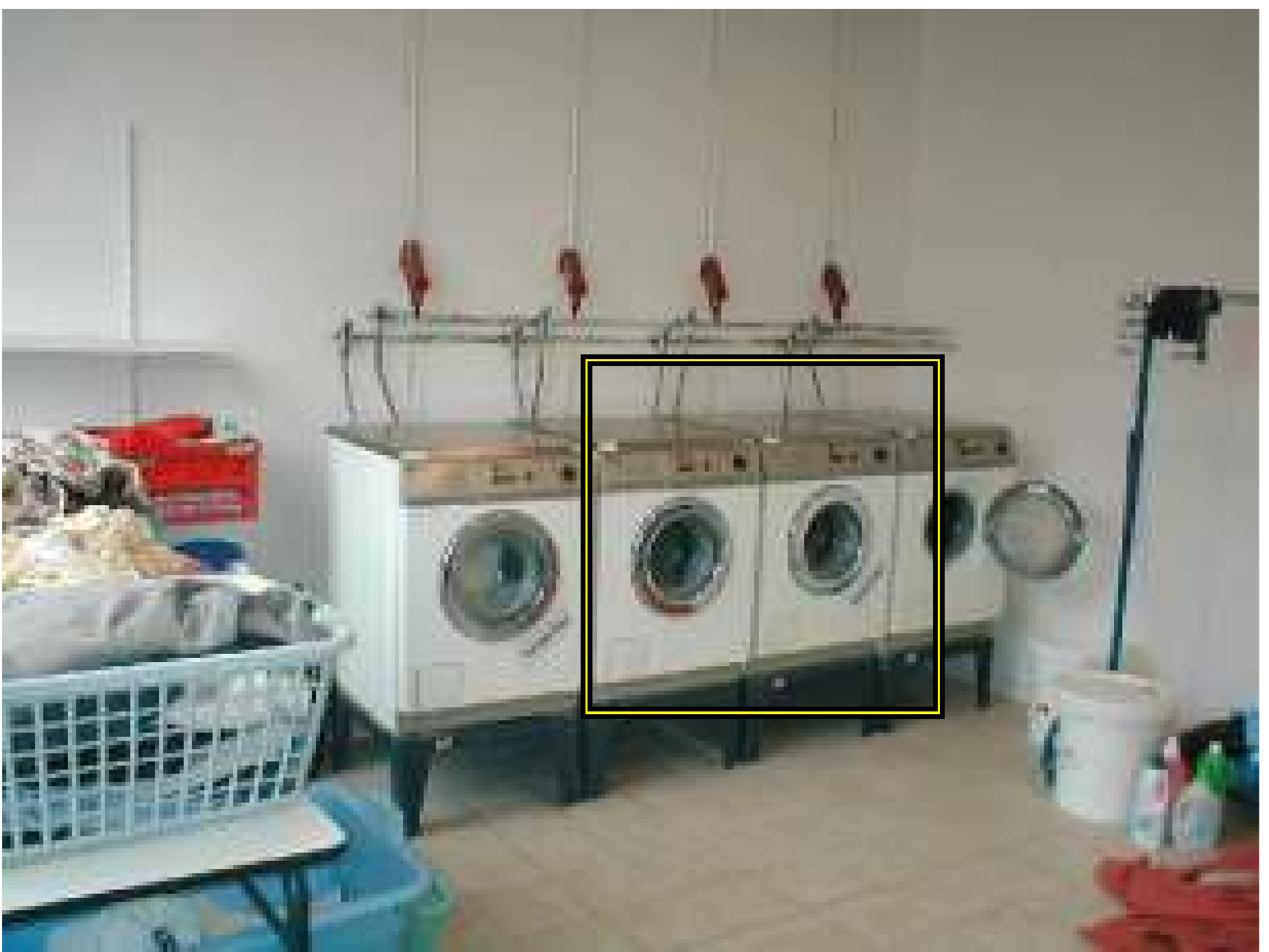} &
				\includegraphics[height=0.63in, width=0.85in]{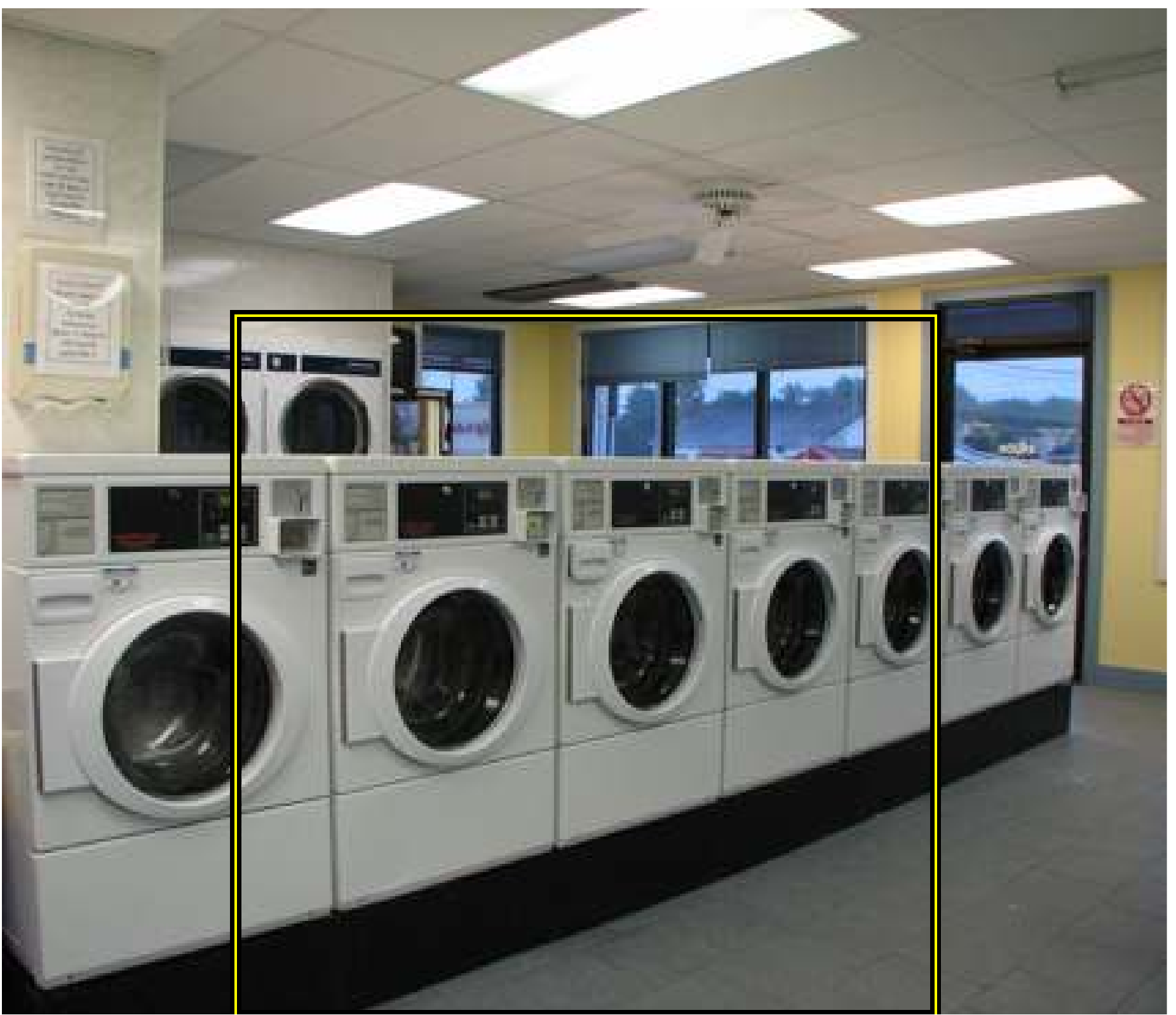} &
				\includegraphics[height=0.63in, width=0.85in]{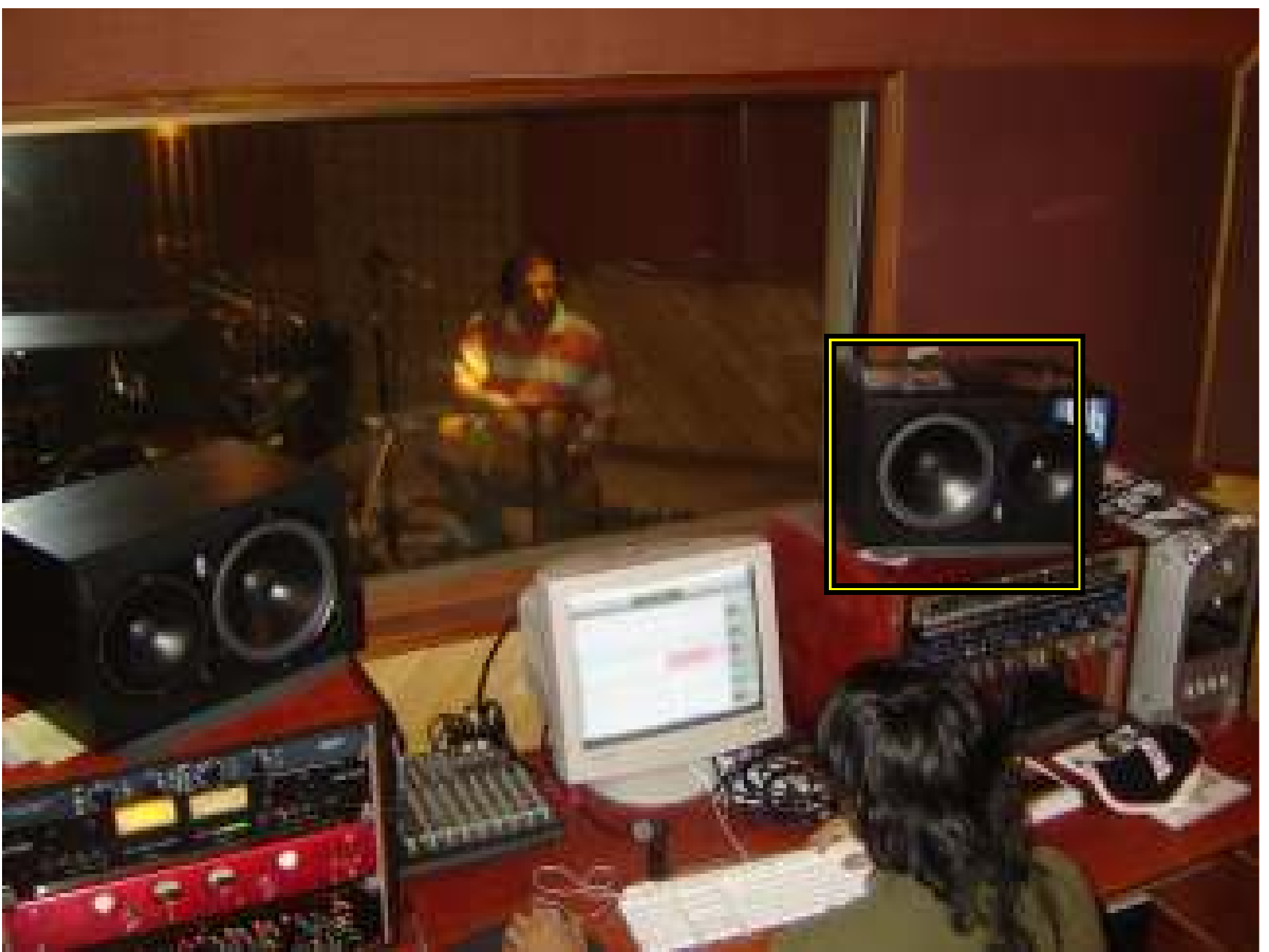} &
				\includegraphics[height=0.63in, width=0.85in]{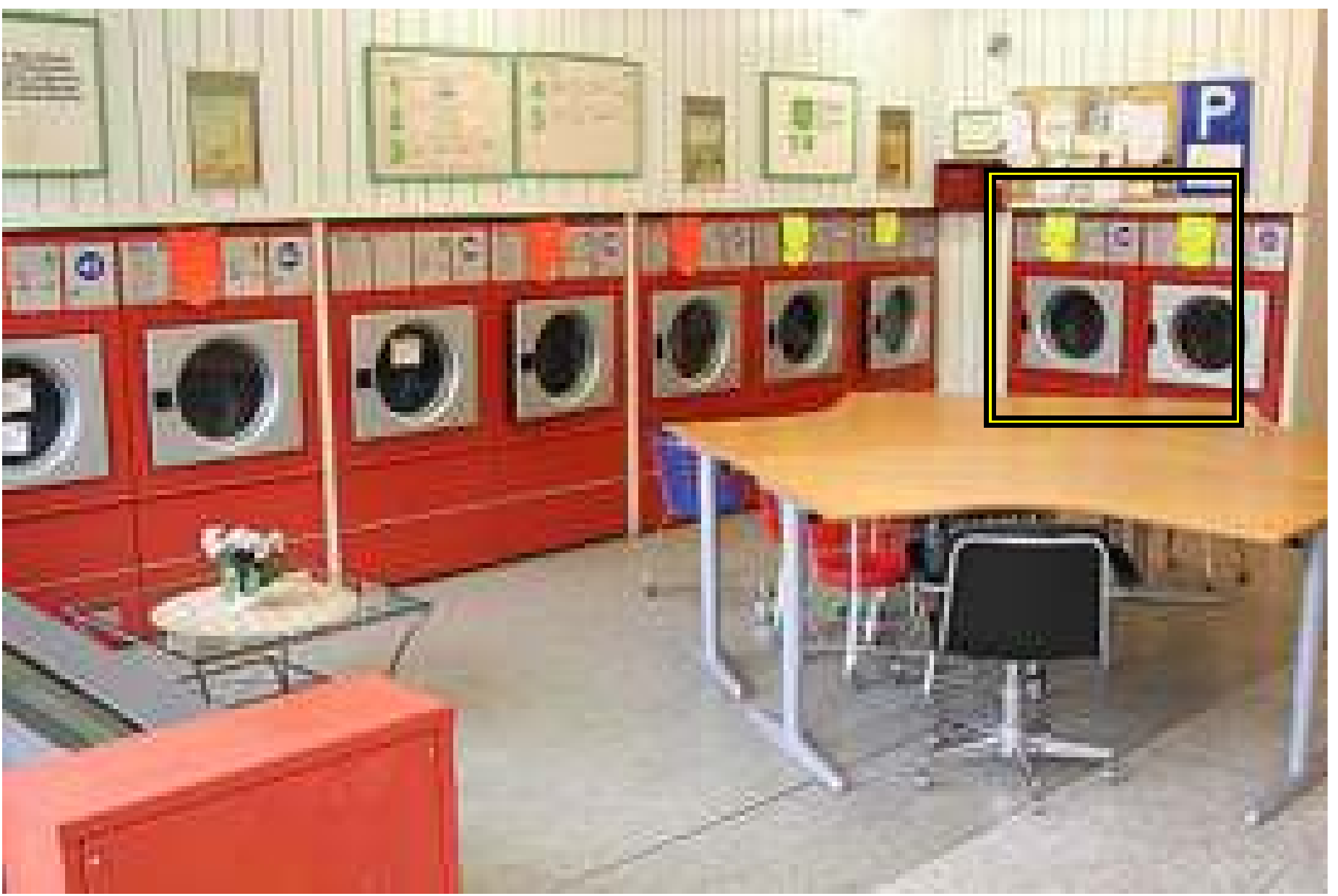} \\ [-0.05cm]
	\rotatebox{90}{\hspace{0.27cm} Part 16}$\;$ &
				\includegraphics[height=0.63in, width=0.85in]{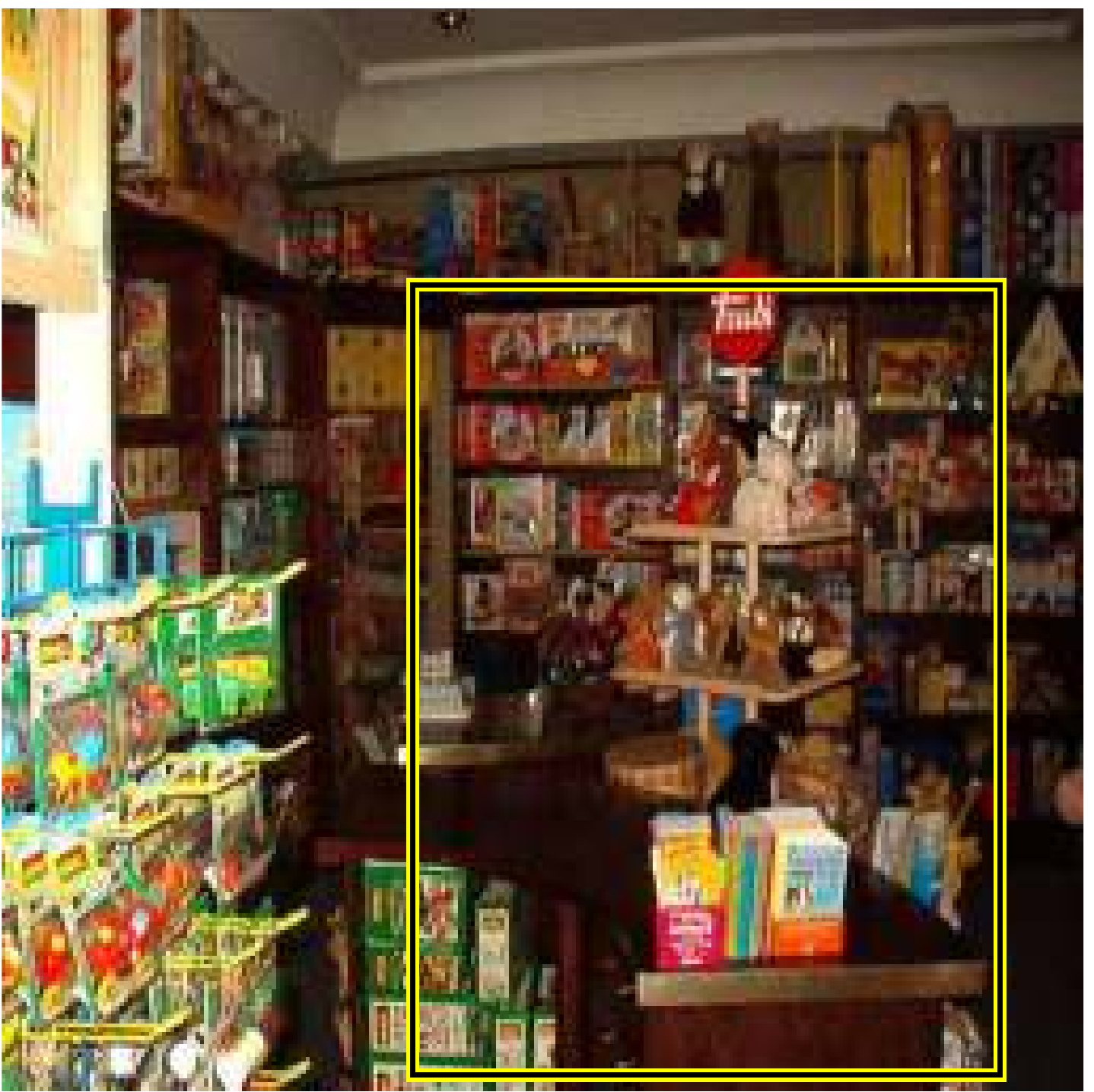} &
				\includegraphics[height=0.63in, width=0.85in]{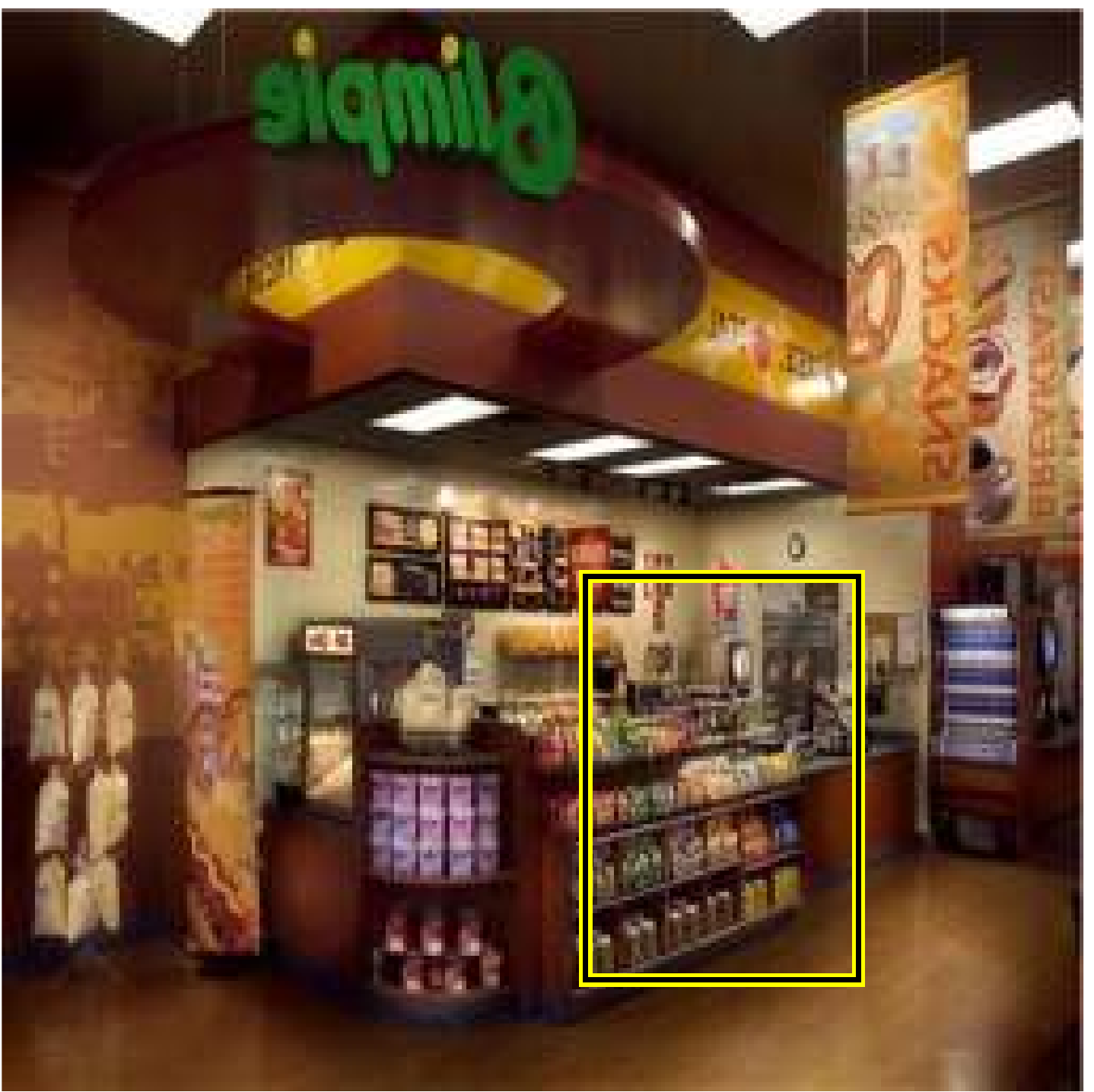} &
				\includegraphics[height=0.63in, width=0.85in]{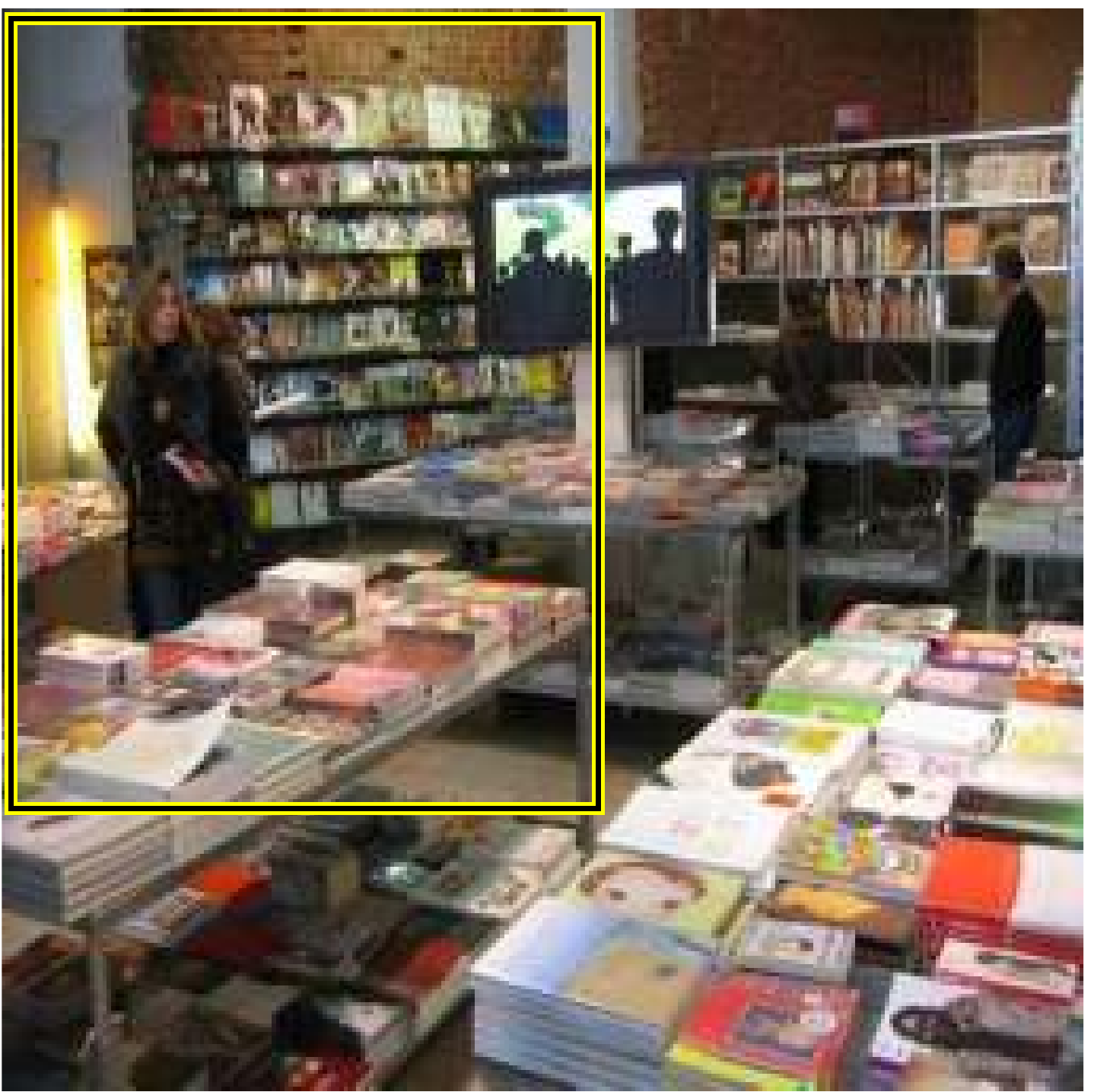} &
				\includegraphics[height=0.63in, width=0.85in]{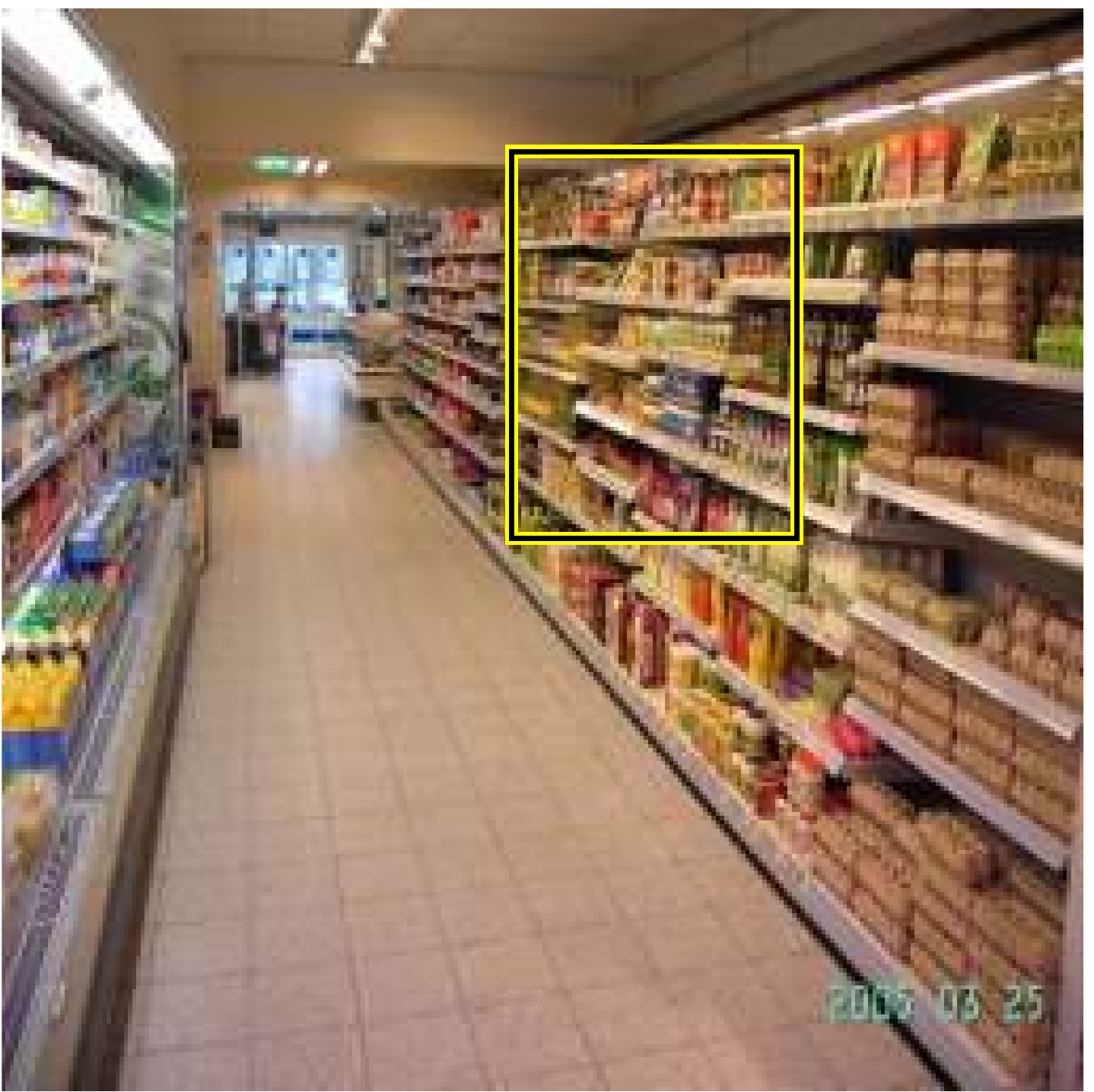} &
				\includegraphics[height=0.63in, width=0.85in]{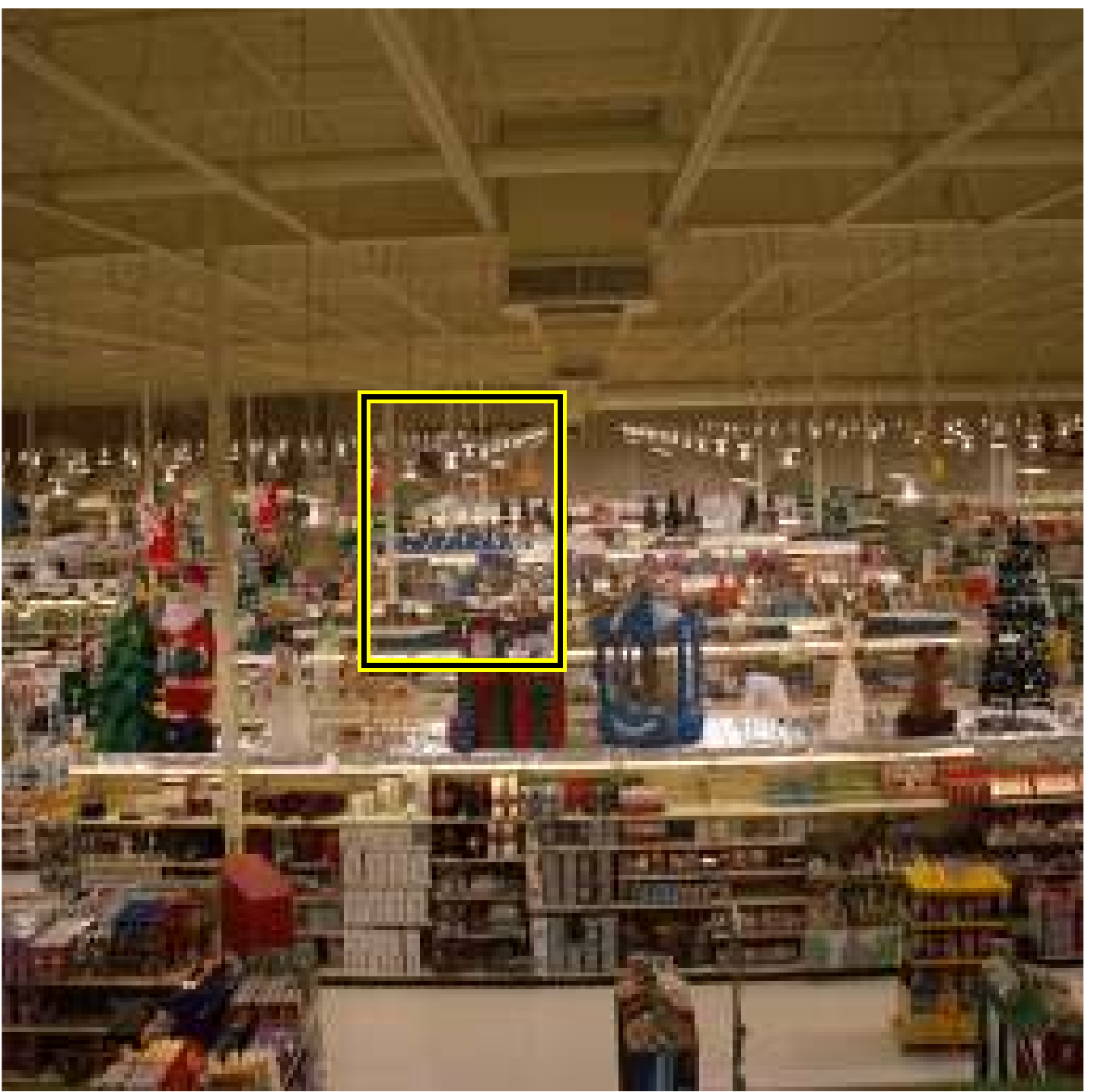} &
				\includegraphics[height=0.63in, width=0.85in]{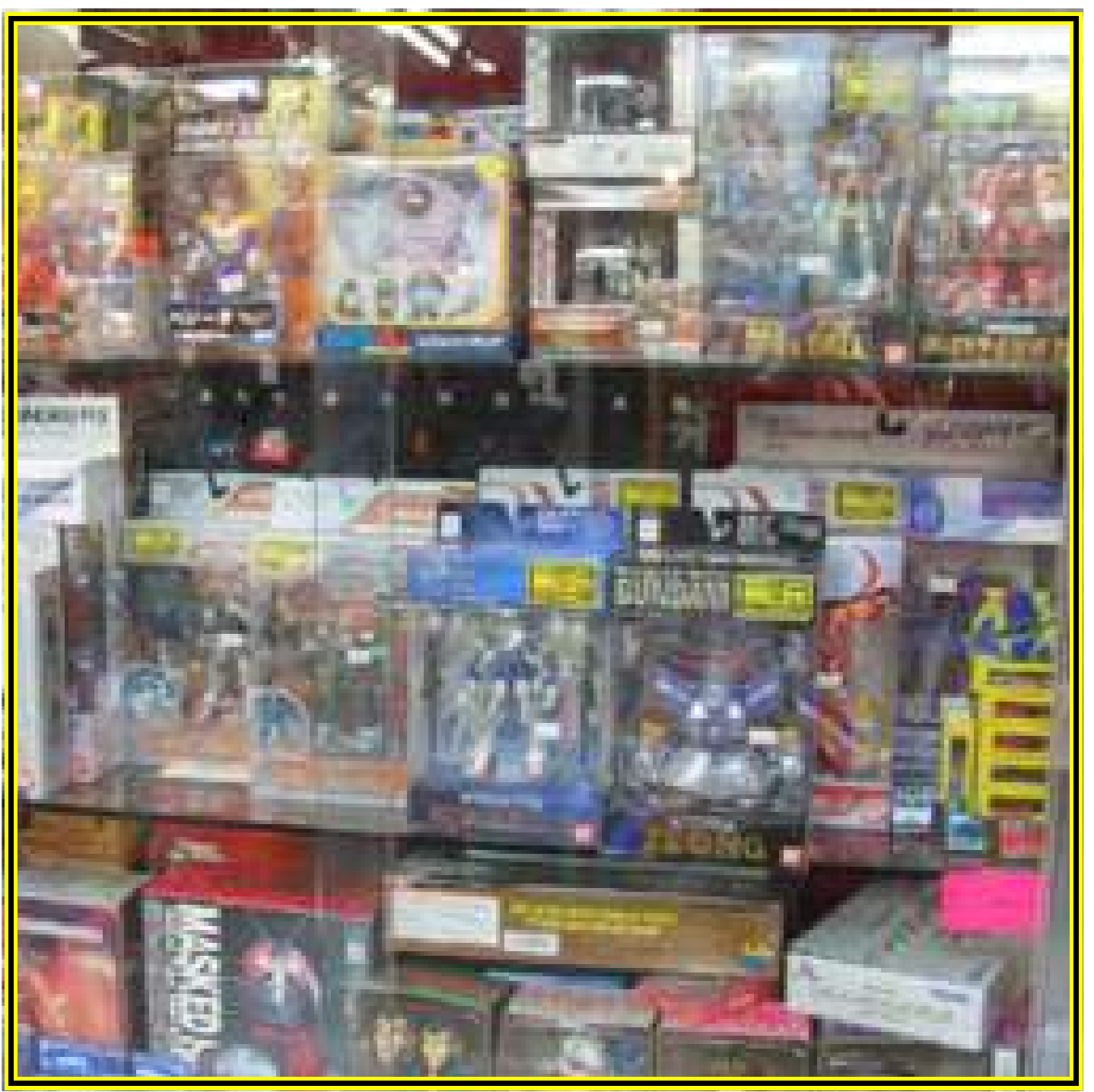} \\ [-0.05cm]
	\rotatebox{90}{\hspace{0.27cm} Part 17}$\;$ &
				\includegraphics[height=0.63in, width=0.85in]{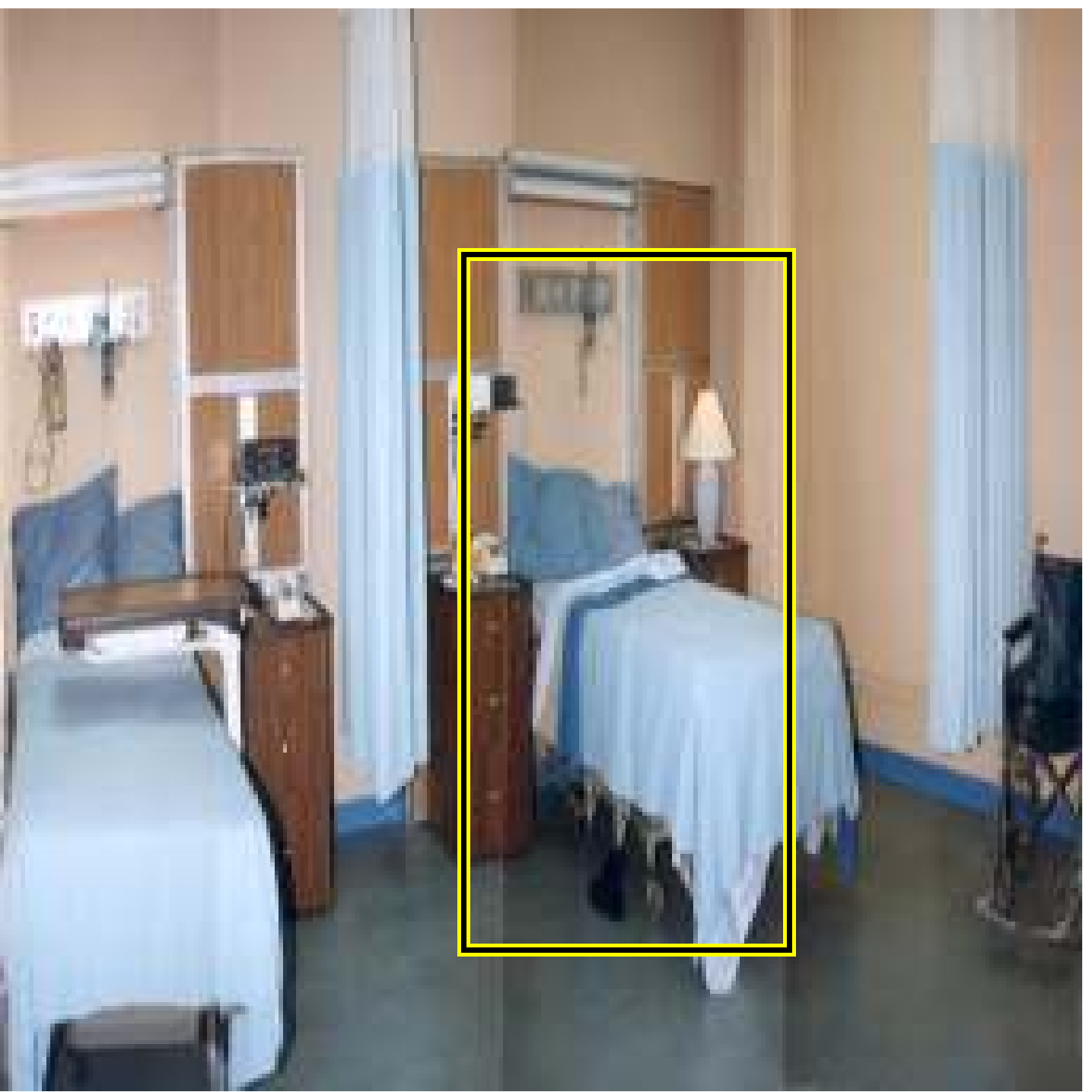} &
				\includegraphics[height=0.63in, width=0.85in]{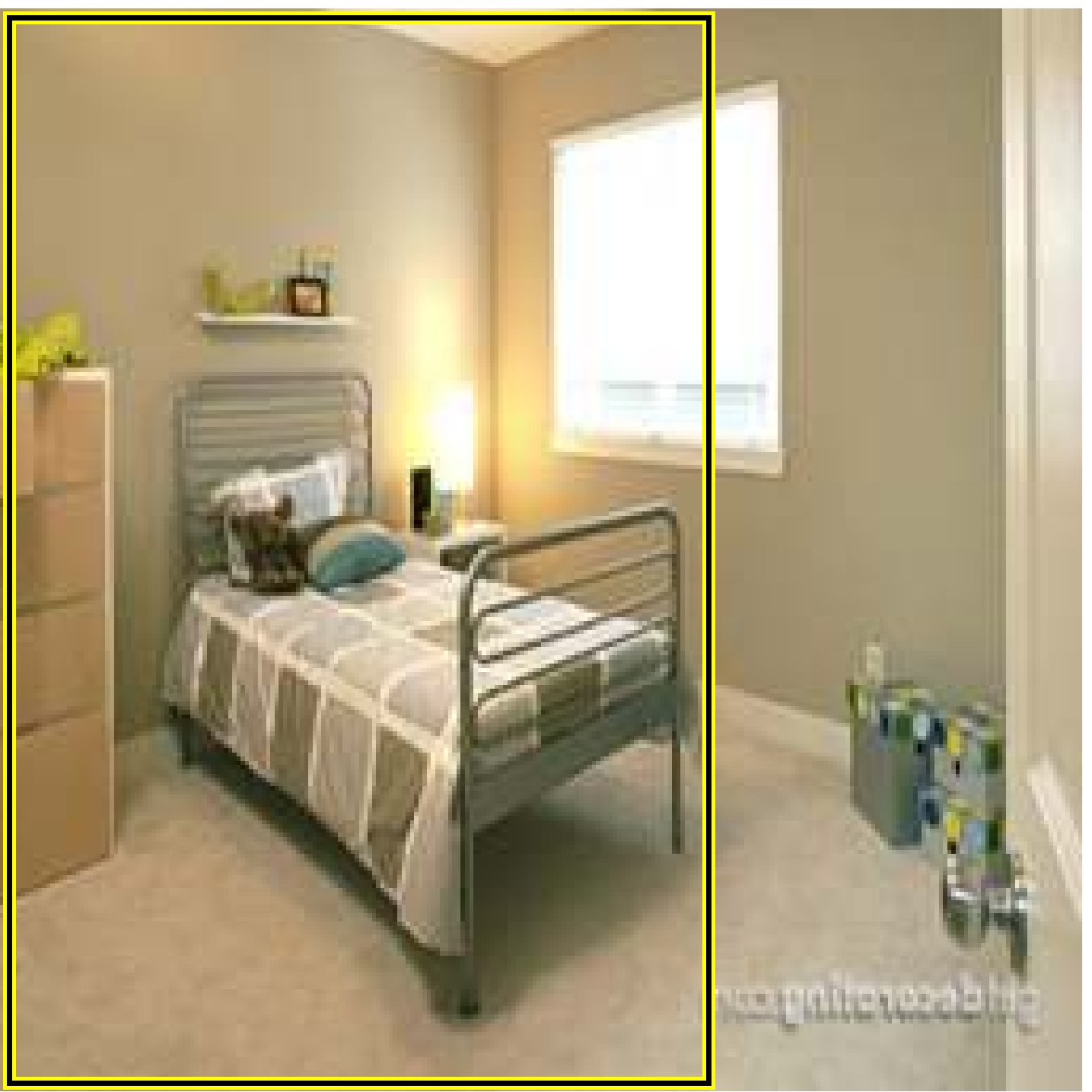} &
				\includegraphics[height=0.63in, width=0.85in]{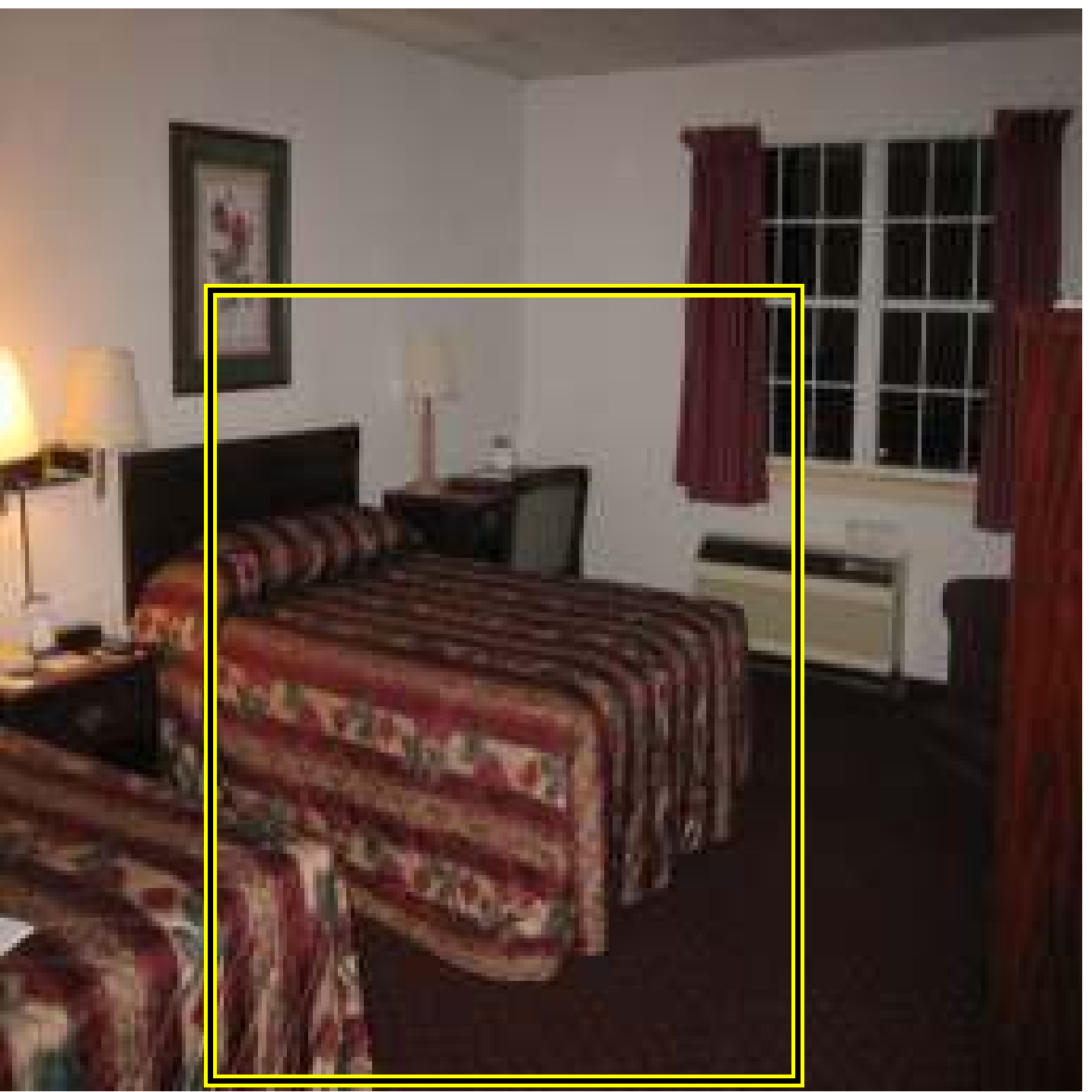} &
				\includegraphics[height=0.63in, width=0.85in]{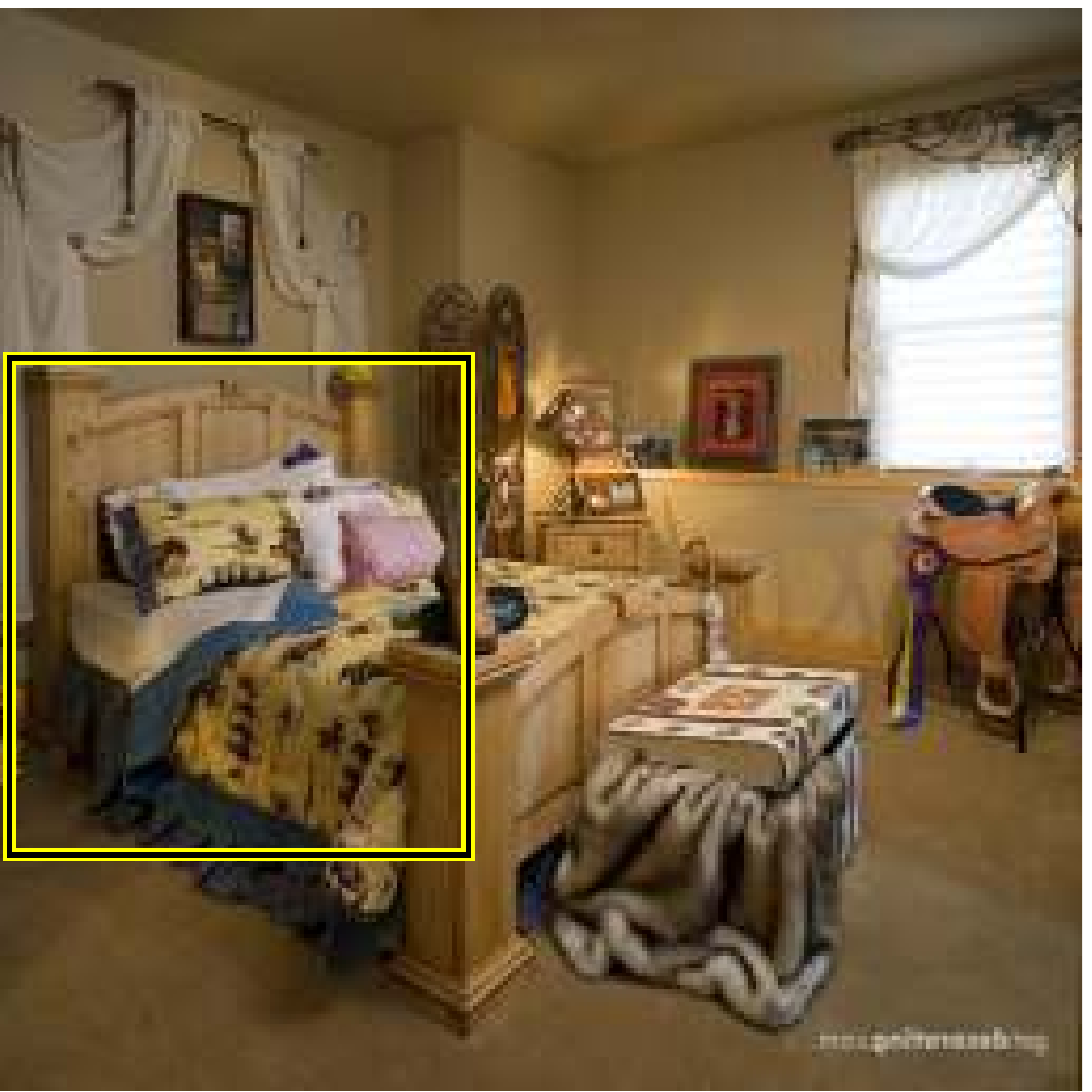} &
				\includegraphics[height=0.63in, width=0.85in]{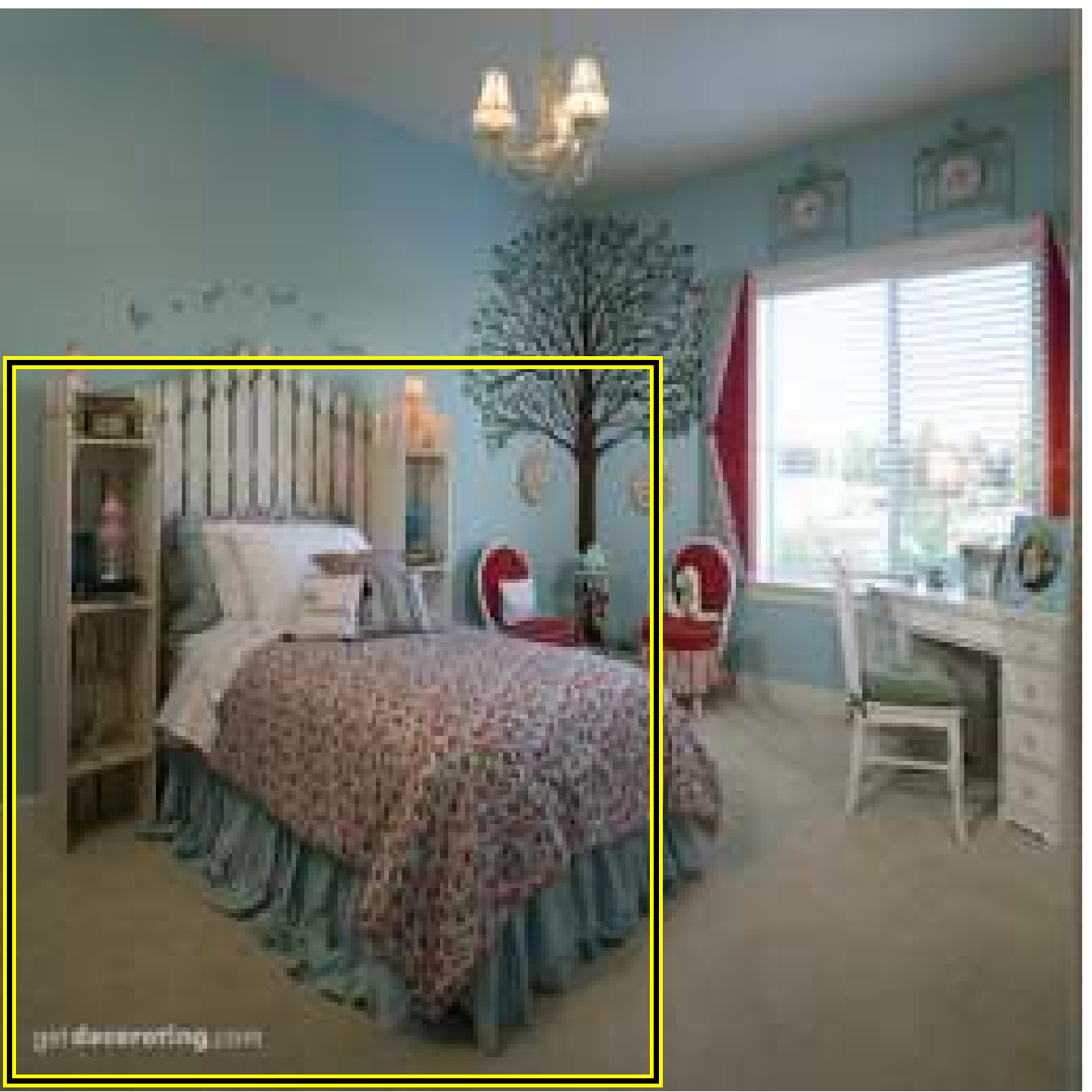} &
				\includegraphics[height=0.63in, width=0.85in]{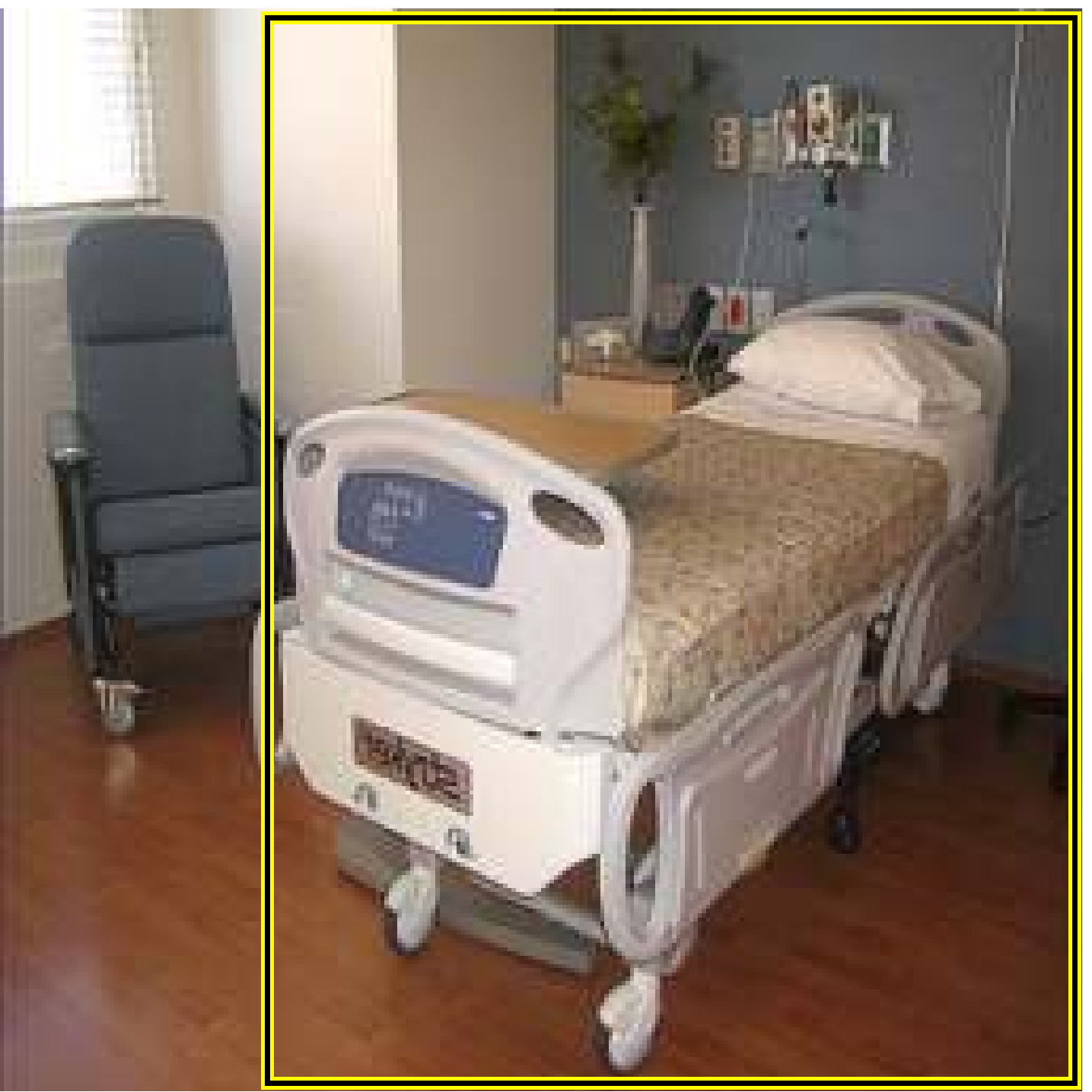} \\ [-0.05cm]
	\rotatebox{90}{\hspace{0.27cm} Part 20}$\;$ &
				\includegraphics[height=0.63in, width=0.85in]{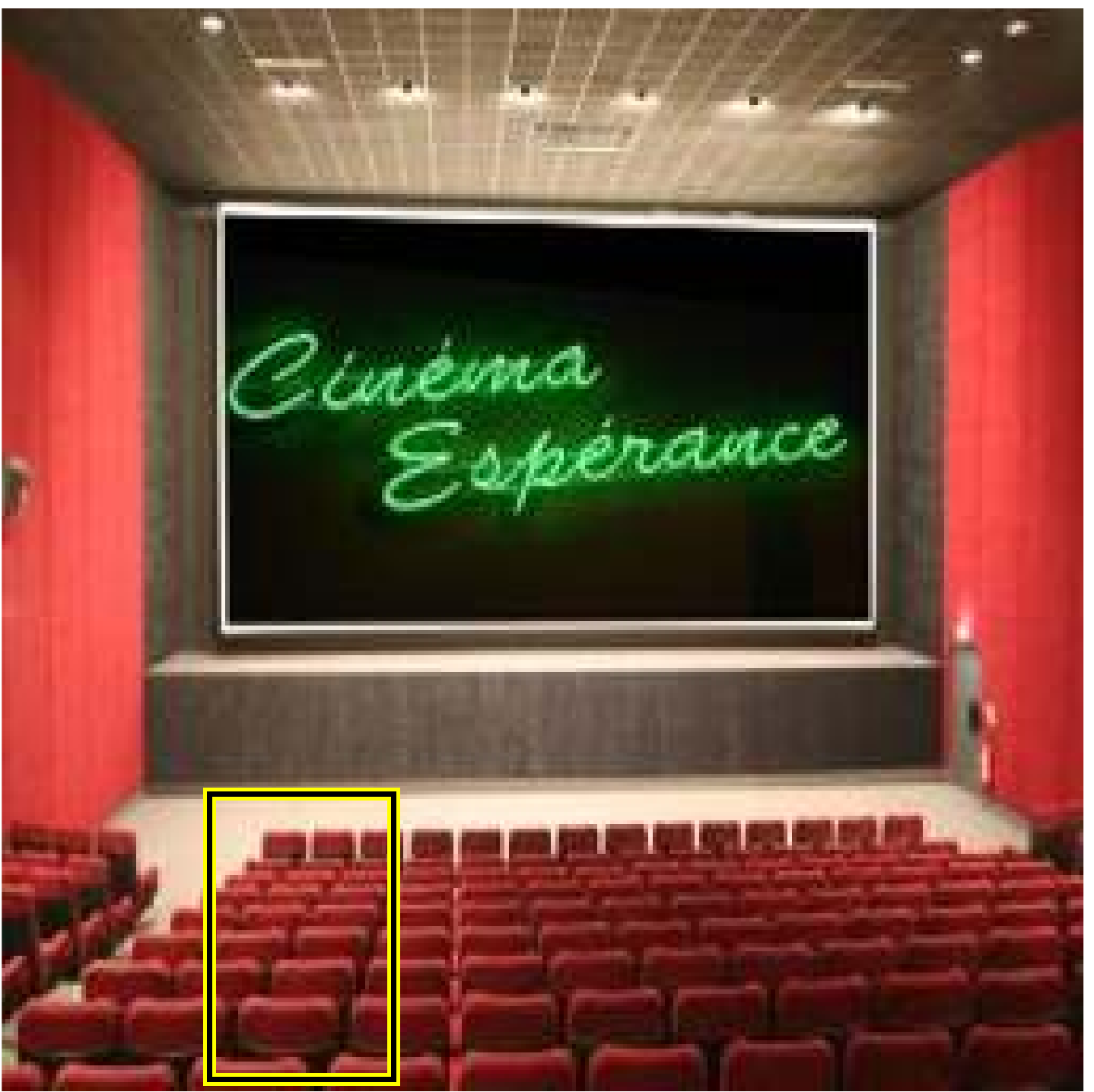} &
				\includegraphics[height=0.63in, width=0.85in]{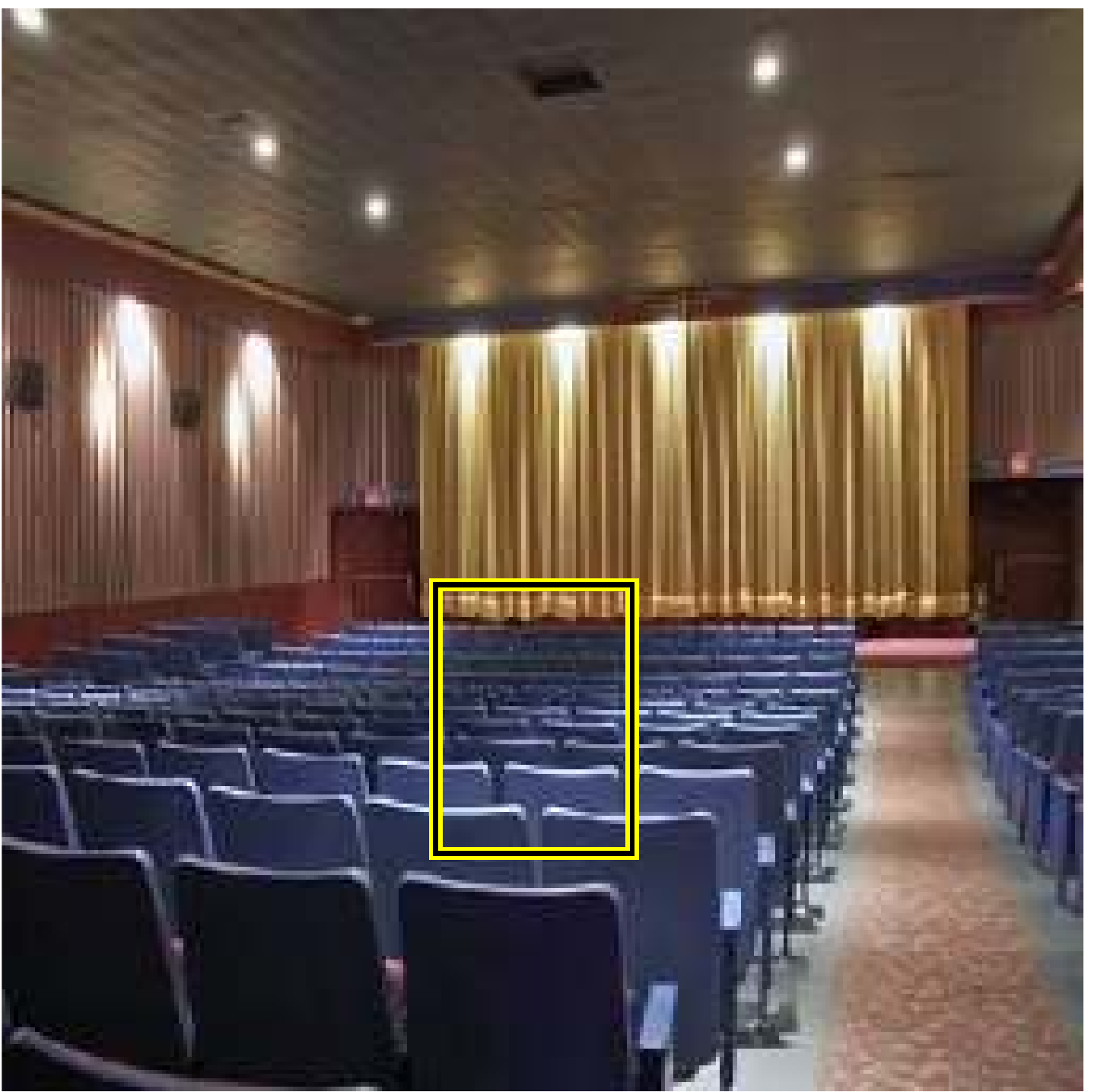} &
				\includegraphics[height=0.63in, width=0.85in]{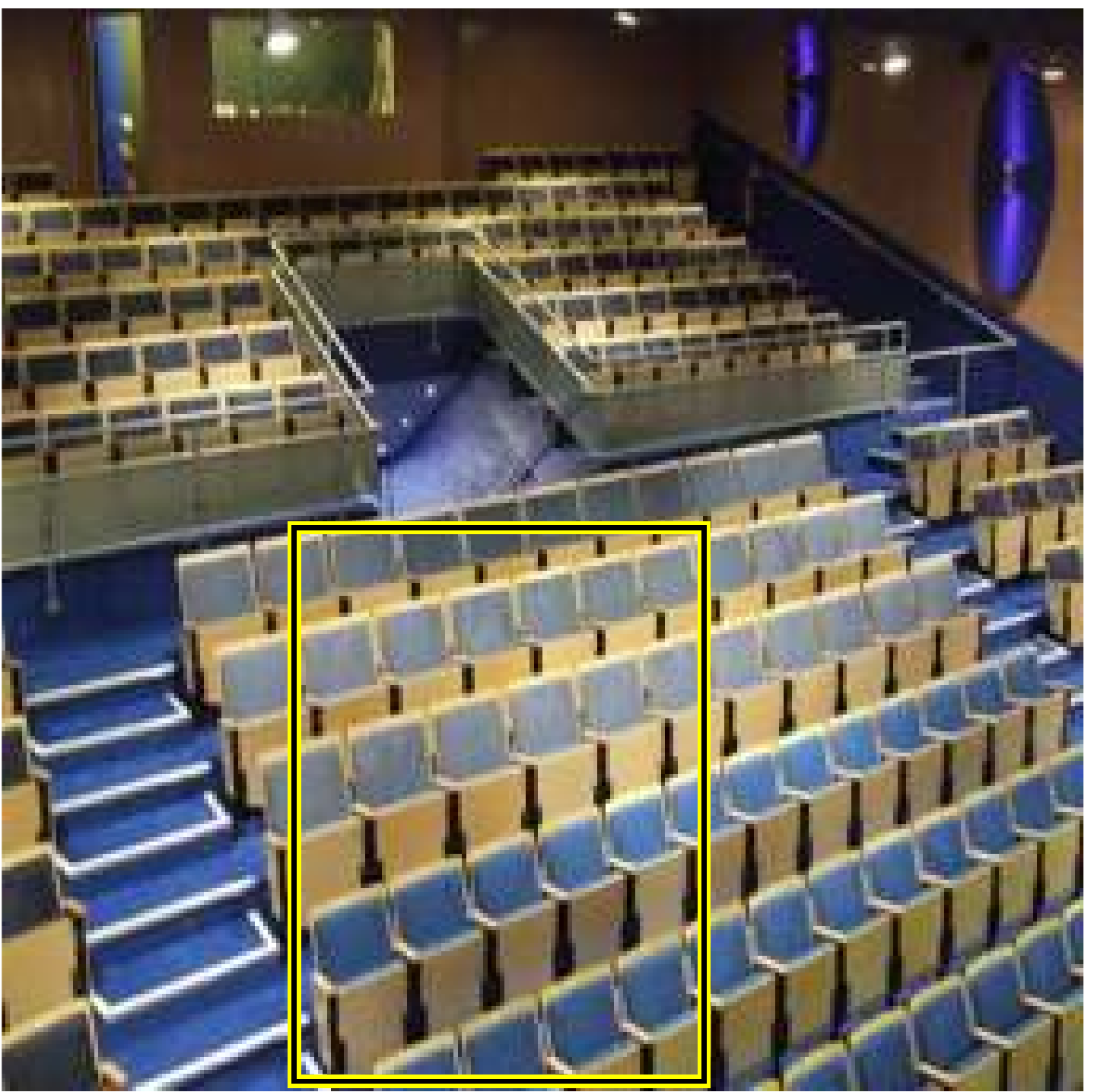} &
				\includegraphics[height=0.63in, width=0.85in]{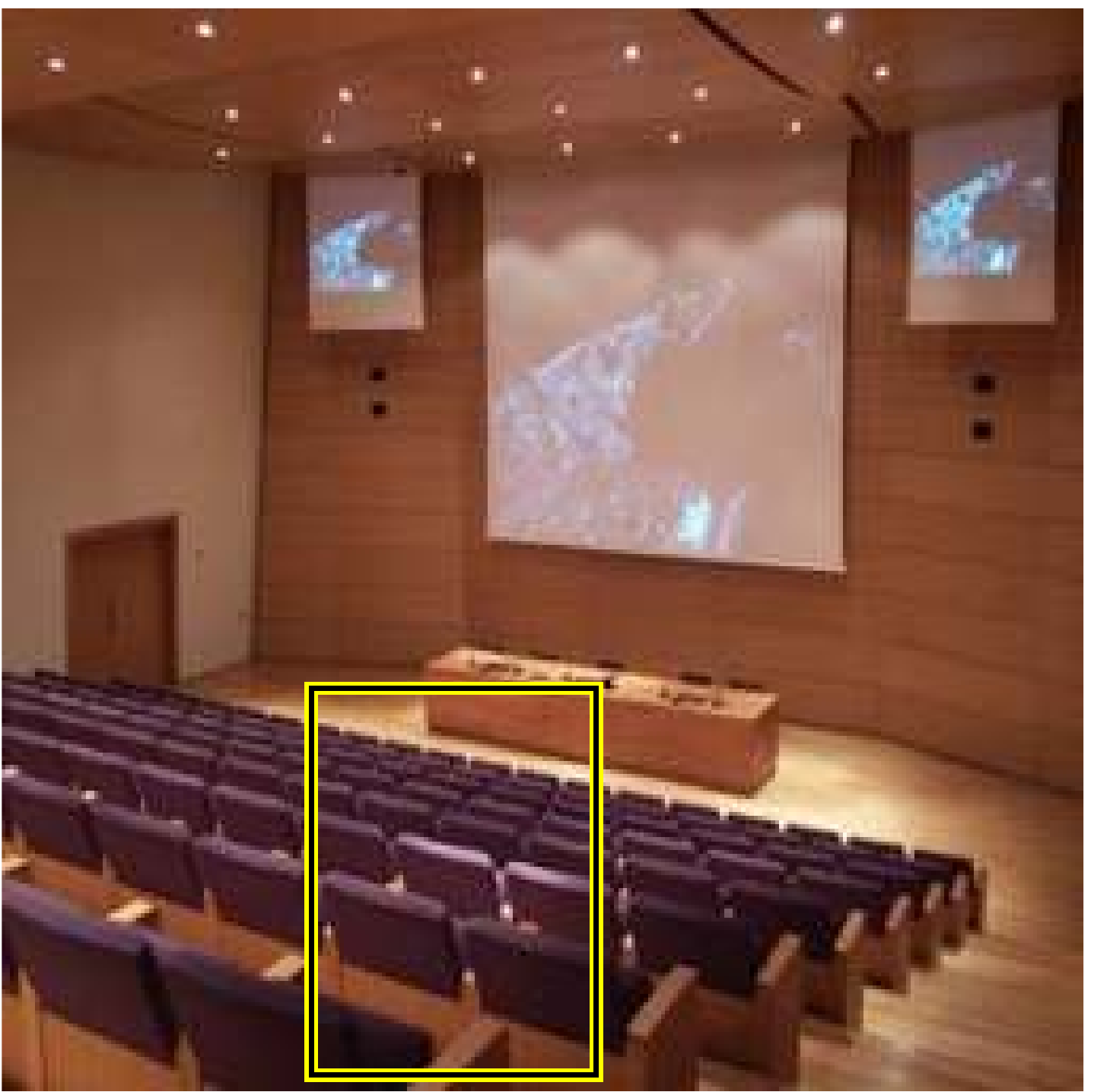} &
				\includegraphics[height=0.63in, width=0.85in]{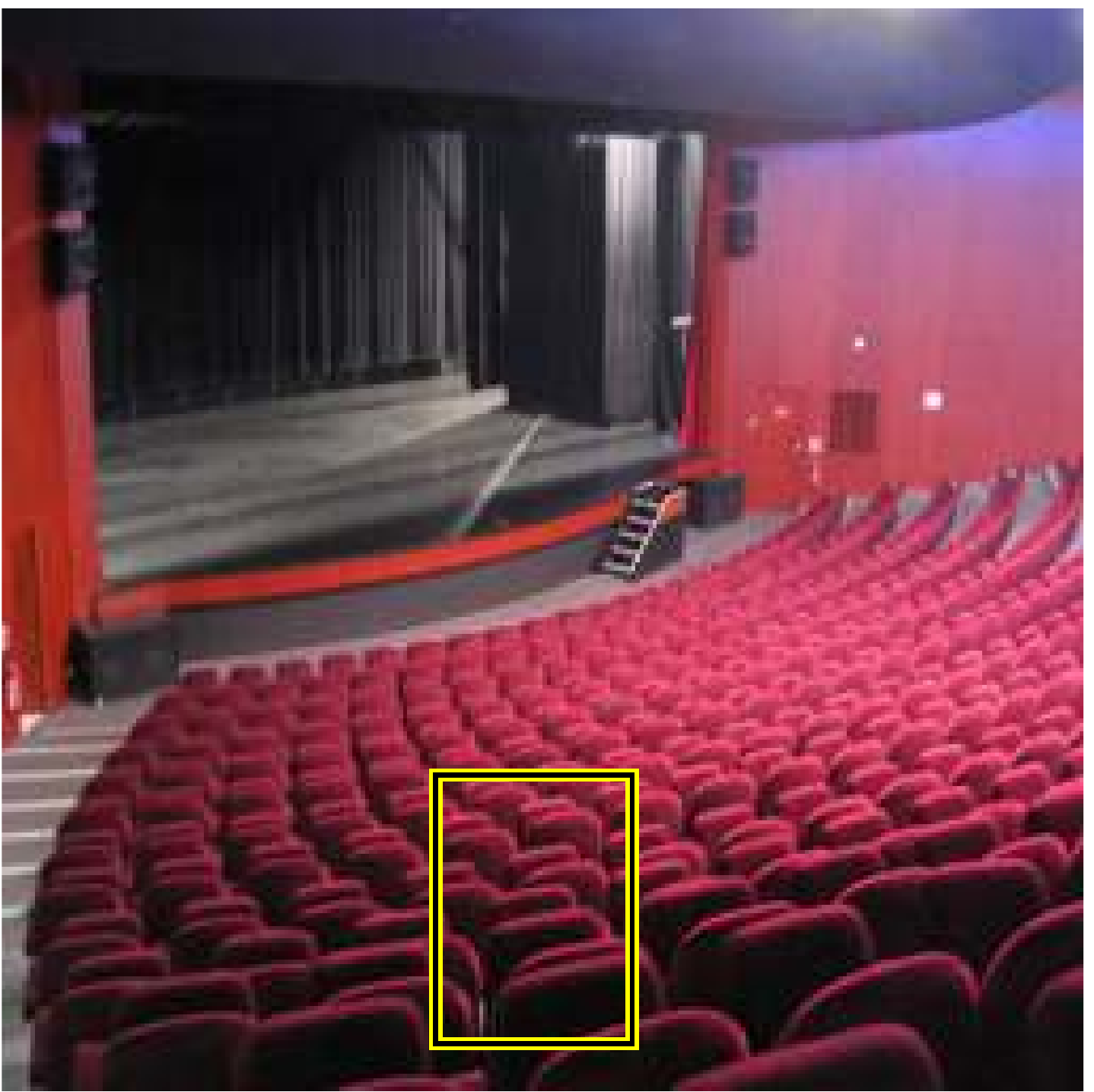} &
				\includegraphics[height=0.63in, width=0.85in]{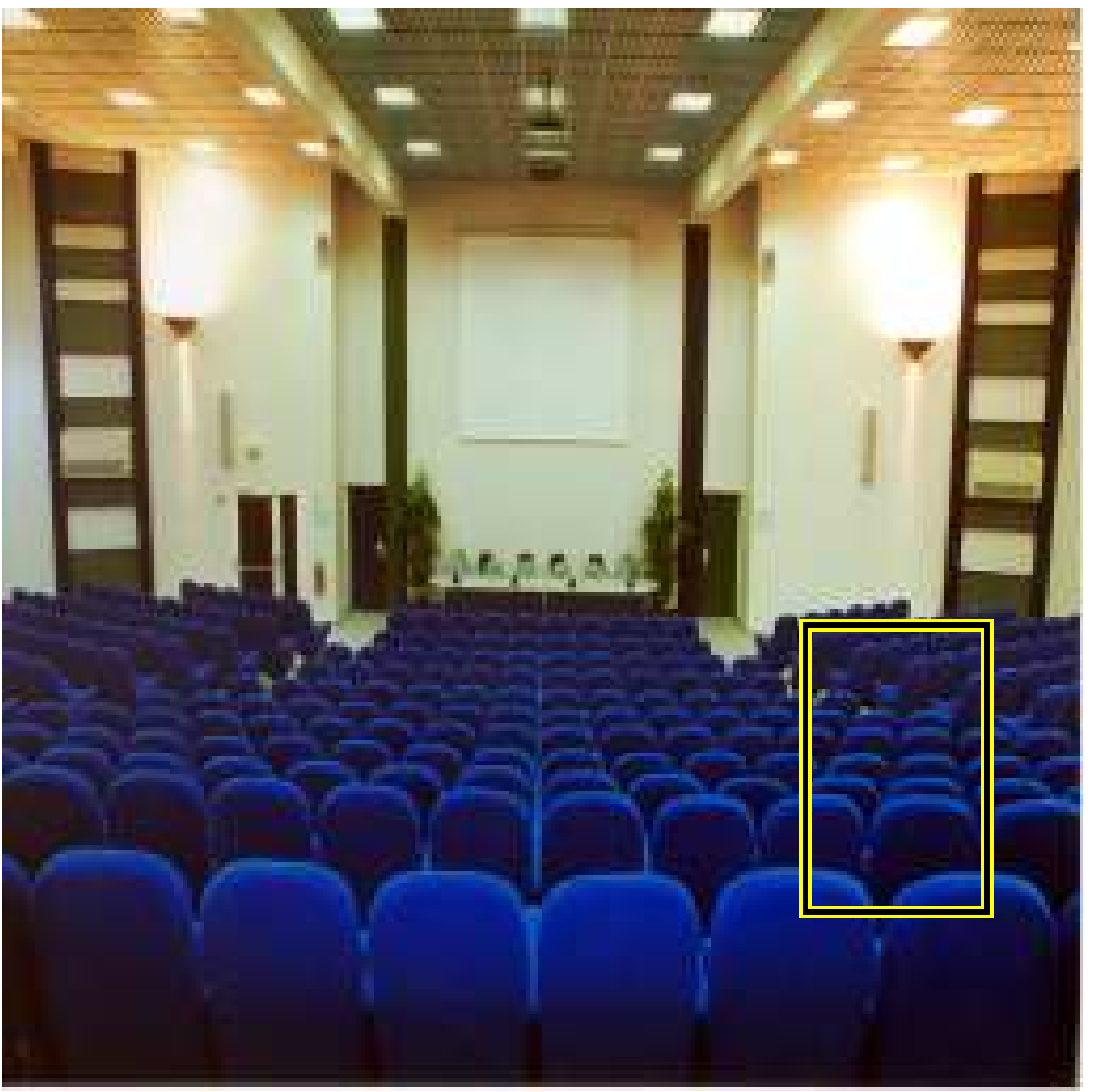}\\ [-0.05cm]
	\rotatebox{90}{\hspace{0.27cm}Part 25}$\;$ &
				\includegraphics[height=0.63in, width=0.85in]{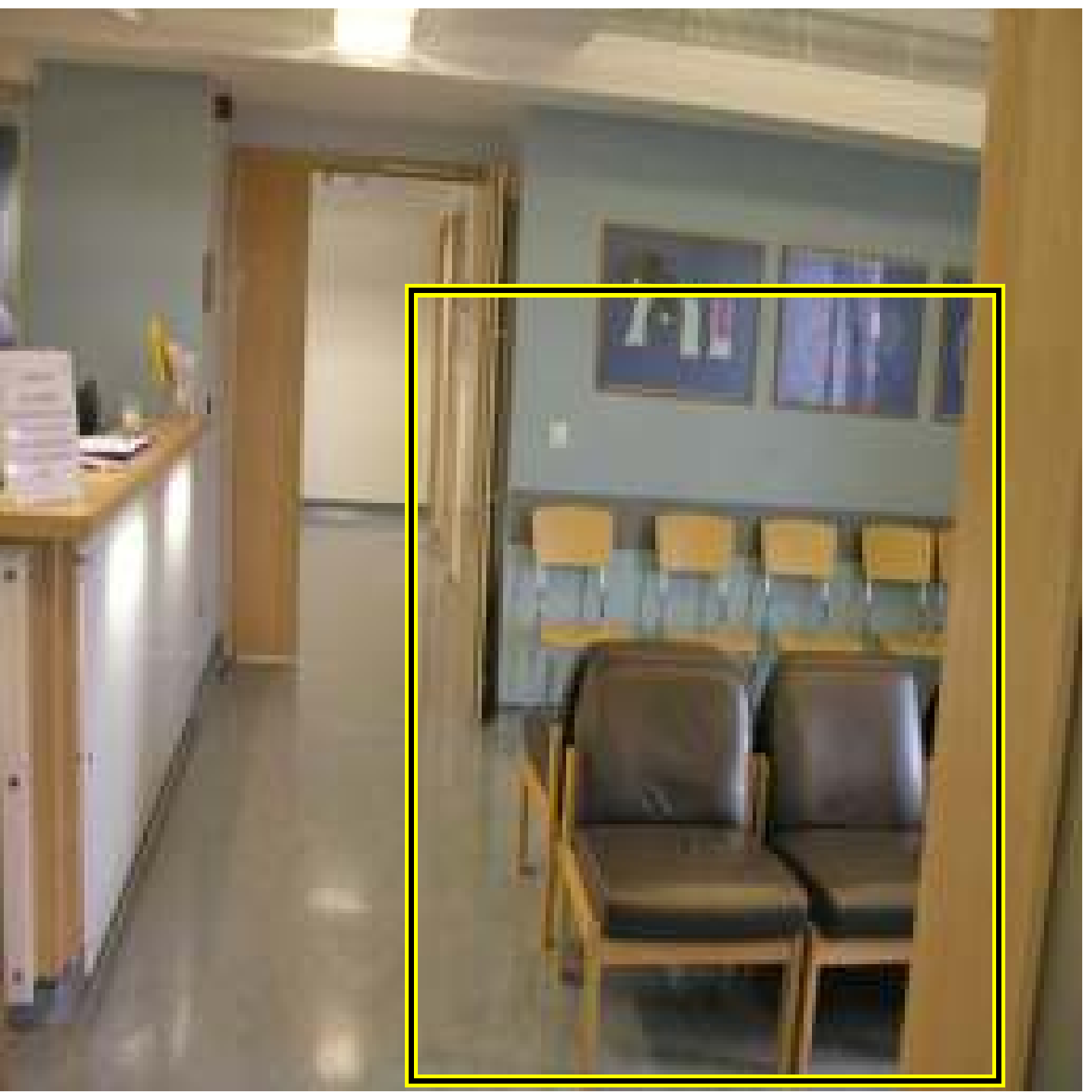} &
				\includegraphics[height=0.63in, width=0.85in]{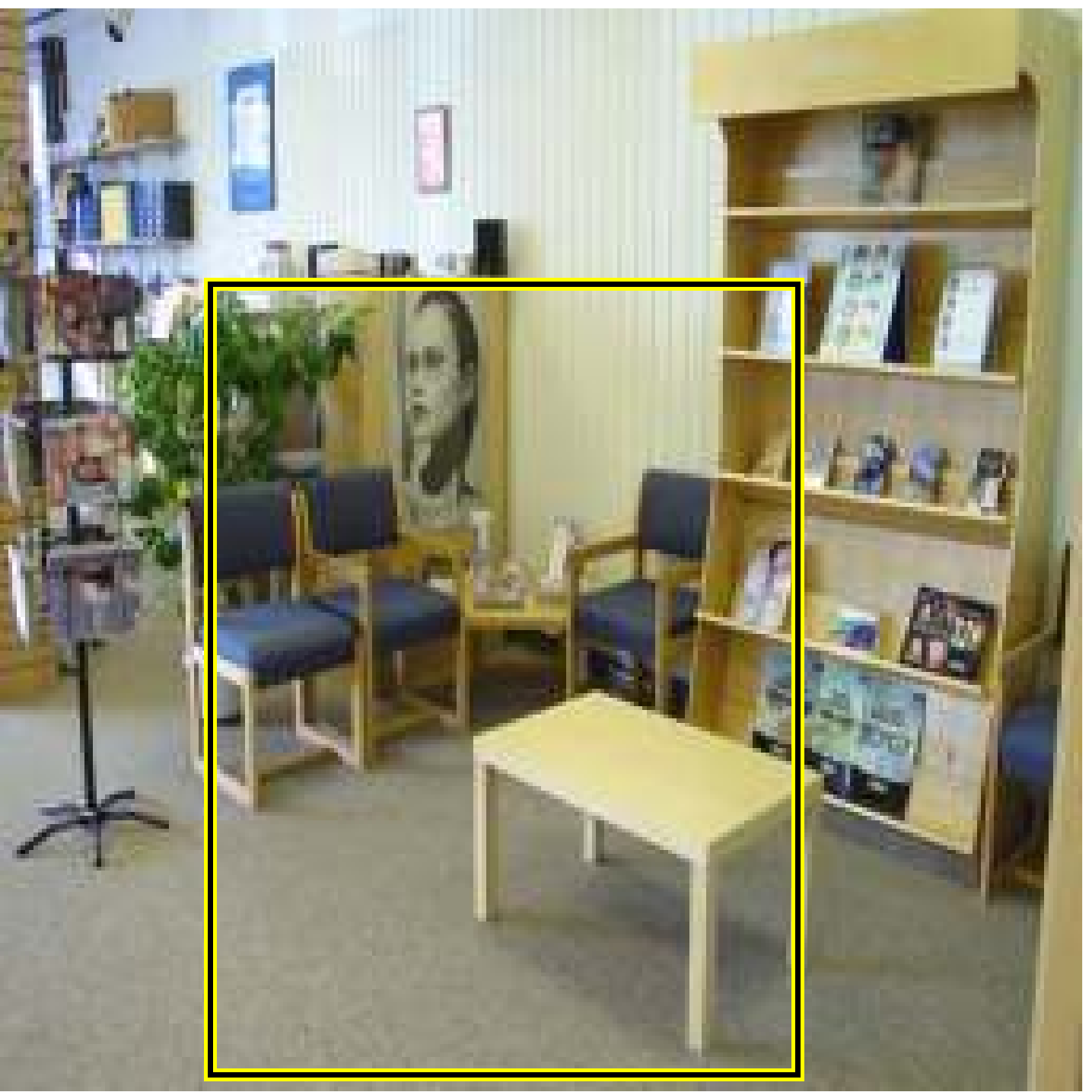} &
				\includegraphics[height=0.63in, width=0.85in]{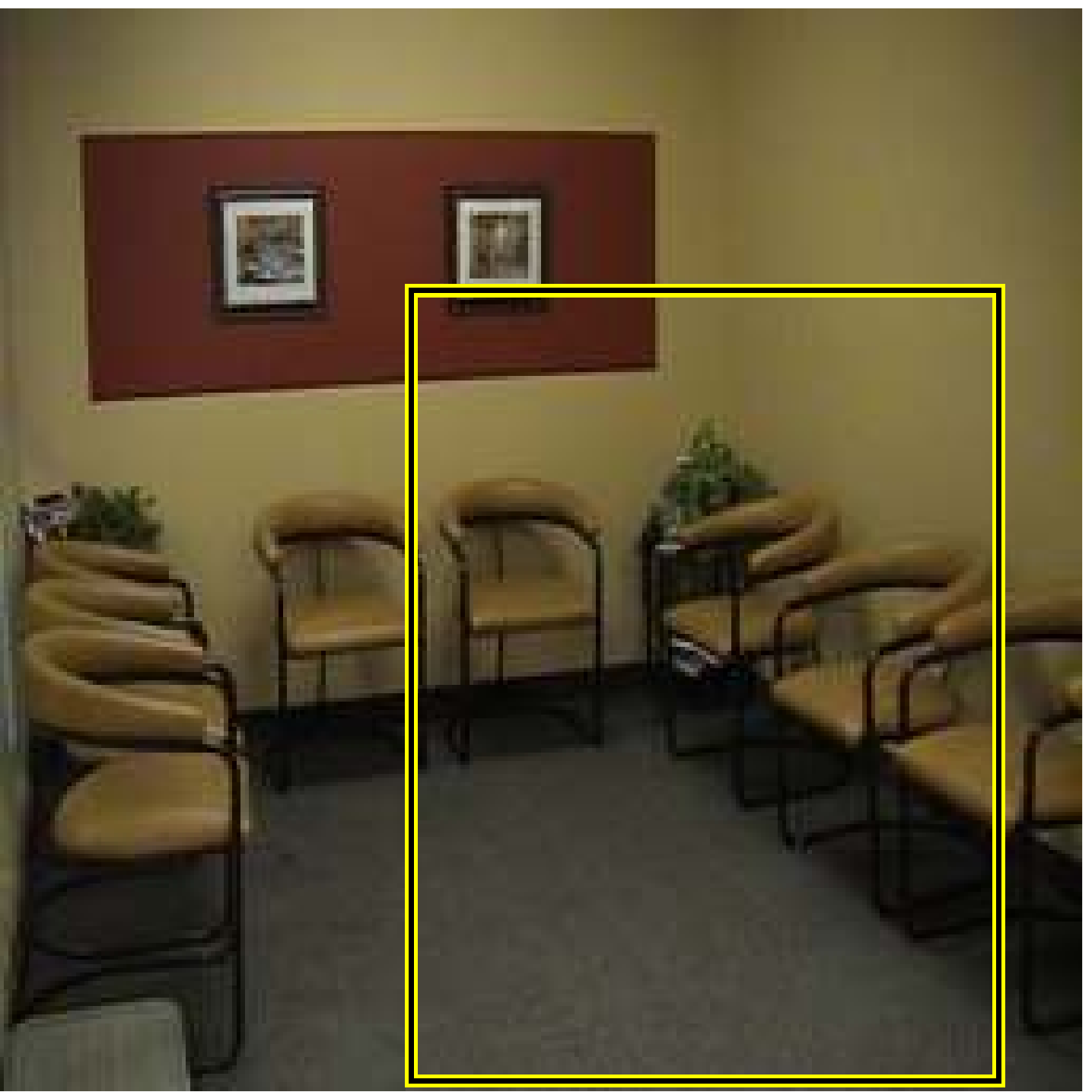} &
				\includegraphics[height=0.63in, width=0.85in]{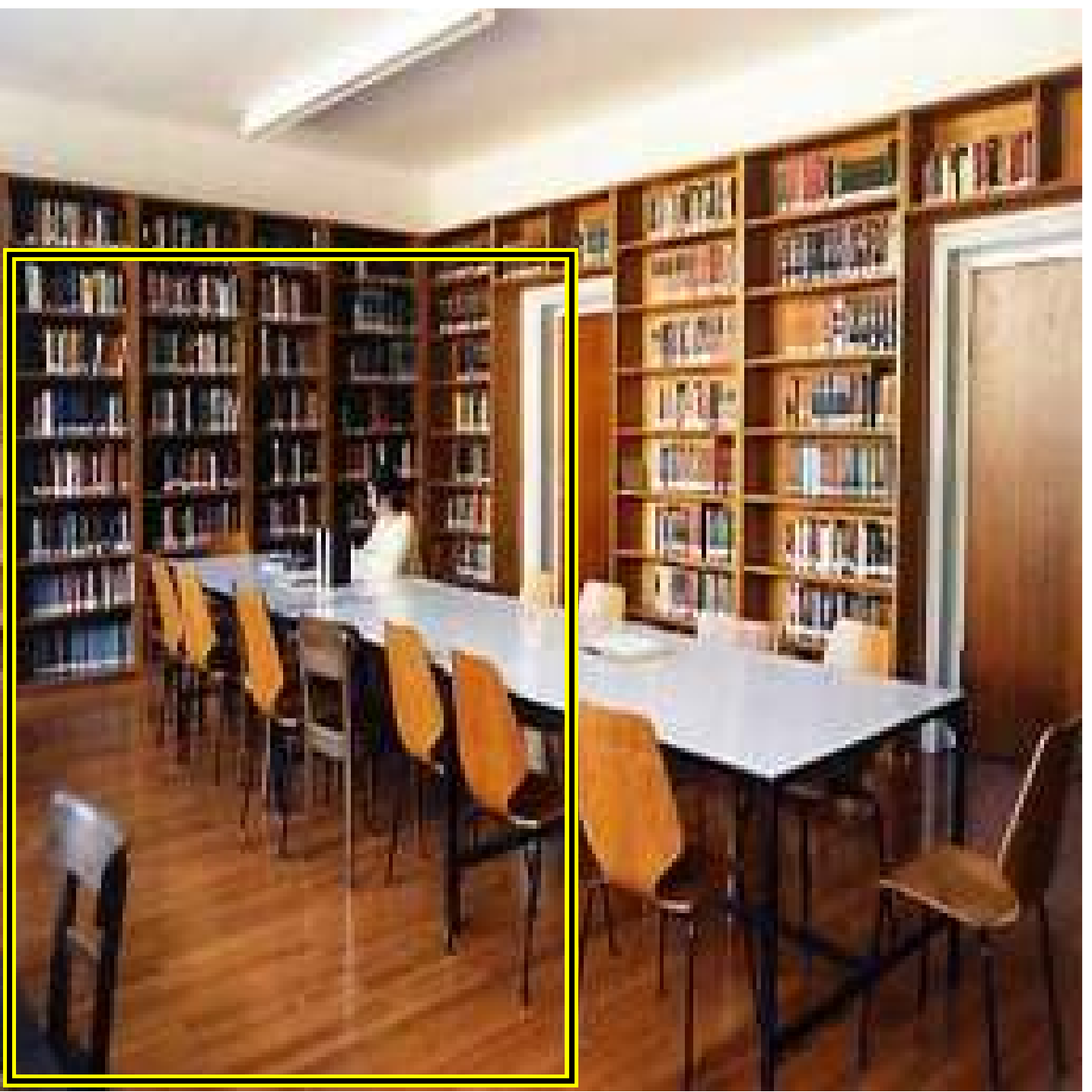} &
				\includegraphics[height=0.63in, width=0.85in]{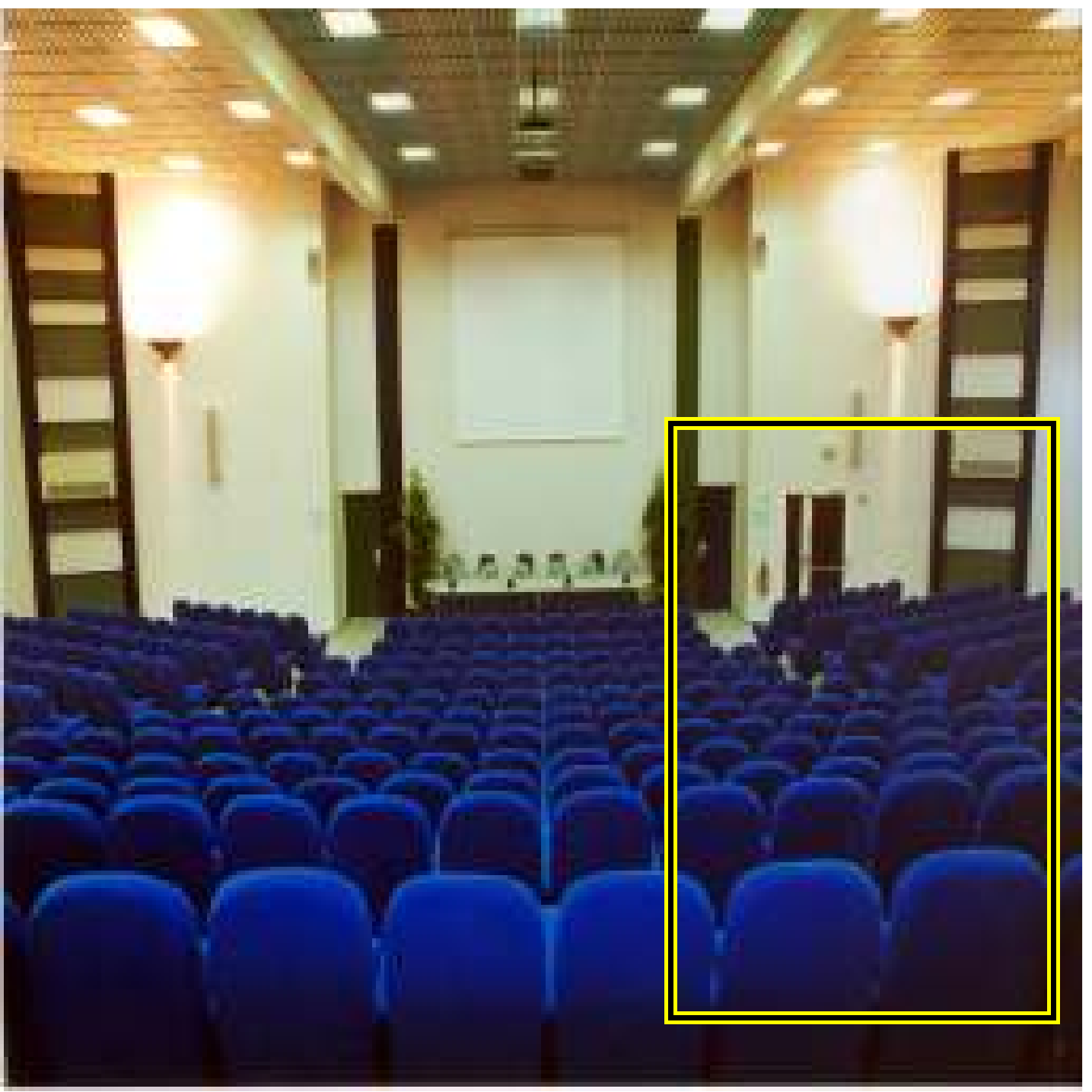} &
				\includegraphics[height=0.63in, width=0.85in]{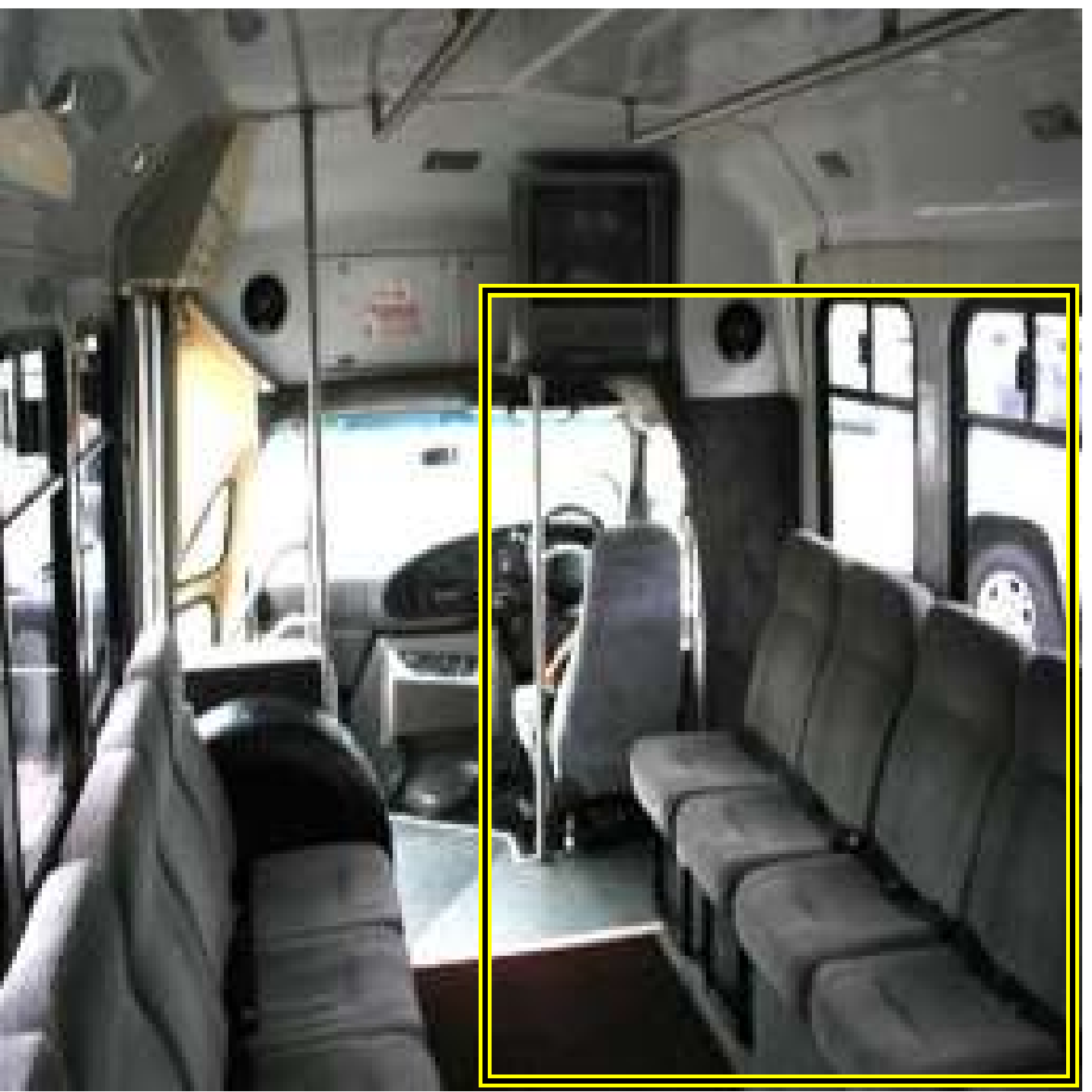}  \\ [-0.05cm]
	\rotatebox{90}{\hspace{0.27cm}Part 27}$\;$ &
				\includegraphics[height=0.63in, width=0.85in]{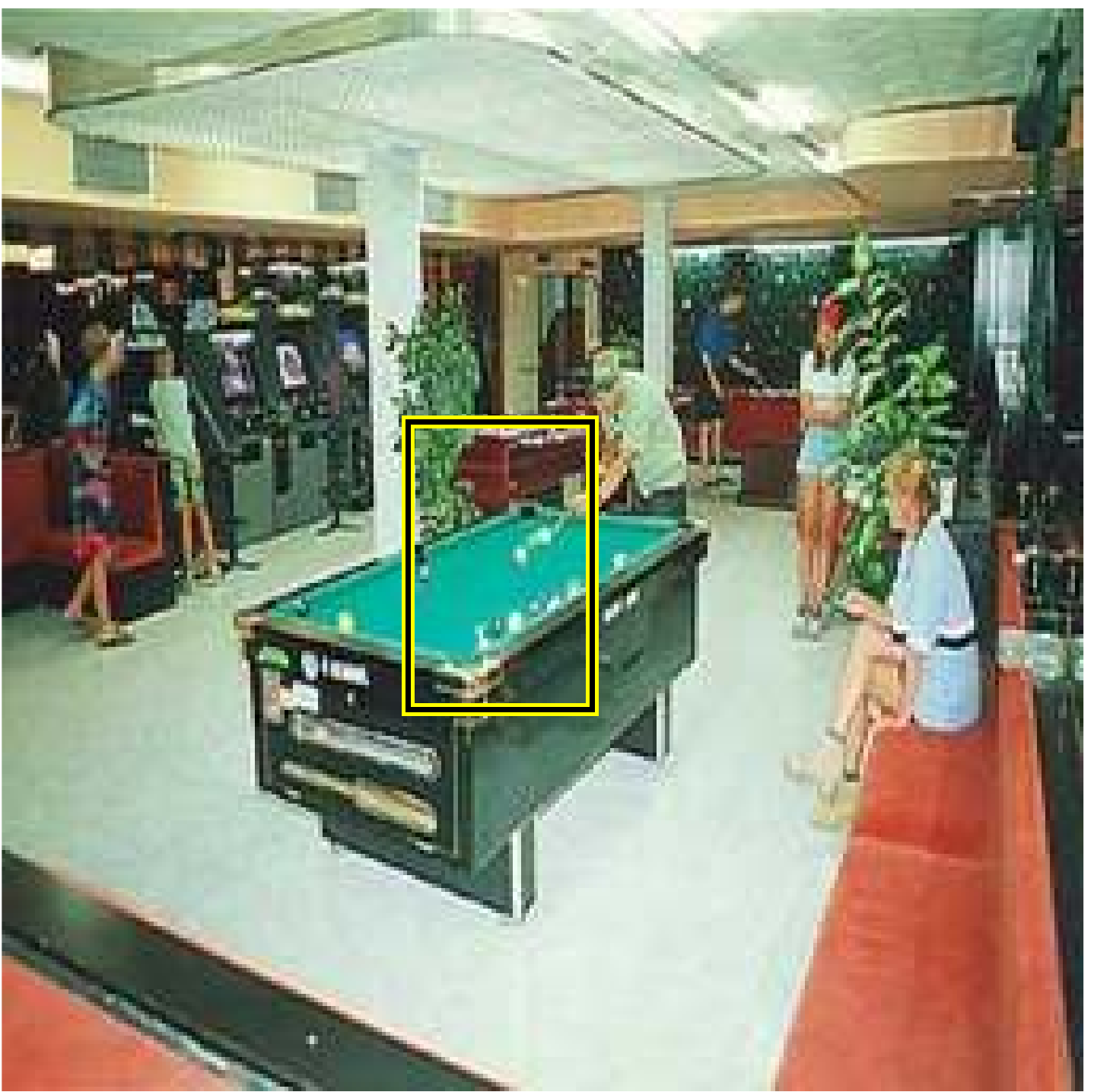} &
				\includegraphics[height=0.63in, width=0.85in]{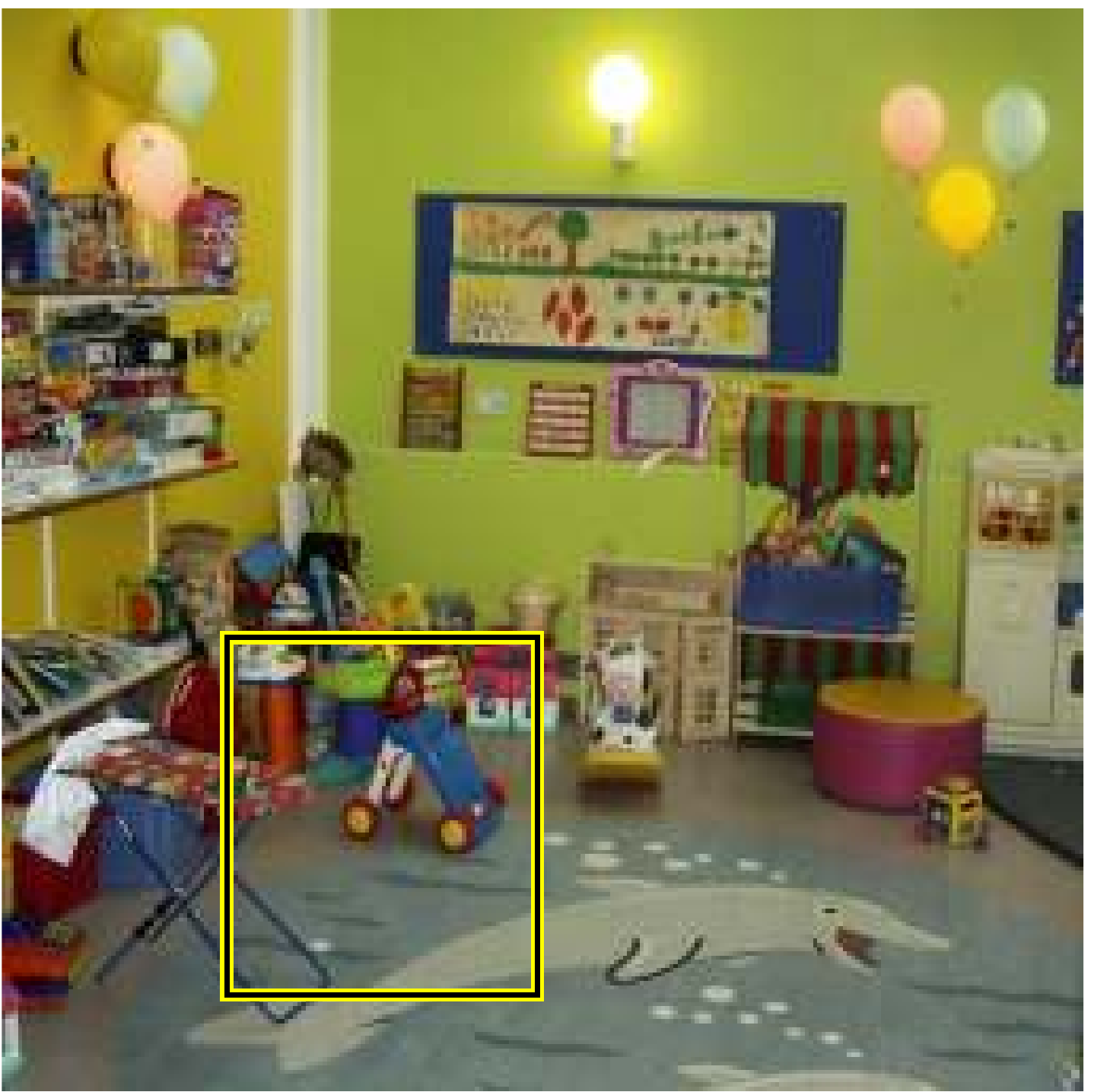} &
				\includegraphics[height=0.63in, width=0.85in]{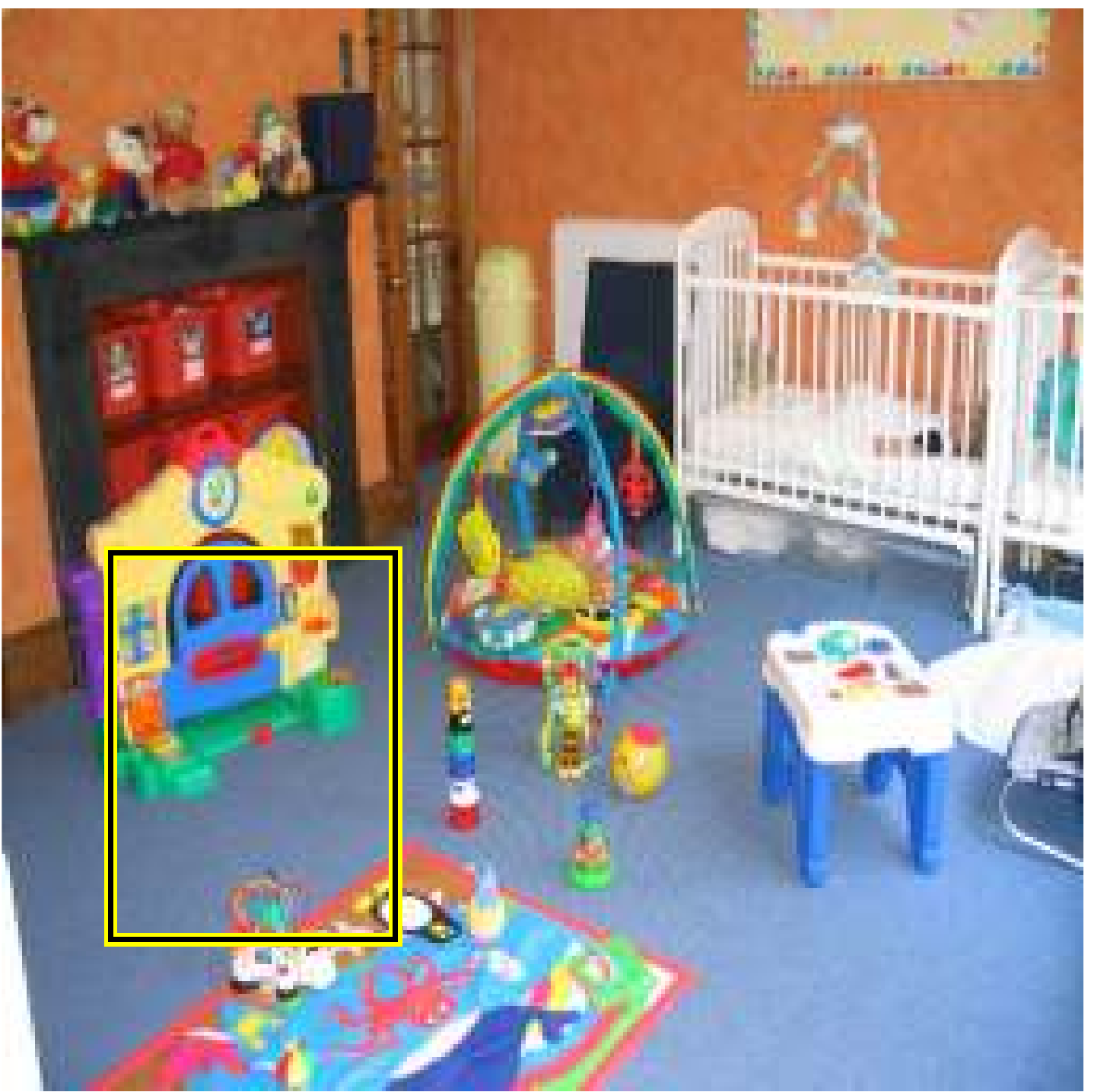} &
				\includegraphics[height=0.63in, width=0.85in]{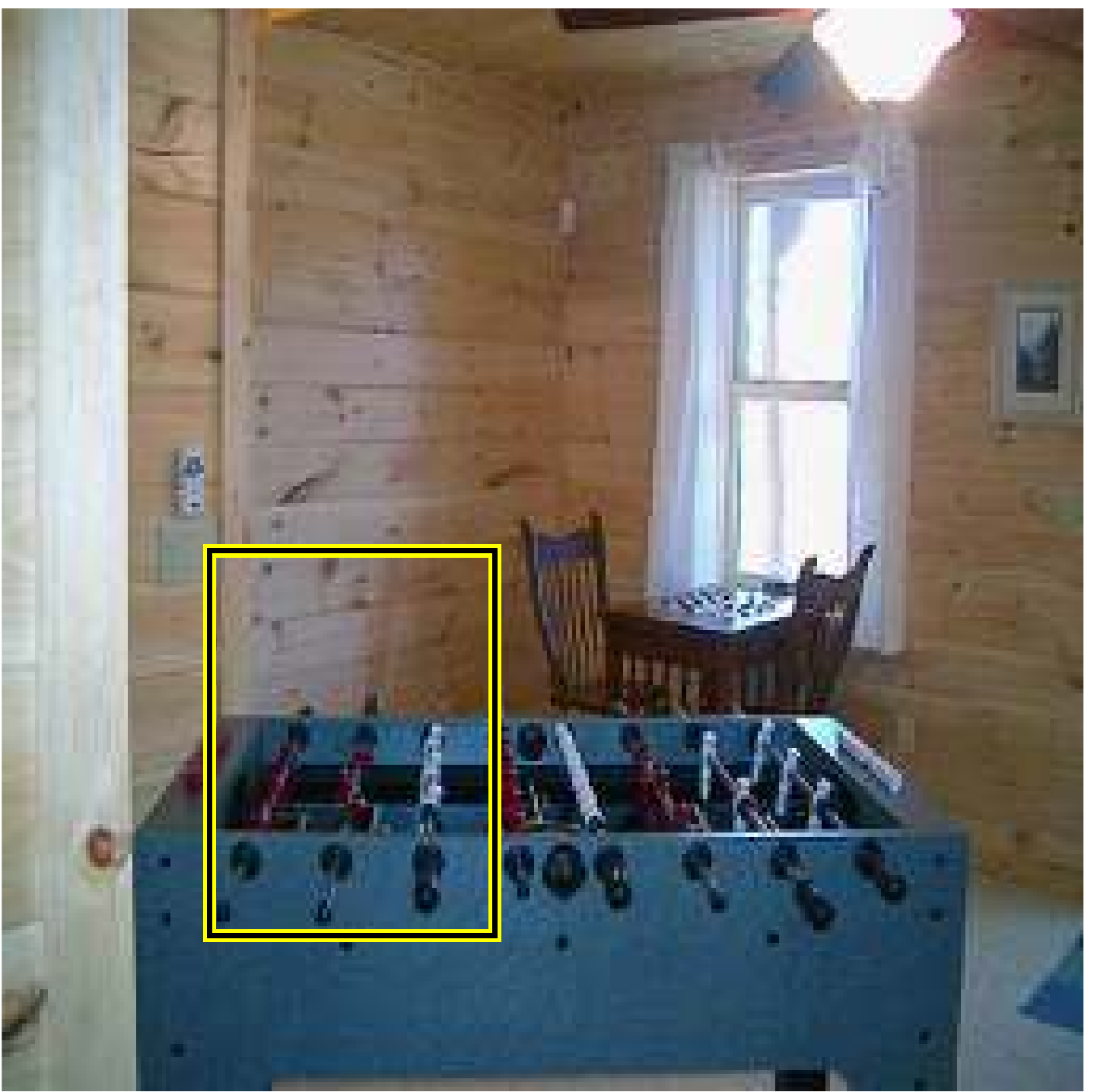} &
				\includegraphics[height=0.63in, width=0.85in]{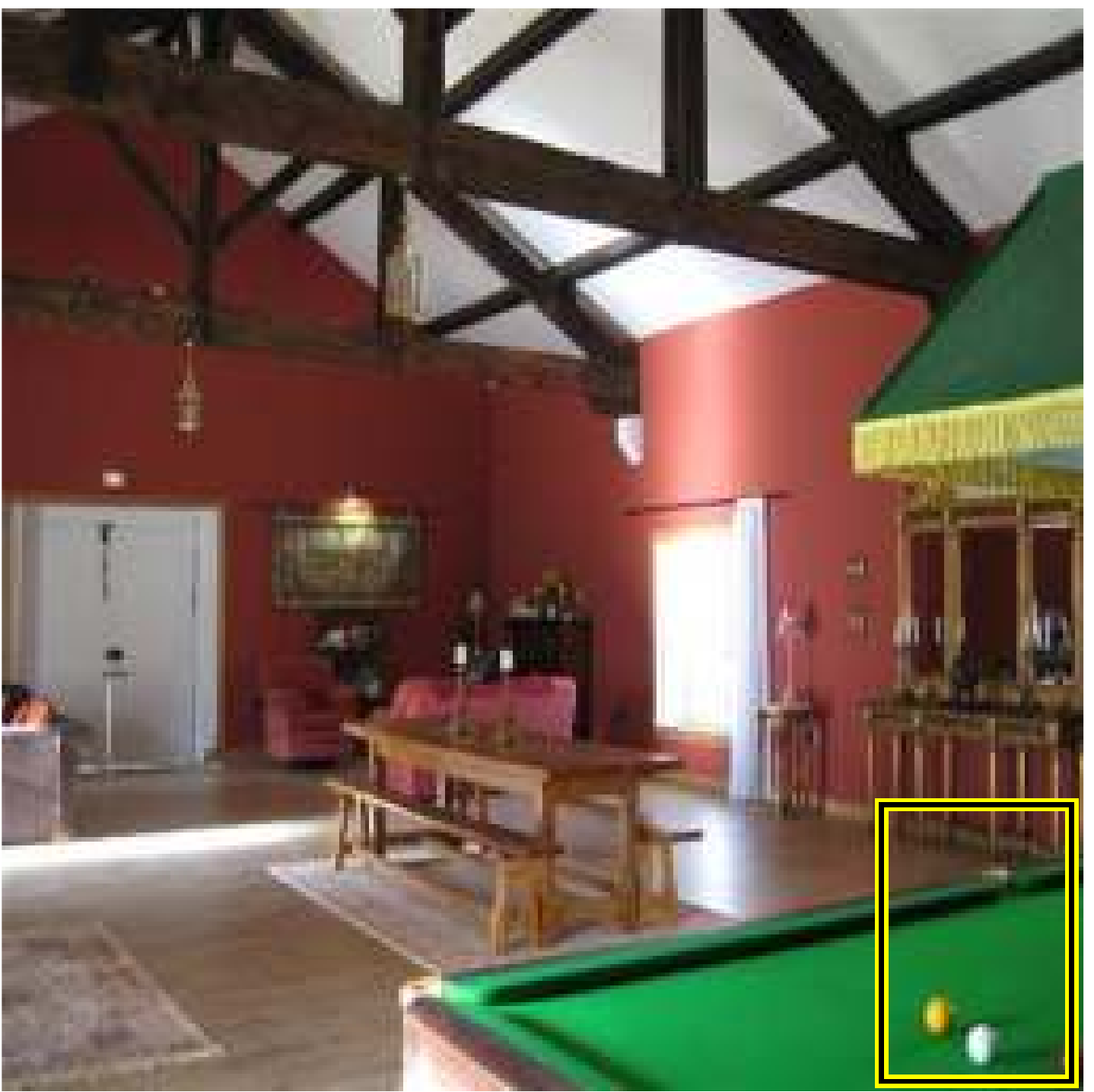} &
				\includegraphics[height=0.63in, width=0.85in]{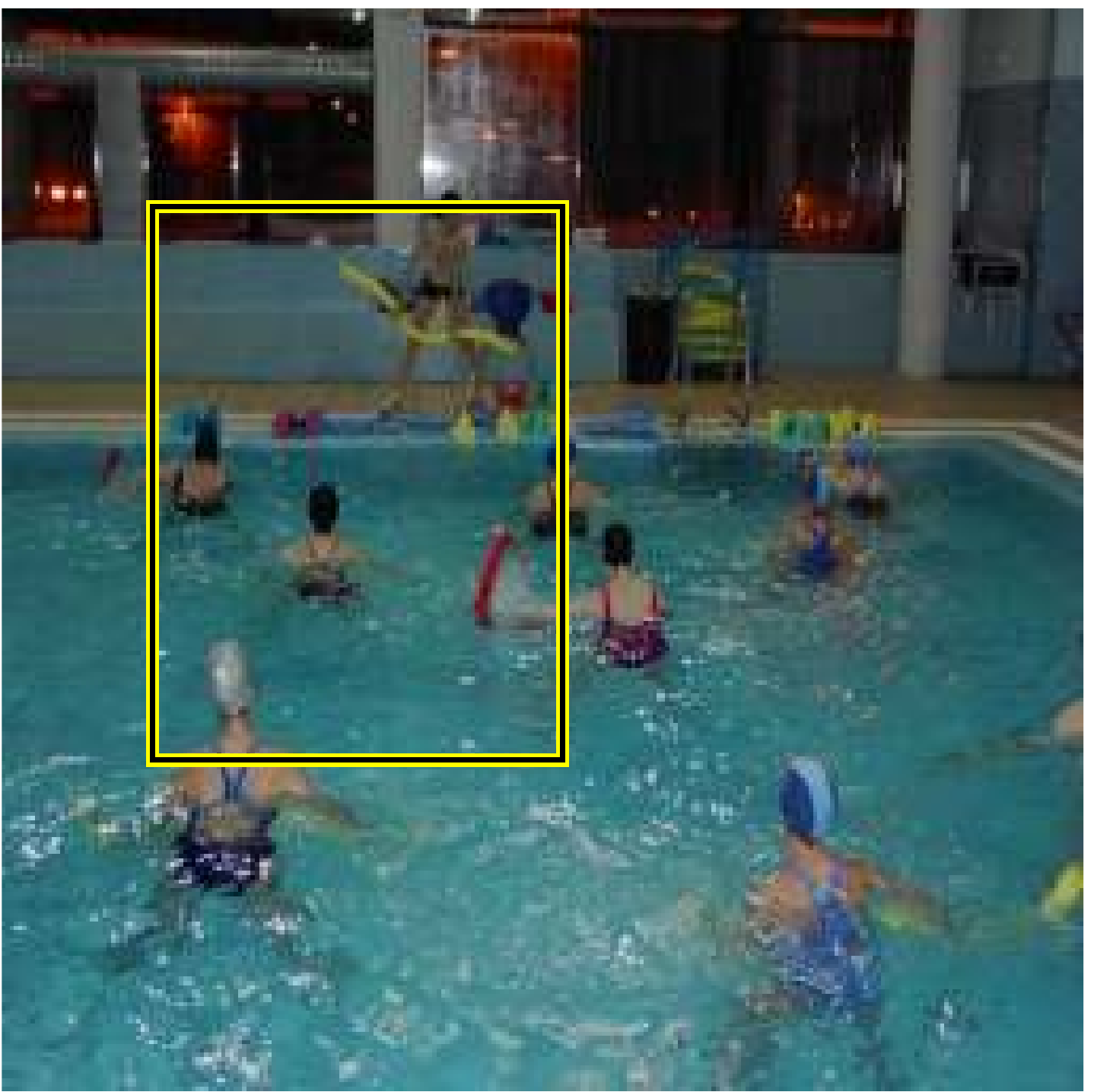} \\ [-0.05cm]
	\rotatebox{90}{\hspace{0.27cm}Part 29}$\;$ &
				\includegraphics[height=0.63in, width=0.85in]{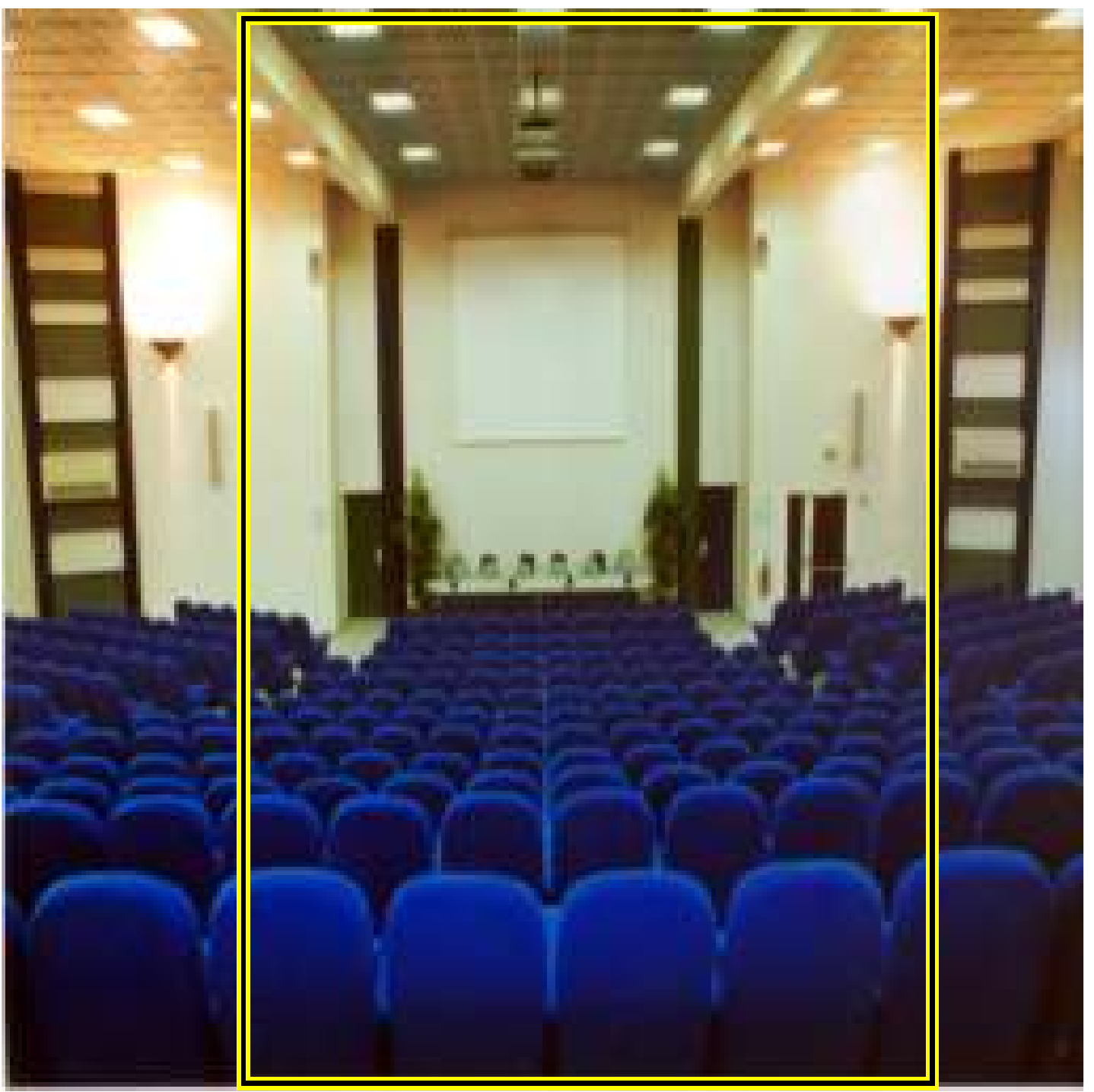} &
				\includegraphics[height=0.63in, width=0.85in]{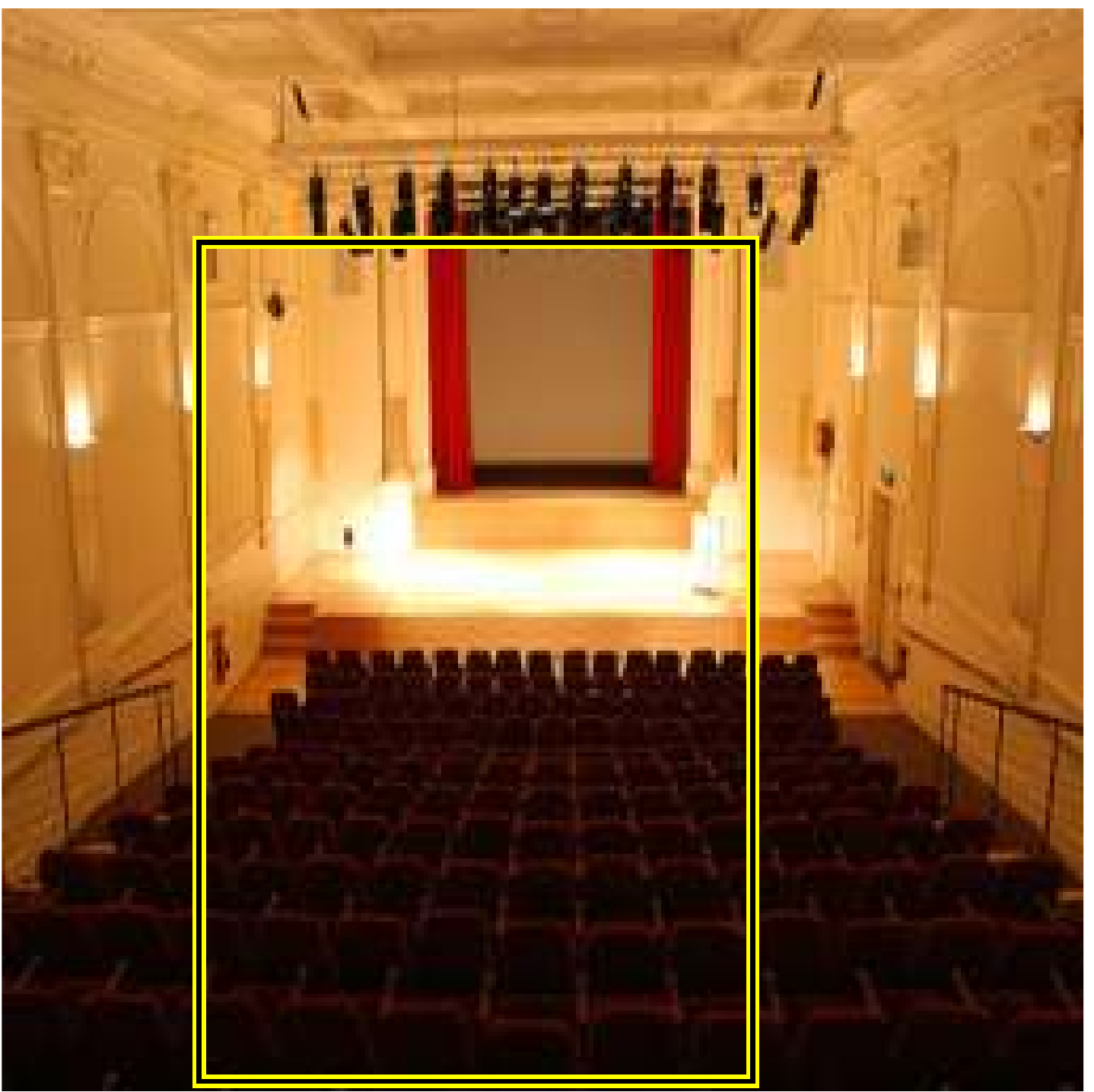} &
				\includegraphics[height=0.63in, width=0.85in]{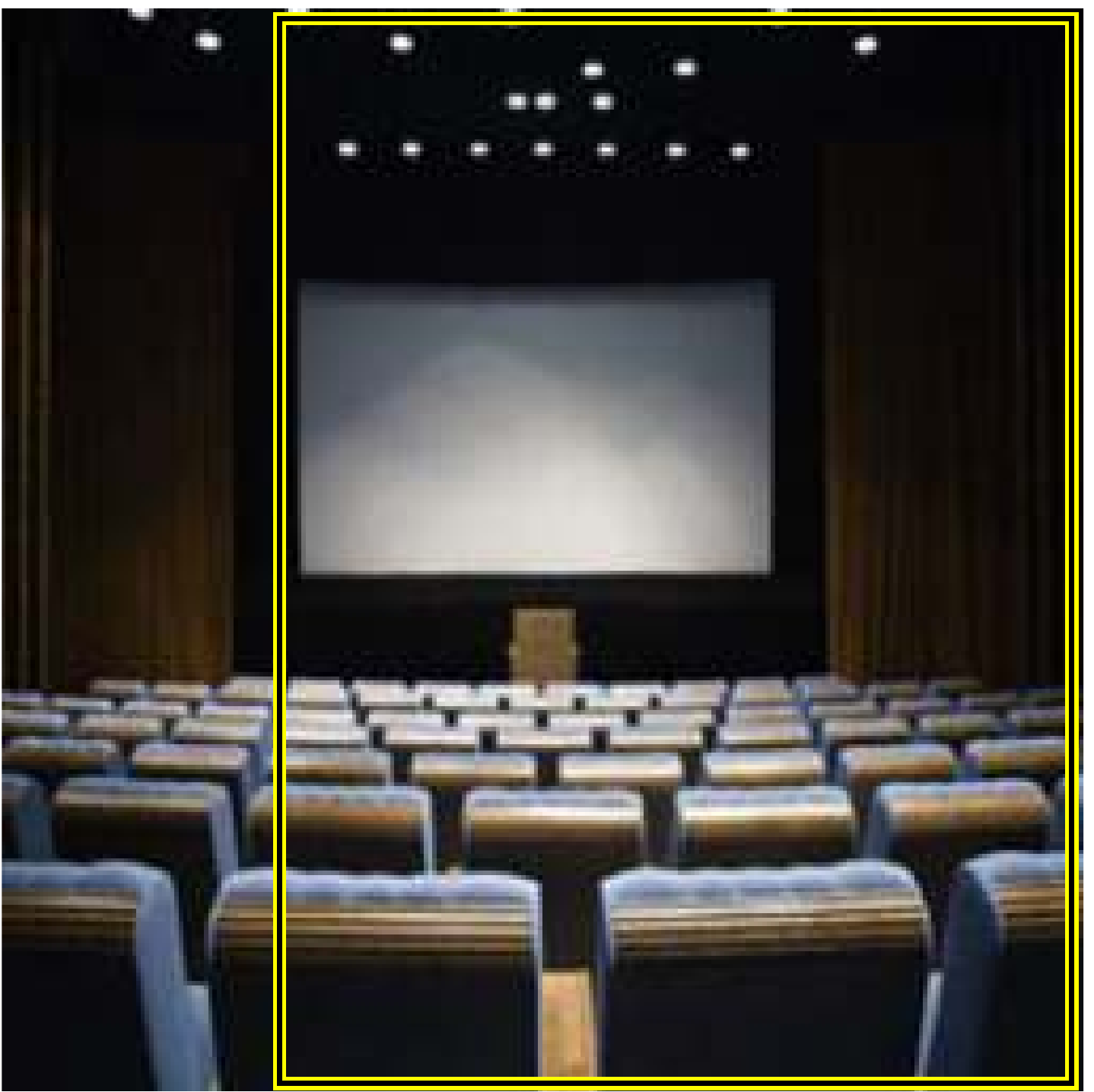} &
				\includegraphics[height=0.63in, width=0.85in]{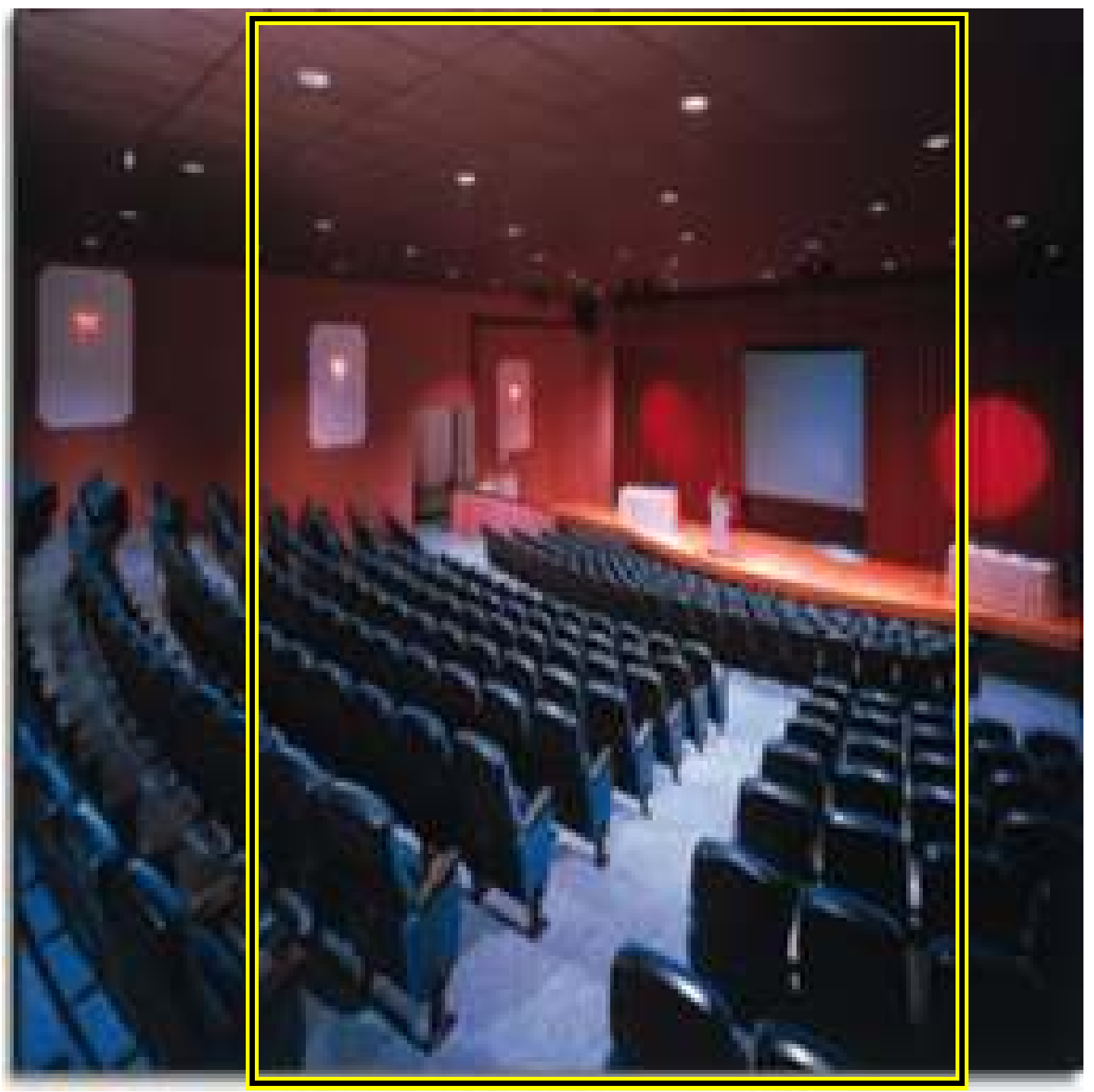} &
				\includegraphics[height=0.63in, width=0.85in]{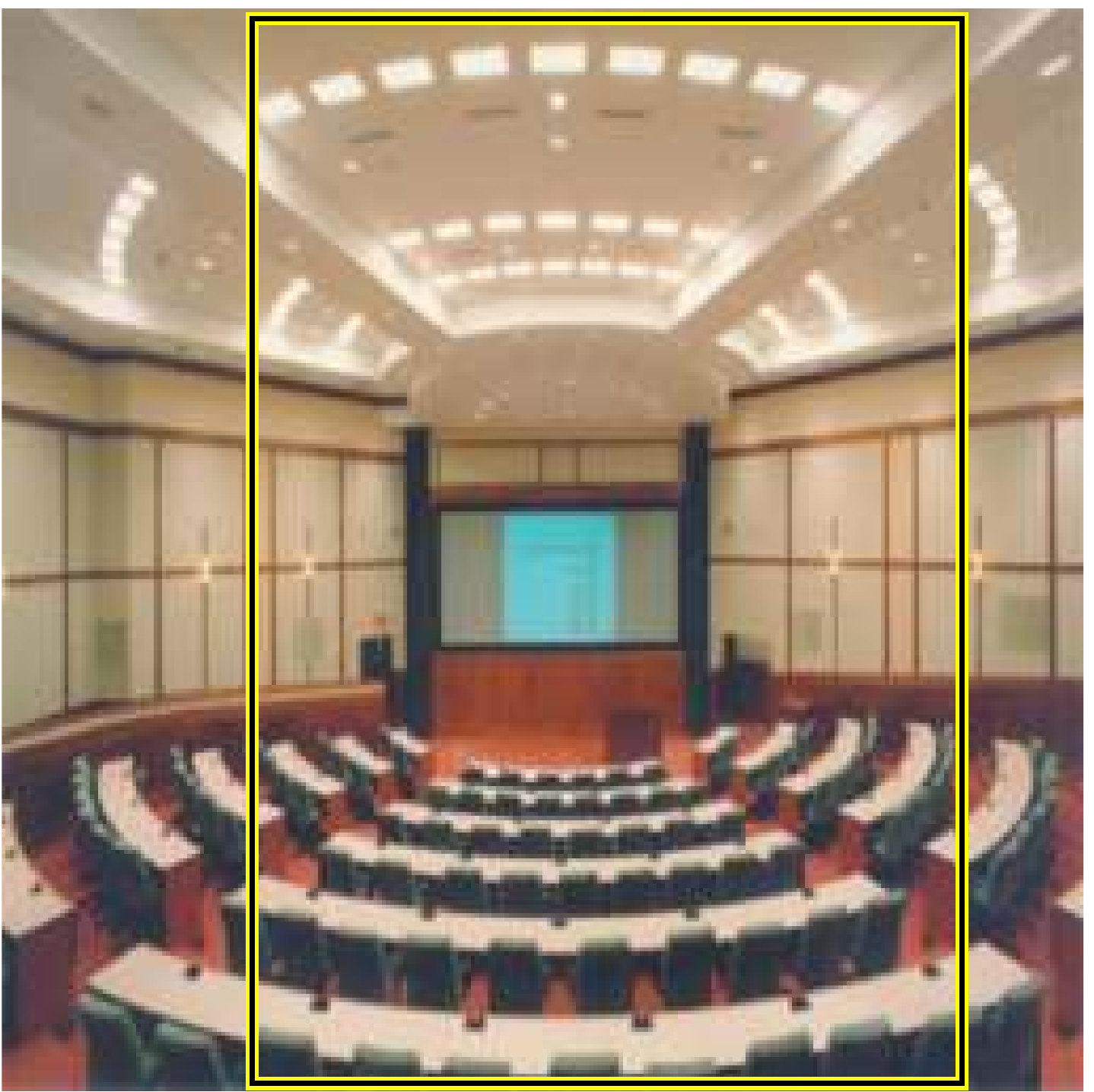} &
				\includegraphics[height=0.63in, width=0.85in]{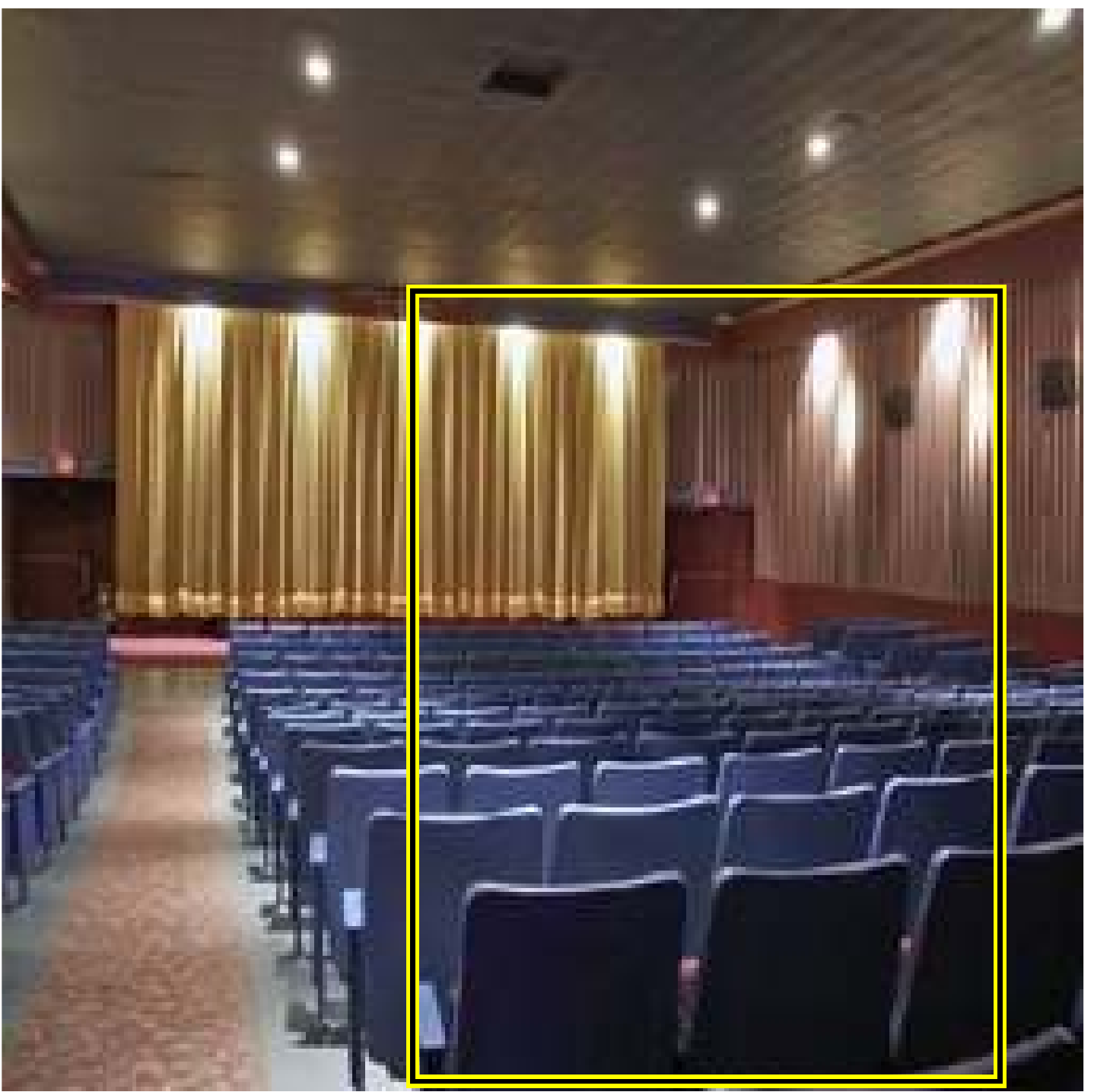} \\ [-0.05cm]
	\rotatebox{90}{\hspace{0.27cm}Part 35}$\;$ &
				\includegraphics[height=0.63in, width=0.85in]{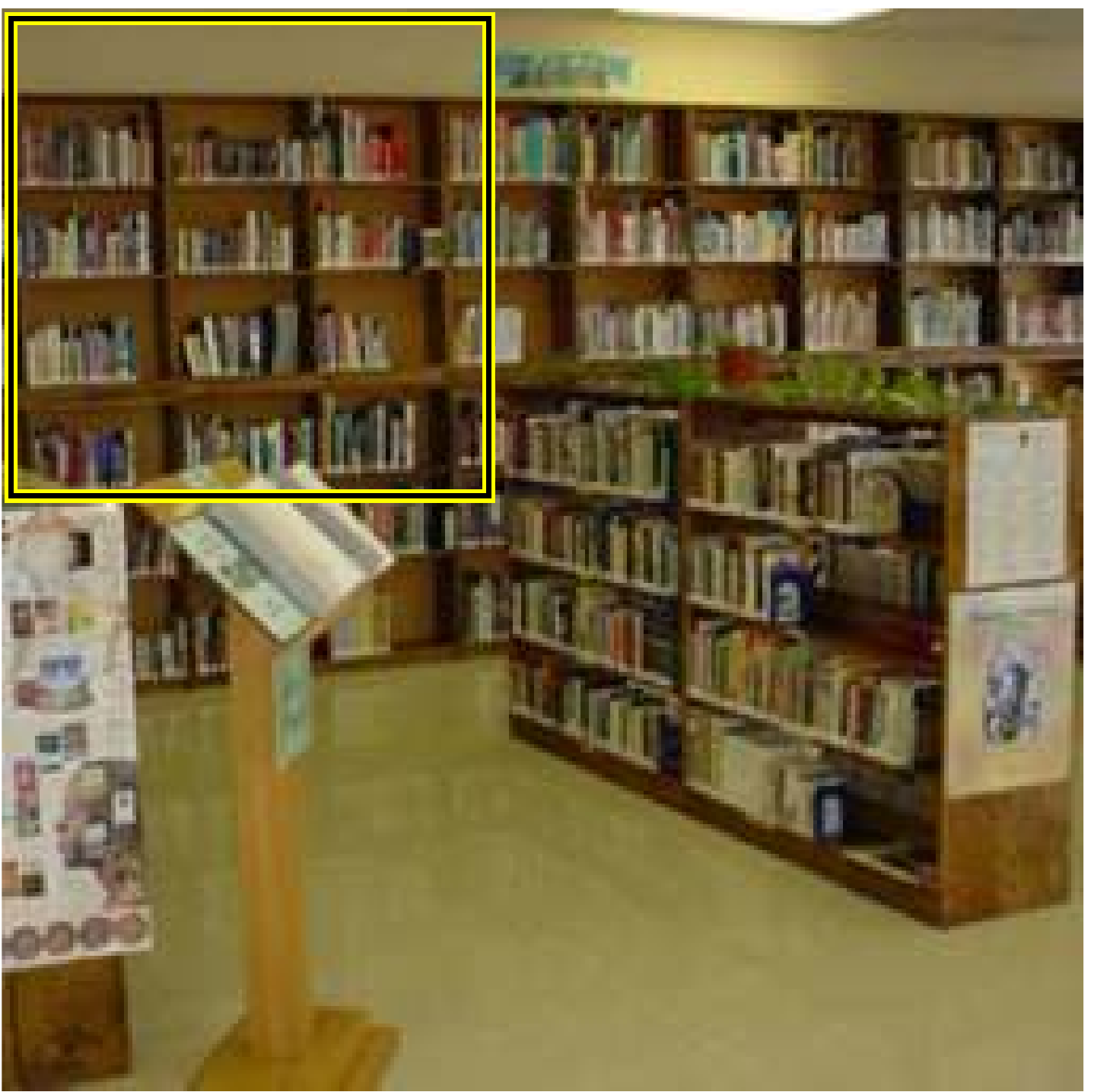} &
				\includegraphics[height=0.63in, width=0.85in]{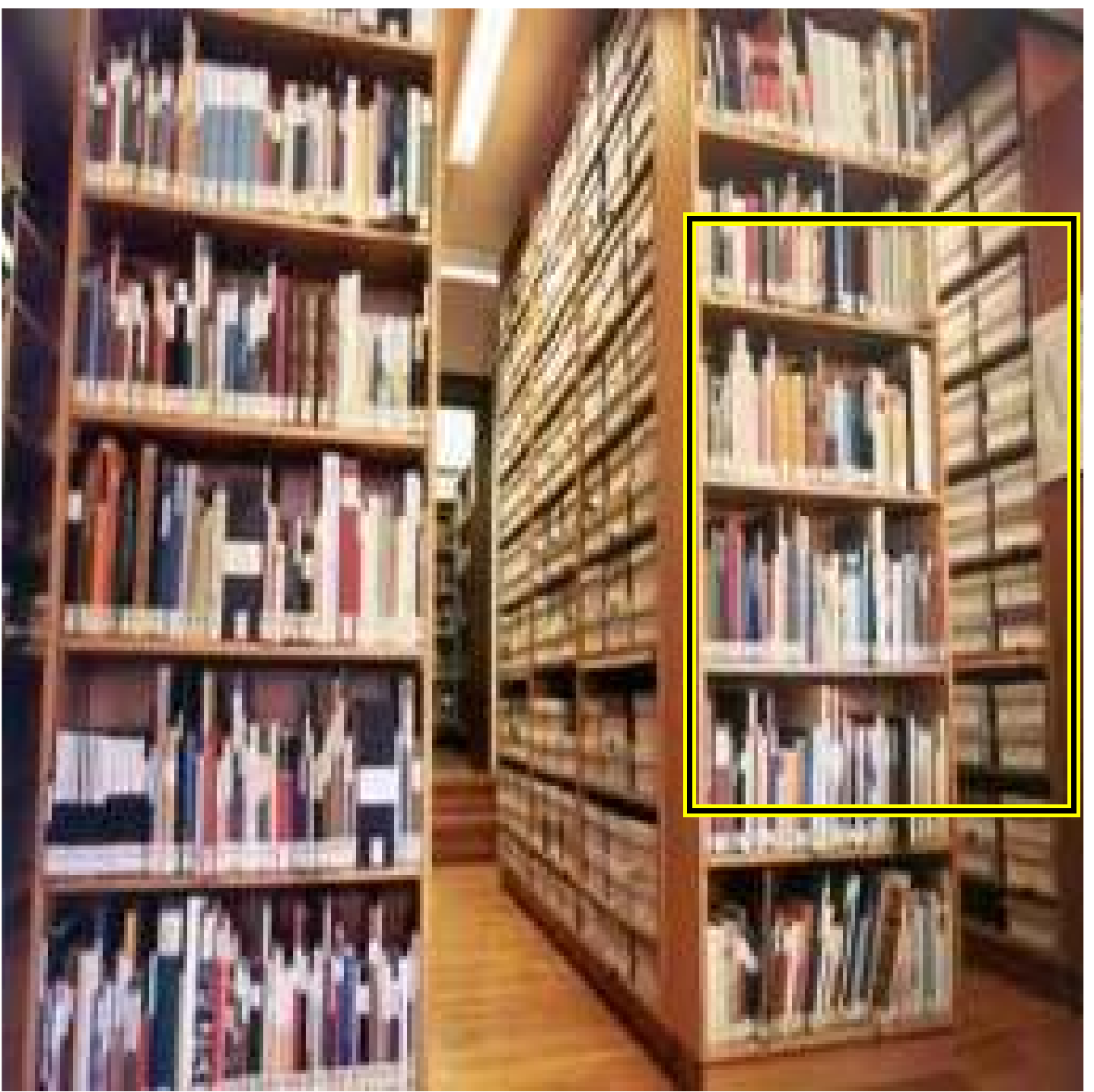} &
				\includegraphics[height=0.63in, width=0.85in]{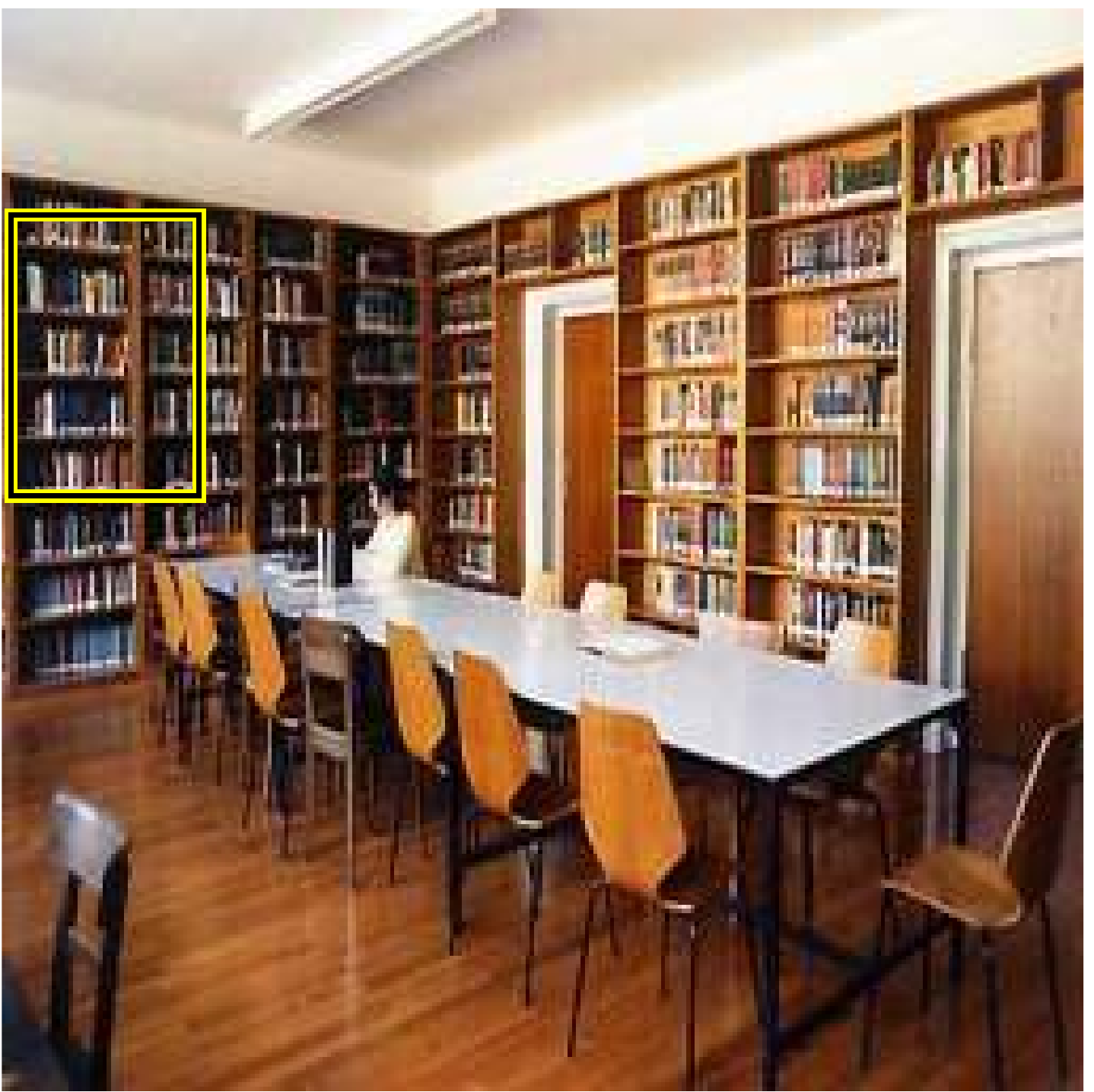} &
				\includegraphics[height=0.63in, width=0.85in]{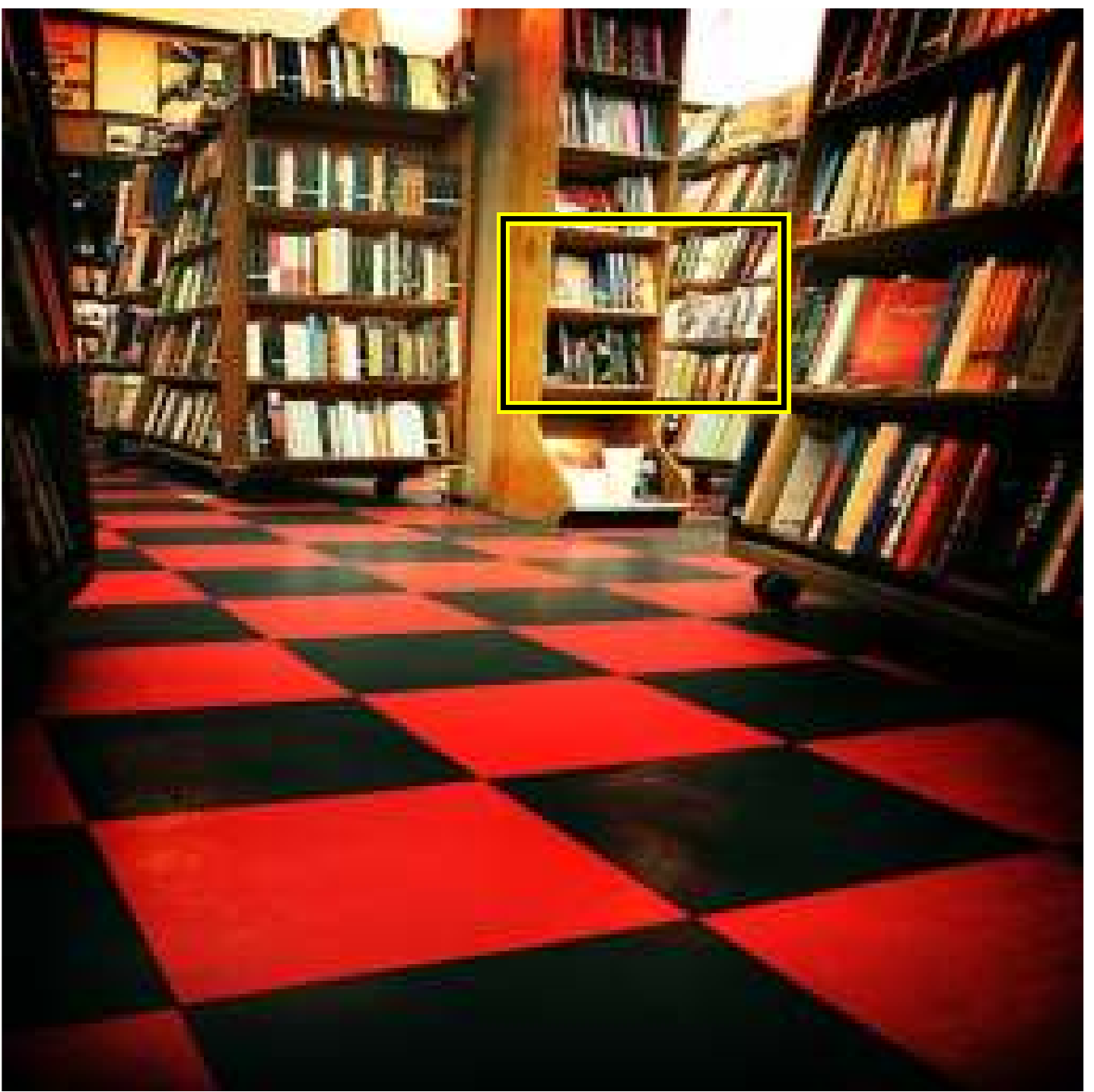} &
				\includegraphics[height=0.63in, width=0.85in]{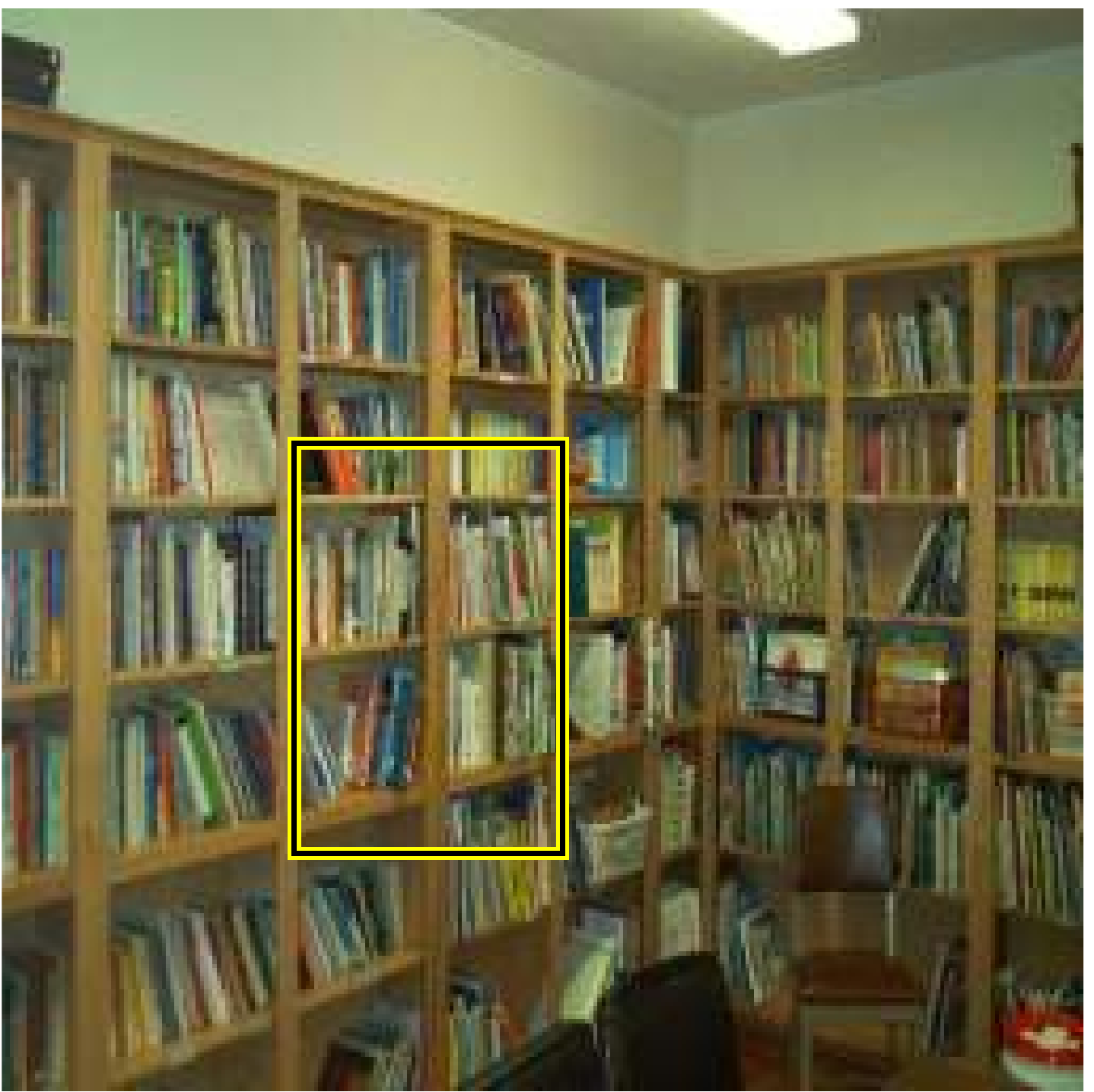} &
				\includegraphics[height=0.63in, width=0.85in]{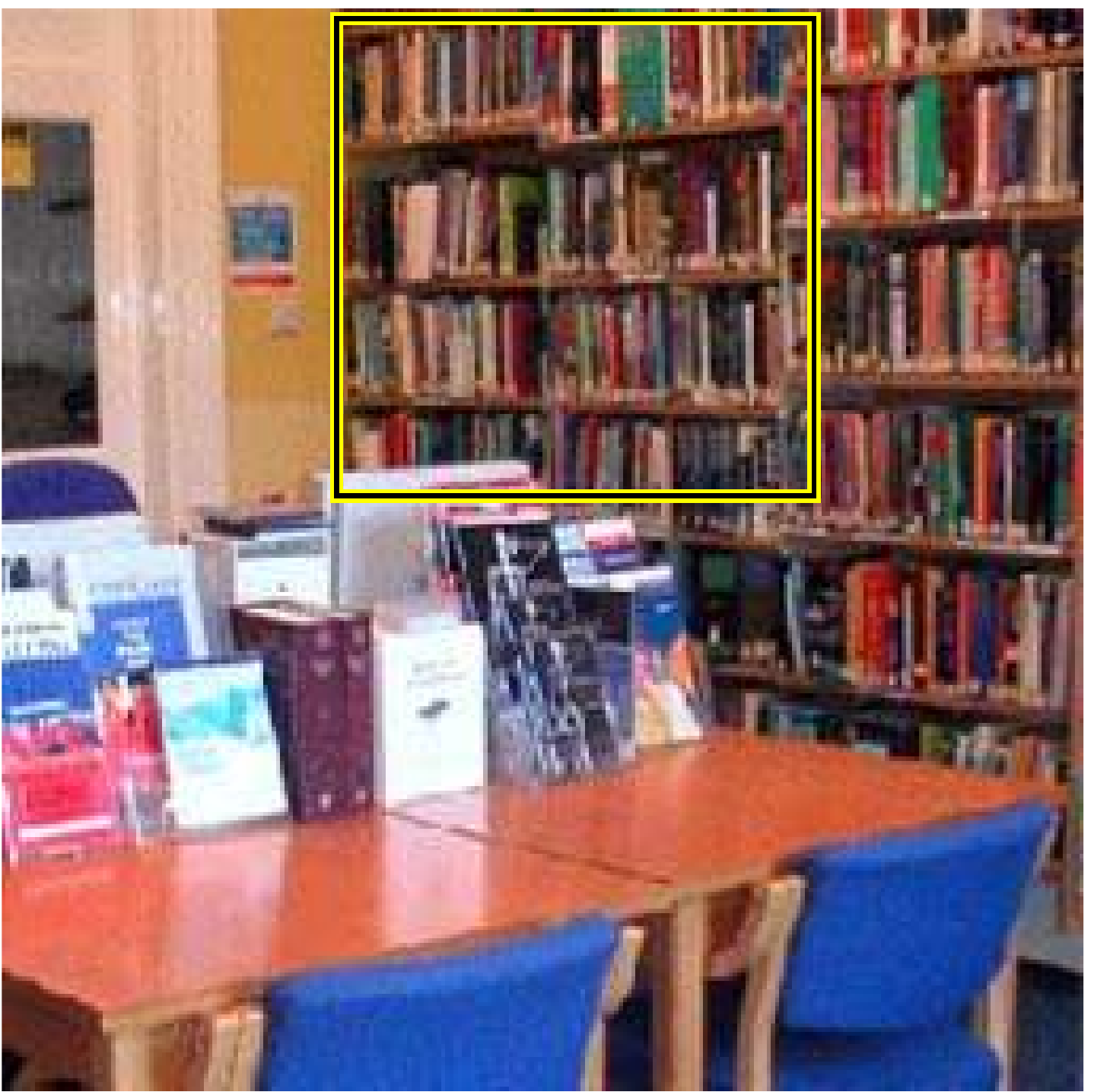} \\ [-0.05cm]
	\rotatebox{90}{\hspace{0.27cm}Part 37}$\;$ &
				\includegraphics[height=0.63in, width=0.85in]{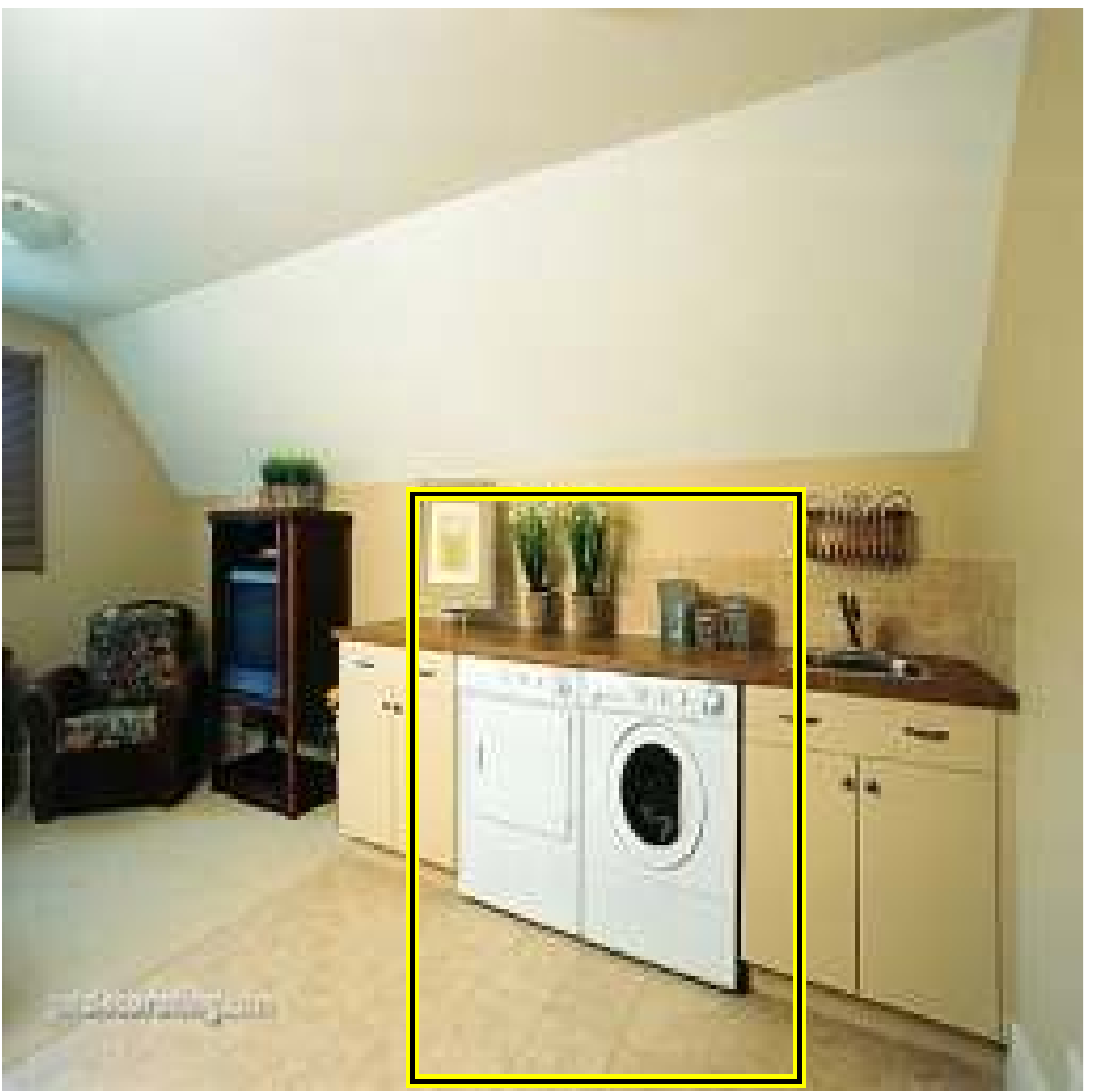} &
				\includegraphics[height=0.63in, width=0.85in]{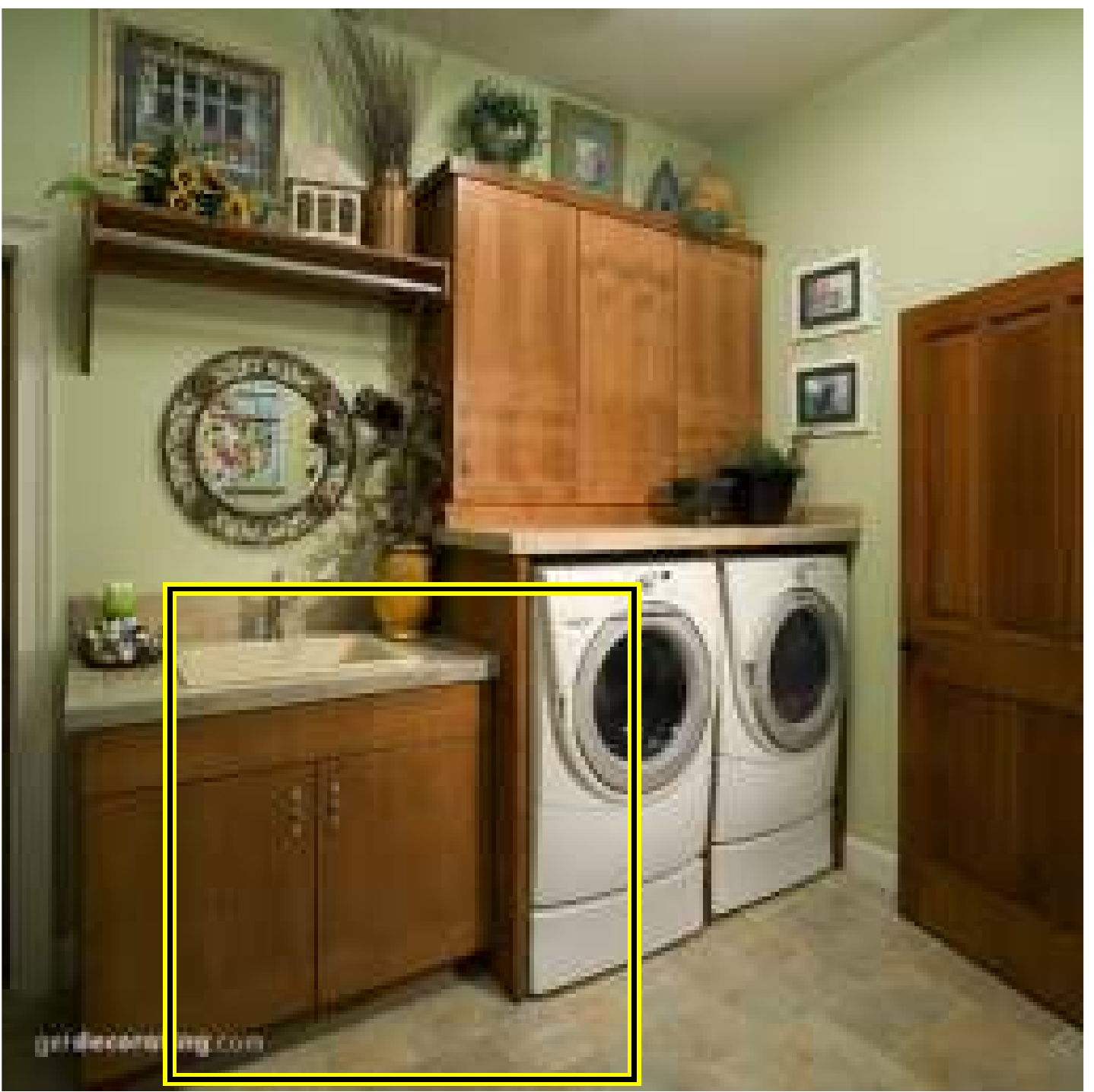} &
				\includegraphics[height=0.63in, width=0.85in]{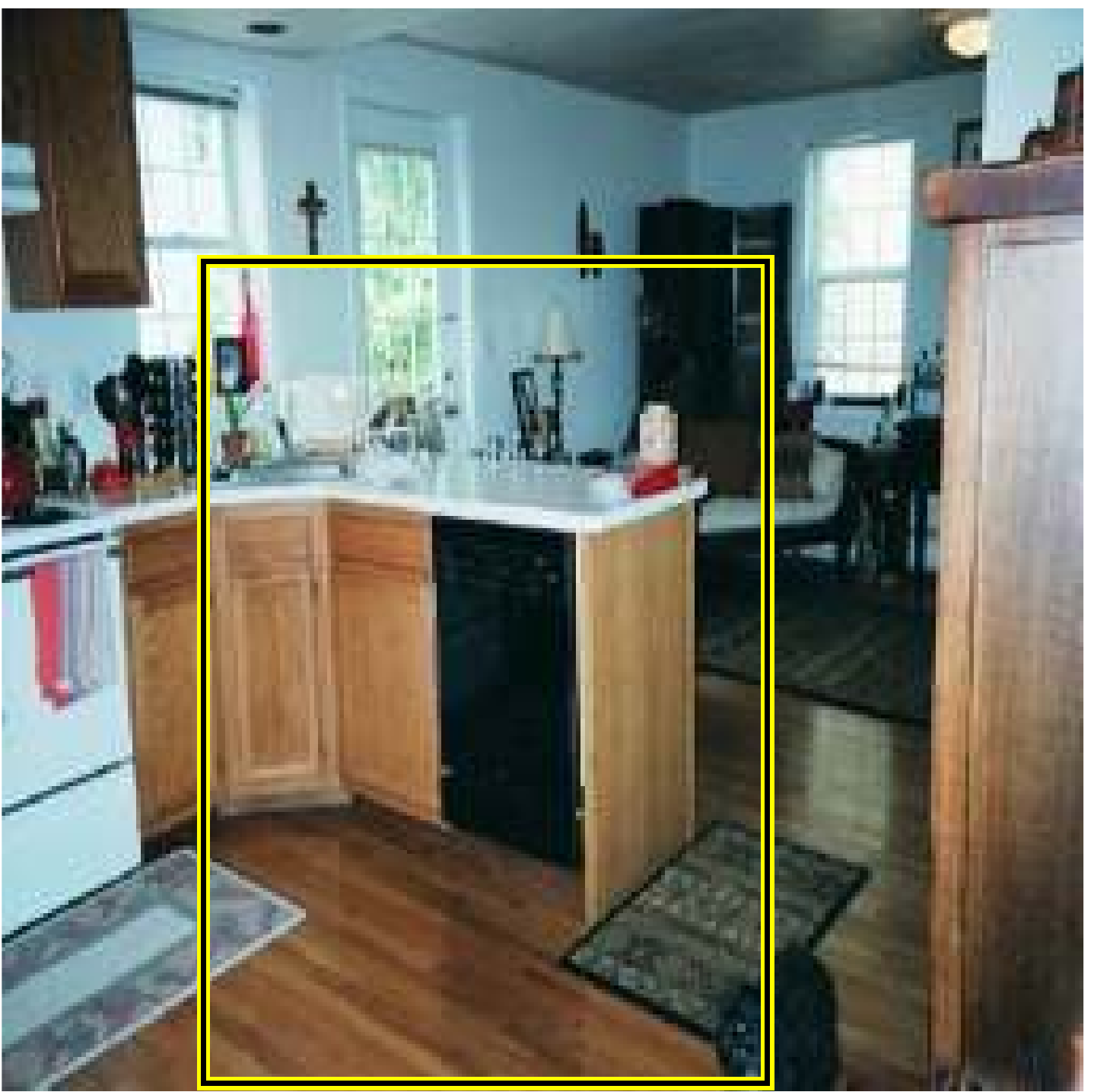} &
				\includegraphics[height=0.63in, width=0.85in]{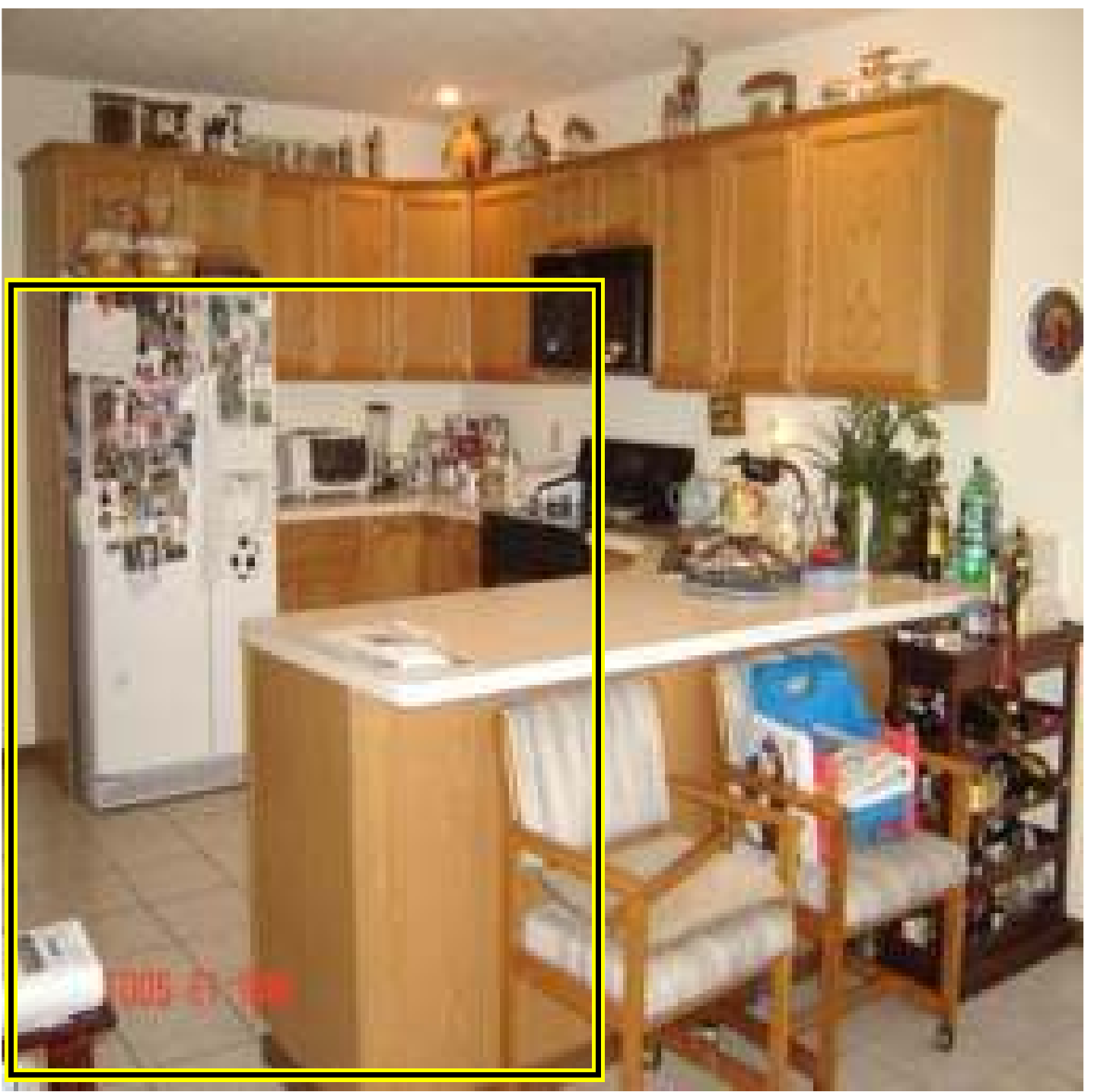} &
				\includegraphics[height=0.63in, width=0.85in]{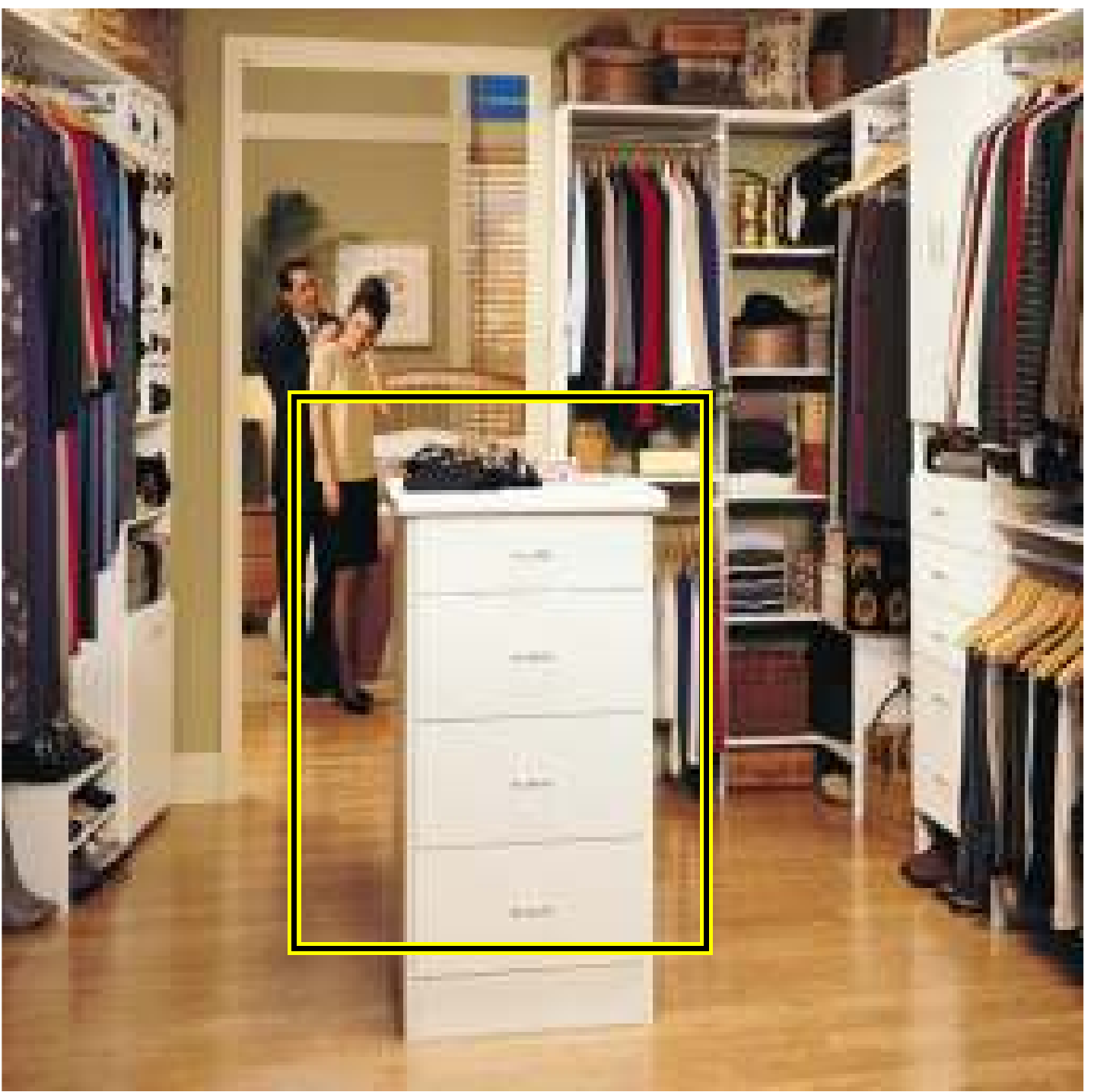} &
				\includegraphics[height=0.63in, width=0.85in]{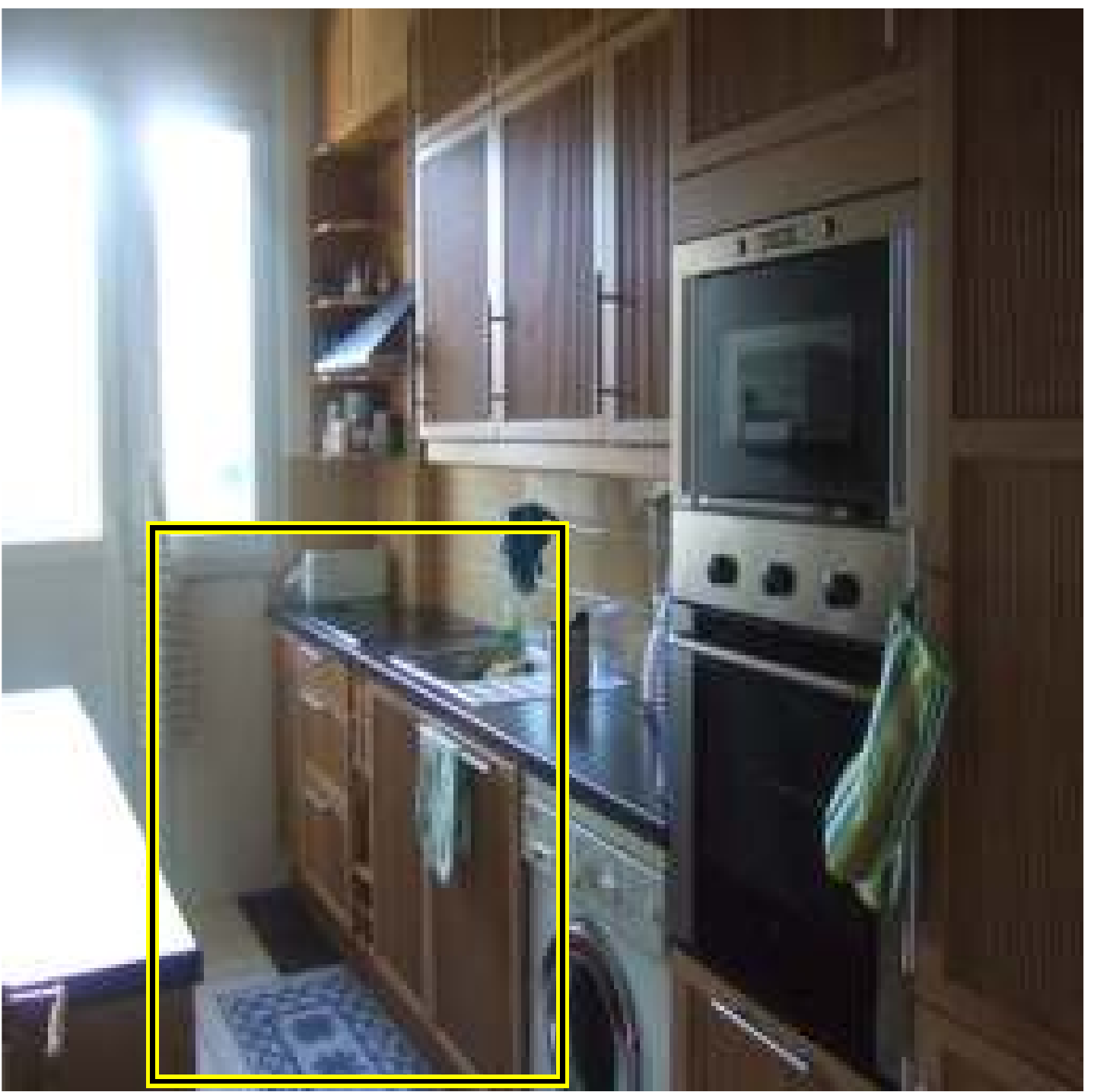} \\ [-0.05cm]
	\rotatebox{90}{\hspace{0.27cm}Part 40}$\;$ &
				\includegraphics[height=0.63in, width=0.85in]{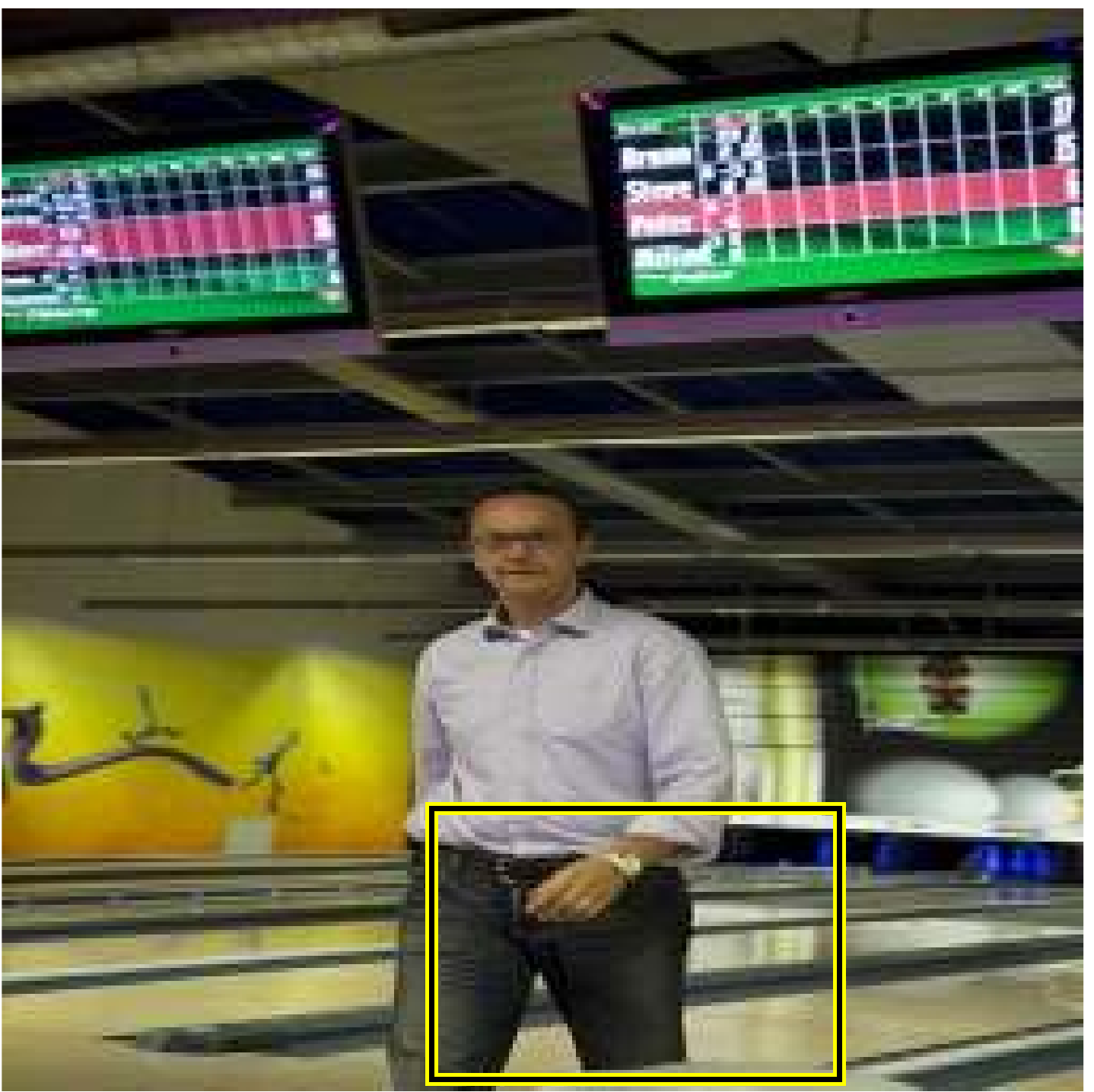} &
				\includegraphics[height=0.63in, width=0.85in]{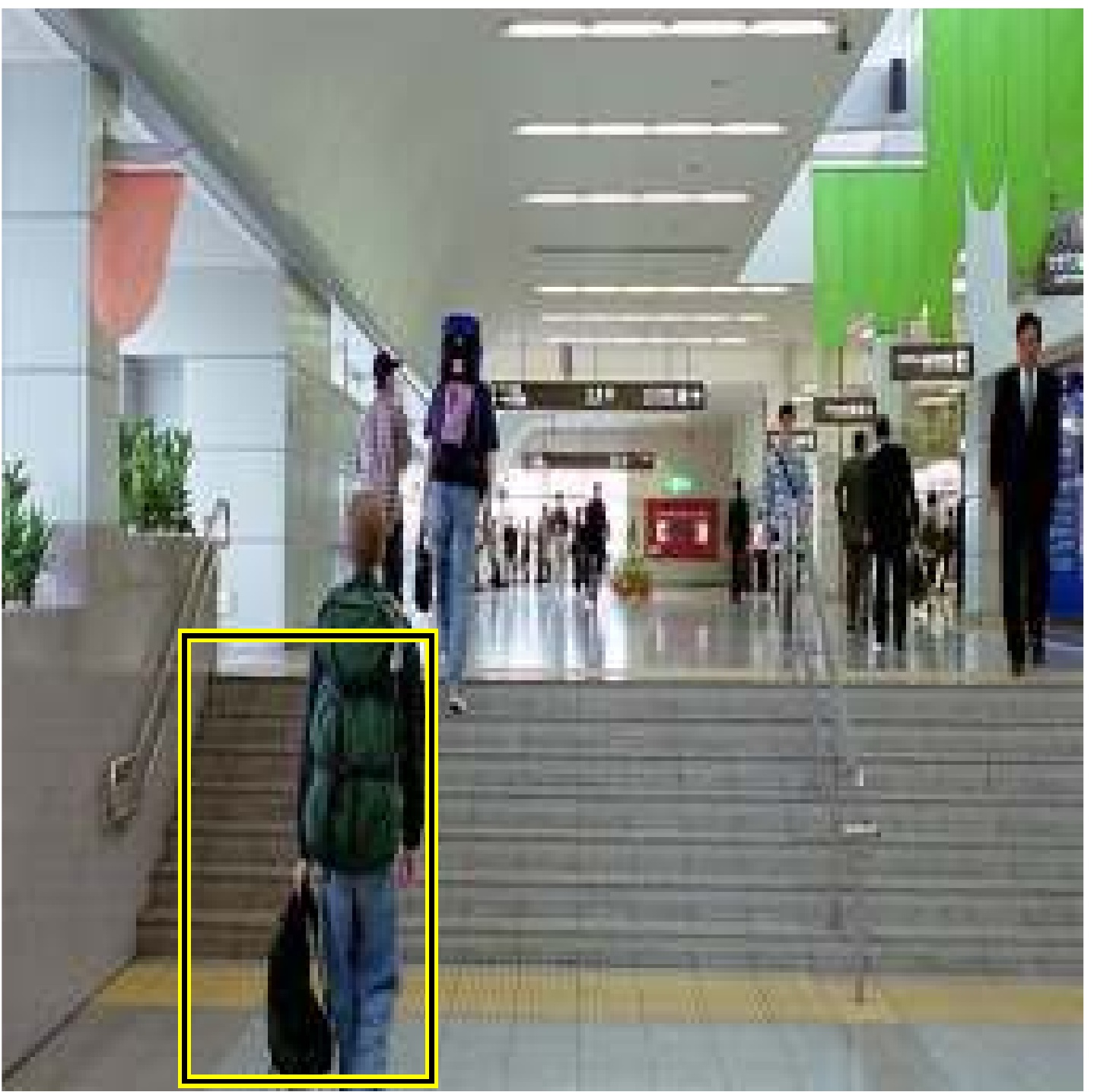} &
				\includegraphics[height=0.63in, width=0.85in]{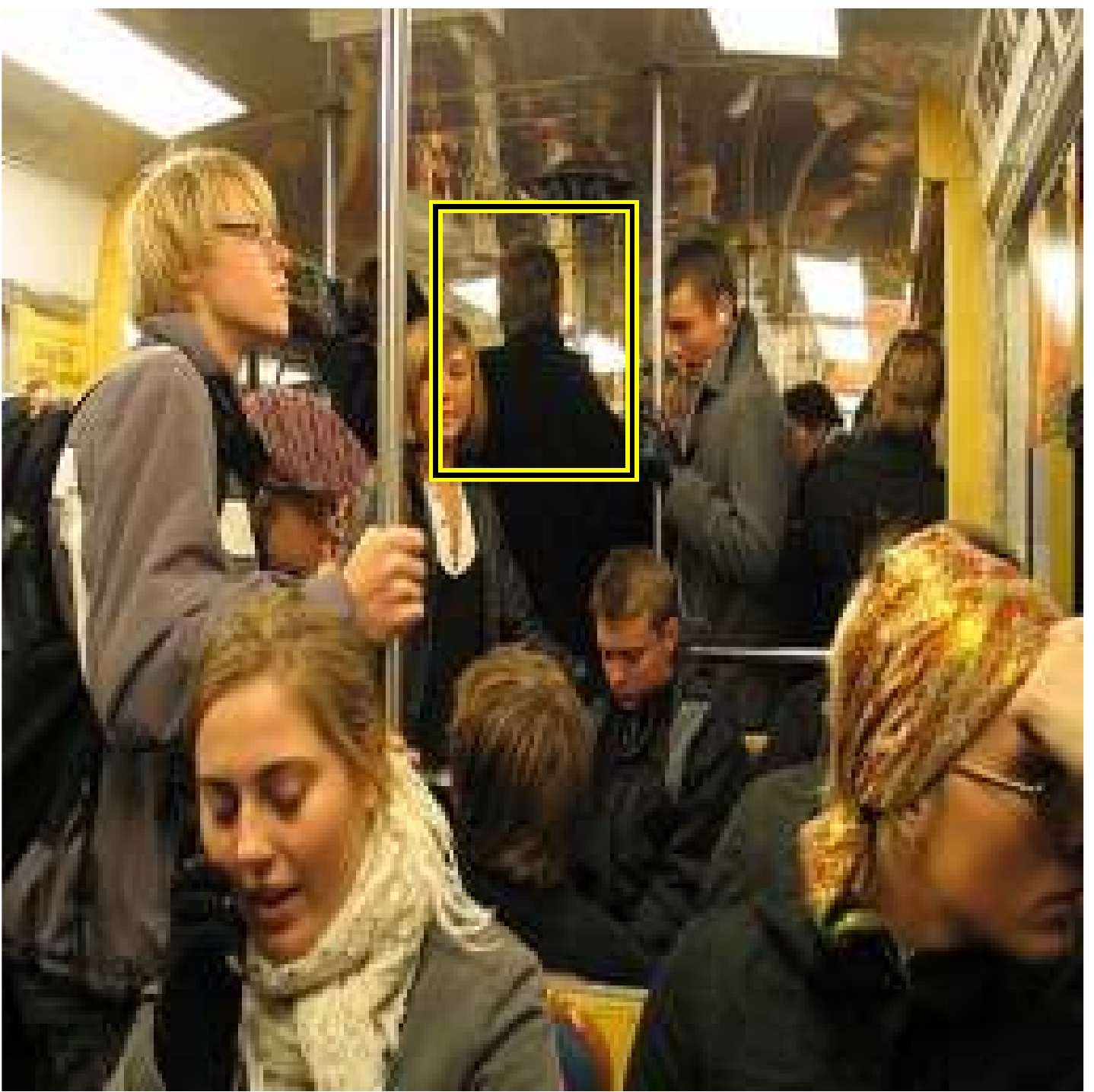} &
				\includegraphics[height=0.63in, width=0.85in]{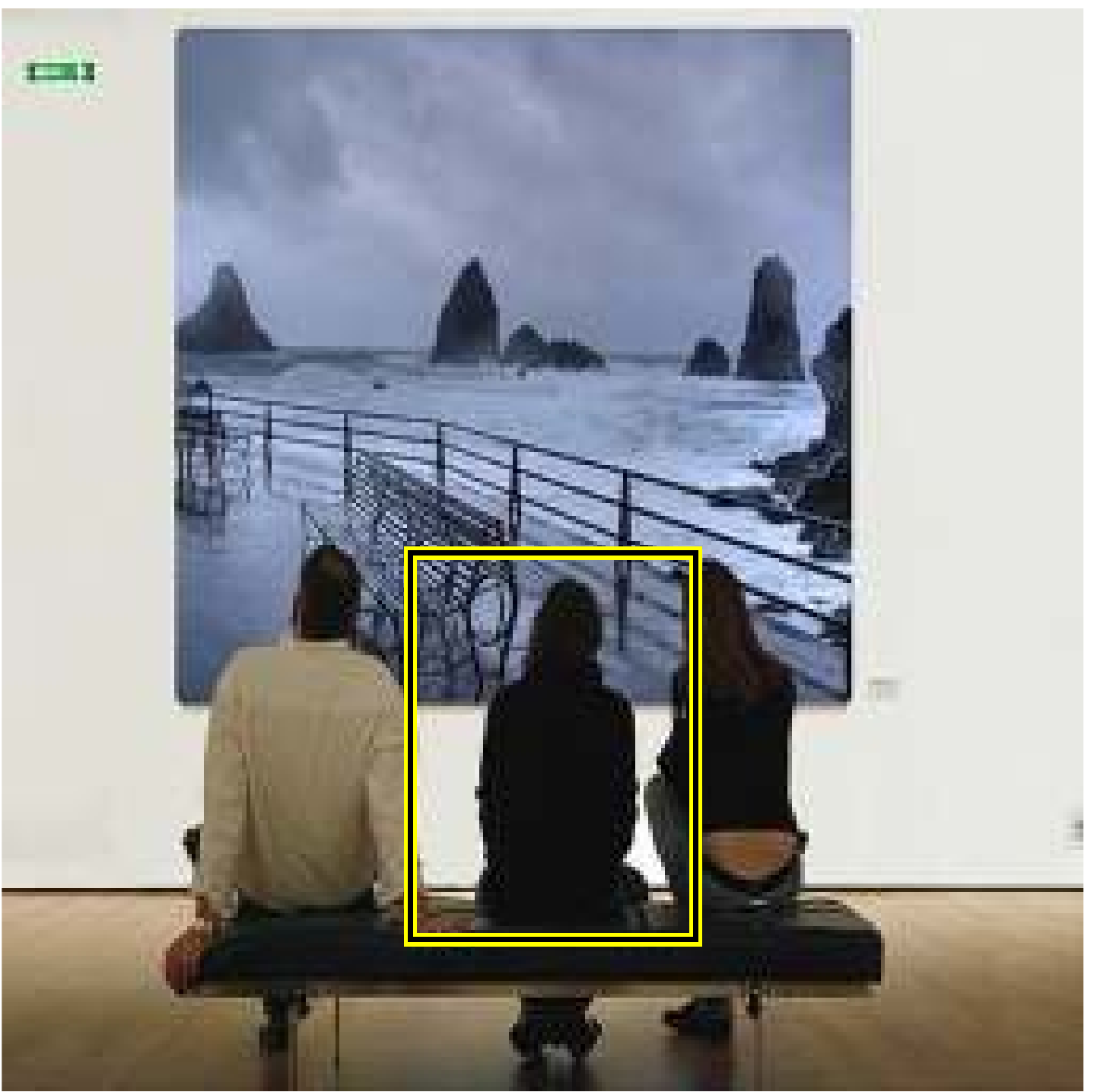} &
				\includegraphics[height=0.63in, width=0.85in]{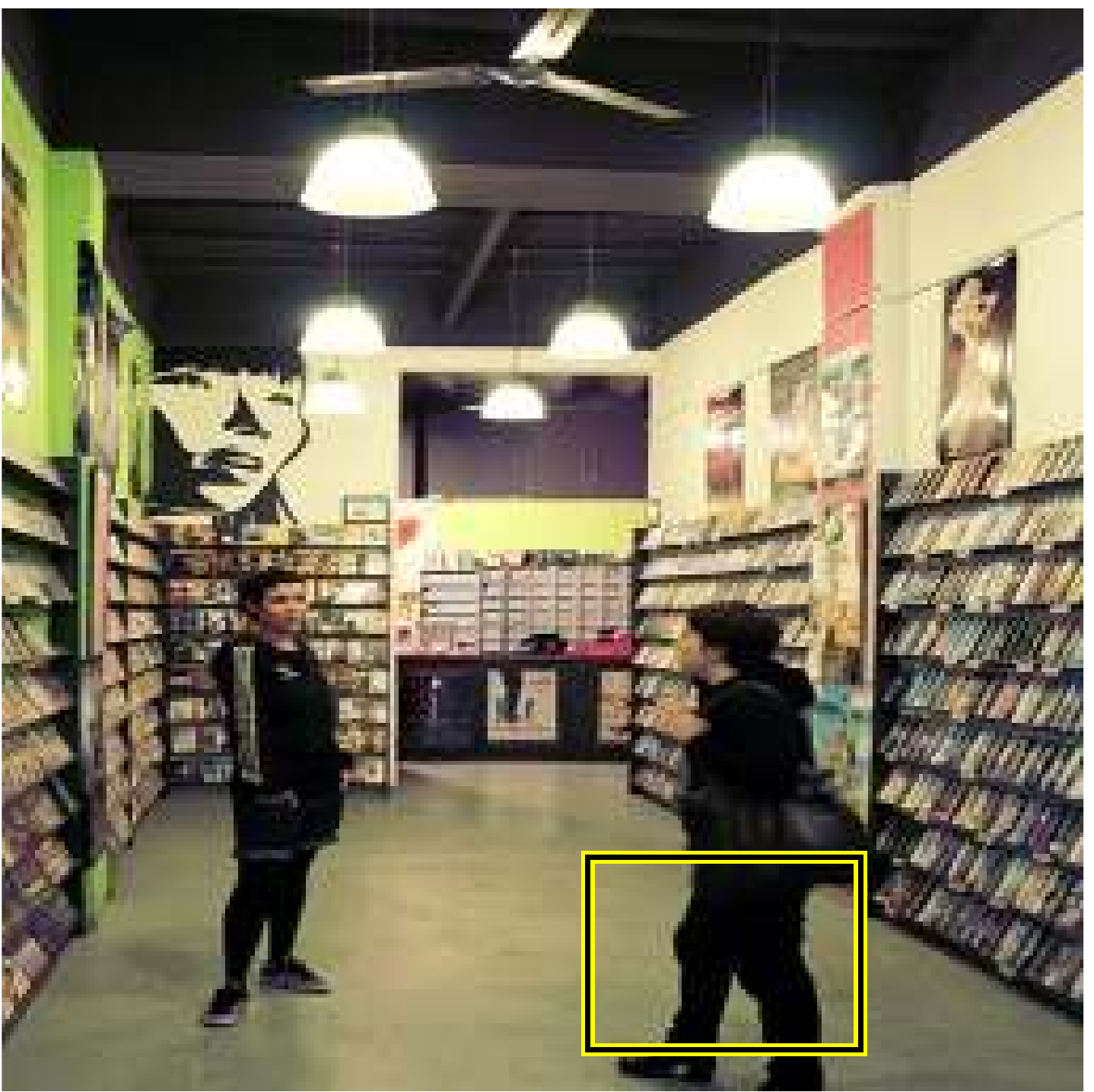} &
				\includegraphics[height=0.63in, width=0.85in]{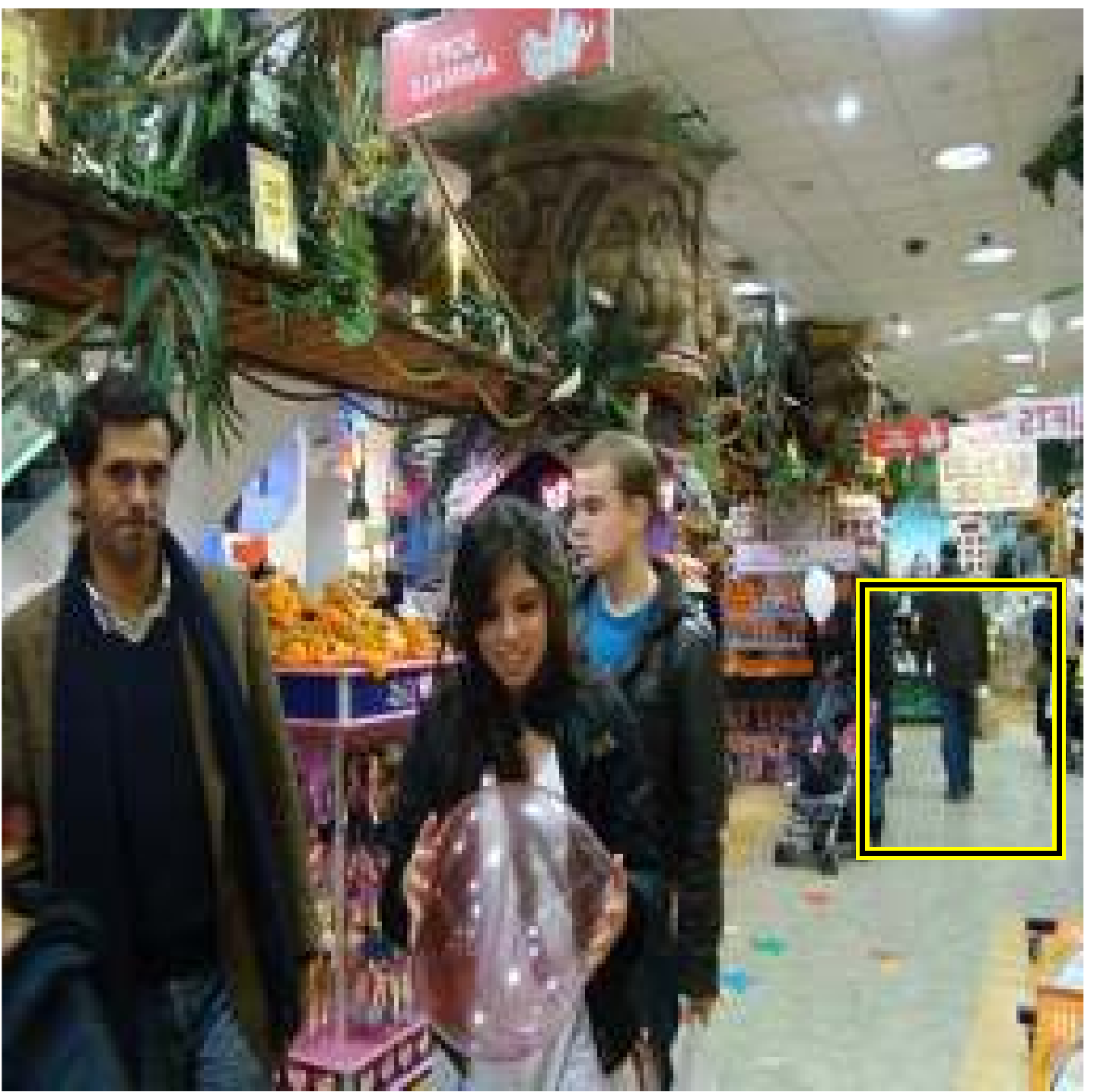} \\ [-0.05cm]
	\rotatebox{90}{\hspace{0.27cm}Part 43}$\;$ &
				\includegraphics[height=0.63in, width=0.85in]{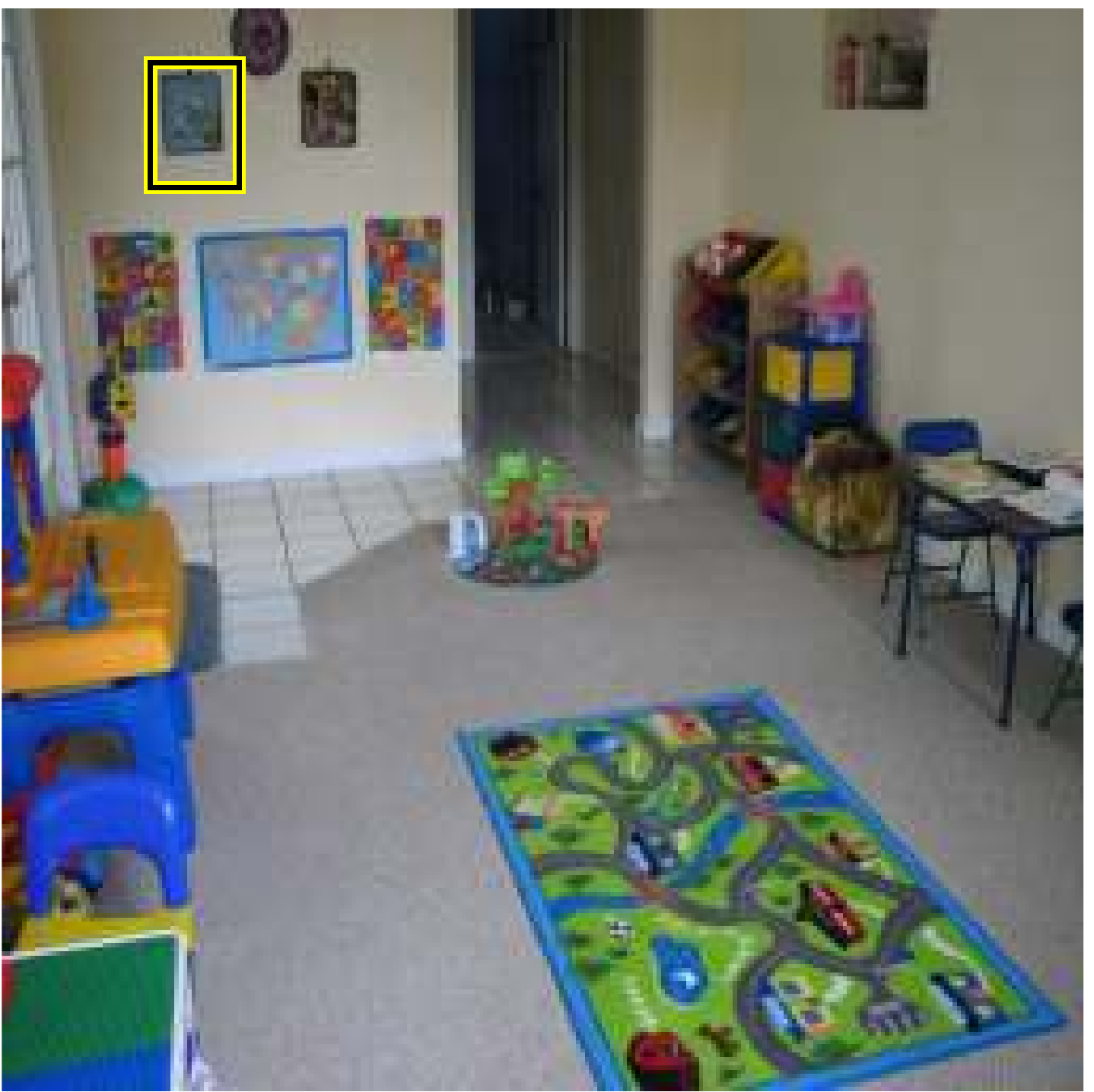} &
				\includegraphics[height=0.63in, width=0.85in]{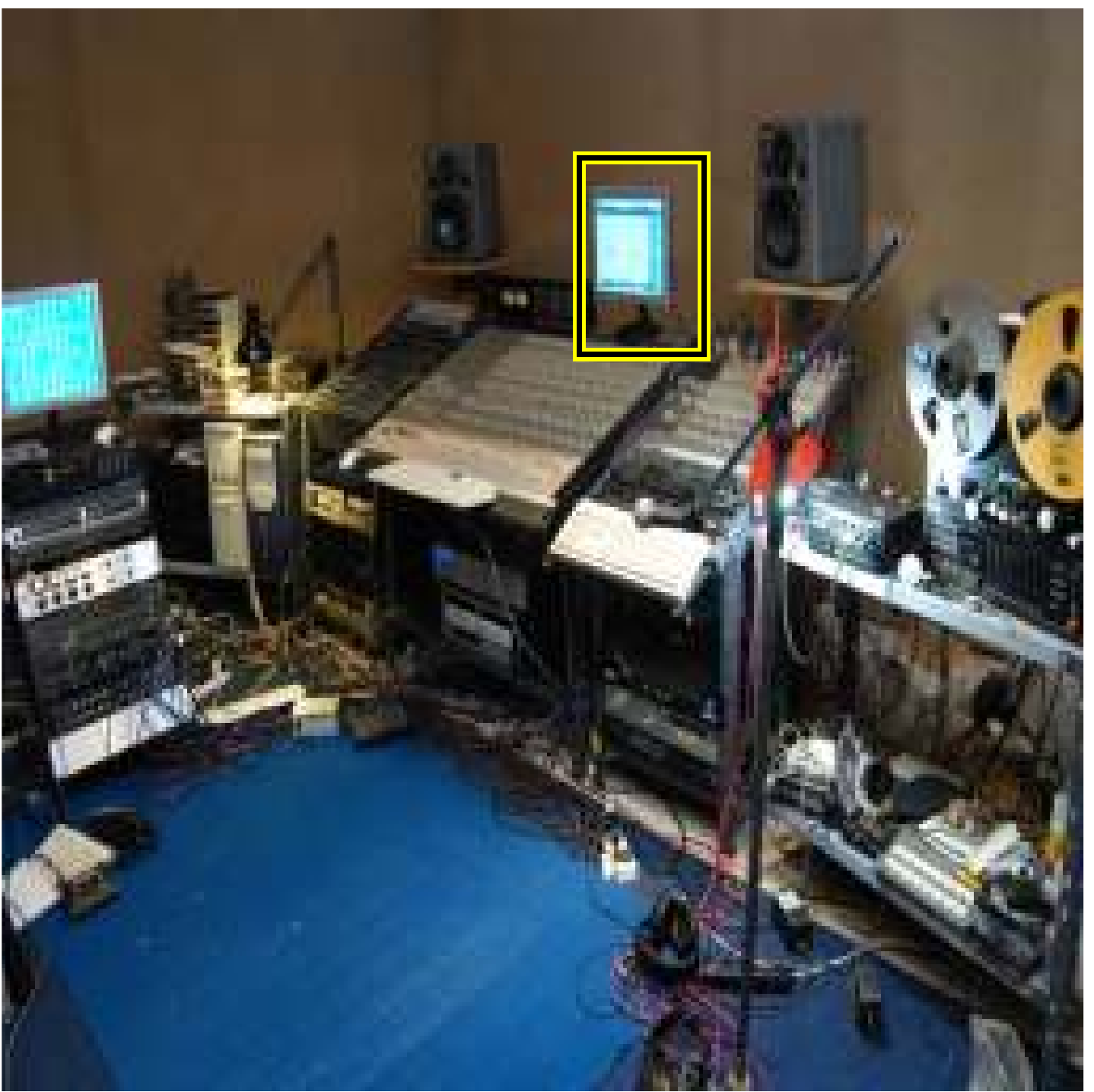} &
				\includegraphics[height=0.63in, width=0.85in]{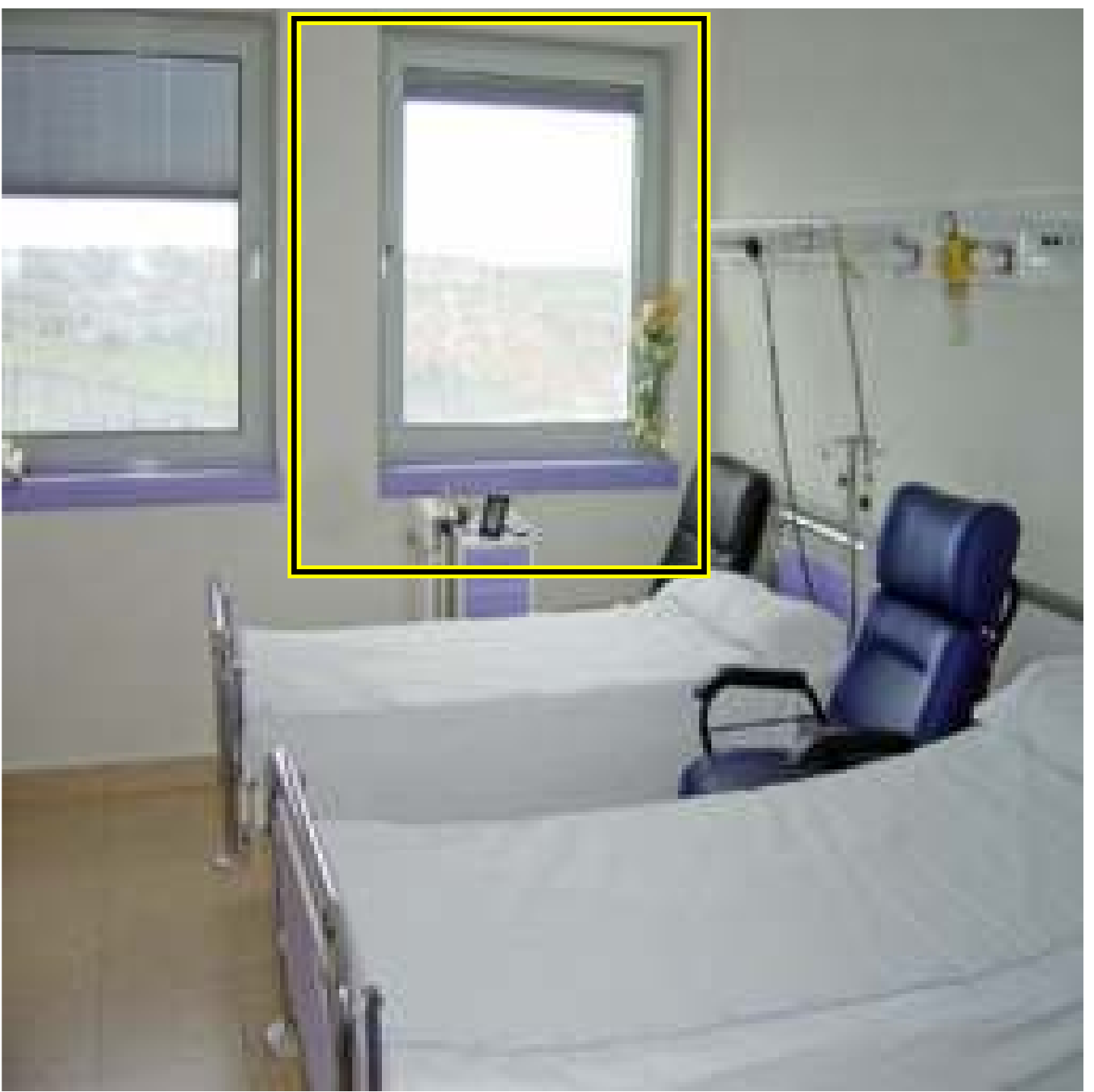} &
				\includegraphics[height=0.63in, width=0.85in]{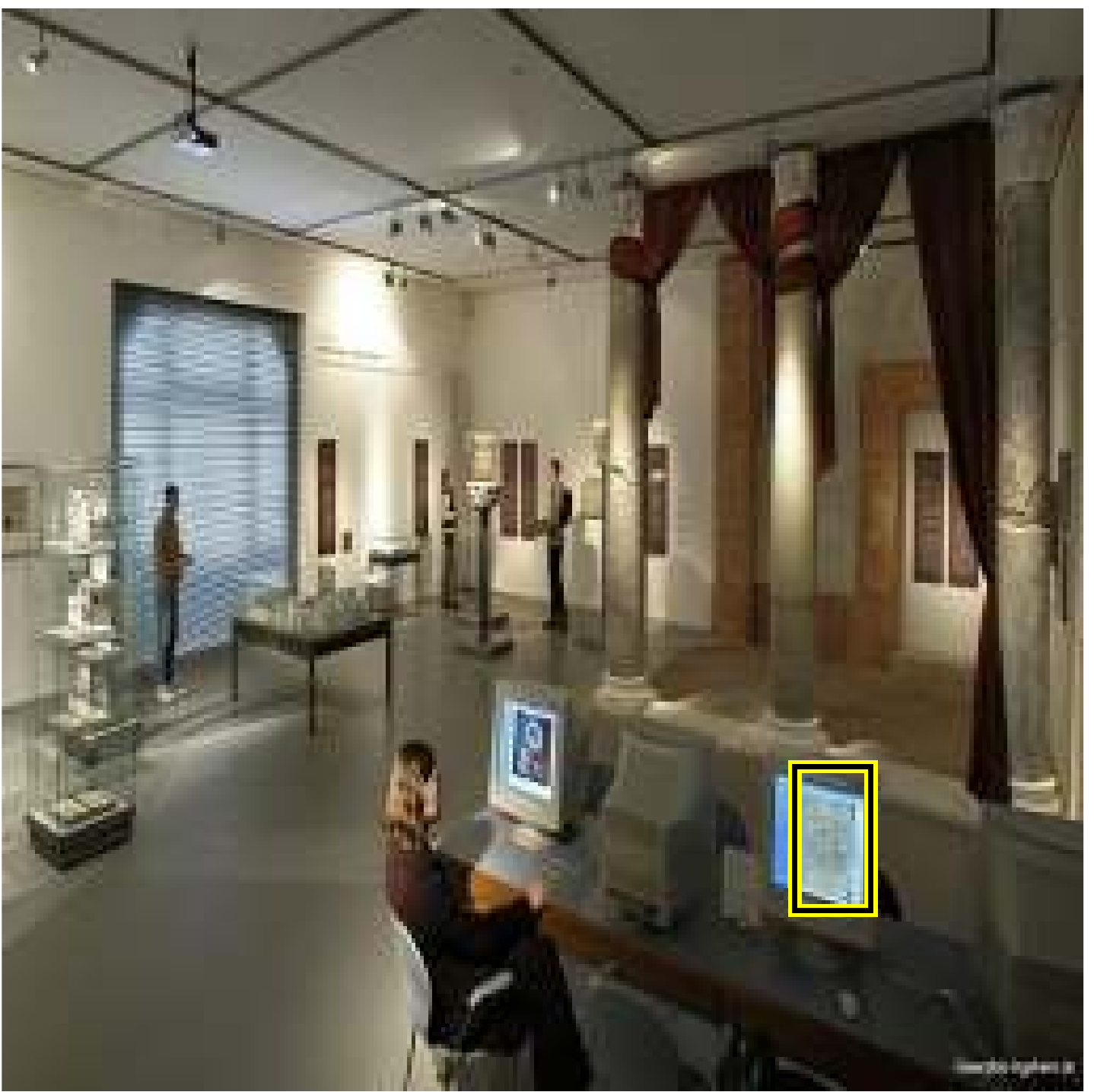} &
				\includegraphics[height=0.63in, width=0.85in]{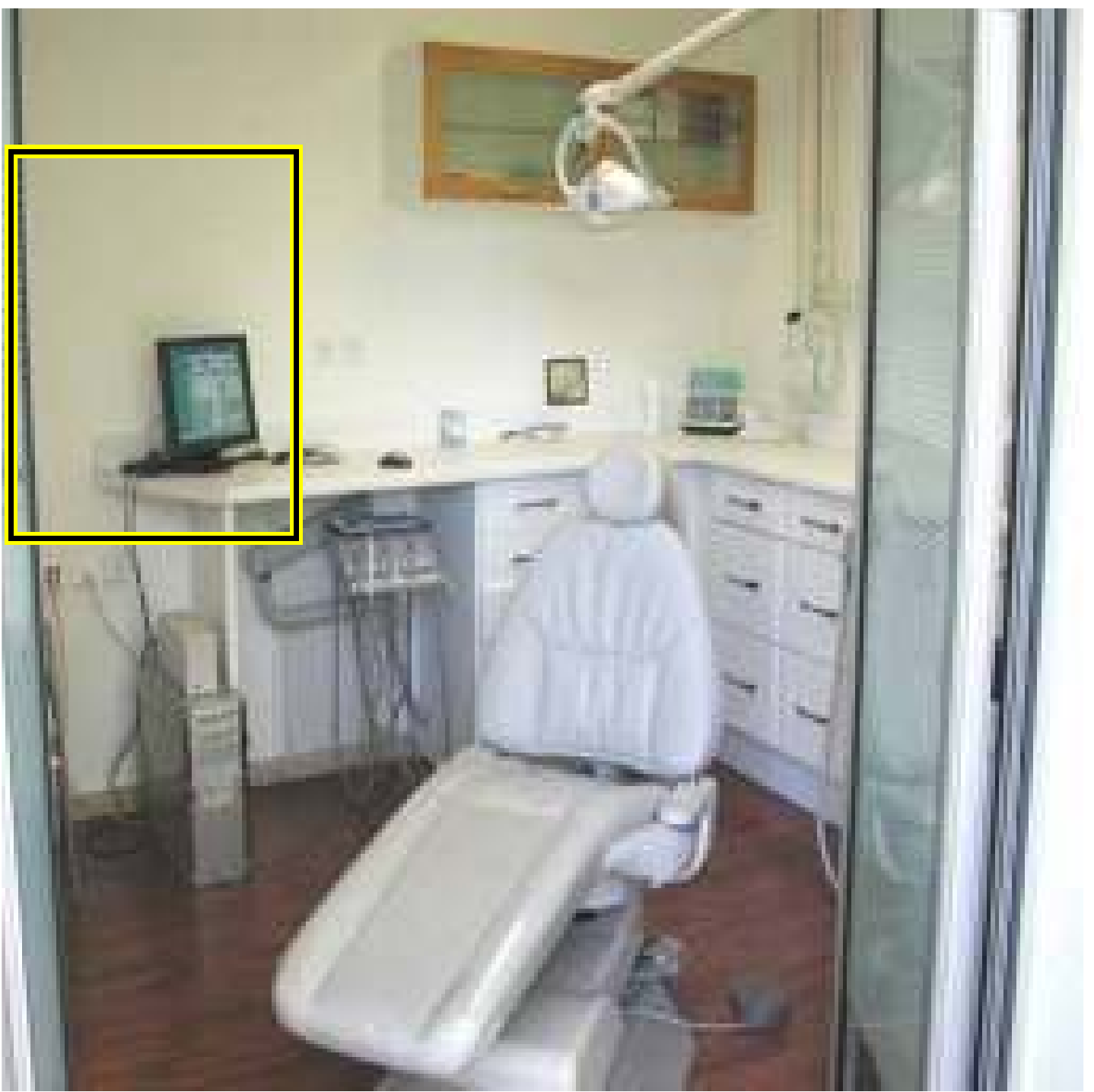} &
				\includegraphics[height=0.63in, width=0.85in]{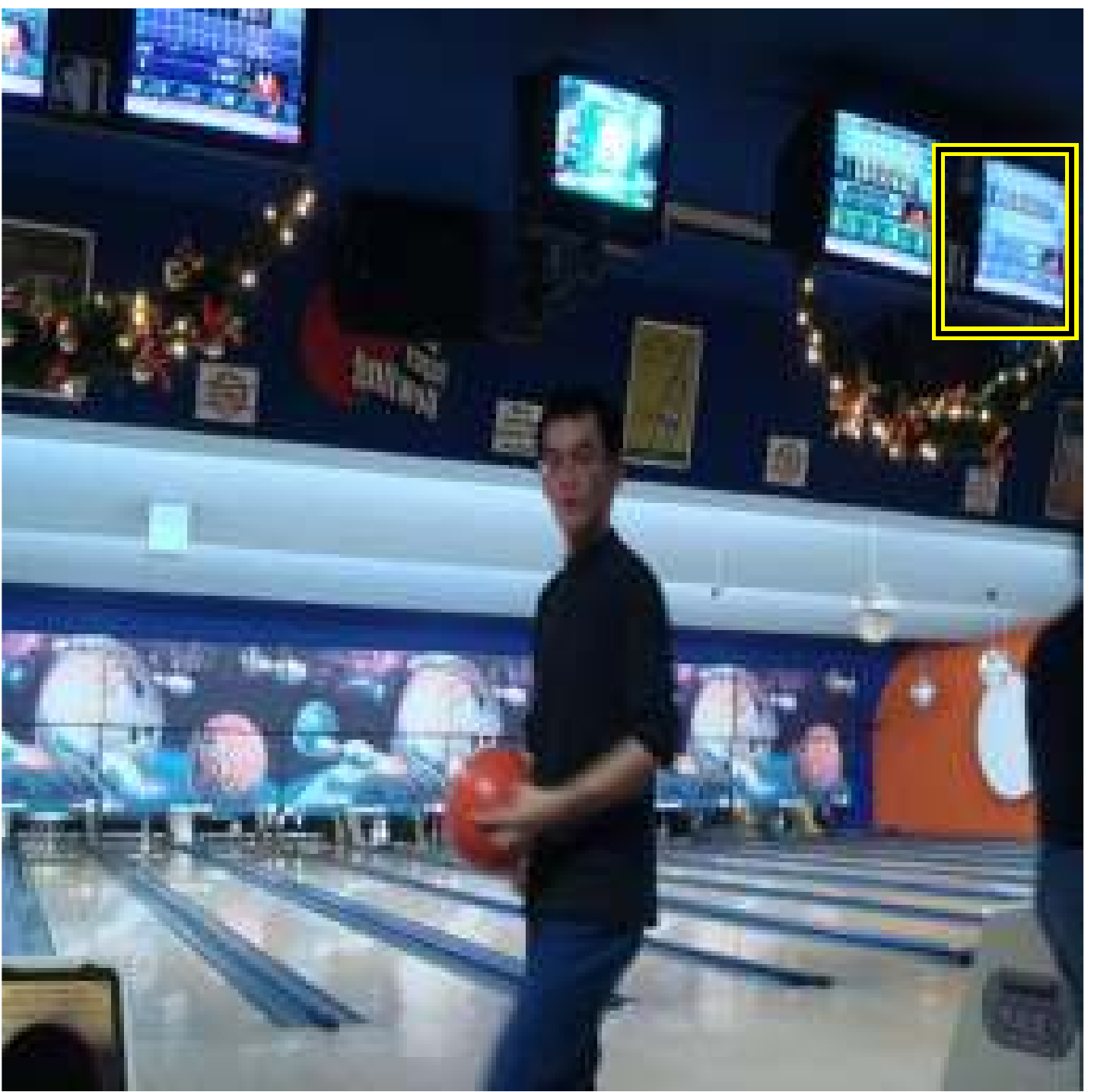} \\ [-0.05cm]
	\rotatebox{90}{\hspace{0.27cm}Part 46}$\;$ &
				\includegraphics[height=0.63in, width=0.85in]{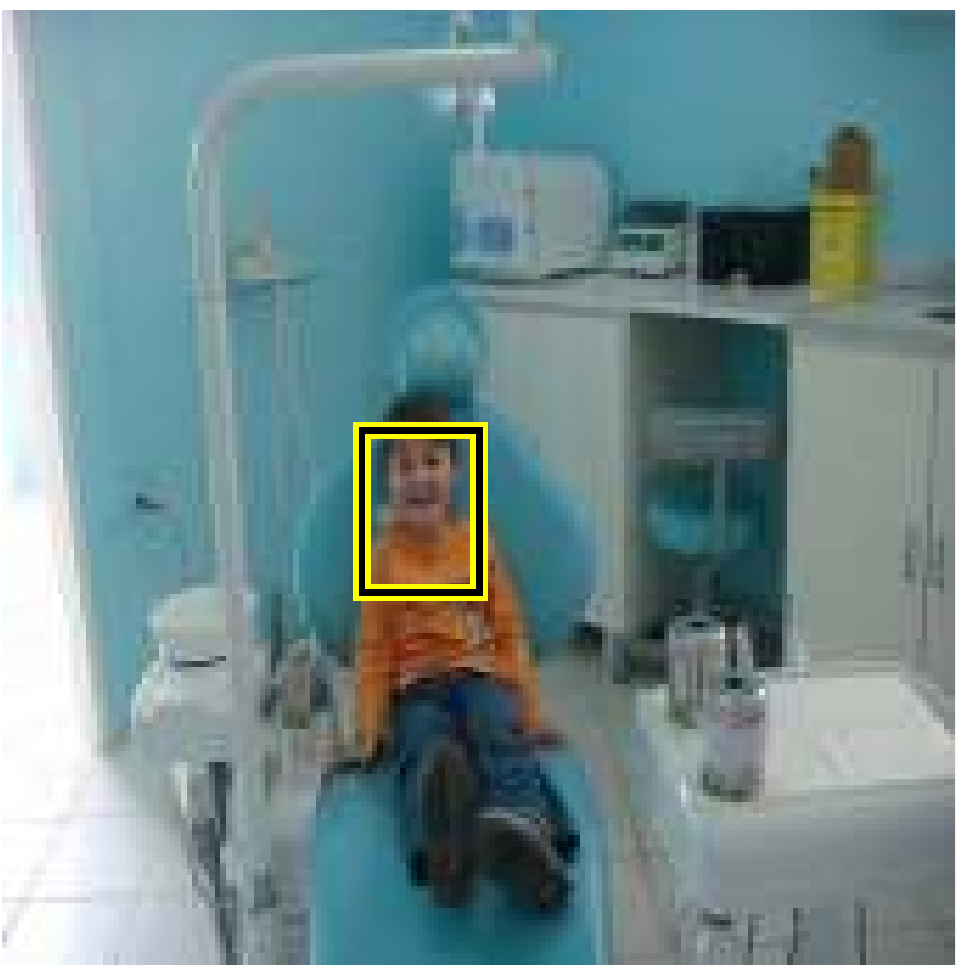} &
				\includegraphics[height=0.63in, width=0.85in]{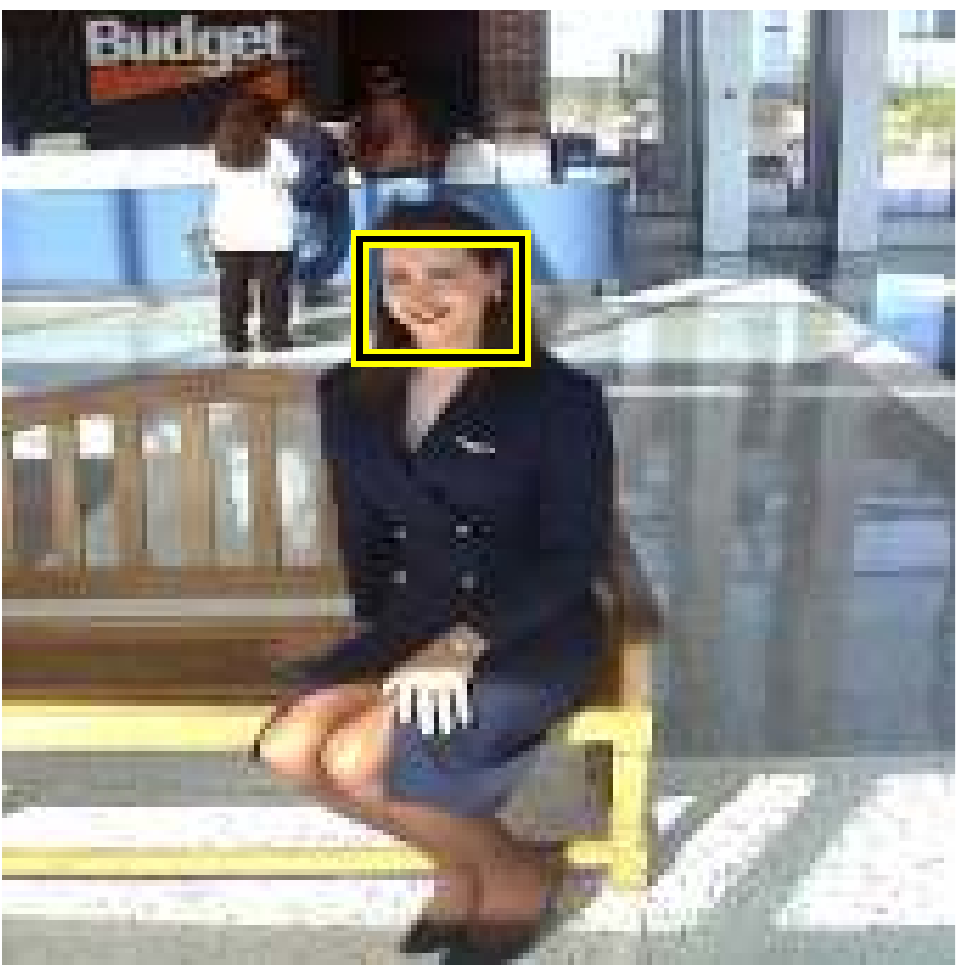} &
				\includegraphics[height=0.63in, width=0.85in]{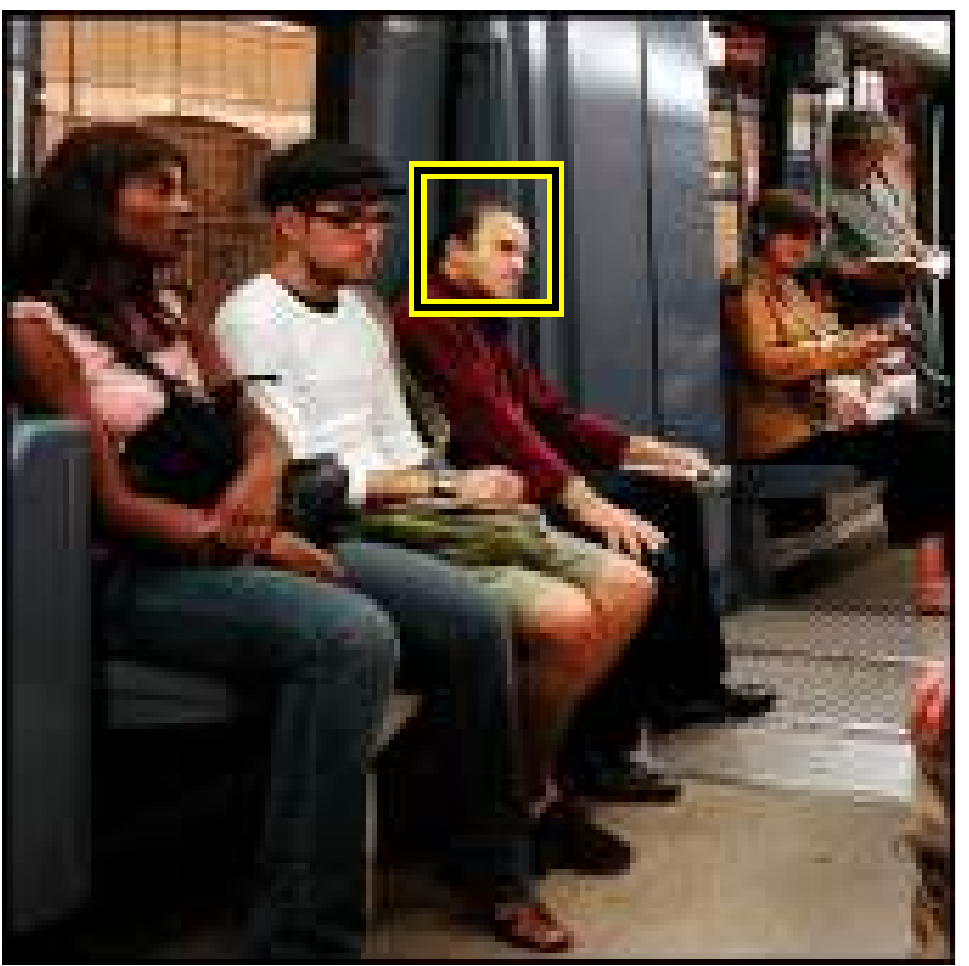} &
				\includegraphics[height=0.63in, width=0.85in]{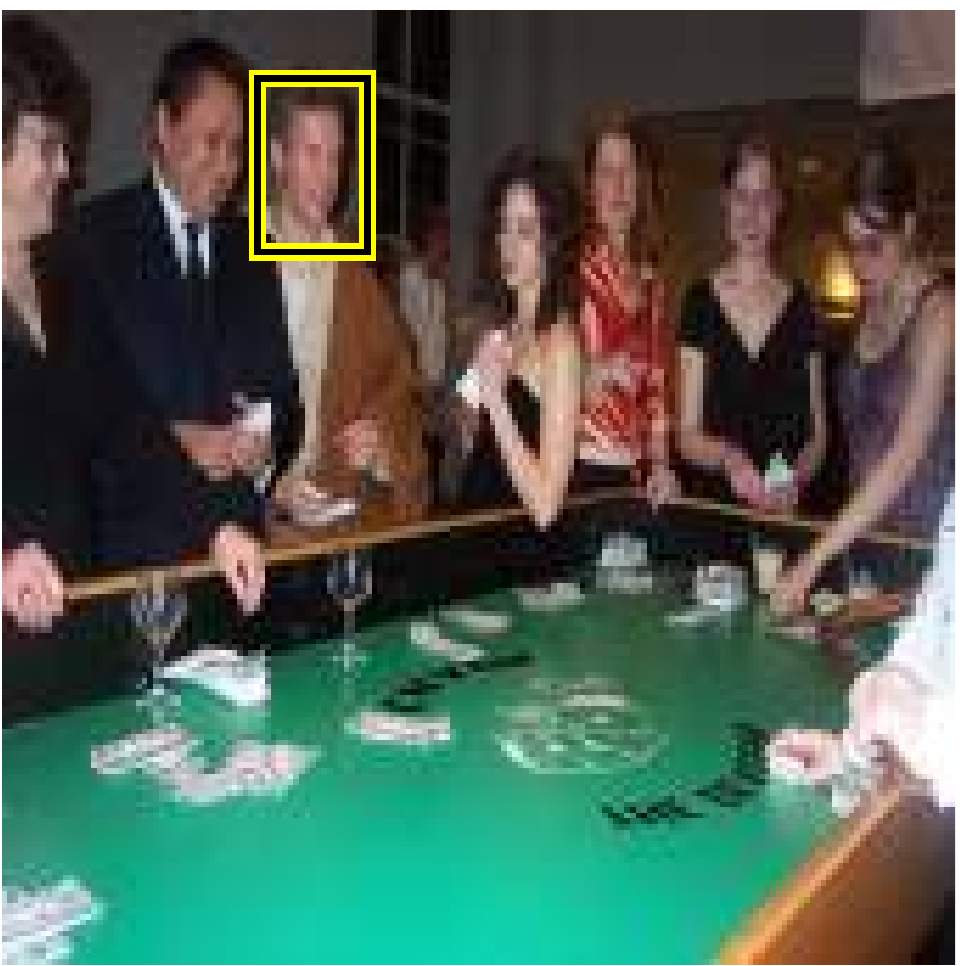} &
				\includegraphics[height=0.63in, width=0.85in]{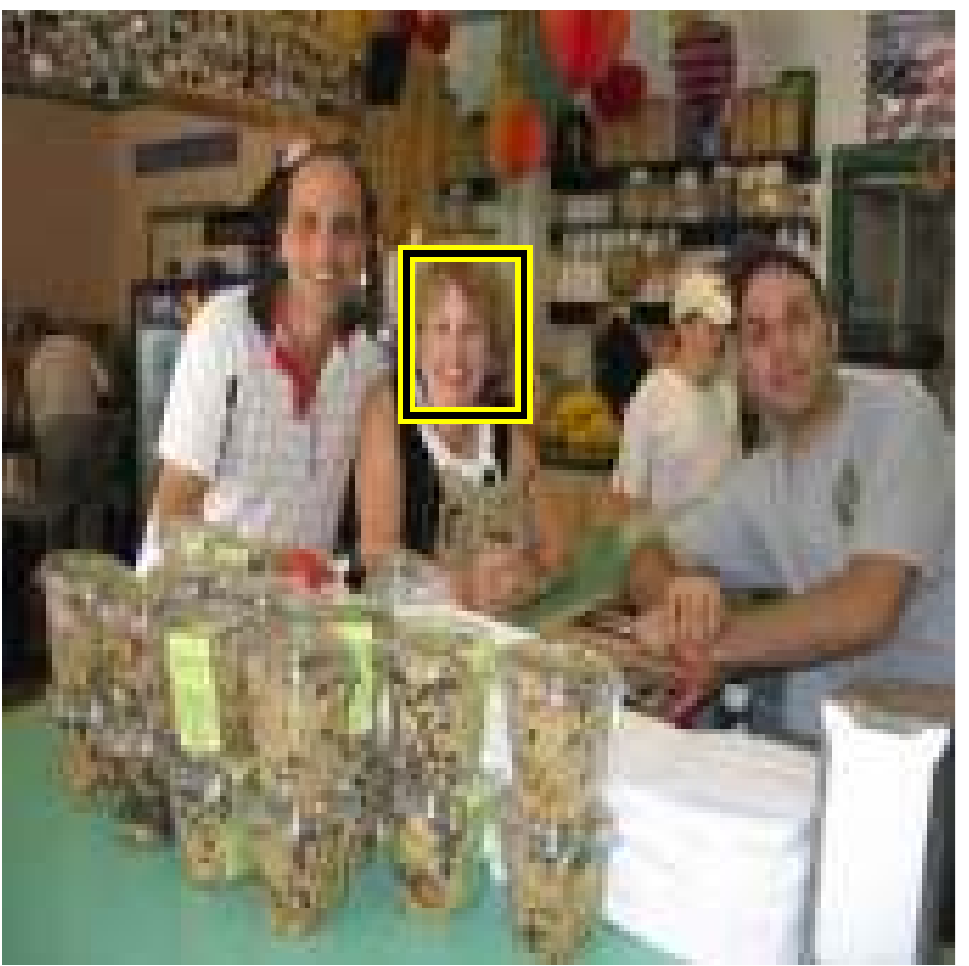} &
				\includegraphics[height=0.63in, width=0.85in]{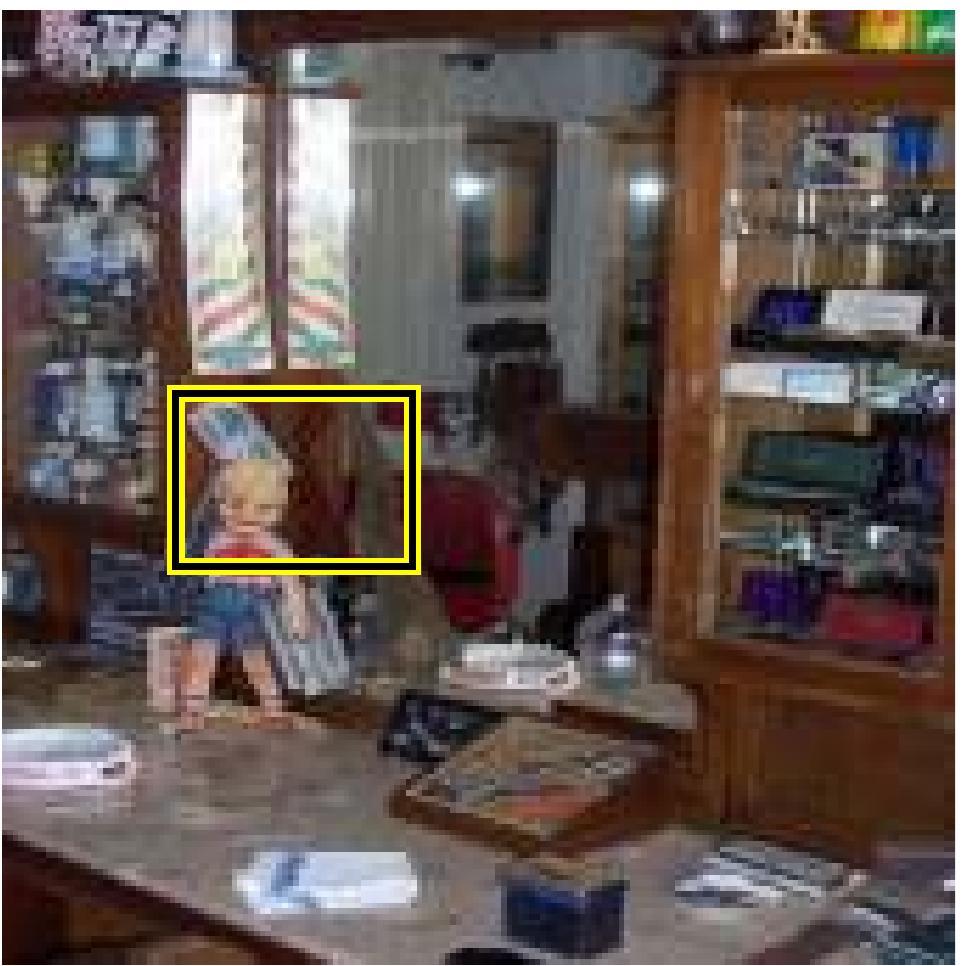}
	\end{tabular}
	\vspace{-1em}
	\caption{Top detections of parts on test images of the full dataset. The numbers in the first column match the part indices in Figure~\ref{fig:u_matrix_full}. Part detection is done in a multi-scale sliding window fashion and using a $\sq{256 \times 256}$ window. For visualization purposes images are stretched to have the same size.}
	\label{fig:top_scoring_patches_CNN}
	\vspace{-1.25em}
\end{figure*}

Figure~\ref{fig:top_scoring_patches_CNN} illustrates selectivity of a few parts. Each row shows the highest scoring detections of a particular part on test images. The part indices in the first column match those of Figure~\ref{fig:u_matrix_full}. Even though most detections look consistently similar the images usually belong to multiple classes demonstrating part sharing across categories. For example, while part 17 appears to capture \emph{bed} the images belong to \emph{hospitalroom}, \emph{childrensroom}, and \emph{bedroom} classes. While part 25 appears to capture \emph{seats} the images belong to \emph{waitingroom}, \emph{library}, \emph{auditorium}, and \emph{insidebus}. Conversely, multiple parts may capture the same semantic concept. For example, parts 3, 16, and 35 appear to capture \emph{shelves} but they seem to be tuned specifically to shelves in \emph{pantry}, \emph{store}, and \emph{book-shelves} respectively. 
%Similarly, parts 20 and 25 both appear to capture \emph{seats} but the former is specific to rows of seats (e.g. in \emph{amphitheater}) whereas the latter responds to a wider range of seat-types. 
%Some parts capture subparts of others. For example, part 22 appears to capture \emph{containers} which are typically found on shelves. 
Some parts respond to a part of an object; \eg part 40 and 46 respond to \emph{leg} and \emph{face}.
Others find entire objects or even composition of multiple objects. For example, parts 6, 17, 37, 43 detect \emph{laundromats}, \emph{beds}, \emph{cabinets}, and \emph{monitor}. Part 29 detects composition of \emph{seats-and-screen}.% in \emph{movietheater} and \emph{auditorium}. %On the other hand, there are parts that capture low-level features such as the mesh pattern of part 31 and the high-frequency horizontal stripes of part 41. Also, there are parts that are selective for certain colors. For example, part 9 appears to be capturing specific \emph{red} patterns (in particular \emph{fruits} and \emph{flowers}). Part 48 is very well tuned to finding \emph{text}. 
%Some parts respond to human body parts. Part 40 appears to capture \emph{human-body} (\emph{legs} in particular). Part 46 appears to be a very accurate \emph{human-face} detector. 

The part weight matrix $u$ (Figure~\ref{fig:u_matrix_full}) helps us better understand how parts assists classification. Part 6 has significantly high weight for class \emph{laundromat} and it appears to be a good detector thereof.
%Part 14 is highly weighted for \emph{nursery} and \emph{staircase} classes and it appears to capture vertical bars. 
Part 27 fires strongly on game/sports-related scenes. The weight matrix reveals that this part is strongly correlated with \emph{gameroom}, \emph{casino}, and \emph{poolinside}. Part 17 fires strongly on \emph{bed} and it has the highest weight for \emph{hospitalroom}, \emph{children\_room}, \emph{bedroom}, and \emph{operating\_room}. 

Weight matrix also identifies negative parts. An interesting example is part 46 (the face detector). 
%One would expect this part to be positive for most classes as humans are usually present in indoor scenes. 
It has the lowest weight for \emph{buffet}, \emph{classroom}, \emph{computerroom}, and \emph{garage}. This suggests that part 46 is a negative part for these classes relative to others.
This is rather surprising because one would expect to find people in scenes such as \emph{classroom} and \emph{computerroom}.
%finding faces in these classes provides counter-evidence for them (hence decreasing their classification score) when compared to any of the other classes.
%Although this may make some sense for the class \emph{garage} but it is counter intuitive for the other three classes. 
We examined all training images of these classes and found no visible faces in them except for 1 image in \emph{classroom} and 3 images in \emph{computerroom} with hardly visible faces and 1 image in \emph{garage} with a clear face in it.

\section{Conclusions}
\label{sec:conclusion}
We presented a simple pipeline to train part-based models for image
classification. All model parameters are trained jointly in our framework; 
this includes shared part filters and class-specific part weights.
All stages of our training pipeline are driven \emph{directly} 
by the same objective namely the classification performance on a training set. 
In particular, our framework does not rely on ad-hoc heuristics for selecting
discriminative and/or diverse parts.
We also introduced the concept of ``negative parts" for part-based models.

Models based on our randomly generated parts perform better than 
almost all previously published work despite the profound simplicity of the method.
%(\cite{doersch13} being the only exception when HOG features are used).
Using CNN features and random parts we obtain $77.1\%$ accuracy on the MIT-indoor dataset,
improving the state-of-the-art.
We also showed that part selection and joint training can be used to
train a model that achieves better or the same level of performance as a 
system with randomly generated parts while using much fewer parts.

Joint training alternates between training part weights and updating part filters. 
This process can be initiated before the first or the second step leading to 
two different initialization schemes.  Currently we use random examples to initialize the part filters.  It would also be possible
to initialize the entries in $u$ based on how a
hypothetical part is correlated with a class; 
negatively, irrelevant, or positively.  Training the part filters would then
learn part models that fit this description.

%Though it may be hard to find good initial parts, 
%it can be easy to initialize the part weight matrix. 
%For example, one can fill entries in $u$ with $-1$, $0$, or $+1$ 
%based on how a

%This information may be obtained, for example, manually or by crowd sourcing. 
%This is an interesting direction for future work. 
%Other methods like~\cite{juneja13,doersch13} are bound to be started from initialized parts.

%We plan to make the joint training procedure faster by implementing it
%in GPU in future. This allows us to train models with a few thousand 
%parts for classification tasks with a few hundred categories.
%\clearpage

\bibliographystyle{iclr2015}
\bibliography{bibfile}
\clearpage

\appendix
\section*{Appendix}
\section{Notes on Optimization of the Bound $B_u$}
\label{sec:notes_on_optimization_supplement}
The joint training procedure outlined in Section~\ref{sec:joint_training} is computationally expensive 
because of two reasons. Firstly, joint training involves optimizing the model parameters for all 
categories simultaneously. This includes all the shared part filters as well as all class-specific 
part weights. Secondly, learning part filters requires convolving them with training images repeatedly 
which is a slow process. Similar to (\cite{dpm}) we use a caching mechanism to make this process tractable 
(Section~\ref{sec:caching_supplement}). It works by breaking the problem of training parts into a series of 
smaller optimization problems. The solution to each sub-problem optimizes the part filters on a limited set 
of candidate locations (instead of all possible locations in a scale pyramid). To expedite optimization of 
the parts over a cache even further we use the cutting-plane method (Section~\ref{sec:cutting_plane_supplement}).

\subsection{Caching Hard Examples}
\label{sec:caching_supplement}
We copy the bound $B_u$ from~(\ref{eq:np_objective_function_bound}) here:
\begin{align}
	\hspace{-0.2cm}
	\sq{B_u(w, w^{\text{old}}) = \lambda_w ||w||^2 
	+ \sum_{i=1}^k \max\left\{0,\ %
	1 + \max_{y \neq y_i} (u_y - u_{y_i}) \cdot
	\left[ 
	S_{y, y_i} r(x_i, w) + 
	\bar{S}_{y, y_i} s(x_i, z_i, w)
	\right]
	\right\}}
\end{align}
Recall that $S_{y, y'}$ and 
$\bar{S}_{y, y'}$ are $\sq{m \times m}$ diagonal 0-1 matrices 
that select $s(x_i, z_{i,j}, w_j)$ when $\sq{u_{y, j} - u_{y_i, j} < 0}$ and $r(x_i, w_j)$ otherwise.
The two functions $r$ and $s$ are defined in~(\ref{eq:part-scores}).

Minimizing $B_u(w, w^{old})$ over $w$ requires repeatedly computing the vector of filter responses $r(x_i, w)$ from the entire scale hierarchy of each image which is very expensive; see (\ref{eq:part-scores}). To make the minimization of $B_u(w, w^{old})$ tractable we use an iterative ``caching" mechanism. In each iteration, we update the content of the cache and find the optimal $w$ subject to the data in the cache. This procedure is guaranteed to converge to the \emph{global} minimum of $B_u(w, w^{old})$. For each part, the cache stores only a few number of \emph{active} part locations from the scale hierarchy of each image. Thus, finding the highest responding location for each part among the active entries in the cache requires only a modest amount of computation. 

Note that here, unlike (\cite{dpm}), there is no notion of hard \emph{negatives} because we use a multi-class classification objective. Instead we have hard \emph{examples}. A hard example is a training example along with the best assignment of its latent variables (with respect to a given model) that either has non-zero loss or lies on the decision boundary. In the following we explain our caching mechanism and prove it converges to the unique global minimum of the bound.

Let $\sq{Z_i= \{(y, \plc) : y \in \mathcal{Y}, \plc \in H(x_i)^m\}}$ be the set of all possible latent configurations of $m$ parts on image $x_i$. Also let $\Phi(x, \plc, \bar{\plc}, a, \bar{a}) = (a_1 \psi(x, \plc_1) + \bar{a}_1 \psi(x, \bar{\plc}_1); \dots; a_m \psi(x, \plc_m) + \bar{a}_m \psi(x, \bar{\plc}_m))$ be some features extracted from placed parts. The feature function takes in an image $x$, two part placement vectors $\plc, \bar{\plc} \in H(x)^m$, and two $m$-dimensional part weight vectors $a, \bar{a}$ as input and outputs an $md$-dimensional feature vector.  Define
$$a_{y, y'}=(u_y - u_{y_i})^T \bar{S}_{y, y_i}$$
$$\bar{a}_{y, y'}=(u_y - u_{y_i})^T \bar{S}_{y, y_i}$$ Note that if $a_{y, y', j} \neq 0$ then $\bar{a}_{y, y', j} = 0$ and vice versa. Finally, we define $\plc^{(i, w)}_j = \argmax_{z_j \in H(x_i)} s(x_i, z_j, w_j)$ to be the best placement of part $j$ on image $x_i$ using the part filter defined by $w$. We use this notation and rewrite $B_u(w, w^{old})$ as follows,
\begin{align}
	B_u(w, w^{old}) &= \lambda_w ||w||^2 + \sum_{i=1}^k \max_{(y, \plc) \in Z_i} w^T \Phi(x_i, \plc, \plc^{(i, w^{old})}, a_{y, y_i}, \bar{a}_{y, y_i})) + \Delta(y_i, y)
\end{align}
When optimizing $B_u(w, w^{old})$, we define a cache $C$ to be a set of triplets $(i, f, \delta)$ where $i$ indicates the $i$-th training example and $f$ and $\delta$ indicate the feature vector $\Phi(x_i, \plc, \plc^{(i, w^{old})}, a_{y, y_i}, \bar{a}_{y, y_i})$ and the loss value $\Delta(y_i, y)$ associated to a particular $(y, \plc) \in Z_i$ respectively. The bound $B_u$ with respect to a cache $C$ can be written as follows: 
\begin{align}
	\mathcal{B}_C(w) = \mathcal{B}_C(w; u, w^{old}) = \lambda_w ||w||^2 + \sum_{i=1}^k \max_{(i, f, \delta) \in C} w^T f + \delta
	\label{eq:of_cache}
\end{align}
Note that $\mathcal{B}_{C_A}(w; u, w^{old}) = B_u(w, w^{old})$ when $C_A$ includes all possible latent configurations; that is $\sq{C_A=\{(i, f, \delta) | \forall i \in \{1, \dots, k\}, \forall (y, \plc) \in Z_i, f=\Phi(x_i, \plc, \plc^{(i, w^{old})}, a_{y, y_i}, \bar{a}_{y, y_i}), \delta=\Delta(y_i, y)\}}$.

We denote the set of \emph{hard} and \emph{easy} examples of a dataset $\mathcal{D}$ with respect to $w$ and $w^{old}$ by $\mathcal{H}(w, w^{old}, \mathcal{D})$ and $\mathcal{E}(w, w^{old}, \mathcal{D})$, respectively, and define them as follows:
\begin{align}
	\mathcal{H}(w, w^{old}, \mathcal{D}) =& \{(i, \Phi(x_i, \plc, \plc^{(i, w^{old})}, a_{y, y_i}, \bar{a}_{y, y_i}), \Delta(y_i, y)) | 1 \leq i \leq k, \notag \\
	& (y, \plc) = \argmax_{(\hat{y}, \hat{\plc}) \in Z_i} w^T \Phi(x_i, \hat{\plc}, \plc^{(i, w^{old})}, a_{\hat{y}, y_i}, \bar{a}_{\hat{y}, y_i}) + \Delta(y_i, \hat{y})\} \label{eq:np_hard_examples} \\
	\mathcal{E}(w, w^{old}, \mathcal{D}) =& \{(i, \Phi(x_i, \plc, \plc^{(i, w^{old})}, a_{y, y_i}, \bar{a}_{y, y_i}), \Delta(y_i, y)) | 1 \leq i \leq k, (y, \plc) \in Z_i, \notag \\
	& w^T \Phi(x_i, \plc, \plc^{(i, w^{old})}, a_{y, y_i}, \bar{a}_{y, y_i}) + \Delta(y_i, y) < 0\}
\end{align}
Note that if $y=y_i$  the term in the $\argmax$ in (\ref{eq:np_hard_examples}) is zero, regardless of the value of $\hat{\plc}$. That is because $\forall y \in \mathcal{Y}: a_{y, y} = \bar{a}_{y, y} = \textbf{0}$. So, $w^T f + \delta$ is non-negative for any $(i, f, \delta) \in \mathcal{H}(w, w^{old}, \mathcal{D})$. 

We use the caching procedure outlined in Algorithm~\ref{alg:np_cache} to optimize the bound $B_u$. The benefit of this algorithm is that instead of direct optimization of $B_u$ (which is quickly becomes intractable as the size of the problem gets large) it solves several tractable auxiliary optimization problems (line 7 of Algorithm \ref{alg:np_cache}).

The algorithm starts with the initial cache $C_0 = \{(i, \textbf{0}, 0) | 1 \leq i \leq k \}$ where $\textbf{0}$ is the all-zero vector. This corresponds to the set of correct classification hypotheses; one for each training image.
It then alternates between updating the cache and finding the $w^*$ that minimizes $\mathcal{B}_C$ until the cache does not change. To update the cache we remove all the easy examples and add new hard examples to it (lines 5,6 of Algorithm \ref{alg:np_cache}). Note that $C_0 \subseteq C$ at all times.
\begin{algorithm}[t]
\begin{algor}[1]
	\item [{*}] Input: $w^{old}$
	\item [{*}] $C_0 := \text{\{}(i, \textbf{0}, 0) | 1 \leq i \leq k \text{\}}$
	\item [{*}] $t := 0$
	\item [while] $\mathcal{H}(w^t, w^{old}, \mathcal{D}) \not \subseteq C_t$
		\item [{*}] $C_t := C_t \setminus \mathcal{E}(w^t, w^{old}, \mathcal{D})$ \label{ln:alg:shrink}
		\item [{*}] $C_t := C_t \cup \mathcal{H}(w^t, w^{old}, \mathcal{D})$ \label{ln:alg:grow}
		\item [{*}] $w^{t+1} := \argmin_w \mathcal{B}_{C_t}(w)$ \label{ln:alg:optimize}
		\item [{*}] $t := t+1$
	\item [endwhile]
	\item [{*}] output $w^t$
\end{algor}
\caption{\label{alg:np_cache} Fast optimization of the convex bound $B_u(w, w^{old})$ using hard example mining}
\end{algorithm}

Depending on the value of $\lambda_w$, in practice, Algorithm~\ref{alg:np_cache} may take up to 10 iterations to converge. However, one can save most of these cache-update rounds by retaining the cache content after convergence and using it to warm-start the next call to the algorithm. With this trick, except for the first call, Algorithm~\ref{alg:np_cache} takes only 2-3 rounds to converge. The reason is that many cache entries remain active even after $w^{old}$ is updated; this happens, in particular, as we get close to the last iterations of the joint training loop (lines 2-8 of Algorithm~\ref{alg:np_optimization}). Note that to employ this trick one has to modify the feature values (i.e. the $f$ field in the triplets $(i, f, \delta)$) of the entries in the retained cache according to the updated value of $w^{old}$.

The following theorem shows that the caching mechanism of Algorithm \ref{alg:np_cache} works; meaning that it converges to $w^* = \argmin_{w'} B_u(w', w^{old})$ for any value of $u, w^{old}$.

\begin{thrm}
	The caching mechanism converges to $w^* = \argmin_{w'} B_u(w', w^{old})$.
	\begin{proof}
		Let $C_A$ be the cache that contains \emph{all} possible latent configurations on $\mathcal{D}$. Assume that Algorithm~\ref{alg:np_cache} converges, after $T$ iterations, to $w^\dagger=\argmin_w \mathcal{B}_{C_T}(w)$. Then since the algorithm converged $C_A \setminus C_T \subseteq \mathcal{E}(w^\dagger, w^{old}, \mathcal{D})$. Consider a small ball around $w^\dagger$ such that for any $w$ in this ball $\mathcal{H}(w, w^{old}, \mathcal{D}) \subseteq C_T$. The two functions $\mathcal{B}_{C_A}(w)$ and $B_u(w, w^{old})$ are equal in this ball and $w^\dagger$ is a local minimum \emph{inside} this region. This implies that $w^\dagger$ is the global minimum of $B_u(w, w^{old})$ because it is a strictly convex function and therefore has a unique local minimum. To complete the proof we only need to show that the algorithm does in fact converge. The key idea is to note that the algorithm does not visit the same cache more than once. So, it has to converge in a finite number of iterations because the number of possible caches is finite.
	\end{proof}
\end{thrm}

\subsection{Optimizing $\mathcal{B}_C$ via Cutting-Plane Method}
\label{sec:cutting_plane_supplement}
In this section we review a fast method that implements line 7 of Algorithm~\ref{alg:np_cache}. The goal is to solve the following optimization problem:
\begin{align}
	w^*_C &= \argmin_w \mathcal{B}_C(w) \notag \\
	&= \argmin_w \lambda_w ||w||^2 + \sum_{i=1}^k \max_{(i,f, \delta) \in C} w^T f + \delta \label{eq:op_cache}
\end{align}
One approach is to treat this as an \emph{unconstrained} optimization problem and search for the optimal $w$ in the entire space that $w$ lives in. Although the form of the objective function (\ref{eq:op_cache}) is complicated (it is piecewise quadratic), one can use \emph{(stochastic) gradient descent} to optimize it. This boils down to starting from an arbitrary point $w$, computing (or estimating) the sub-gradient vector and updating $w$ in the appropriate direction accordingly. Each gradient step requires finding the cache-entry with the highest value (because of the $\max$ operation in Equation \ref{eq:op_cache}) and convergence requires numerous gradient steps. We found this to be prohibitively slow in practice. 

Another approach optimizes a sequence auxiliary objective functions instead. The auxiliary objective functions are simple but constrained. The approach proceeds by gradually adding constraints to make the solution converge to that of the original objective function. We can turn the complex objective function (\ref{eq:op_cache}) into a simple quadratic function but it comes at the price of introducing an extremely large set of constraints. However, our experiments show that the optimization problem becomes tractable by using the 1-slack formulation of \cite{joachims09} and maintaining a \emph{working set} of active constraints.

Let $\sq{\mathcal{W}=\{(e_1, e_2, \dots, e_k) : e_i = (i, f, \delta) \in C\}}$ be the set of constraints to be satisfied. Each constraint $\omega \in \mathcal{W}$ is an $k$-tuple whose $i$-th element is a cache-entry $e_i=(i, f, \delta) \in C$ that corresponds to the $i$-th training example. Note that each constraint $\omega$ specifies a complete latent configuration on the set of training samples (that is the category label and all part locations for \emph{all} training samples). Therefore, each constraint is a linear function of $w$. There is a total training loss value associated to each constraint $\omega=(e_{\omega, 1}, \dots, e_{\omega, k})$ which we refer to as $\loss(\omega, w) = \sum_{i=1}^k w^T f_{\omega, i} + \delta_{\omega, i}$ where $f_{\omega, i}$ and $\delta_{\omega, i}$ are given by $e_{\omega, i}=(i, f_{\omega, i}, \delta_{\omega, i})$.
Solving the unconstrained minimization problem in (\ref{eq:op_cache}) is equivalent to solving the following quadratic programming (QP) problem subject to a set of constraints specified by $\mathcal{W}$:
\begin{align}
	w^*_C =& \argmin_w \lambda_w ||w||^2 + \xi \label{eq:op_cache_1slack_primal}\\
	& \text{s.t. } \forall \omega \in \mathcal{W}: \loss(\omega, w) \leq \xi \notag
\end{align}
In practice, we cannot optimize \ref{eq:op_cache_1slack_primal} over the entire constraint set explicitly since $|\mathcal{W}|$ is too large. Instead, we optimize it over an \emph{active} subset of the constraint set $W \subseteq \mathcal{W}$. More specifically, we start from $W=\emptyset$, solve the QP with respect to the current $W$, add the \emph{most violated constraint}\footnote{The most violated constraint is the one with the highest loss value i.e. $\argmax_{\omega \in \mathcal{W}} \loss(\omega, w)$.} to $W$, and repeat until addition of no new constraint increases the value of the objective function by more than a given $\epsilon$. \cite{joachims09} showed that the number of iterations it takes for this algorithm to converge is inversely proportional to the value of epsilon and is also independent of the size of the training set. They also showed that the dual of the QP in \ref{eq:op_cache_1slack_primal} has an extremely sparse solution in the sense that most of the dual variables turn out to be zero. Since dual variables correspond to constraints in the primal form the observation implies that only a tiny fraction of the primal constraints will be \emph{active}. This is the key behind efficiency of the algorithm in practice.

\subsection{Dual Formulation}
Let $M$ be a $\sq{|W| \times |W|}$ kernel matrix for the feature function $\phi(\omega) = (2 \lambda_w)^{-\frac{1}{2}} \sum_{i=1}^k f_{\omega, i}$. Also, let $b$ be a vector of length $|W|$ where $b_\omega = - \sum_{i=1}^k \delta_{\omega, i}$ for $\omega \in W$.  We derive the dual of \ref{eq:op_cache_1slack_primal} and write it in the following simple form:
\begin{align}
	\alpha^* =& \argmin_{\alpha_\omega \geq 0} \frac{1}{2} \alpha^T M \alpha + \alpha^T b \label{eq:op_cache_dual_simple} \\
	& \text{s.t. } \sum_{\omega \in W}\alpha_\omega \leq 1 \notag
\end{align}
The solution of the dual and primal are related through the following equation:
\begin{align}
	w^* = -\frac{1}{2 \lambda_w} \sum_{\omega \in W} \alpha^*_\omega \sum_{i=1}^k f_{\omega, i} \label{eq:op_cache_dual_kkt_1}
\end{align}
Since we start from an empty set of constraints $W=\emptyset$ and gradually add constraints to it, after enough iterations, many of the $\alpha_\omega$'s become (and remain) zero for the rest of the optimization process. This happens in particular for the constraints that were added in the earlier rounds of growing $W$. This observation suggests that we can prune the constraint set $W$ in each iteration by discarding $\omega$'s for which $\alpha_\omega$ has remained zero for a certain number of consecutive iterations. We use Algorithm~\ref{alg:qp_solver} to solve the optimization problem of Equation~\ref{eq:op_cache} in practice.
\begin{algorithm}[t]
\begin{algor}[1]
	\item [{*}] Input: Cache $C$, convergence precision $\epsilon$
	\item [{*}] $W := \emptyset$
	\item [repeat] ~
		\item [{*}] construct $|W| \times |W|$ matrix $M$ and vector $b$ (see text)
		\item [{*}] solve the QP in Equation \ref{eq:op_cache_dual_simple} to find $\alpha^*$ and compute $\xi$
		\item [{*}] $w := \frac{-1}{2 \lambda_w} \sum_{\omega \in W} \alpha^*_\omega \sum_{i=1}^k f_{\omega, i}$ (see Equation \ref{eq:op_cache_dual_kkt_1})
		\item [{*}] prune $W$
		\item [{*}] find most violated constraint $\omega^* = \argmax_{\omega \in \mathcal{W}} \loss(\omega, w)$
		\item [{*}] $W := W  \bigcup $\{$\omega^*$\}
	\item [until] $\loss(\omega^*, w) \leq \xi + \epsilon$
	\item [{*}] output $w$
\end{algor}
\caption{\label{alg:qp_solver} Fast QP solver for optimizing $\mathcal{B}_C$}
\end{algorithm}

\section{Part Selection}
\begin{figure}
	\centering
	\includegraphics[width=0.6\textwidth]{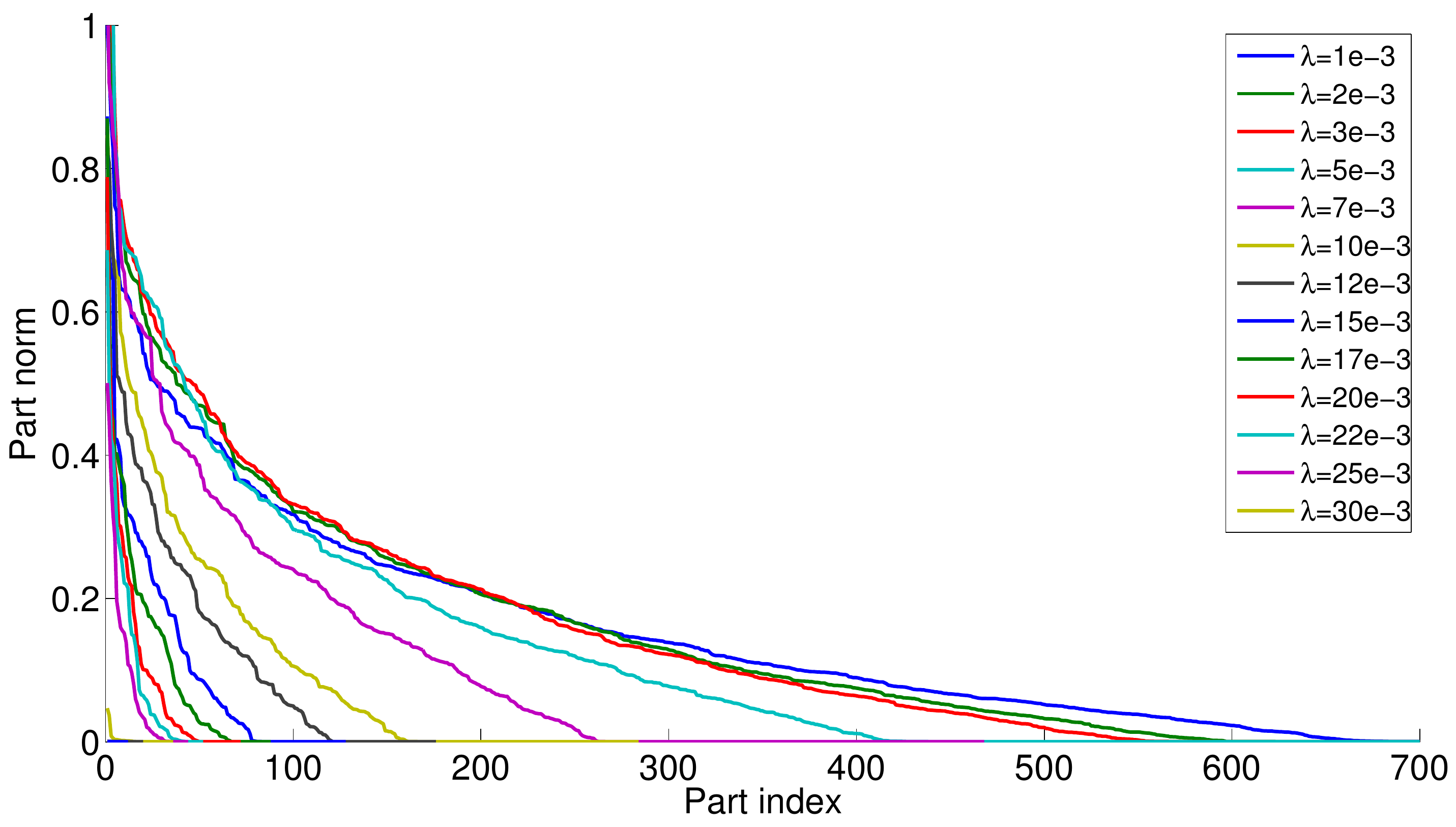}
	\vspace{-0.5em}
	\caption{Effect of $\lambda$ on part norms. Each plot shows sorted $\rho_j$ values for a particular choice of $\lambda$.}\label{fig:L1L2}
	\label{fig:L1L2_norms}
\end{figure}

As mentioned in Section~\ref{sec:init} of the paper, we use group sparsity to select useful parts from a pool of randomly initialized parts. We use the same formulation as in~\cite{sun13}.
Part selection is done by optimizing the following objective function:
\begin{align}
	 \lambda \sum_{j=1}^m \rho_j + \sum_{i=1}^k  \max\{0, \max_{y \neq y_i} (u_y - u_{y_i}) \cdot r(x_i, w) + 1 \}
\end{align}
where $\rho_j = \sqrt{\sum_{y}u_{y, j}^2}$ is the $\ell 2$-norm of the column of $u$ that corresponds to part $j$.  This objective function is convex. We minimize it using stochastic gradient descent. This requires repeatedly taking a small step in the opposite direction of a sub-gradient of the function.  Let $R_g(u) = \lambda \sum_{j=1}^m \rho_j$.
The partial derivative $\frac{\partial R_g}{\partial u_y} = \frac{u_y}{\rho_j}$ explodes as $\rho_j$ goes to zero. 
Thus, we round the $\rho_j$'s as they approach zero. We denote the rounded version by $\tau_j$ and define them as follows
\[ \tau_j = \left\{ 
  \begin{array}{l l}
    \rho_j & \quad \text{if $\rho_j > \epsilon$} \\
    \frac{\rho_j^2}{2 \epsilon} +  \frac{\epsilon}{2} & \quad \text{if $\rho_j \leq \epsilon$}
  \end{array} \right.\]
The constants in the second case are set so that $\tau_j$ is continuous; that is $\frac{\rho_j^2}{2 \epsilon} +  \frac{\epsilon}{2} = \rho_j \text{ when } \rho_j = \epsilon$.
In summary, part selection from an initial pool of parts $\sq{w=(w_1, \dots, w_m)}$ involves optimizing the following objective function:
\begin{align}
	 \lambda \sum_{j=1}^m \tau_j + \sum_{i=1}^k  \max\{0, \max_{y \neq y_i} (u_y - u_{y_i}) \cdot r(x_i, w) + 1 \}
\end{align}
We can control the sparsity of the solution to this optimization problem by changing the value of $\lambda$.
In Figure \ref{fig:L1L2_norms} we plot $\rho_j$ for all parts in decreasing order. Each curve corresponds to the result obtained with a different $\lambda$ value. These plots suggest that the number of selected parts (i.e. parts whose $\rho_j$ is larger than a threshold that depends on $\epsilon$) decreases monotonically as $\lambda$ increases. We adjust $\lambda$ to obtain a target number of selected parts.

\section{More on Visualization of the Model}
\label{sec:visualization_supplement}
\begin{figure*}[t]
	\centering
	\renewcommand{\tabcolsep}{1pt}
	\begin{tabular}{ c c c c c c c c c c c c c }
	1 & 2 & 3 & 4 & 5 & 6 & 7 & 8 & 9 & 10 & 11 & 12 & 13 \\ [-0.05cm]
	\includepart{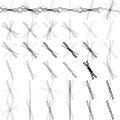} &
	\includepart{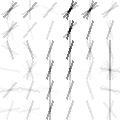} &
	\includepart{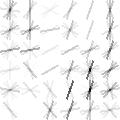} &
	\includepart{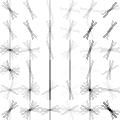} &
	\includepart{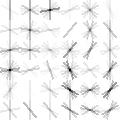} &
	\includepart{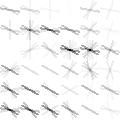} &
	\includepart{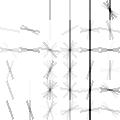} &
	\includepart{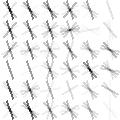} &
	\includepart{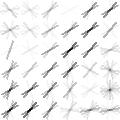} &
	\includepart{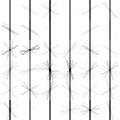} &
	\includepart{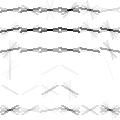} &
	\includepart{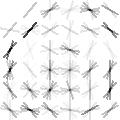} &
	\includepart{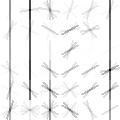} \\
	14 & 15 & 16 & 17 & 18 & 19 & 20 & 21 & 22 & 23 & 24 & 25 & 26 \\ [-0.05cm]
	\includepart{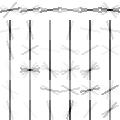} &
	\includepart{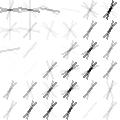} &
	\includepart{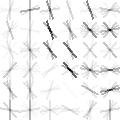} &
	\includepart{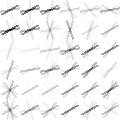} &
	\includepart{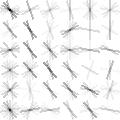} &
	\includepart{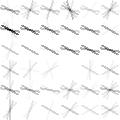} &
	\includepart{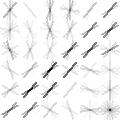} &
	\includepart{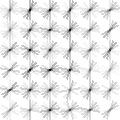} &
	\includepart{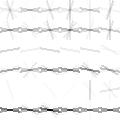} &
	\includepart{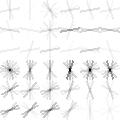} &
	\includepart{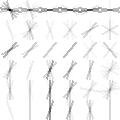} &
	\includepart{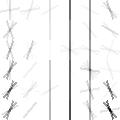} &
	\includepart{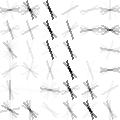} \\
	27 & 28 & 29 & 30 & 31 & 32 & 33 & 34 & 35 & 36 & 37 & 38 & 39 \\ [-0.05cm]
	\includepart{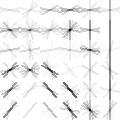} &
	\includepart{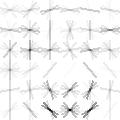} &
	\includepart{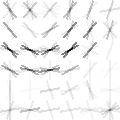} &
	\includepart{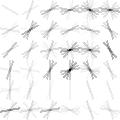} &
	\includepart{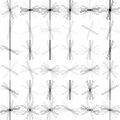} &
	\includepart{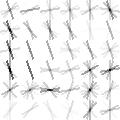} &
	\includepart{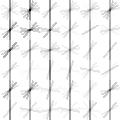} &
	\includepart{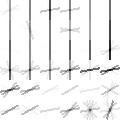} &
	\includepart{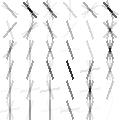} &
	\includepart{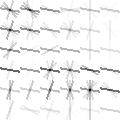} &
	\includepart{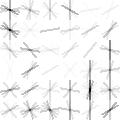} &
	\includepart{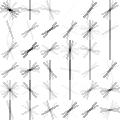} &
	\includepart{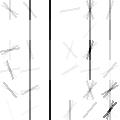} \\
	40 & 41 & 42 & 43 & 44 & 45 & 46 & 47 & 48 & 49 & 50 & 51 & 52\\ [-0.05cm]
	\includepart{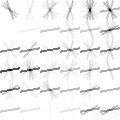} &
	\includepart{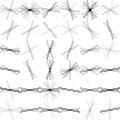} &
	\includepart{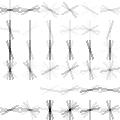} &
	\includepart{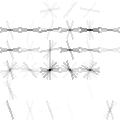} &
	\includepart{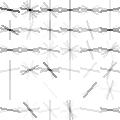} &
	\includepart{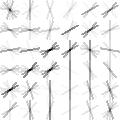} &
	\includepart{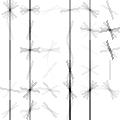} &
	\includepart{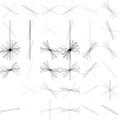} &
	\includepart{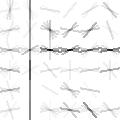} &
	\includepart{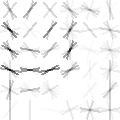} &
	\includepart{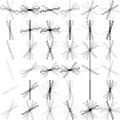} &
	\includepart{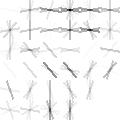} &
	\includepart{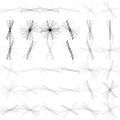}
	\end{tabular}
	\begin{tabular}{ c }
	\includegraphics[width=1.17\textwidth]{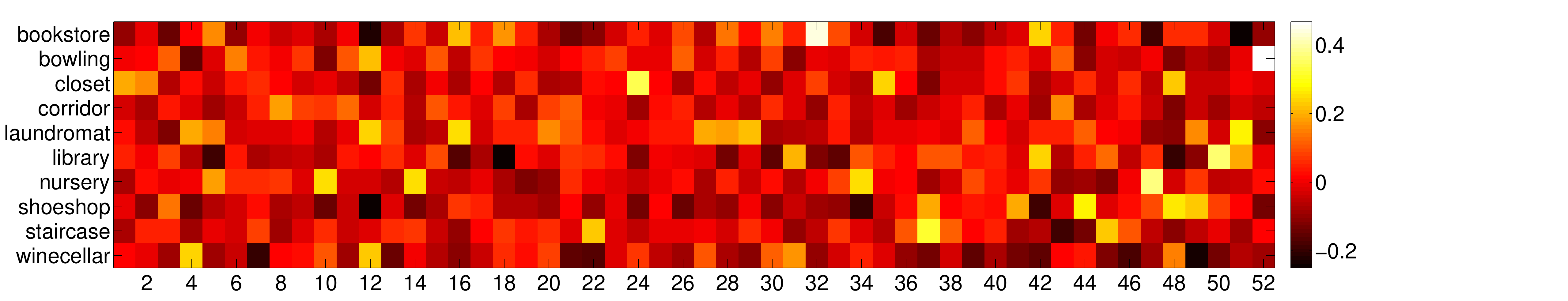}
	\end{tabular}
	\vspace{-1.5em}
	\caption{Part filters (top) and part weights (bottom) after joint training a model with 52 parts on the 10-class dataset. Here we use HOG features. Although the model uses 5 pooling regions (corresponding to cells in $\sq{1 \times 1} + \sq{2 \times 2}$ grids) here we show the part weights only for the first pooling region corresponding the entire image.}
	\label{fig:u_matrix_and_parts}
	\vspace{2.5em}
	\centering
	\renewcommand{\tabcolsep}{0.75pt}
	\begin{tabular}{ c   c c c c c   c c c c c   c c c c c }
	 & \multicolumn{15}{c}{Top scoring patches on test images (1 patch per image)} \\
	\multirow{3}{*}{\rotatebox{90}{Part 42}$\;$} &
				\includegraphics[height=0.33in, width=0.33in]{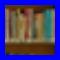} &
				\includegraphics[height=0.33in, width=0.33in]{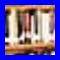} &
				\includegraphics[height=0.33in, width=0.33in]{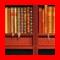} &
				\includegraphics[height=0.33in, width=0.33in]{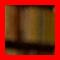} &
				\includegraphics[height=0.33in, width=0.33in]{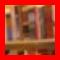} &
				\includegraphics[height=0.33in, width=0.33in]{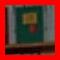} &
				\includegraphics[height=0.33in, width=0.33in]{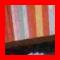} &
				\includegraphics[height=0.33in, width=0.33in]{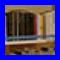} &
				\includegraphics[height=0.33in, width=0.33in]{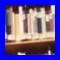} &
				\includegraphics[height=0.33in, width=0.33in]{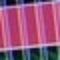} &
				\includegraphics[height=0.33in, width=0.33in]{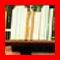} &
				\includegraphics[height=0.33in, width=0.33in]{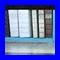} &
				\includegraphics[height=0.33in, width=0.33in]{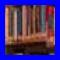} &
				\includegraphics[height=0.33in, width=0.33in]{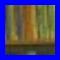} &
				\includegraphics[height=0.33in, width=0.33in]{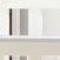} \\ [-0.05cm]
			&	\includegraphics[height=0.33in, width=0.33in]{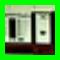} &
				\includegraphics[height=0.33in, width=0.33in]{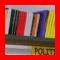} &
				\includegraphics[height=0.33in, width=0.33in]{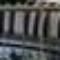} &
				\includegraphics[height=0.33in, width=0.33in]{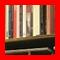} &
				\includegraphics[height=0.33in, width=0.33in]{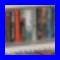} &
				\includegraphics[height=0.33in, width=0.33in]{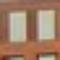} &
				\includegraphics[height=0.33in, width=0.33in]{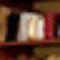} &
				\includegraphics[height=0.33in, width=0.33in]{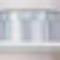} &
				\includegraphics[height=0.33in, width=0.33in]{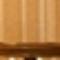} &
				\includegraphics[height=0.33in, width=0.33in]{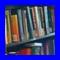} &
				\includegraphics[height=0.33in, width=0.33in]{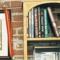} &
				\includegraphics[height=0.33in, width=0.33in]{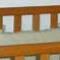} &
				\includegraphics[height=0.33in, width=0.33in]{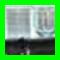} &
				\includegraphics[height=0.33in, width=0.33in]{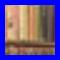} &
				\includegraphics[height=0.33in, width=0.33in]{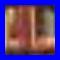} \\ [-0.05cm]
			&	\includegraphics[height=0.33in, width=0.33in]{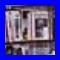} &
				\includegraphics[height=0.33in, width=0.33in]{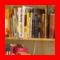} &
				\includegraphics[height=0.33in, width=0.33in]{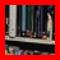} &
				\includegraphics[height=0.33in, width=0.33in]{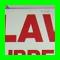} &
				\includegraphics[height=0.33in, width=0.33in]{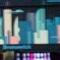} &
				\includegraphics[height=0.33in, width=0.33in]{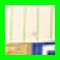} &
				\includegraphics[height=0.33in, width=0.33in]{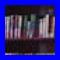} &
				\includegraphics[height=0.33in, width=0.33in]{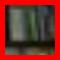} &
				\includegraphics[height=0.33in, width=0.33in]{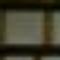} &
				\includegraphics[height=0.33in, width=0.33in]{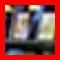} &
				\includegraphics[height=0.33in, width=0.33in]{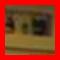} &
				\includegraphics[height=0.33in, width=0.33in]{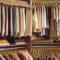} &
				\includegraphics[height=0.33in, width=0.33in]{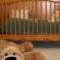} &
				\includegraphics[height=0.33in, width=0.33in]{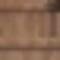} &
				\includegraphics[height=0.33in, width=0.33in]{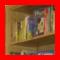}
				\\ [0.2cm]
	\multirow{3}{*}{\rotatebox{90}{Part 37}$\;$} &
				\includegraphics[height=0.33in, width=0.33in]{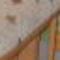} &
				\includegraphics[height=0.33in, width=0.33in]{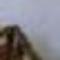} &
				\includegraphics[height=0.33in, width=0.33in]{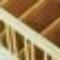} &
				\includegraphics[height=0.33in, width=0.33in]{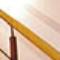} &
				\includegraphics[height=0.33in, width=0.33in]{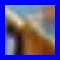} &
				\includegraphics[height=0.33in, width=0.33in]{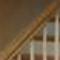} &
				\includegraphics[height=0.33in, width=0.33in]{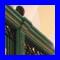} &
				\includegraphics[height=0.33in, width=0.33in]{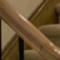} &
				\includegraphics[height=0.33in, width=0.33in]{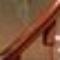} &
				\includegraphics[height=0.33in, width=0.33in]{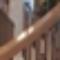} &
				\includegraphics[height=0.33in, width=0.33in]{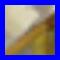} &
				\includegraphics[height=0.33in, width=0.33in]{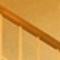} &
				\includegraphics[height=0.33in, width=0.33in]{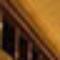} &
				\includegraphics[height=0.33in, width=0.33in]{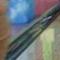} &
				\includegraphics[height=0.33in, width=0.33in]{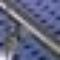} \\ [-0.05cm]
			&	\includegraphics[height=0.33in, width=0.33in]{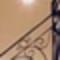} &
				\includegraphics[height=0.33in, width=0.33in]{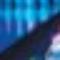} &
				\includegraphics[height=0.33in, width=0.33in]{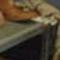} &
				\includegraphics[height=0.33in, width=0.33in]{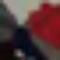} &
				\includegraphics[height=0.33in, width=0.33in]{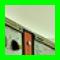} &
				\includegraphics[height=0.33in, width=0.33in]{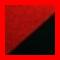} &
				\includegraphics[height=0.33in, width=0.33in]{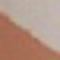} &
				\includegraphics[height=0.33in, width=0.33in]{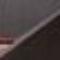} &
				\includegraphics[height=0.33in, width=0.33in]{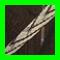} &
				\includegraphics[height=0.33in, width=0.33in]{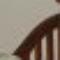} &
				\includegraphics[height=0.33in, width=0.33in]{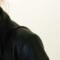} &
				\includegraphics[height=0.33in, width=0.33in]{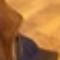} &
				\includegraphics[height=0.33in, width=0.33in]{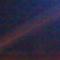} &
				\includegraphics[height=0.33in, width=0.33in]{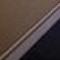} &
				\includegraphics[height=0.33in, width=0.33in]{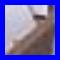} \\ [-0.05cm]
			&	\includegraphics[height=0.33in, width=0.33in]{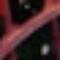} &
				\includegraphics[height=0.33in, width=0.33in]{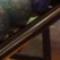} &
				\includegraphics[height=0.33in, width=0.33in]{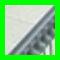} &
				\includegraphics[height=0.33in, width=0.33in]{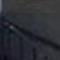} &
				\includegraphics[height=0.33in, width=0.33in]{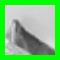} &
				\includegraphics[height=0.33in, width=0.33in]{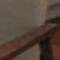} &
				\includegraphics[height=0.33in, width=0.33in]{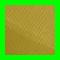} &
				\includegraphics[height=0.33in, width=0.33in]{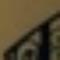} &
				\includegraphics[height=0.33in, width=0.33in]{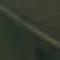} &
				\includegraphics[height=0.33in, width=0.33in]{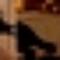} &
				\includegraphics[height=0.33in, width=0.33in]{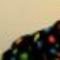} &
				\includegraphics[height=0.33in, width=0.33in]{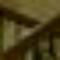} &
				\includegraphics[height=0.33in, width=0.33in]{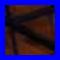} &
				\includegraphics[height=0.33in, width=0.33in]{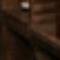} &
				\includegraphics[height=0.33in, width=0.33in]{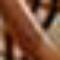}
				\\ [0.2cm]
	\multirow{3}{*}{\rotatebox{90}{Part 29}$\;$} &
				\includegraphics[height=0.33in, width=0.33in]{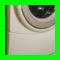} &
				\includegraphics[height=0.33in, width=0.33in]{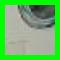} &
				\includegraphics[height=0.33in, width=0.33in]{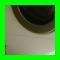} &
				\includegraphics[height=0.33in, width=0.33in]{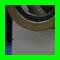} &
				\includegraphics[height=0.33in, width=0.33in]{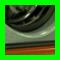} &
				\includegraphics[height=0.33in, width=0.33in]{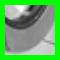} &
				\includegraphics[height=0.33in, width=0.33in]{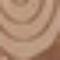} &
				\includegraphics[height=0.33in, width=0.33in]{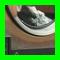} &
				\includegraphics[height=0.33in, width=0.33in]{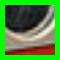} &
				\includegraphics[height=0.33in, width=0.33in]{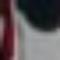} &
				\includegraphics[height=0.33in, width=0.33in]{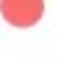} &
				\includegraphics[height=0.33in, width=0.33in]{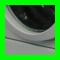} &
				\includegraphics[height=0.33in, width=0.33in]{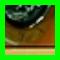} &
				\includegraphics[height=0.33in, width=0.33in]{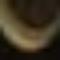} &
				\includegraphics[height=0.33in, width=0.33in]{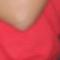} \\ [-0.05cm]
			&	\includegraphics[height=0.33in, width=0.33in]{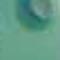} &
				\includegraphics[height=0.33in, width=0.33in]{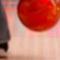} &
				\includegraphics[height=0.33in, width=0.33in]{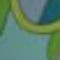} &
				\includegraphics[height=0.33in, width=0.33in]{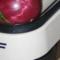} &
				\includegraphics[height=0.33in, width=0.33in]{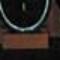} &
				\includegraphics[height=0.33in, width=0.33in]{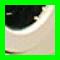} &
				\includegraphics[height=0.33in, width=0.33in]{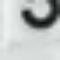} &
				\includegraphics[height=0.33in, width=0.33in]{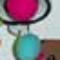} &
				\includegraphics[height=0.33in, width=0.33in]{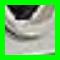} &
				\includegraphics[height=0.33in, width=0.33in]{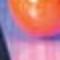} &
				\includegraphics[height=0.33in, width=0.33in]{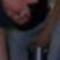} &
				\includegraphics[height=0.33in, width=0.33in]{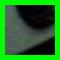} &
				\includegraphics[height=0.33in, width=0.33in]{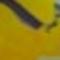} &
				\includegraphics[height=0.33in, width=0.33in]{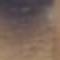} &
				\includegraphics[height=0.33in, width=0.33in]{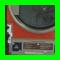} \\ [-0.05cm]
			&	\includegraphics[height=0.33in, width=0.33in]{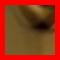} &
				\includegraphics[height=0.33in, width=0.33in]{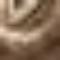} &
				\includegraphics[height=0.33in, width=0.33in]{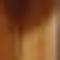} &
				\includegraphics[height=0.33in, width=0.33in]{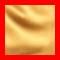} &
				\includegraphics[height=0.33in, width=0.33in]{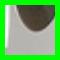} &
				\includegraphics[height=0.33in, width=0.33in]{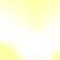} &
				\includegraphics[height=0.33in, width=0.33in]{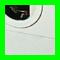} &
				\includegraphics[height=0.33in, width=0.33in]{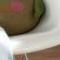} &
				\includegraphics[height=0.33in, width=0.33in]{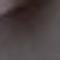} &
				\includegraphics[height=0.33in, width=0.33in]{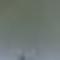} &
				\includegraphics[height=0.33in, width=0.33in]{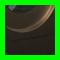} &
				\includegraphics[height=0.33in, width=0.33in]{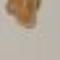} &
				\includegraphics[height=0.33in, width=0.33in]{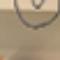} &
				\includegraphics[height=0.33in, width=0.33in]{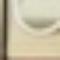} &
				\includegraphics[height=0.33in, width=0.33in]{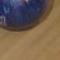}
	\end{tabular}
	\vspace{-1em}
	\caption{Top detections of three parts on test images of the 10-class dataset. The numbers in the first column match the part indices in Figure~\ref{fig:u_matrix_and_parts}. Patches from \emph{bookstore}, \emph{laundromat}, and \emph{library} images are highlighted in red, green, and blue respectively (best viewed in color).} \label{fig:top_scoring_patches_HOG}
	\vspace{-1.25em}
\end{figure*}

We complement Section~\ref{sec:visualization} of the paper by providing more visualizations of jointly trained parts. Figure~\ref{fig:u_matrix_and_parts} shows the part filters and the weight 
matrix after joint training a model with 52 parts on the 10-class subset of MIT-indoor dataset.
This model uses HOG features.
The part weight matrix determines whether a part is positive or negative 
with respect to two categories. For example, part 42 is positive for \emph{bookstore} 
and \emph{library} relative to \emph{laundromat}.  Part 29 is positive for \emph{laundromat}
relative to \emph{bookstore} and \emph{library}. 
Part 37 is positive for \emph{library} relative to \emph{bookstore} so it can be 
used in combination with the other two parts to distinguish between all three categories 
\emph{bookstore}, \emph{library}, and \emph{laundromat}. 
Figure~\ref{fig:top_scoring_patches_HOG} illustrates the top scoring patches for these three parts. 

%For example, part 5 is positive for \emph{bookstore} when compared to \emph{library}. 
%To understand what part 5 captures we have to look at the highest scoring patches of the 
%part in the images of the two categories. Figure \ref{fig:bookstore_vs_library} reveals that 
%part 5 acts as a human detector in \emph{bookstore} images. Furthermore, it turns out that
% human presence is significantly higher in images of \emph{bookstore} than \emph{library} in 
%this dataset ($55\%$ vs $10\%$). The figure also shows that part 42 consistently detects 
%book-shelves in both \emph{bookstore} and $\emph{library}$ and that the weight of this 
%part is high (and about the same) in both categories. Finally, part 37 is negative for 
%\emph{bookstore} when compared to \emph{library}. Although it is not clear from the high 
%scoring patches alone what discriminative aspect of the two categories is captured by this
% part but the pattern is clearly more consistent in \emph{library} images than it is in 
%\emph{bookstore} images.

\begin{figure*}[t]
	\renewcommand{\tabcolsep}{0.75pt}
	\begin{tabular}{ c   c c c c c c  }
	\rotatebox{90}{\hspace{0.25cm} Part 1}$\;$ &
				\includegraphics[height=0.65in, width=0.85in]{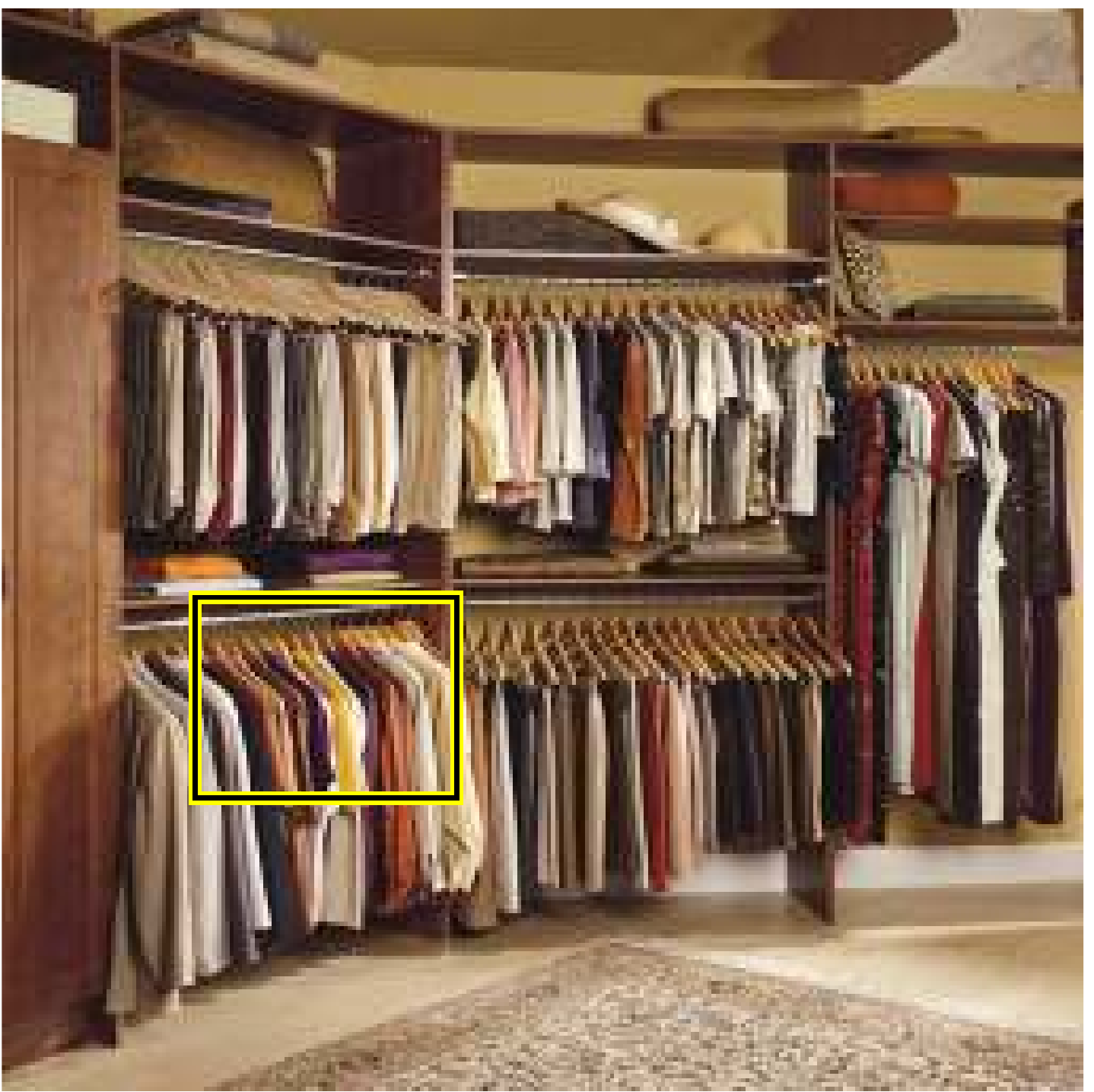} &
				\includegraphics[height=0.65in, width=0.85in]{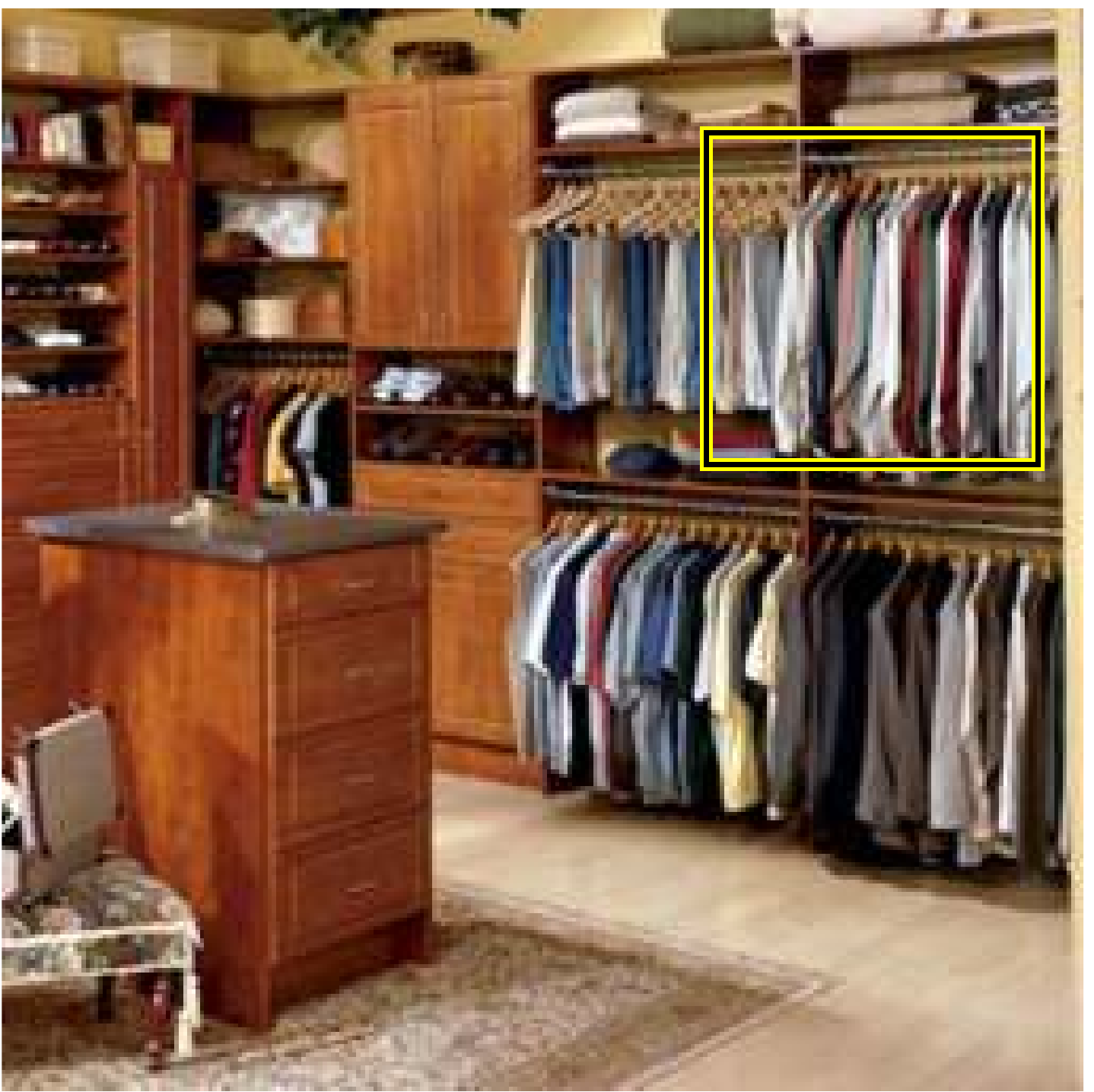} &
				\includegraphics[height=0.65in, width=0.85in]{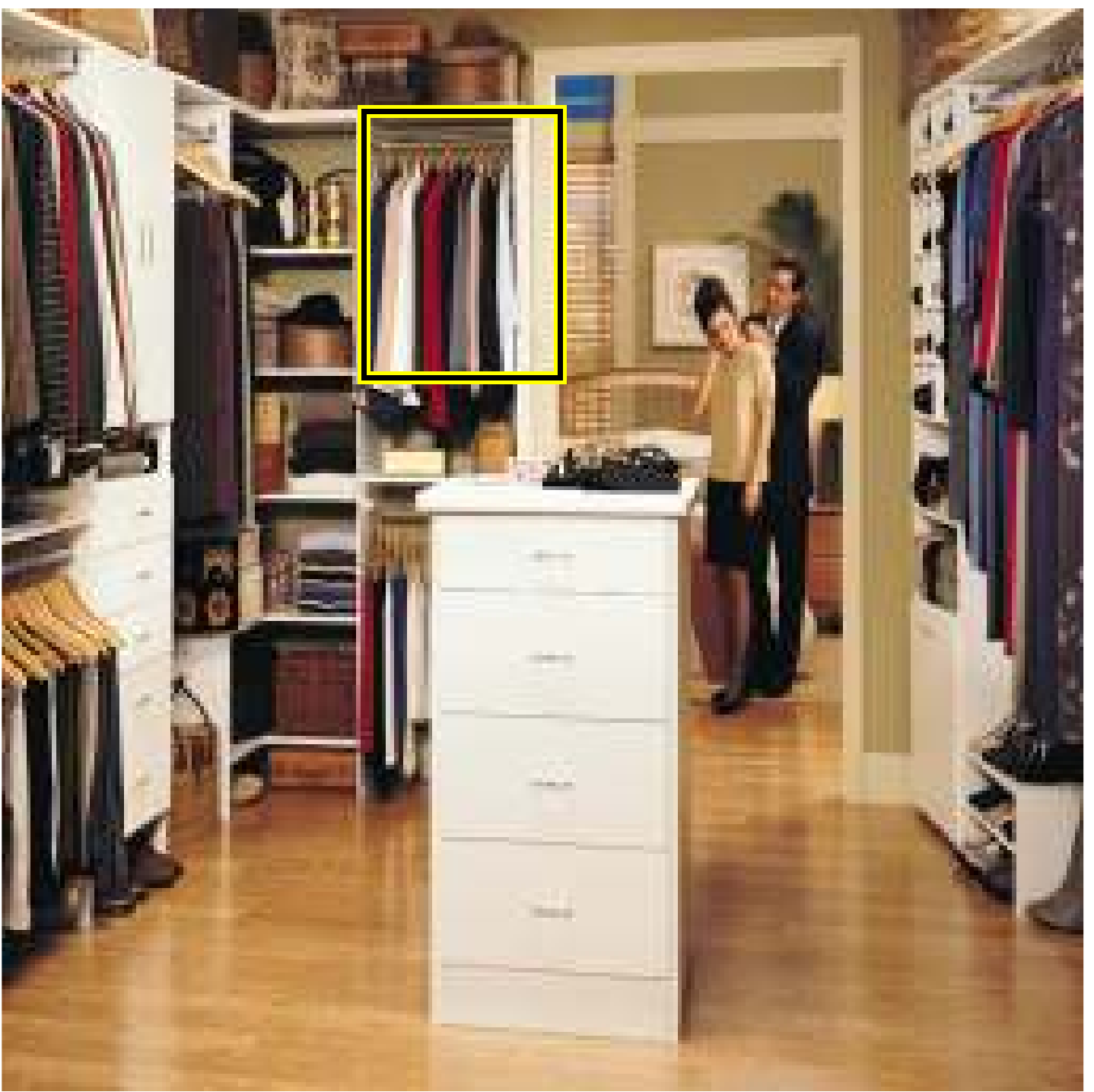} &
				\includegraphics[height=0.65in, width=0.85in]{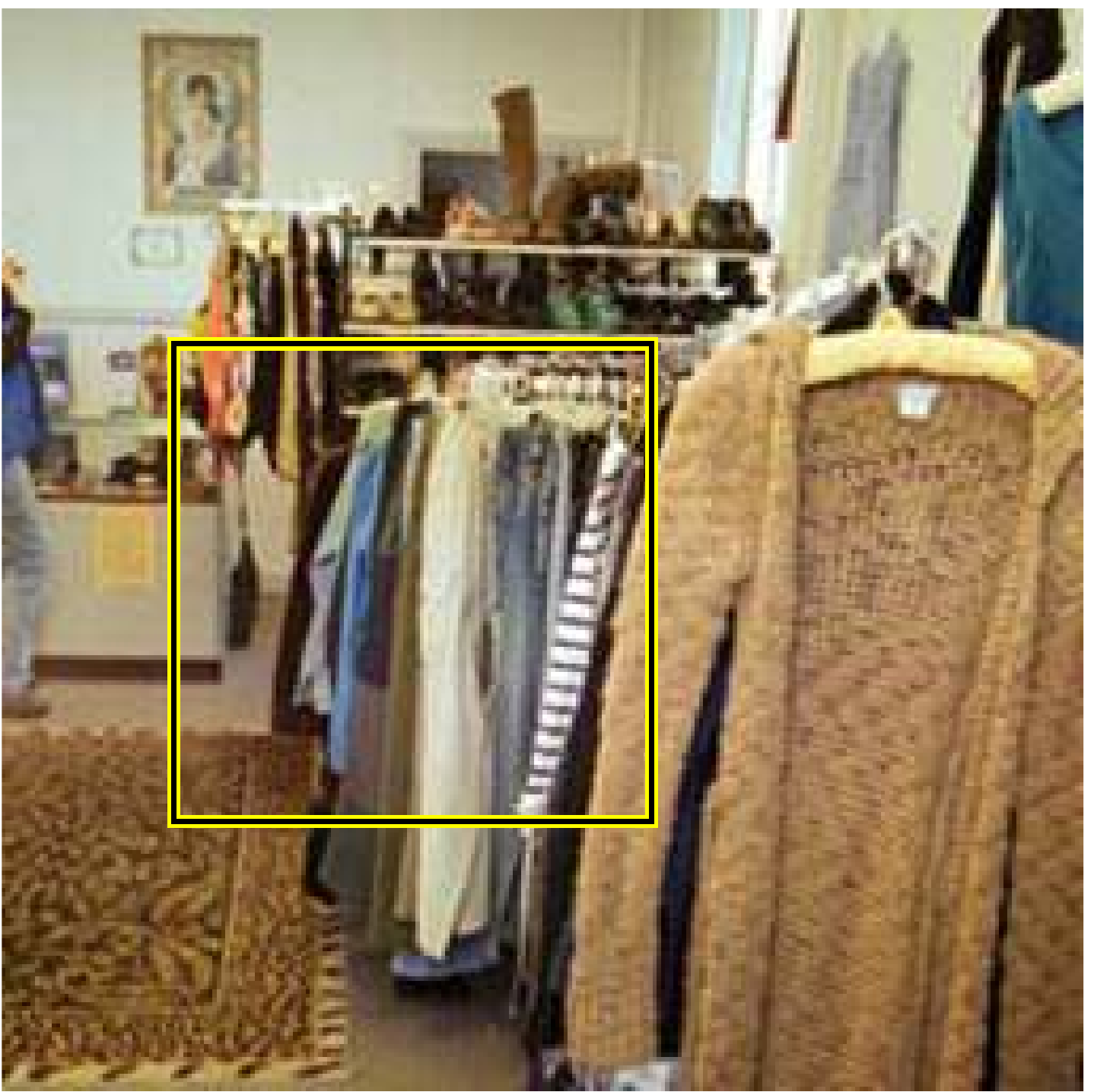} &
				\includegraphics[height=0.65in, width=0.85in]{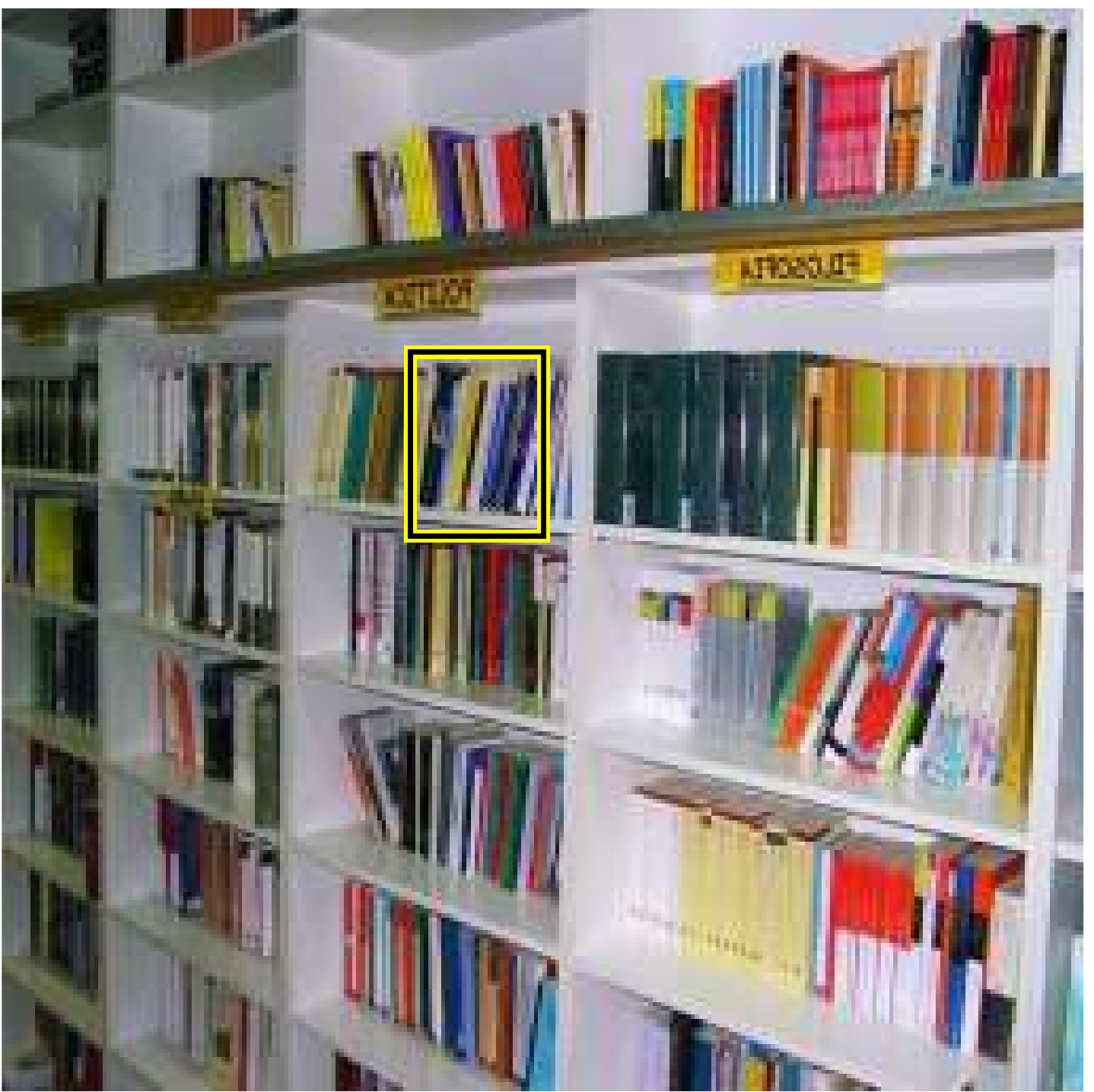} &
				\includegraphics[height=0.65in, width=0.85in]{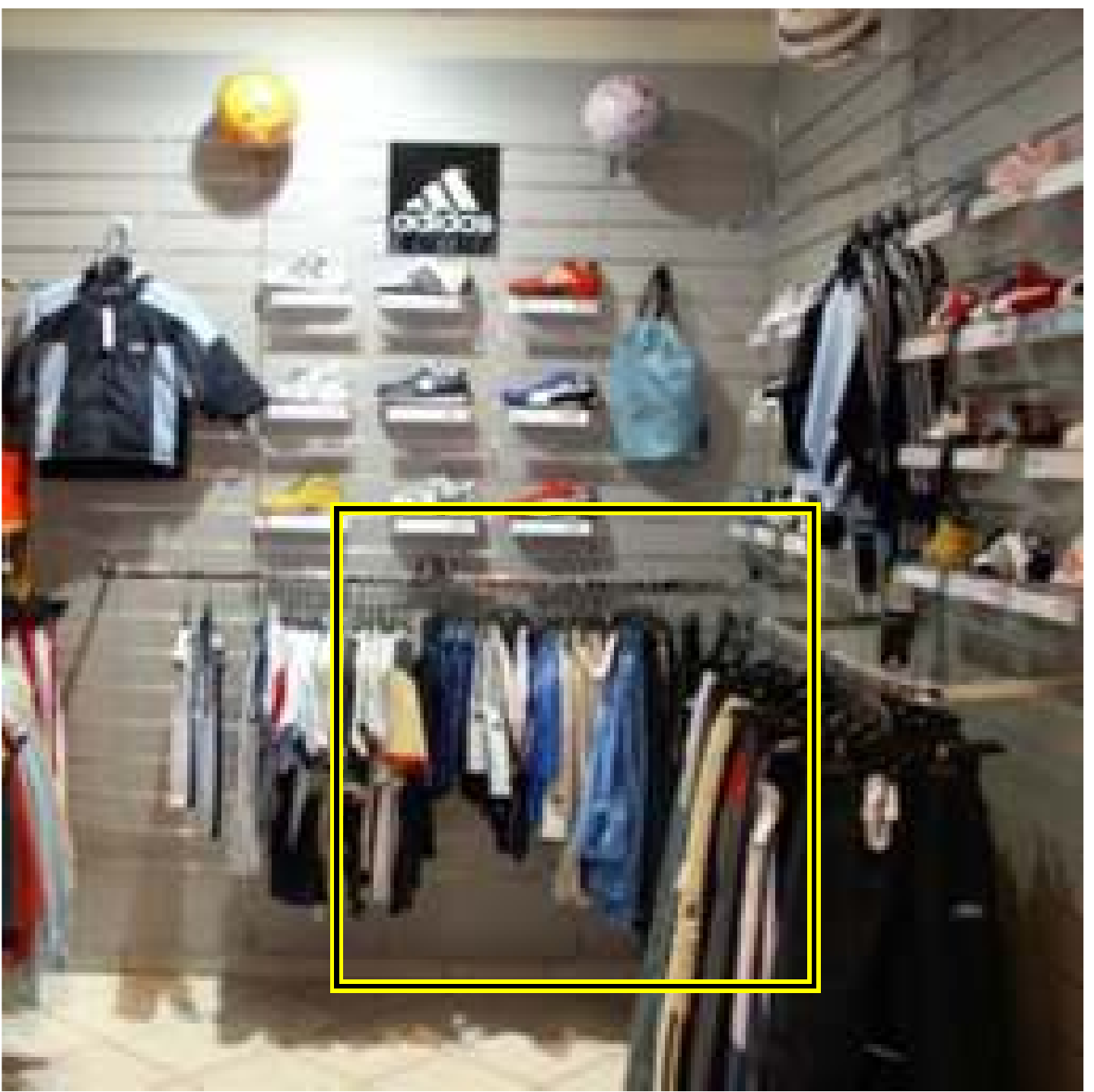} \\ [-0.05cm]
	\rotatebox{90}{\hspace{0.27cm} Part 5}$\;$ &
				\includegraphics[height=0.65in, width=0.85in]{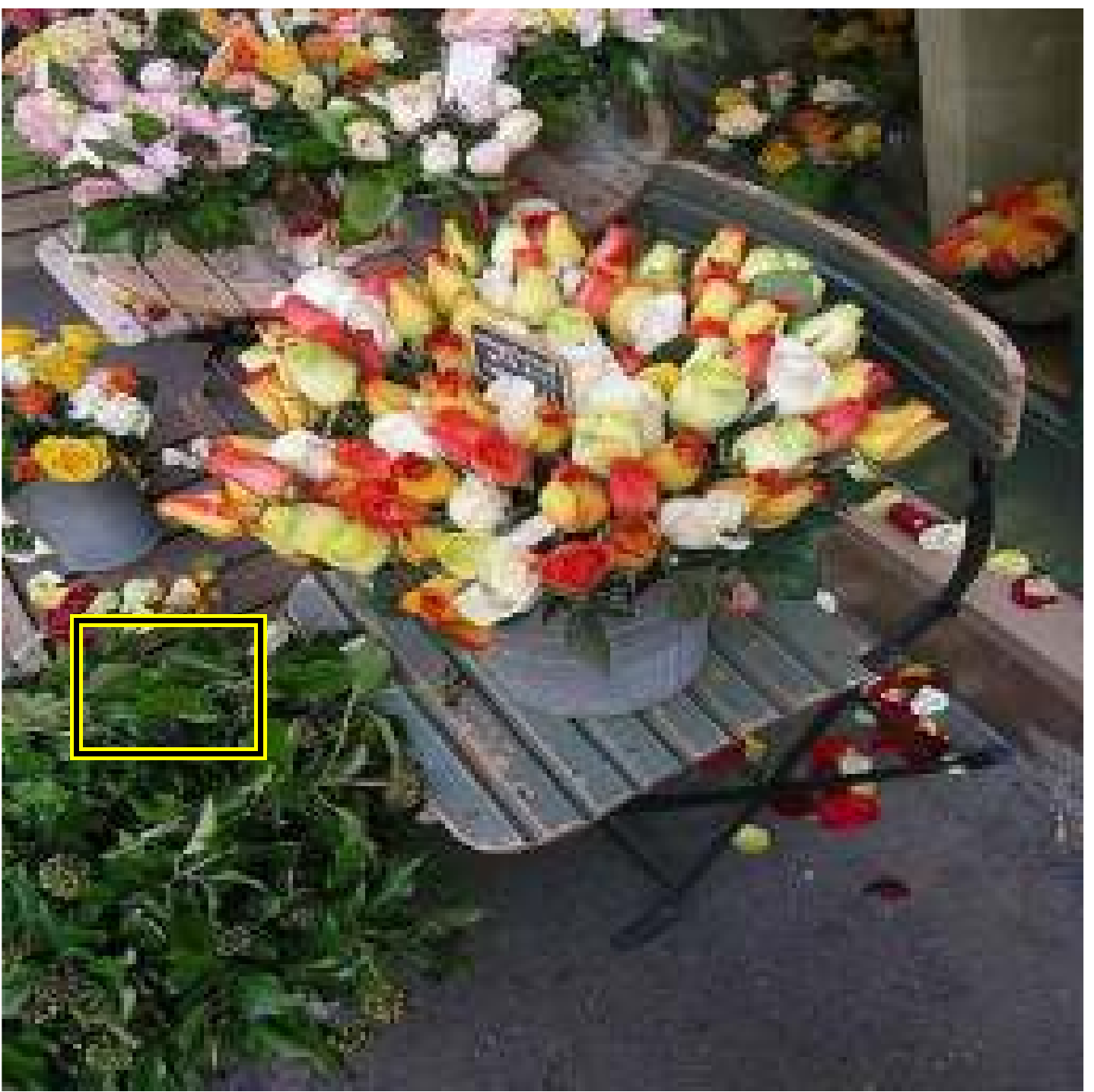} &
				\includegraphics[height=0.65in, width=0.85in]{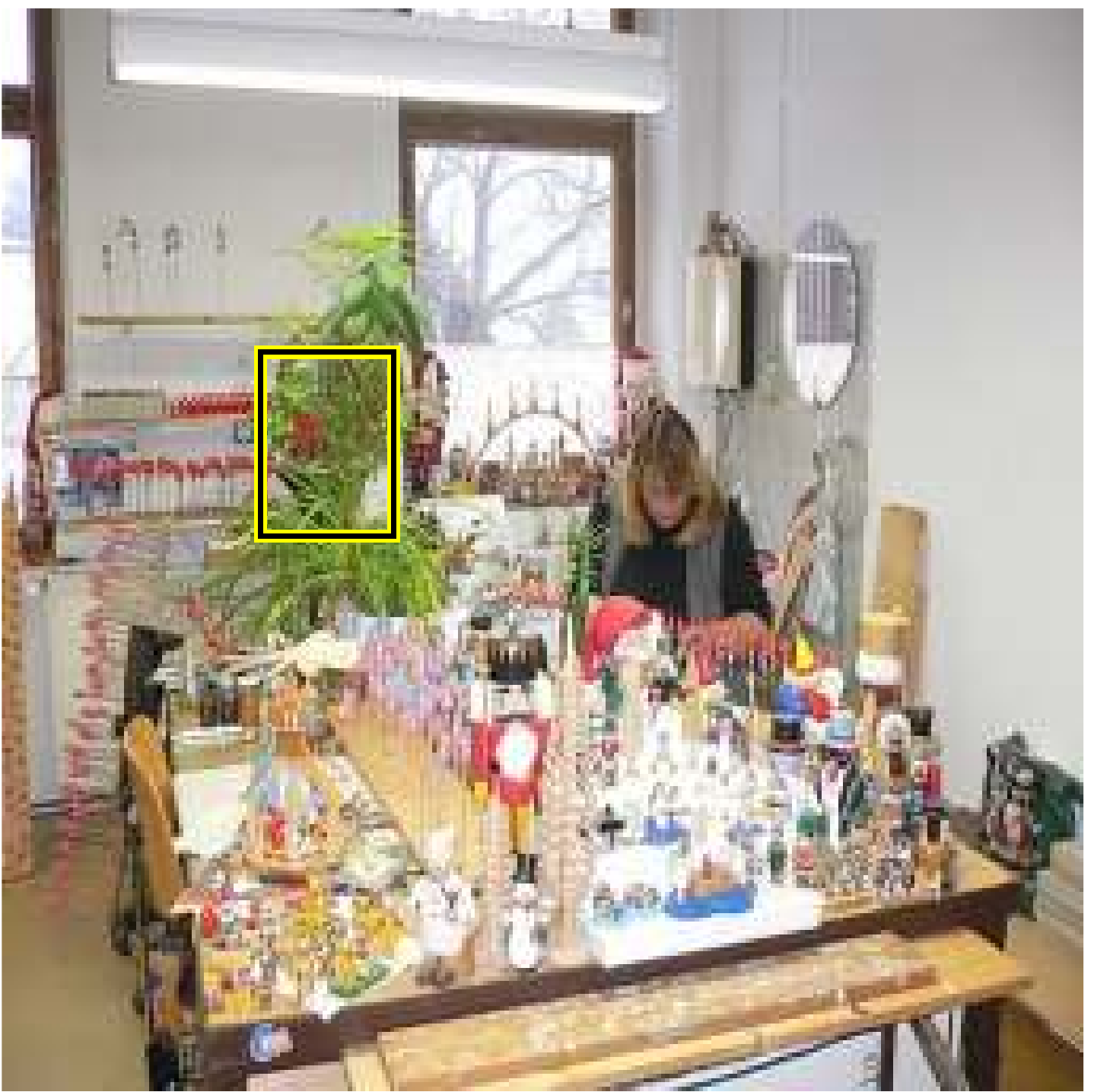} &
				\includegraphics[height=0.65in, width=0.85in]{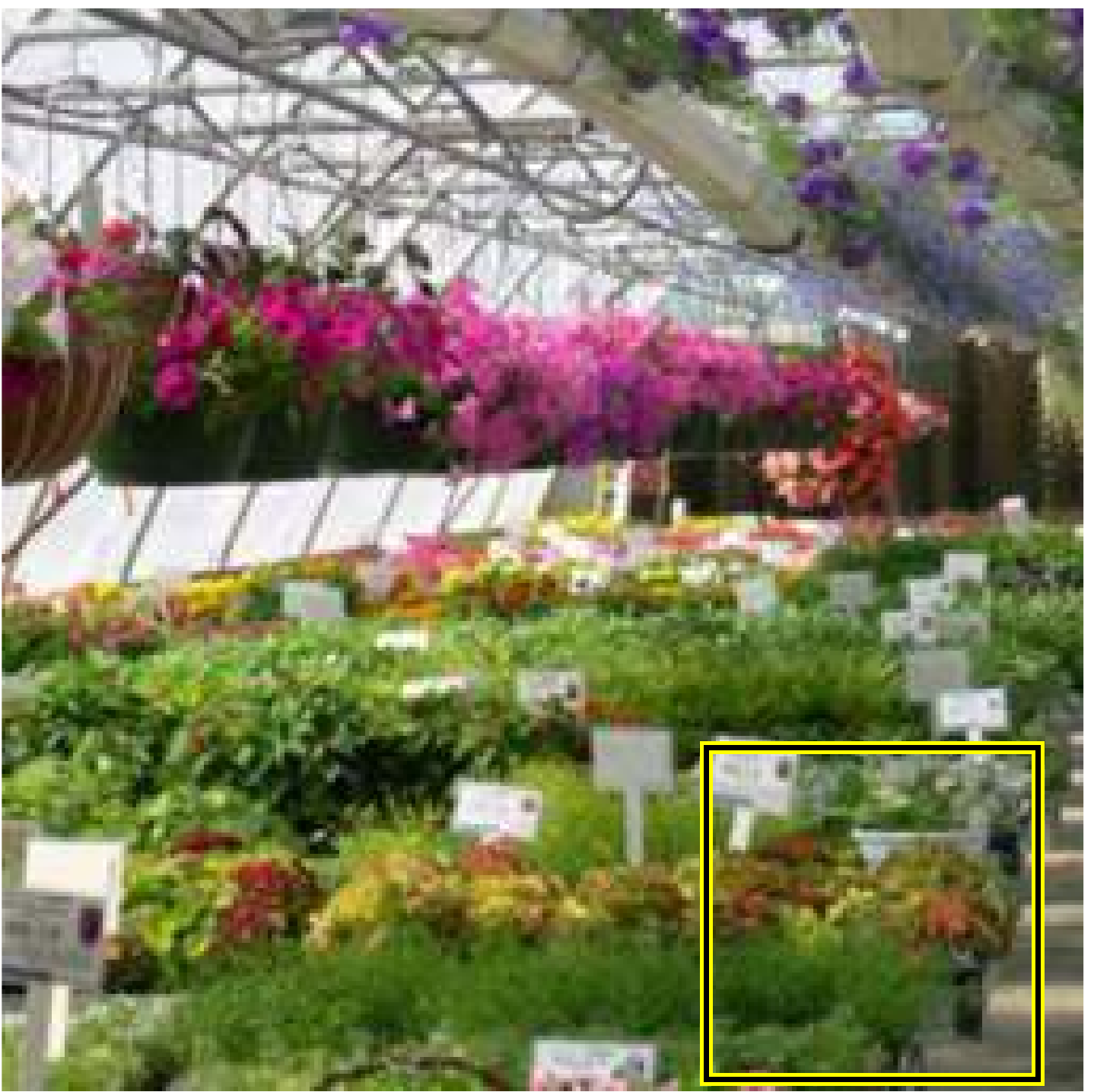} &
				\includegraphics[height=0.65in, width=0.85in]{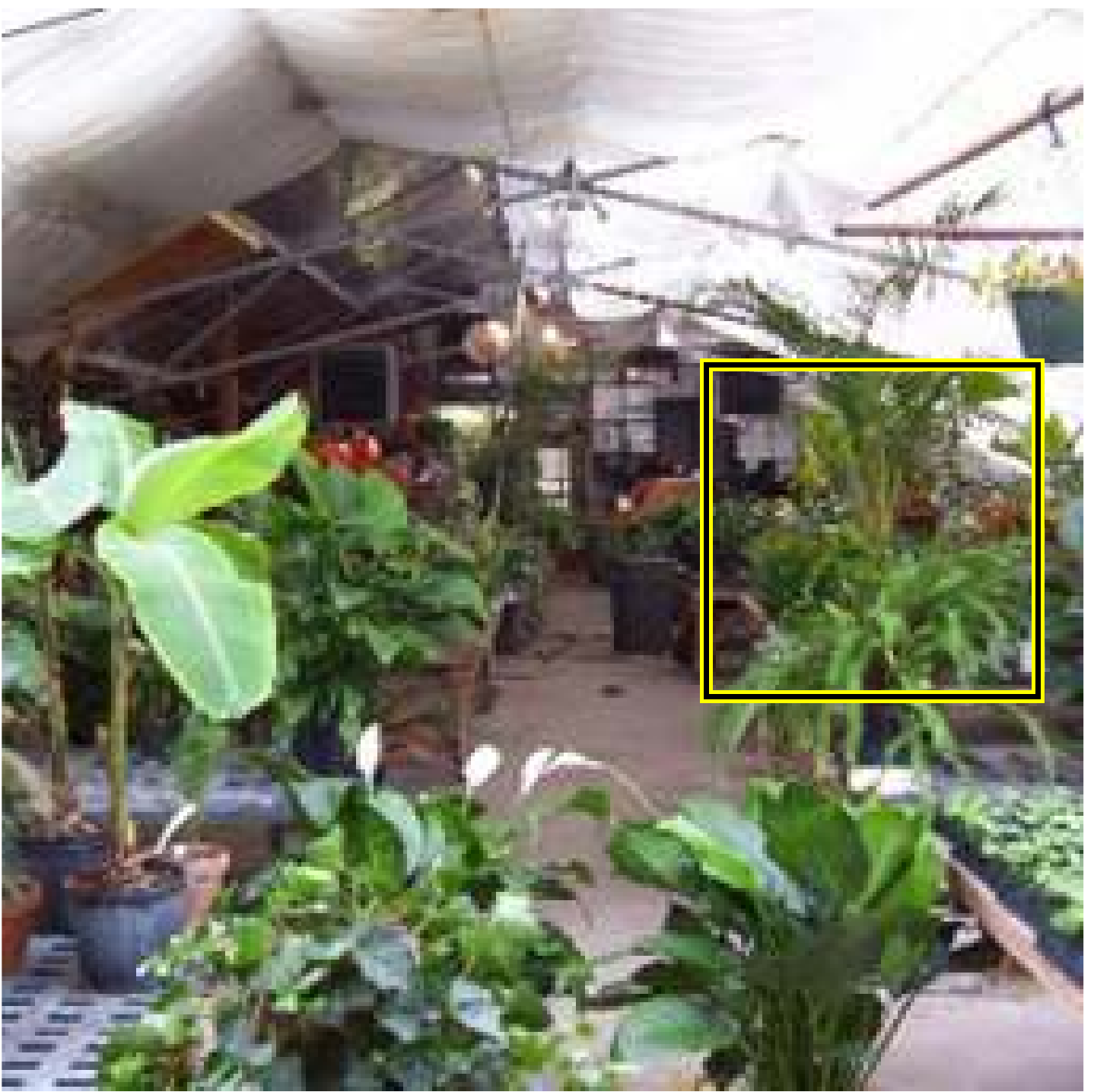} &
				\includegraphics[height=0.65in, width=0.85in]{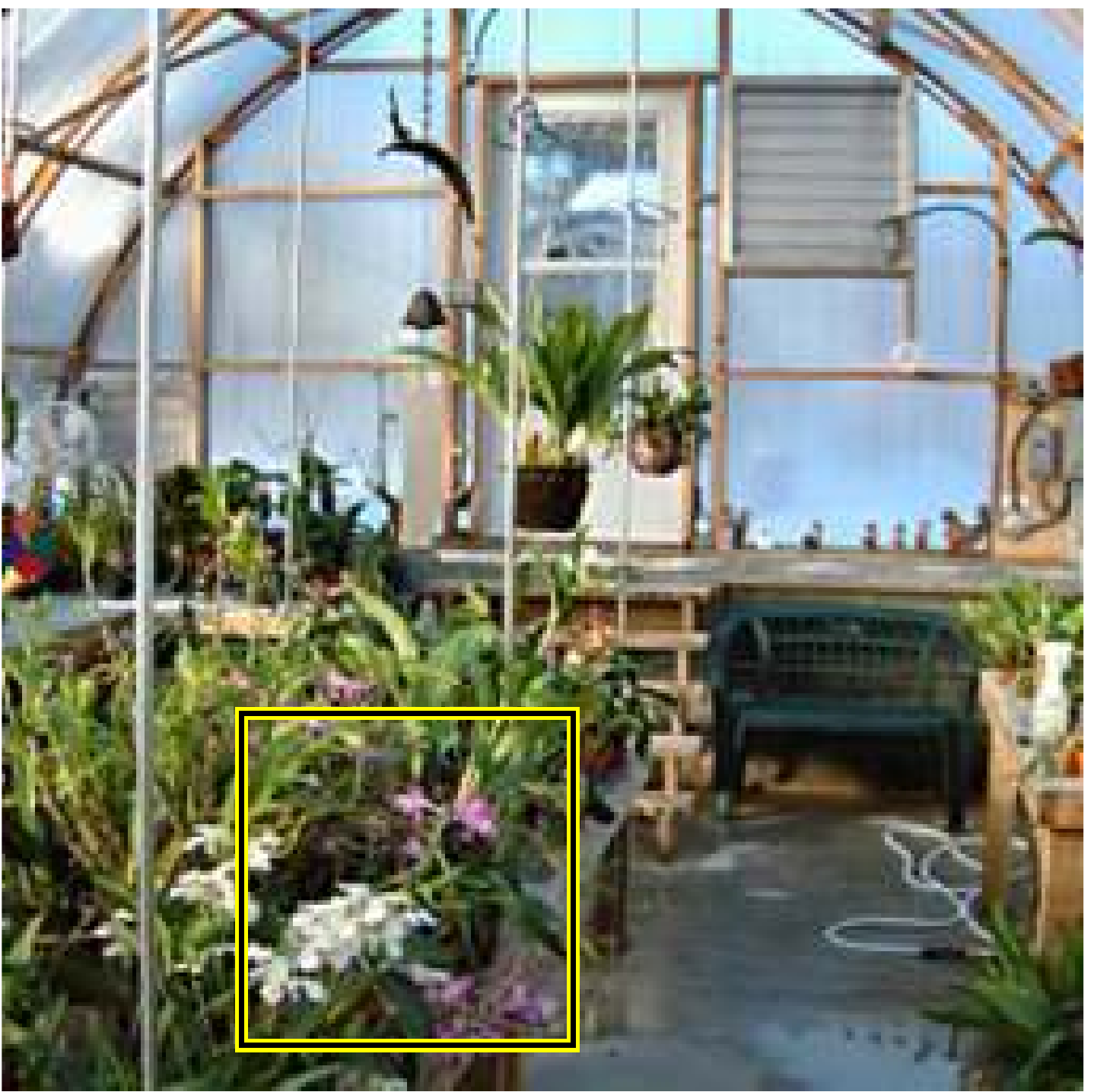} &
				\includegraphics[height=0.65in, width=0.85in]{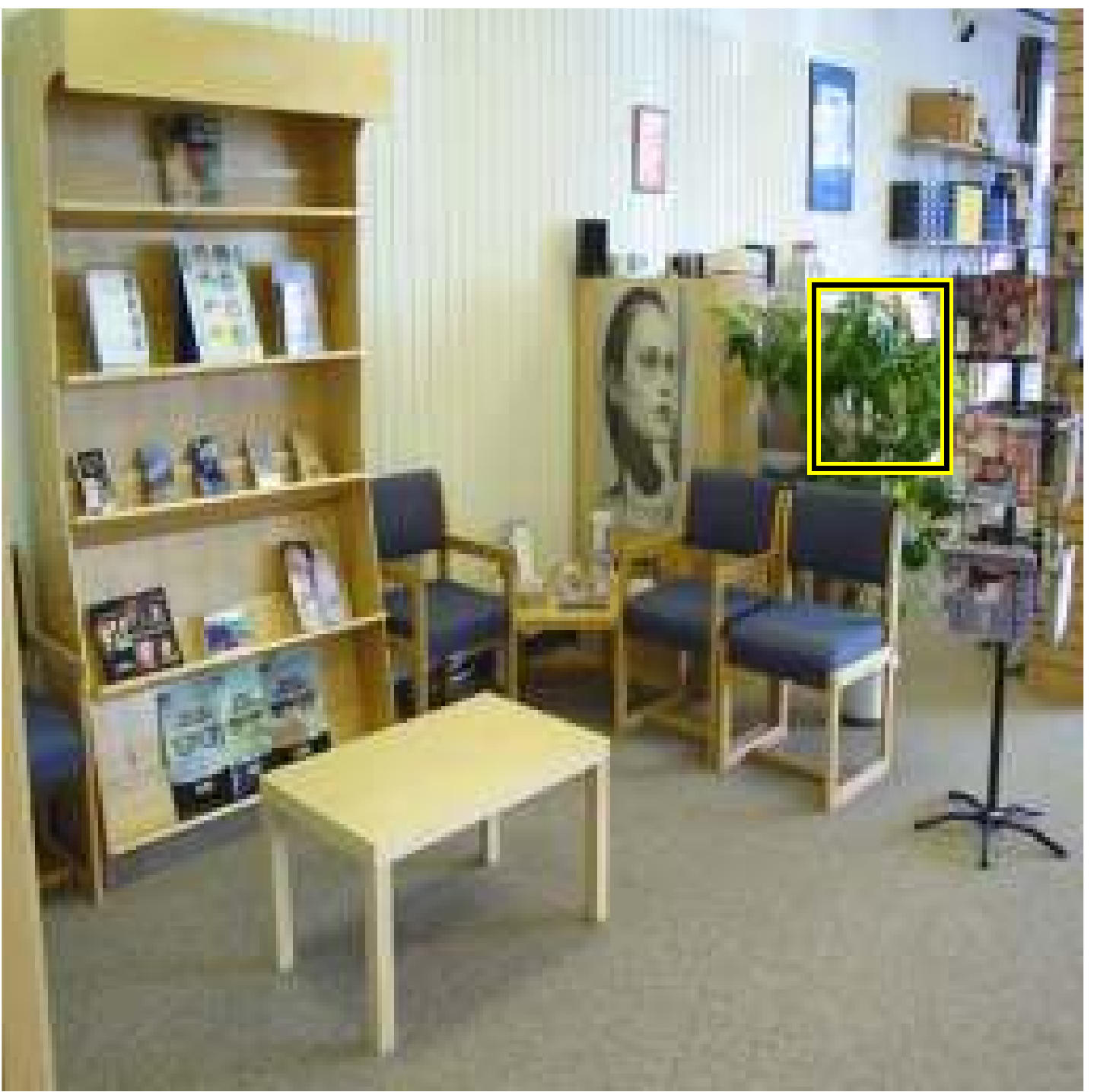} \\ [-0.05cm]
	\rotatebox{90}{\hspace{0.27cm} Part 9}$\;$ &
				\includegraphics[height=0.65in, width=0.85in]{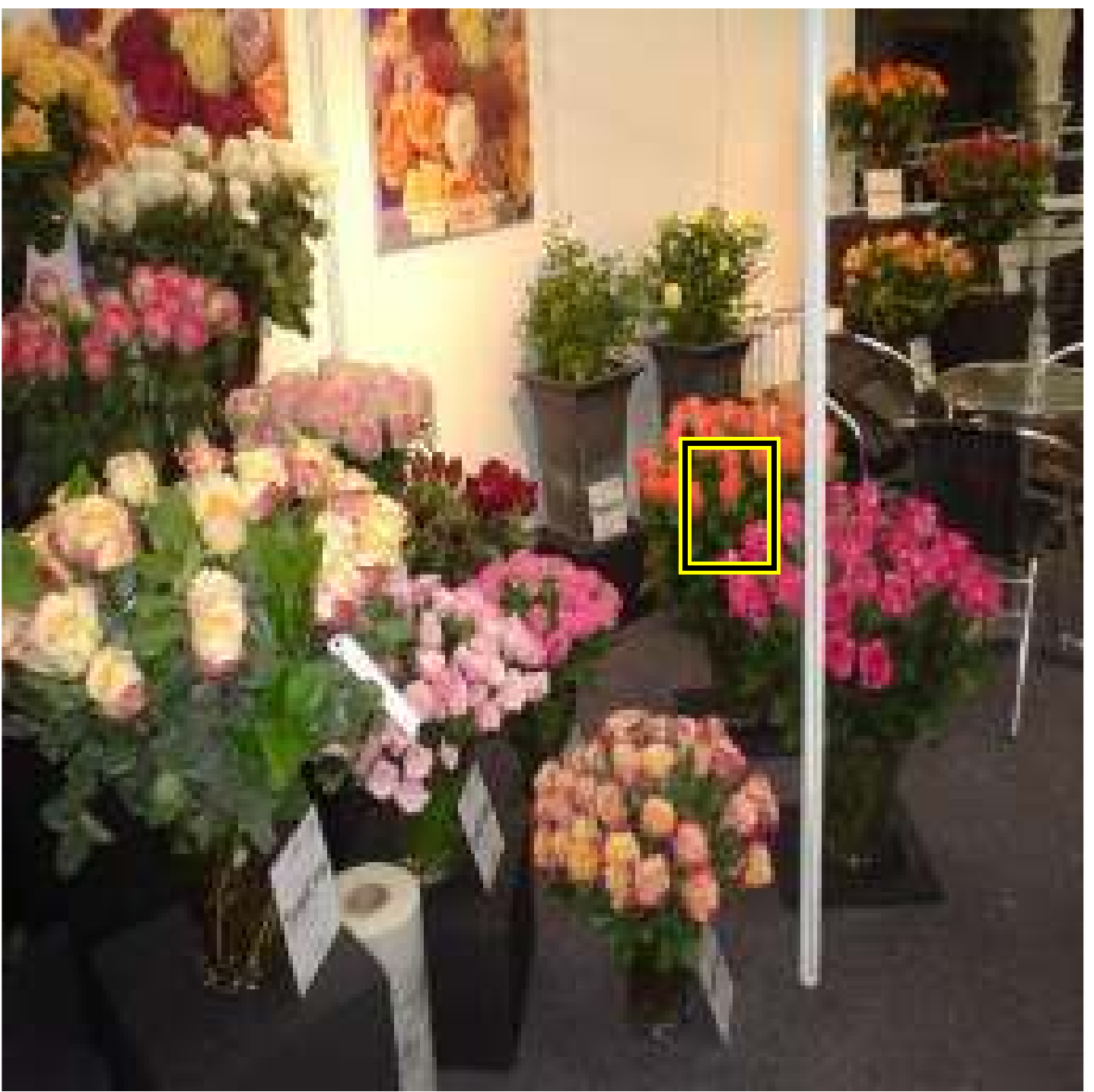} &
				\includegraphics[height=0.65in, width=0.85in]{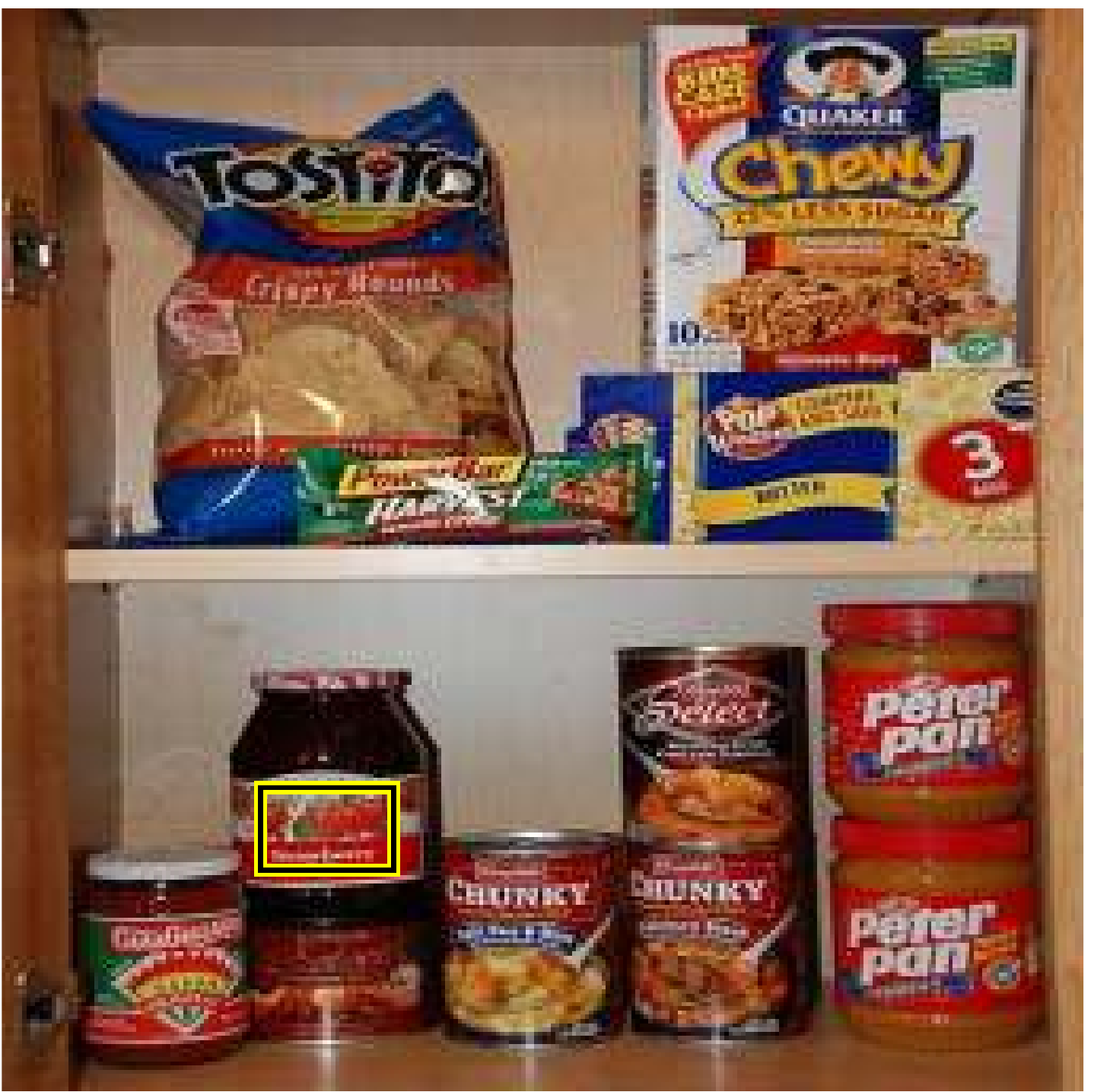} &
				\includegraphics[height=0.65in, width=0.85in]{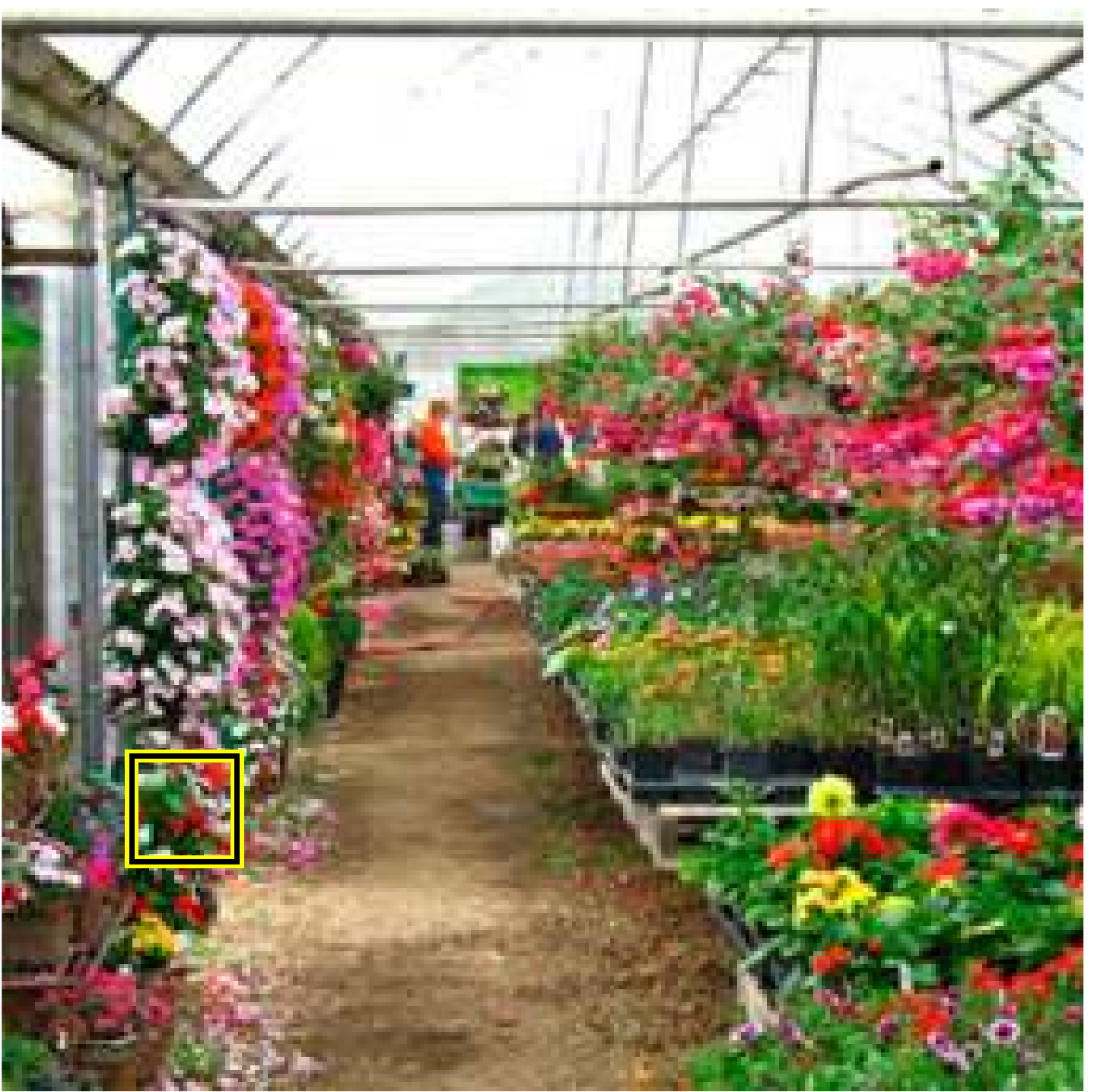} &
				\includegraphics[height=0.65in, width=0.85in]{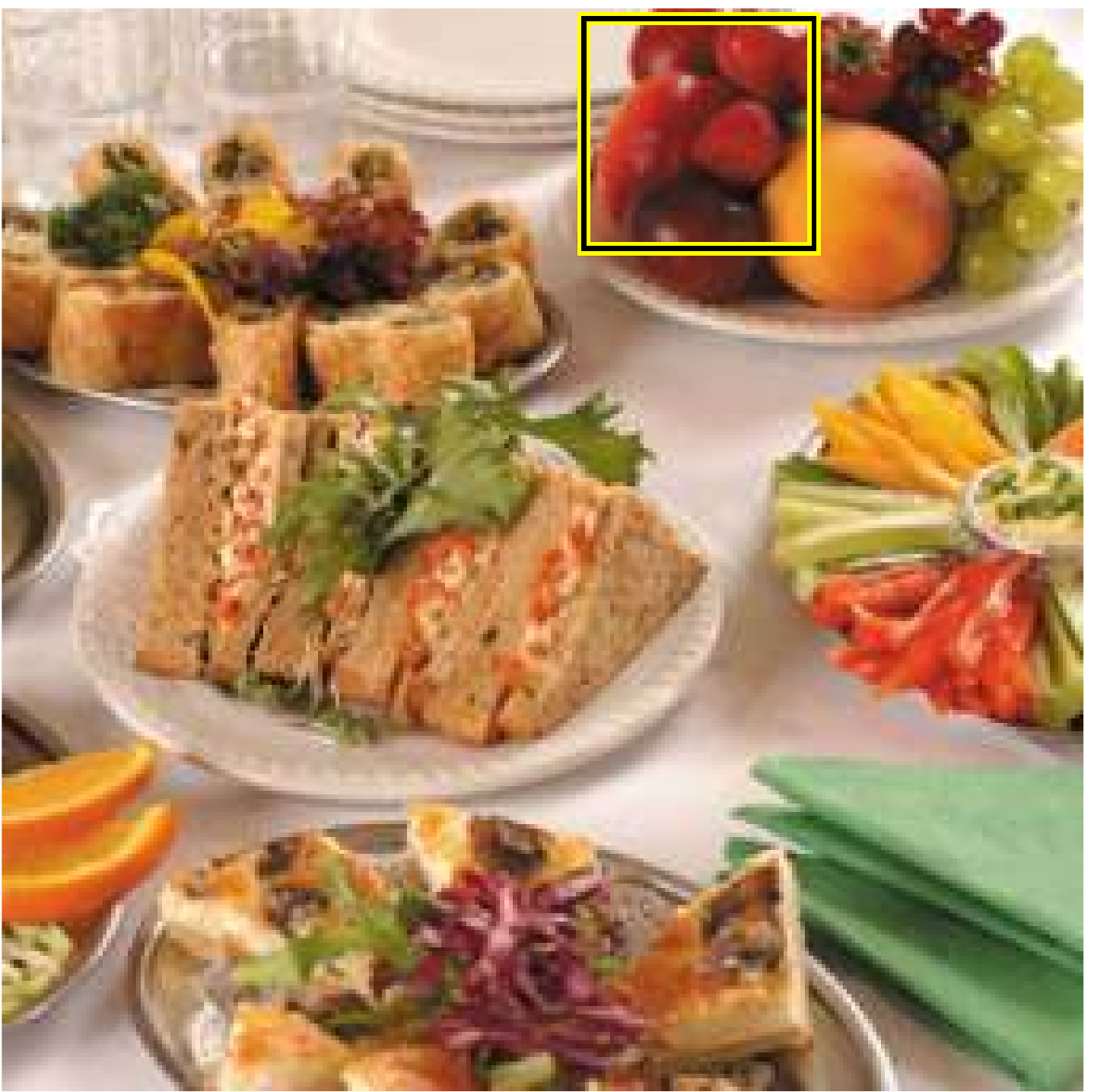} &
				\includegraphics[height=0.65in, width=0.85in]{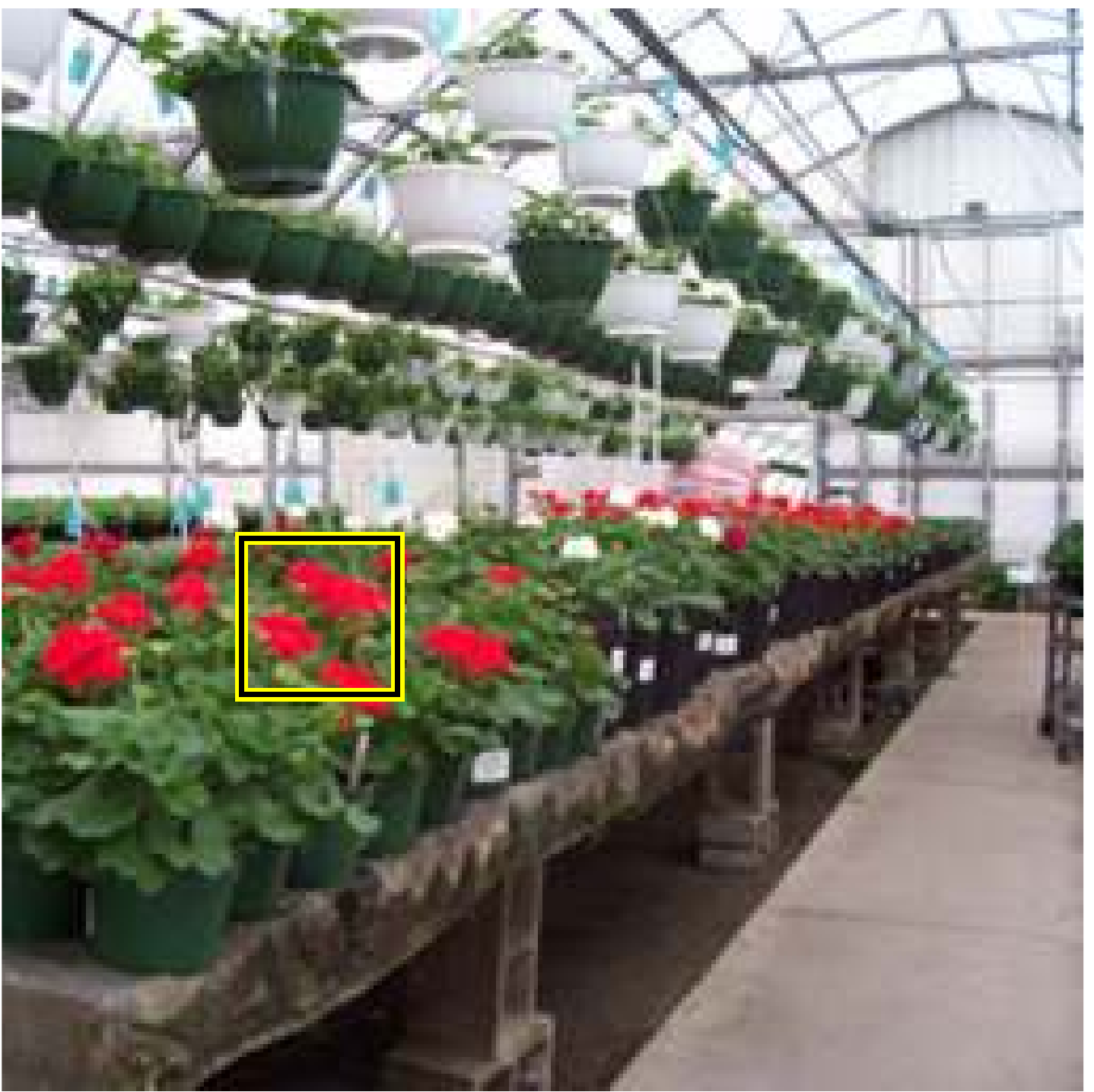} &
				\includegraphics[height=0.65in, width=0.85in]{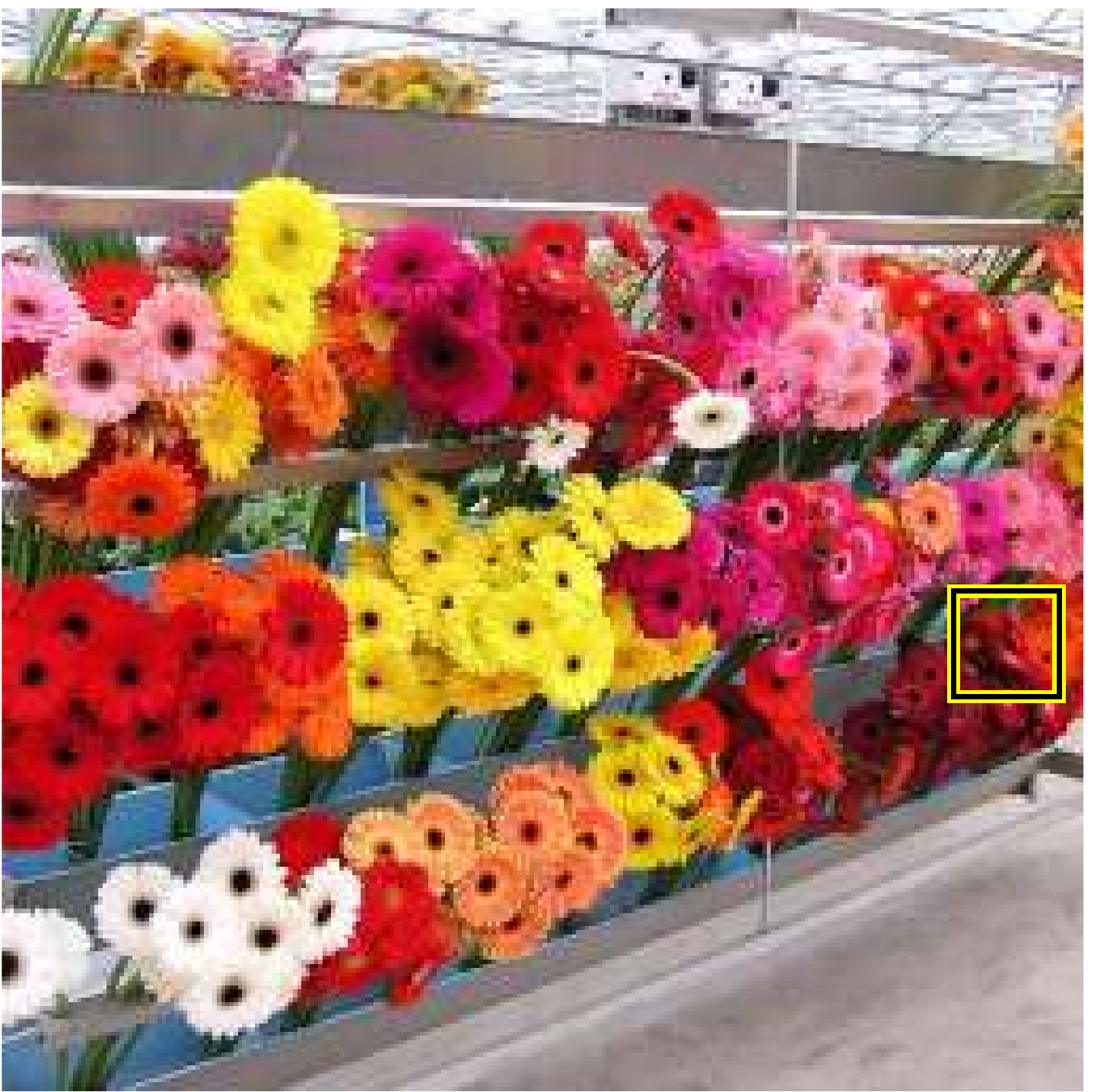} \\ [-0.05cm]
%	\rotatebox{90}{\hspace{0.27cm} Part 11}$\;$ &
%				\includegraphics[height=0.65in, width=0.85in]{images/CNN/placements_compressed/part-index-011_rank-01_image-index-0733} &
%				\includegraphics[height=0.65in, width=0.85in]{images/CNN/placements_compressed/part-index-011_rank-02_image-index-0541} &
%				\includegraphics[height=0.65in, width=0.85in]{images/CNN/placements_compressed/part-index-011_rank-03_image-index-0318} &
%				\includegraphics[height=0.65in, width=0.85in]{images/CNN/placements_compressed/part-index-011_rank-04_image-index-0232} &
%				\includegraphics[height=0.65in, width=0.85in]{images/CNN/placements_compressed/part-index-011_rank-06_image-index-1305} &
%				\includegraphics[height=0.65in, width=0.85in]{images/CNN/placements_compressed/part-index-011_rank-08_image-index-0695} \\ [-0.05cm]
	\rotatebox{90}{\hspace{0.27cm} Part 14}$\;$ &
				\includegraphics[height=0.65in, width=0.85in]{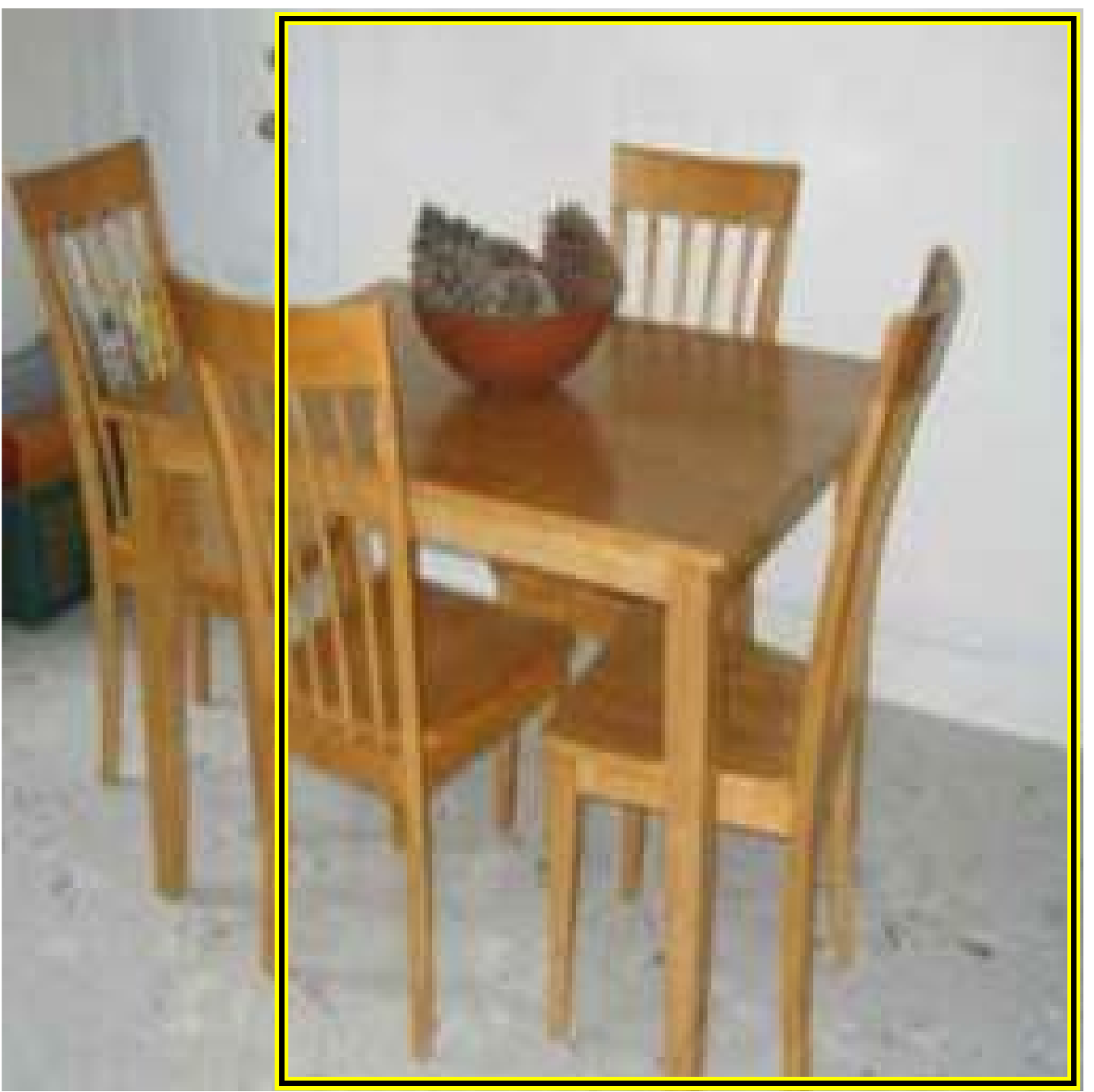} &
				\includegraphics[height=0.65in, width=0.85in]{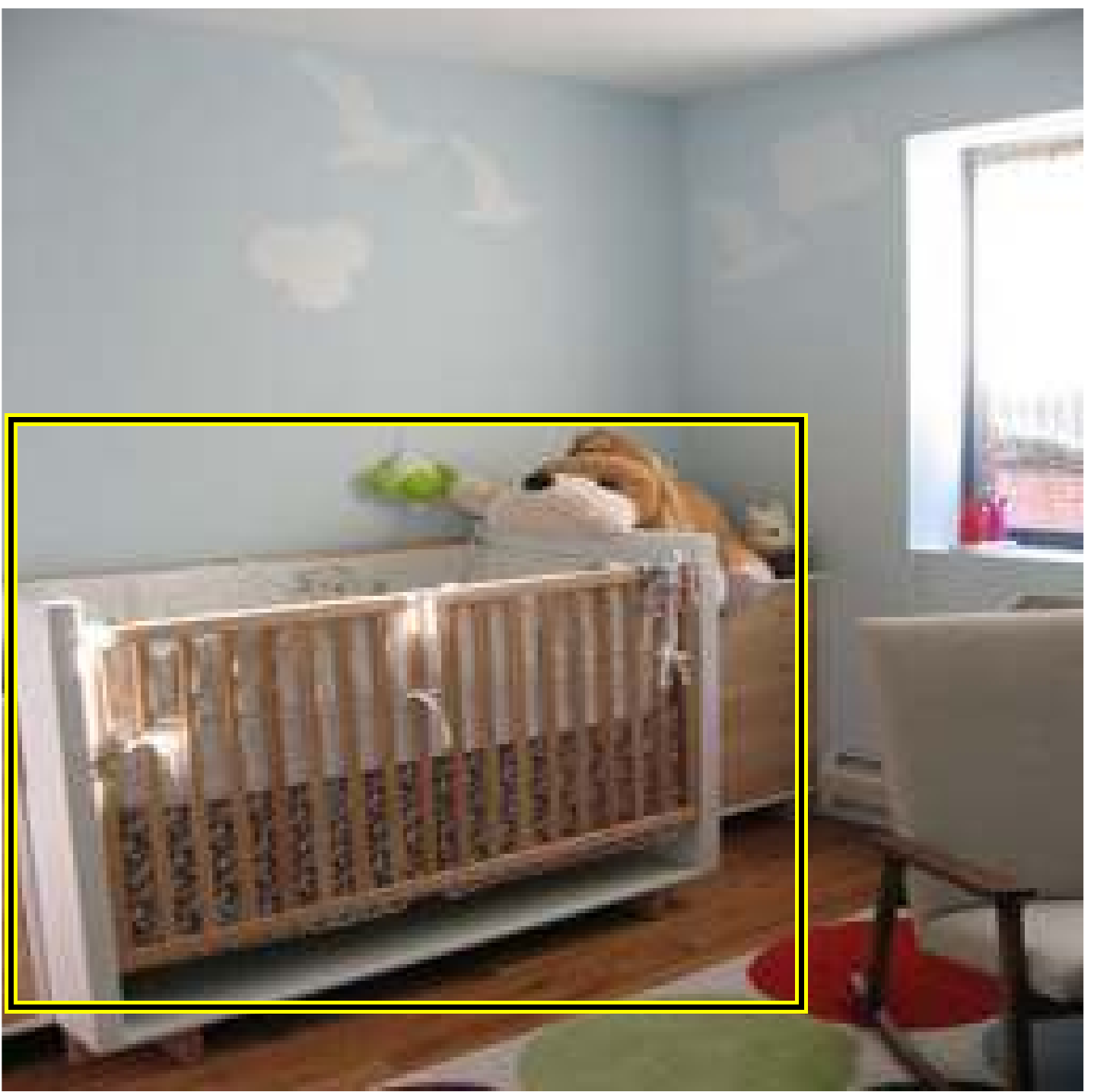} &
				\includegraphics[height=0.65in, width=0.85in]{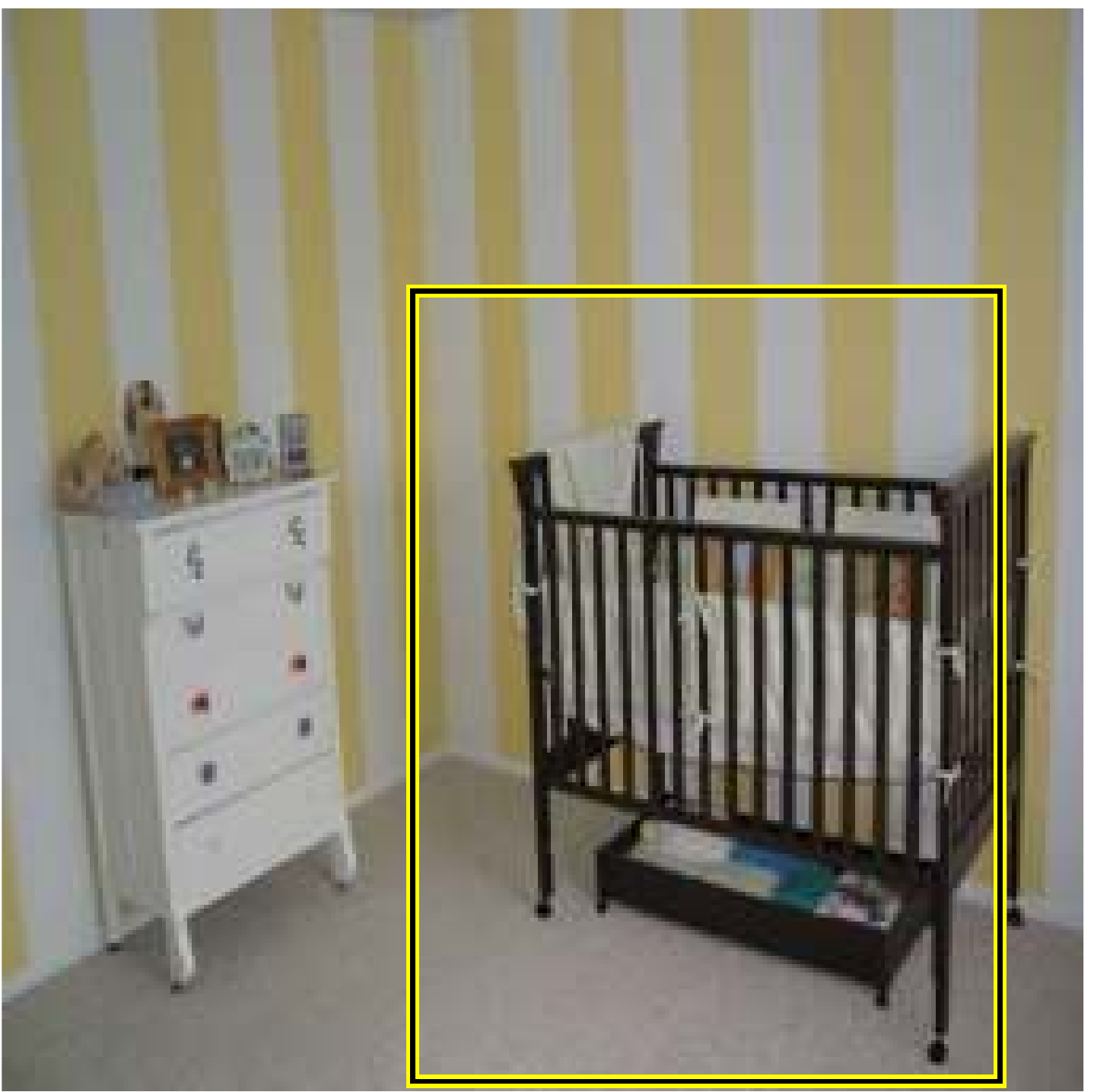} &
				\includegraphics[height=0.65in, width=0.85in]{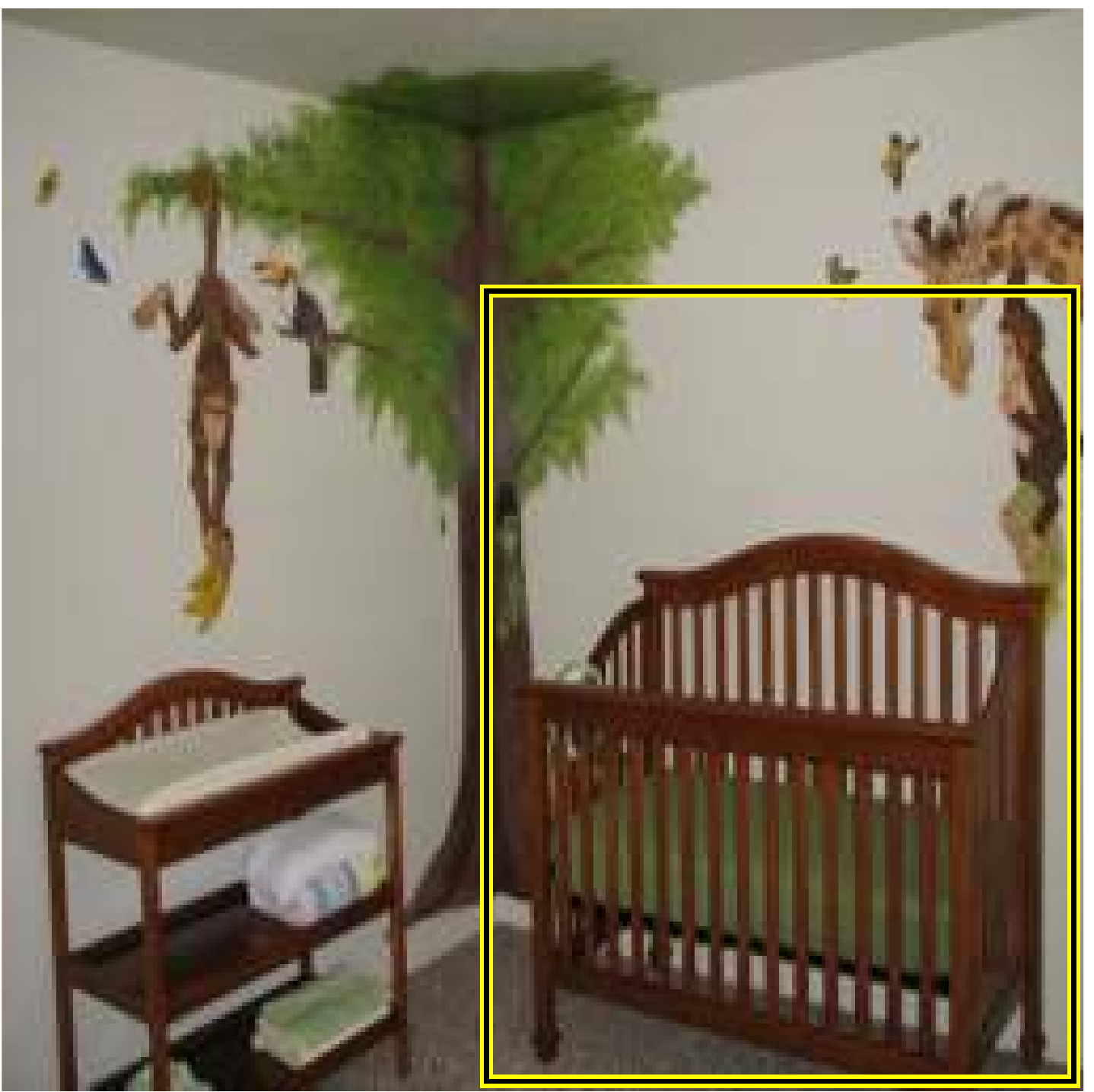} &
				\includegraphics[height=0.65in, width=0.85in]{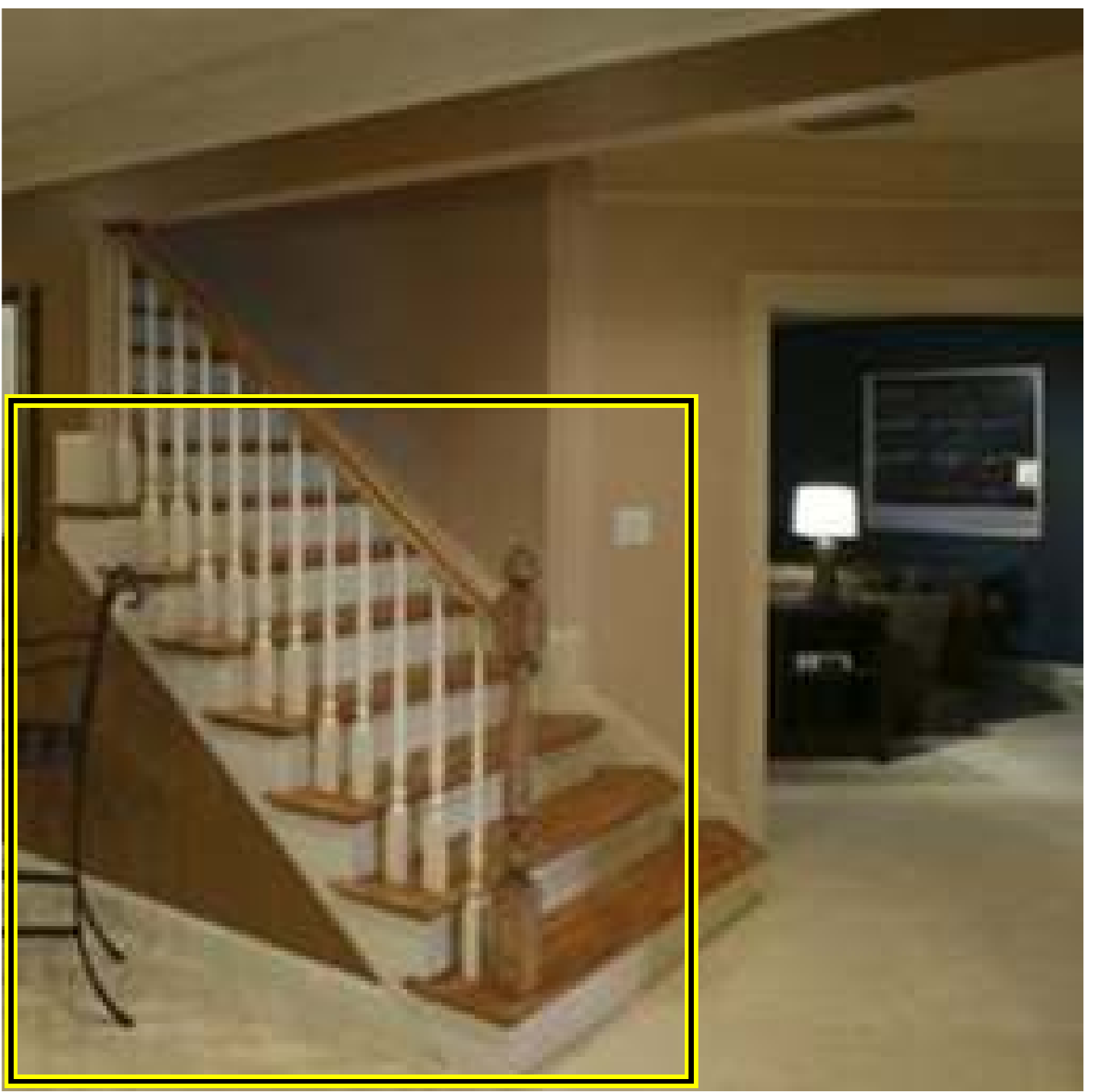} &
				\includegraphics[height=0.65in, width=0.85in]{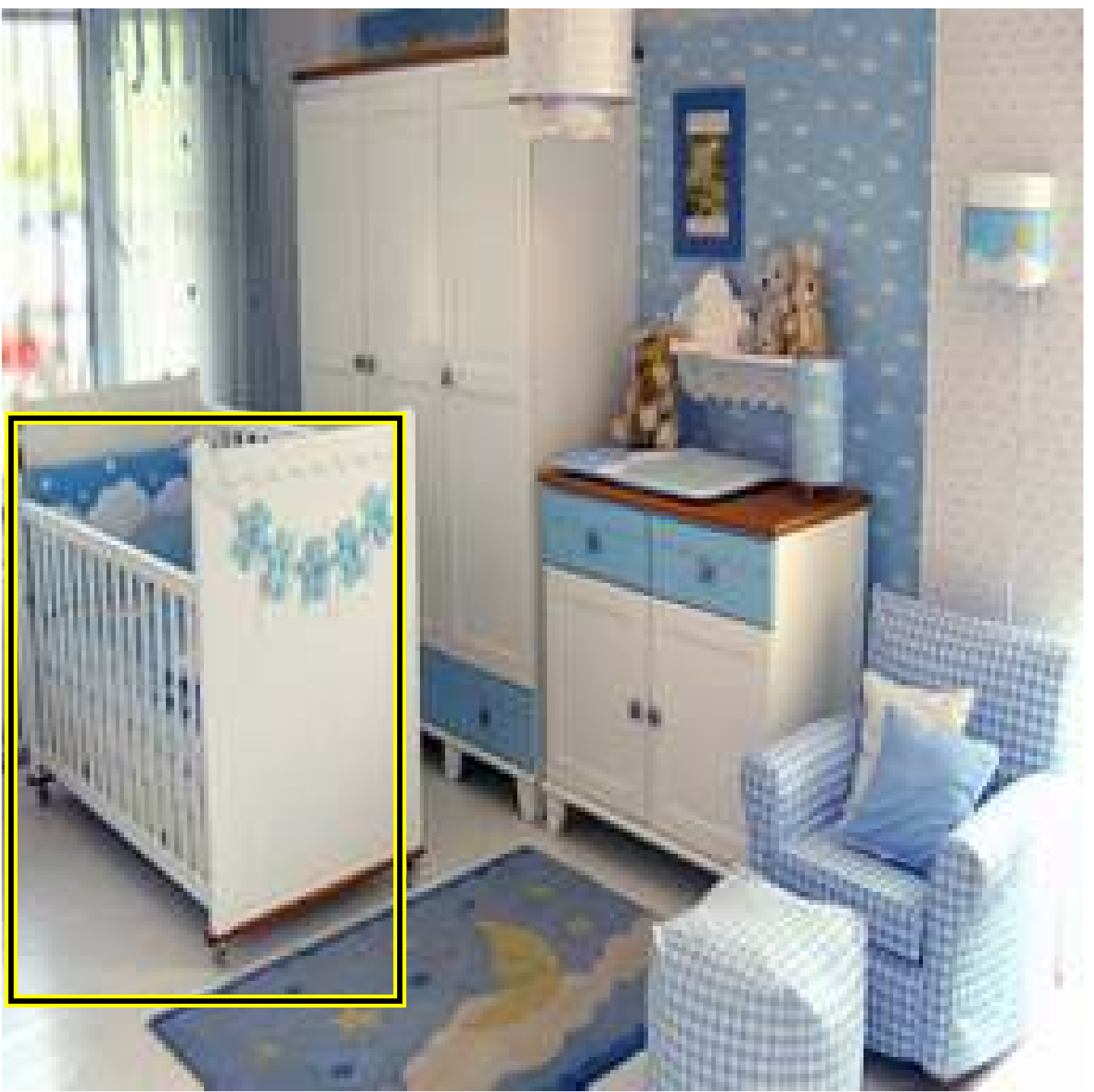} \\ [-0.05cm]
	\rotatebox{90}{\hspace{0.27cm} Part 19}$\;$ &
				\includegraphics[height=0.65in, width=0.85in]{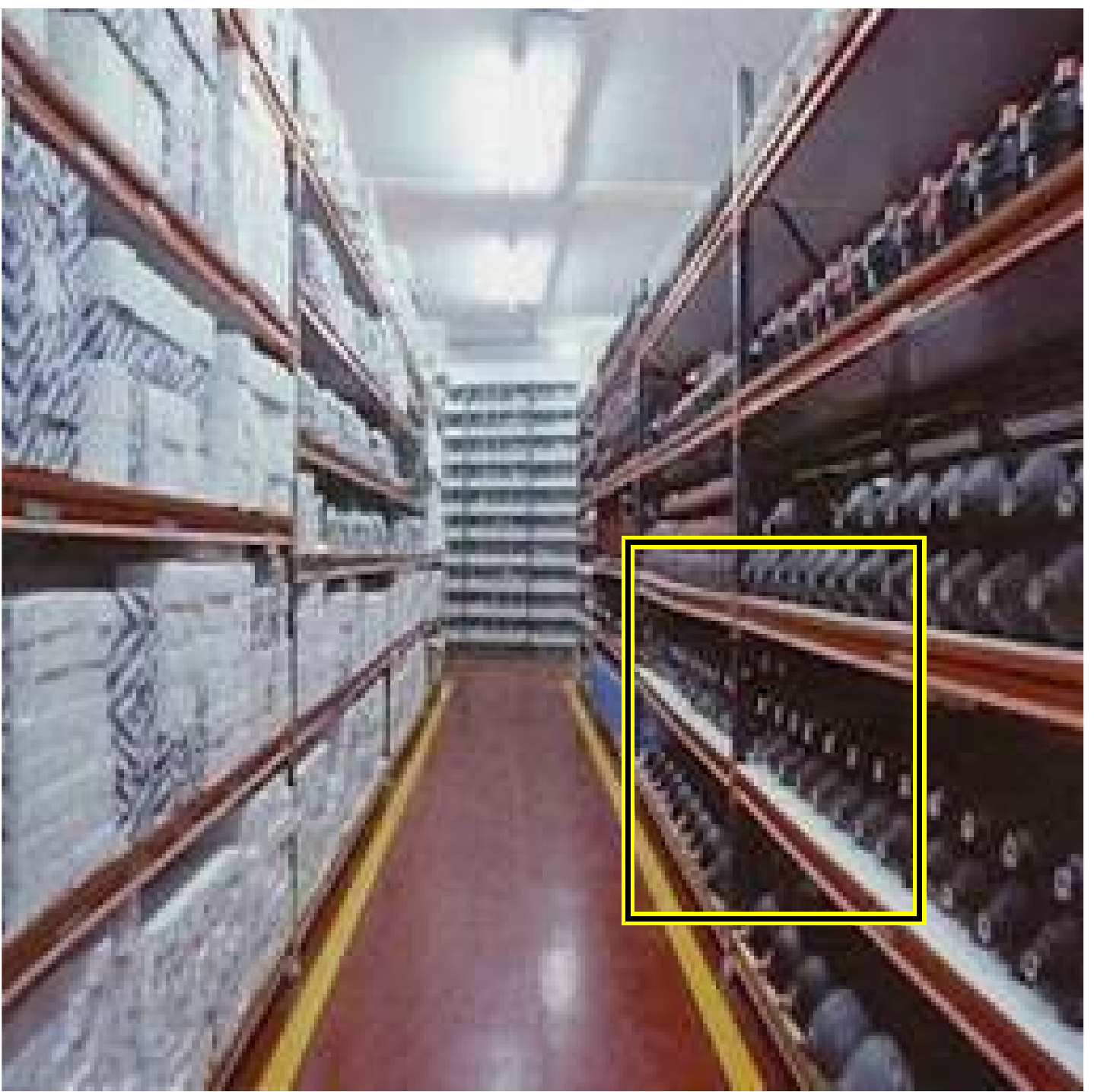} &
				\includegraphics[height=0.65in, width=0.85in]{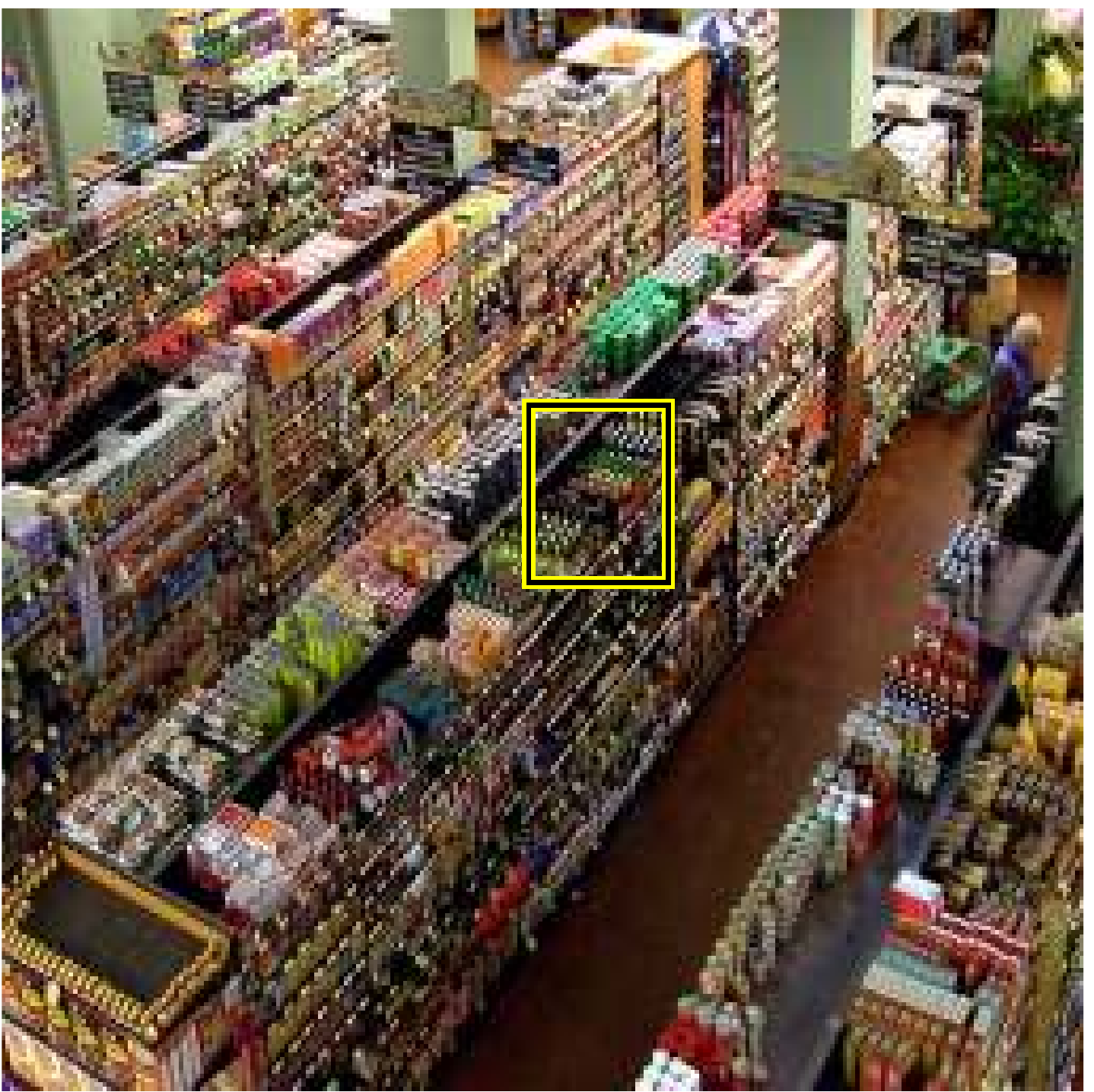} &
				\includegraphics[height=0.65in, width=0.85in]{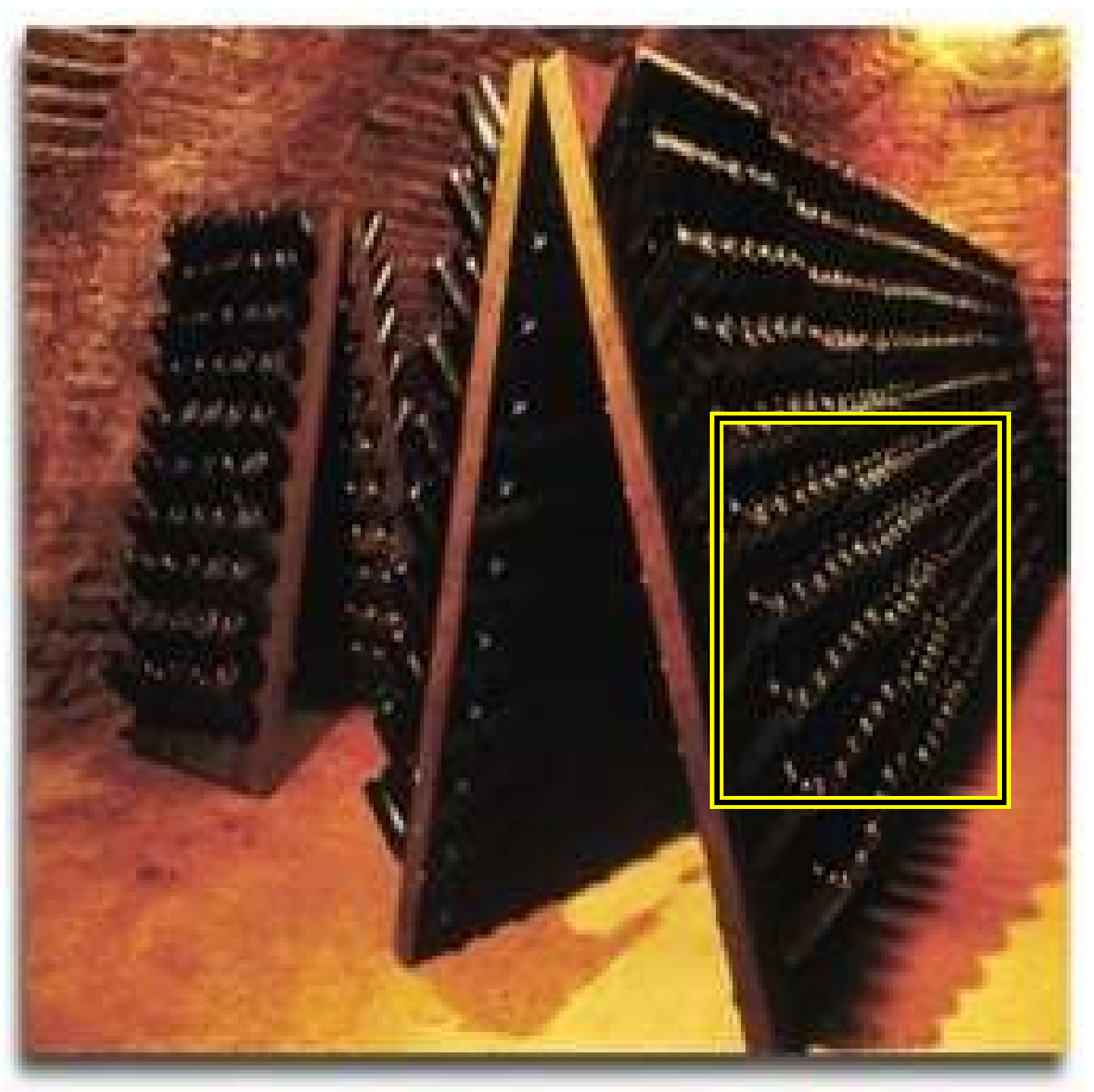} &
				\includegraphics[height=0.65in, width=0.85in]{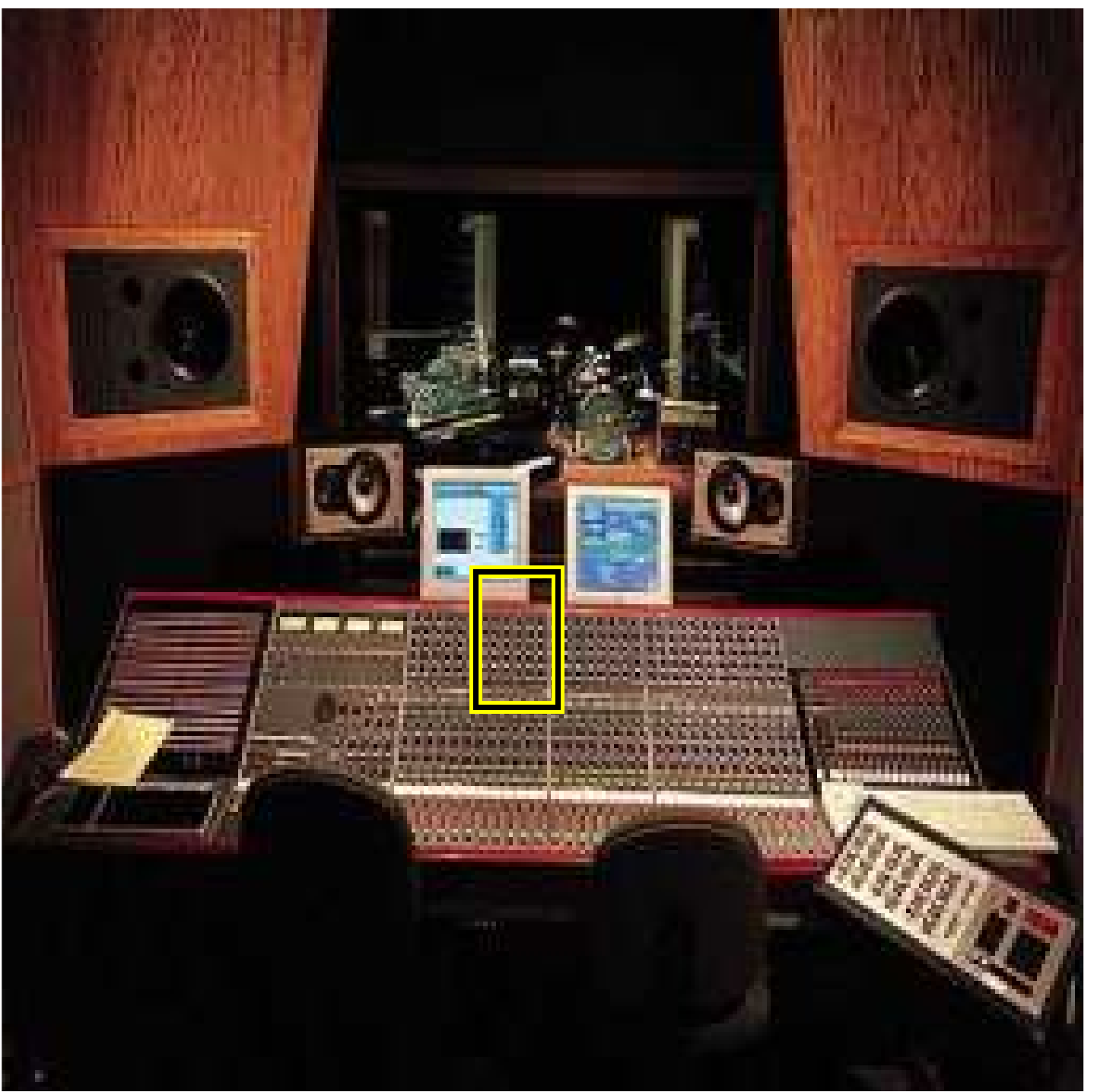} &
				\includegraphics[height=0.65in, width=0.85in]{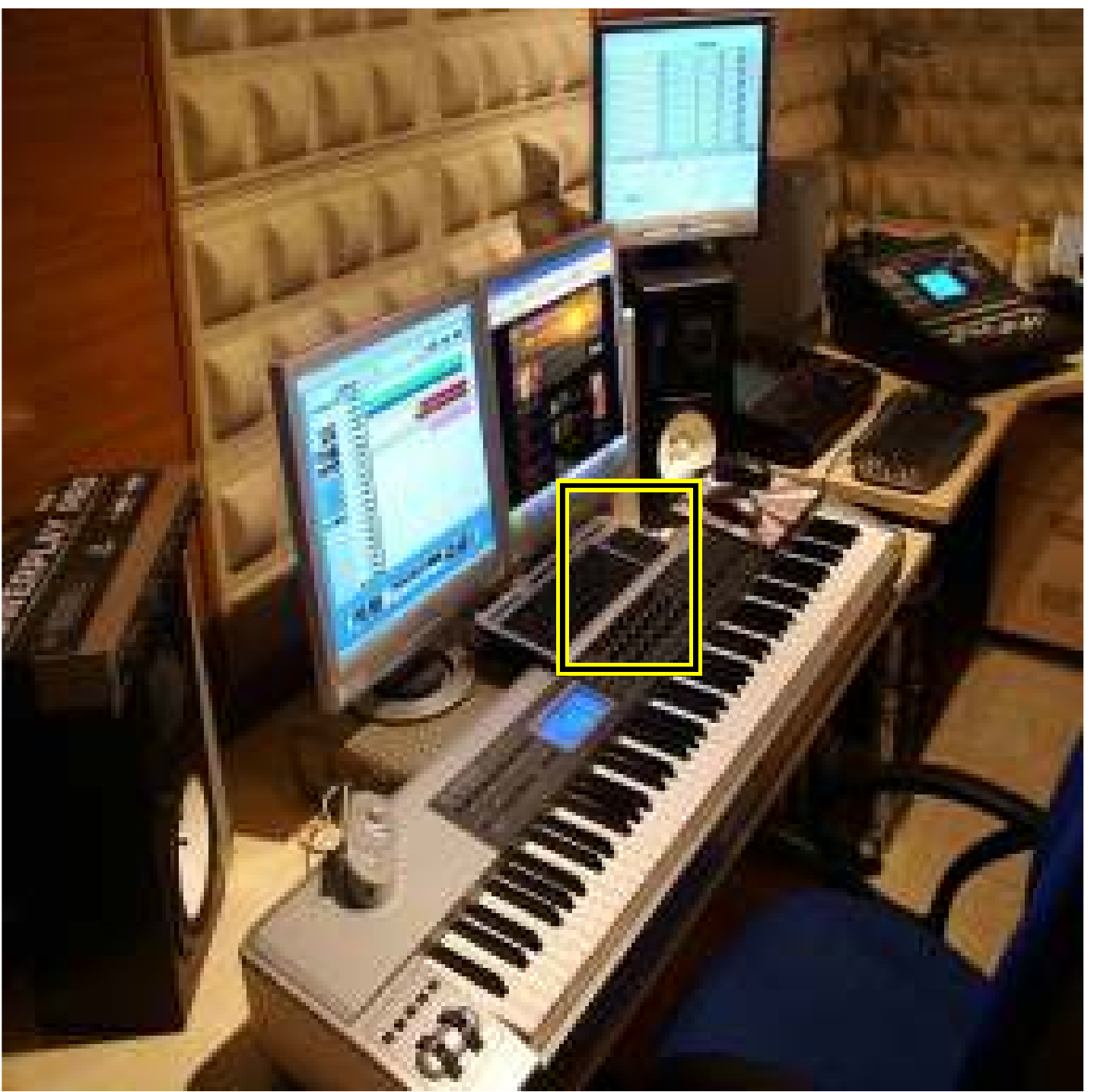} &
				\includegraphics[height=0.65in, width=0.85in]{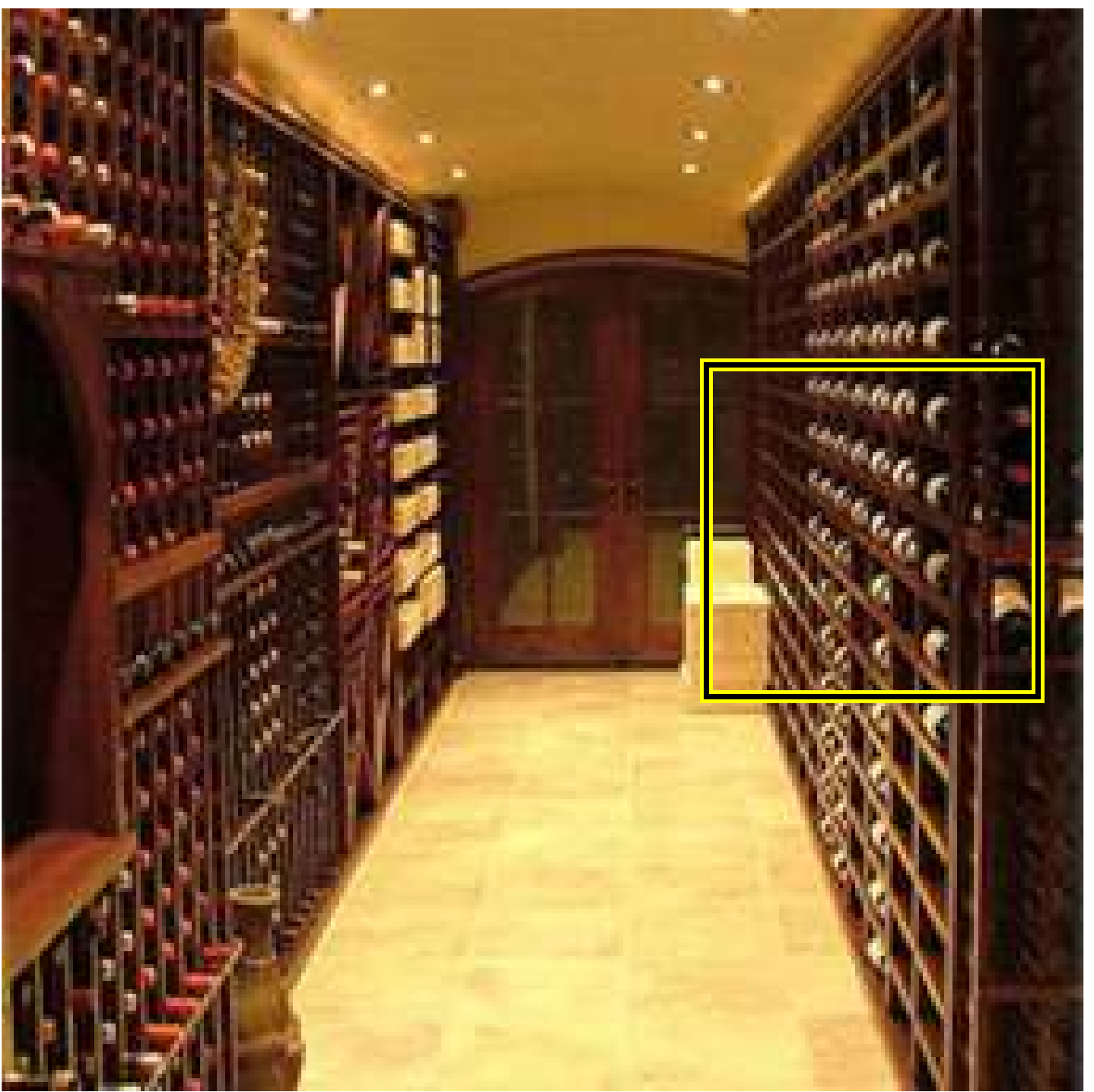} \\ [-0.05cm]
	\rotatebox{90}{\hspace{0.27cm}Part 21}$\;$ &
				\includegraphics[height=0.65in, width=0.85in]{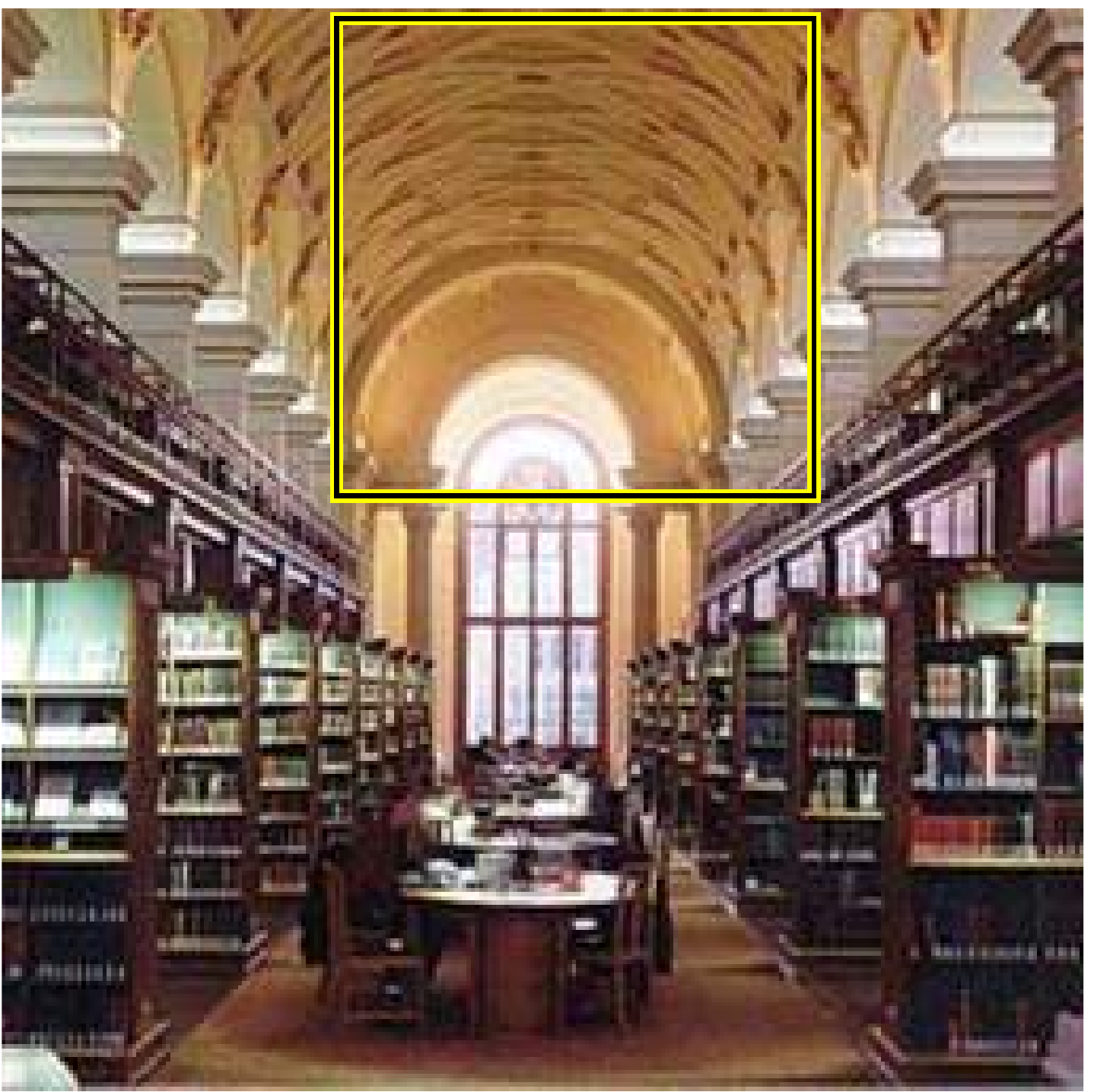} &
				\includegraphics[height=0.65in, width=0.85in]{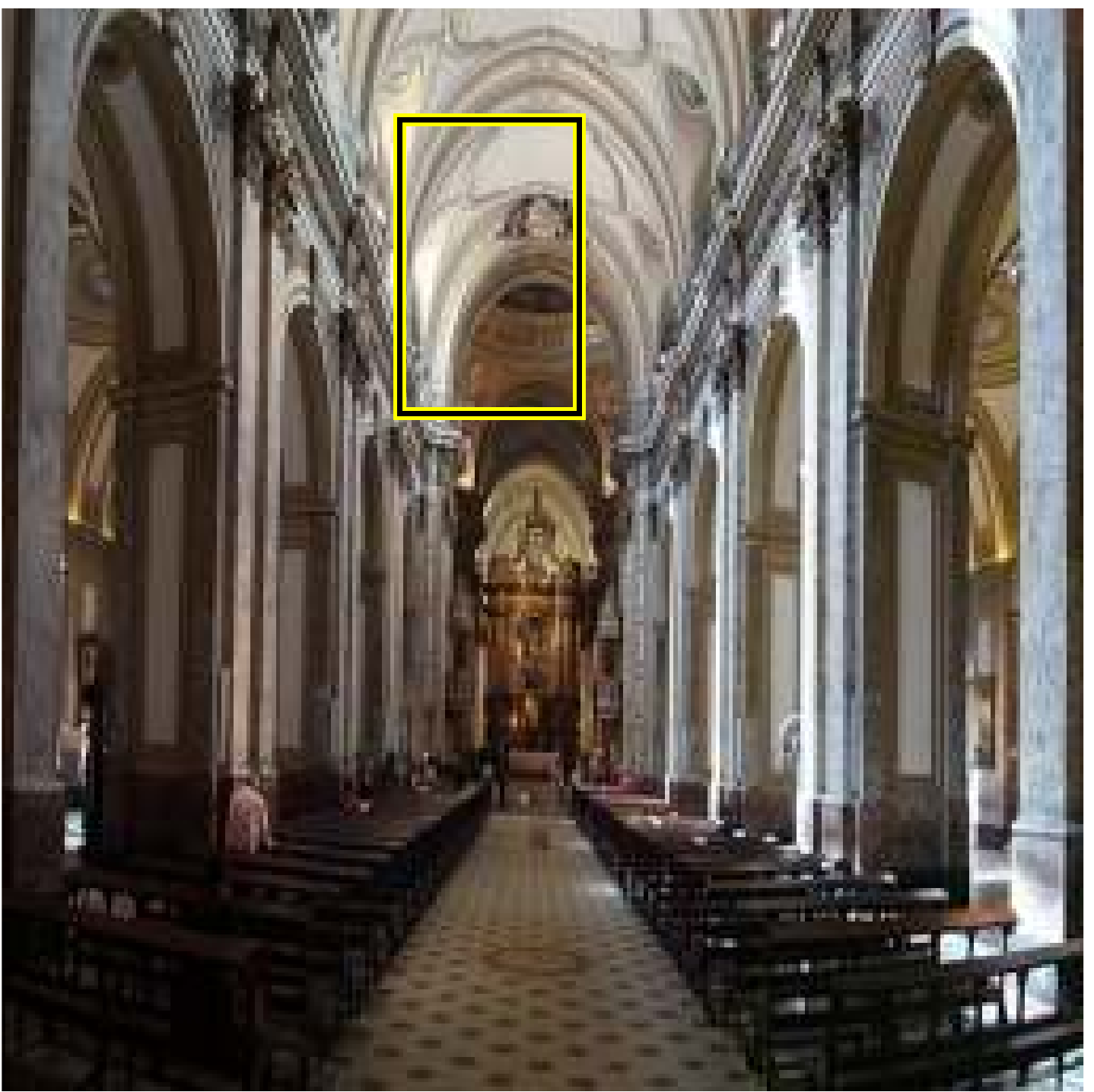} &
				\includegraphics[height=0.65in, width=0.85in]{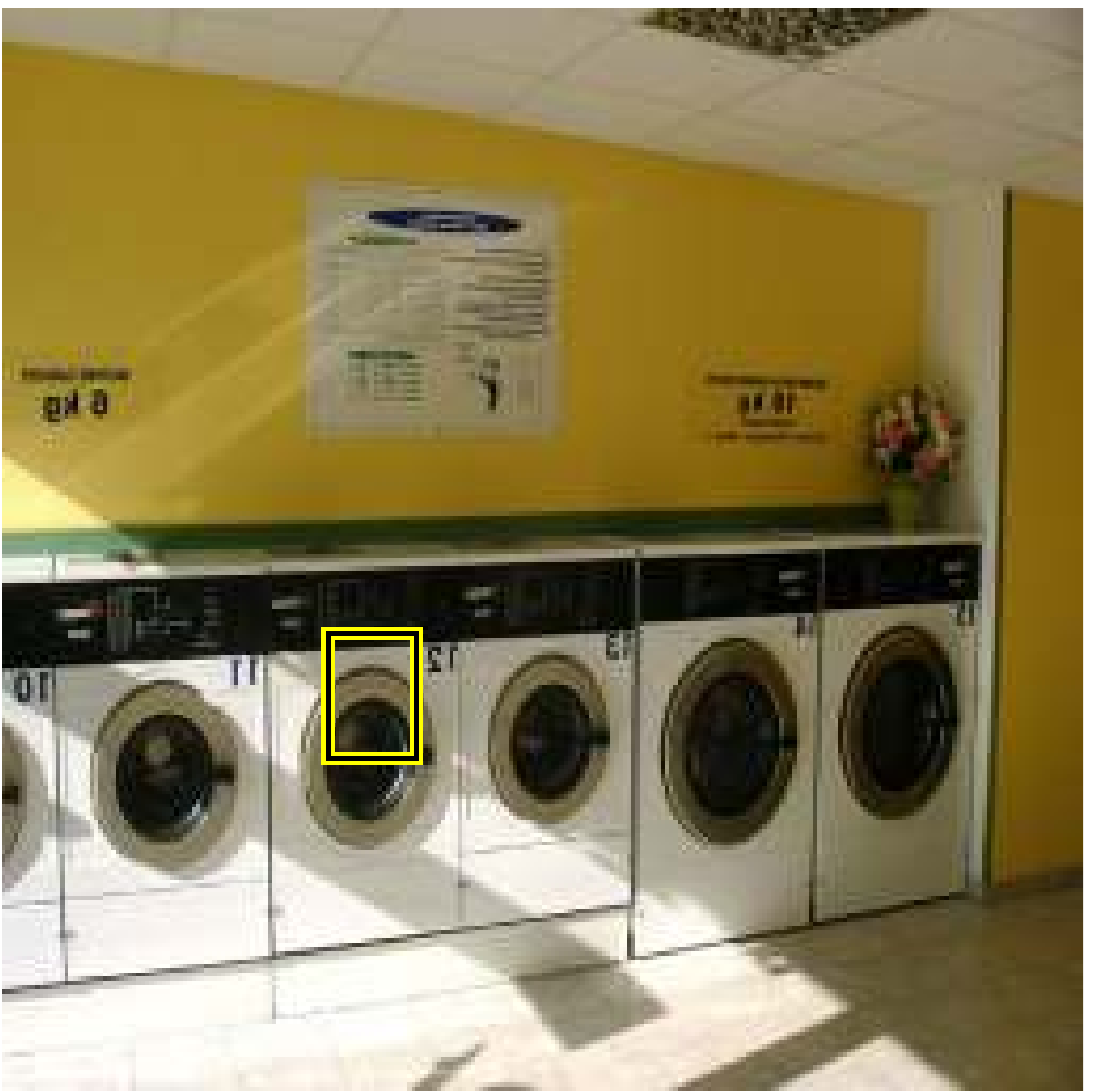} &
				\includegraphics[height=0.65in, width=0.85in]{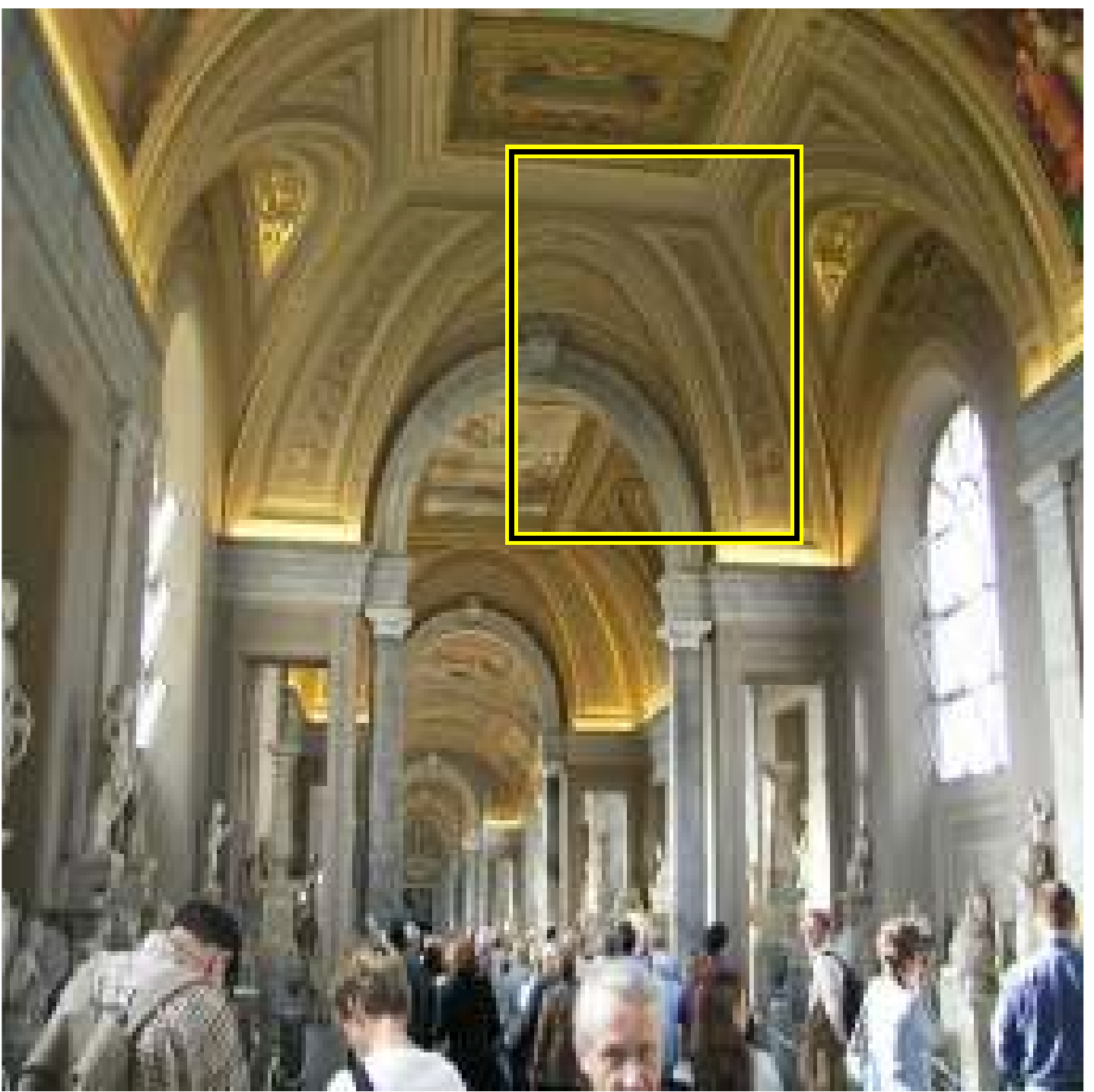} &
				\includegraphics[height=0.65in, width=0.85in]{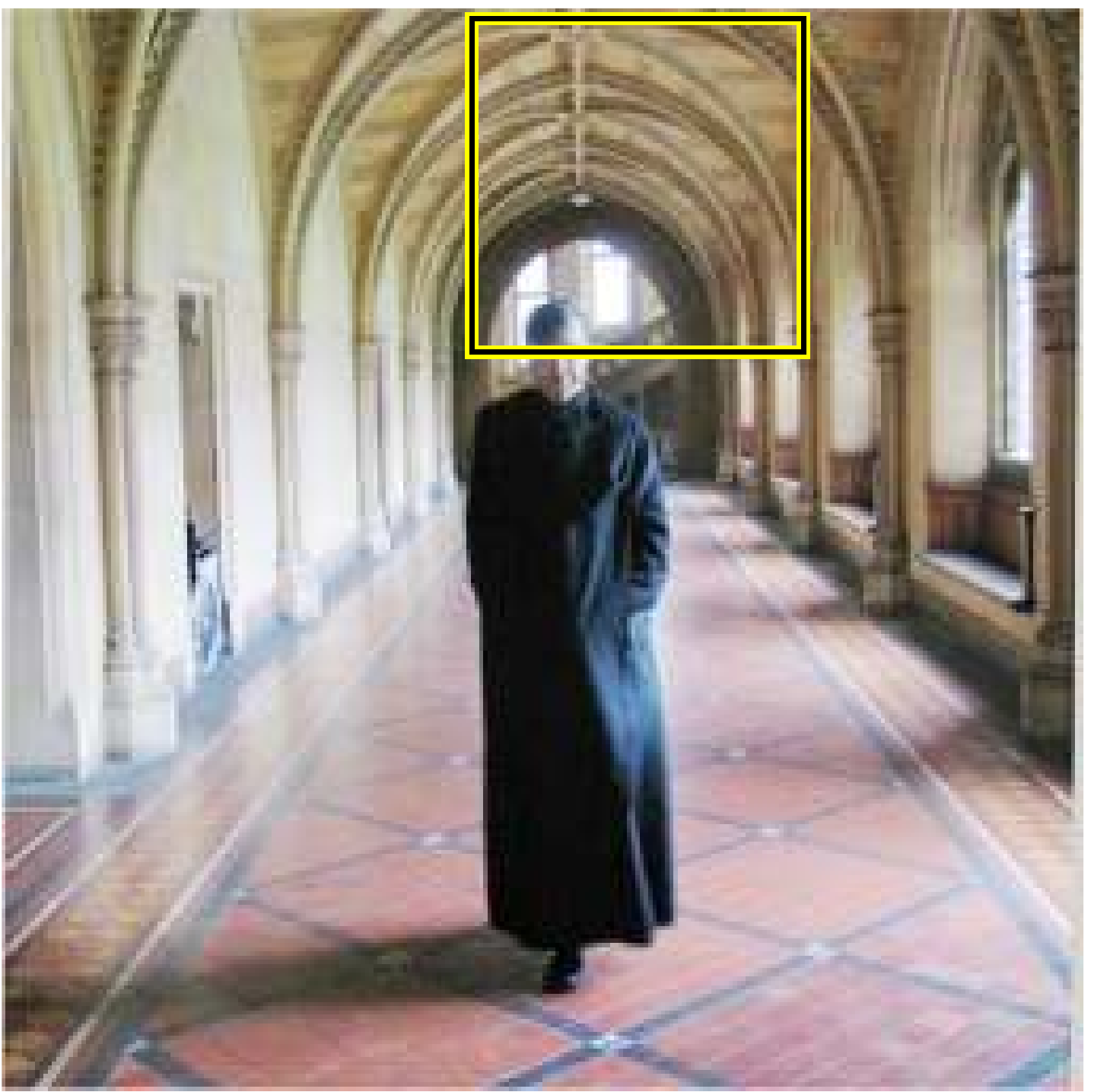} &
				\includegraphics[height=0.65in, width=0.85in]{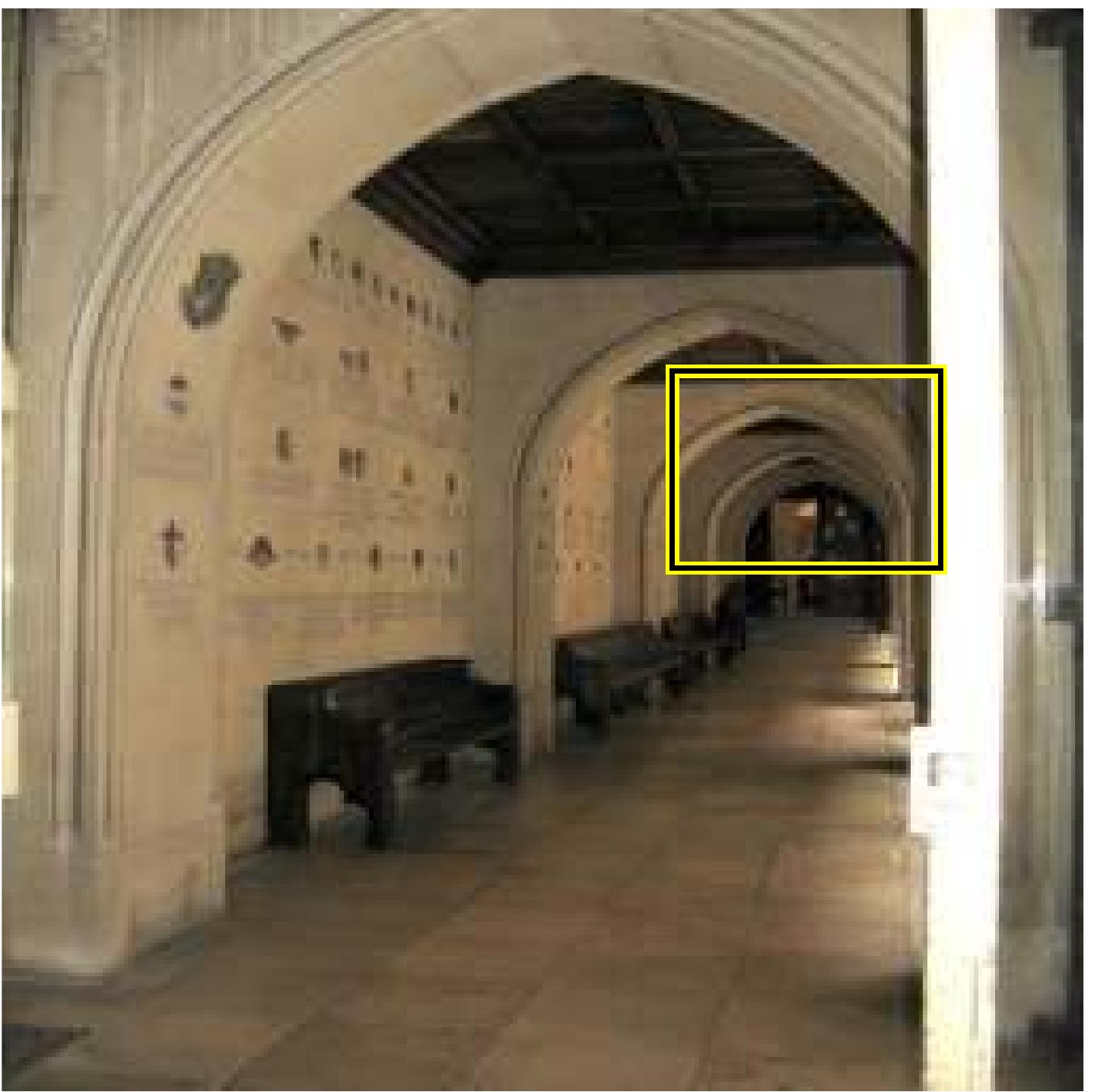} \\ [-0.05cm]
	\rotatebox{90}{\hspace{0.27cm}Part 22}$\;$ &
				\includegraphics[height=0.65in, width=0.85in]{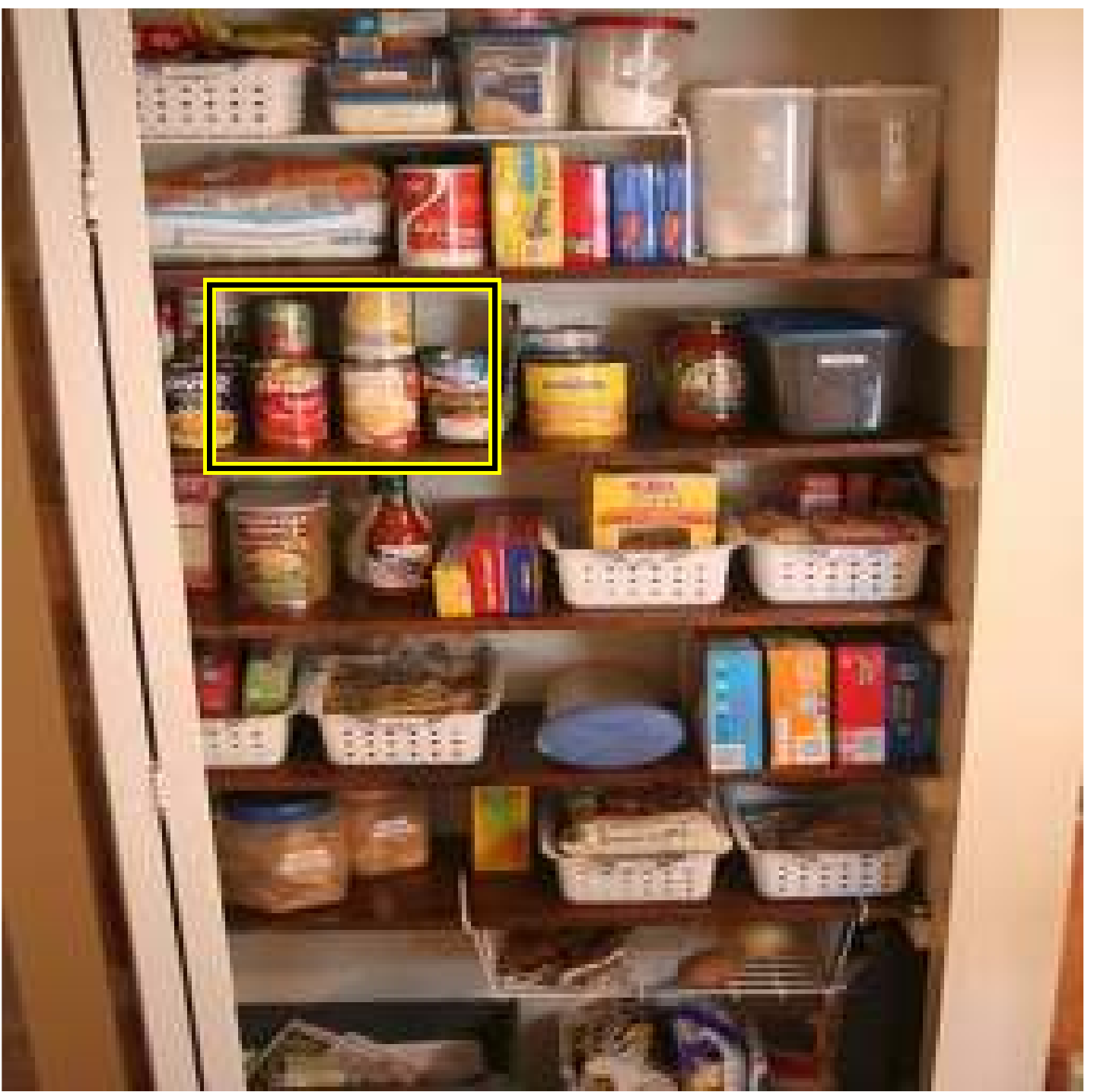} &
				\includegraphics[height=0.65in, width=0.85in]{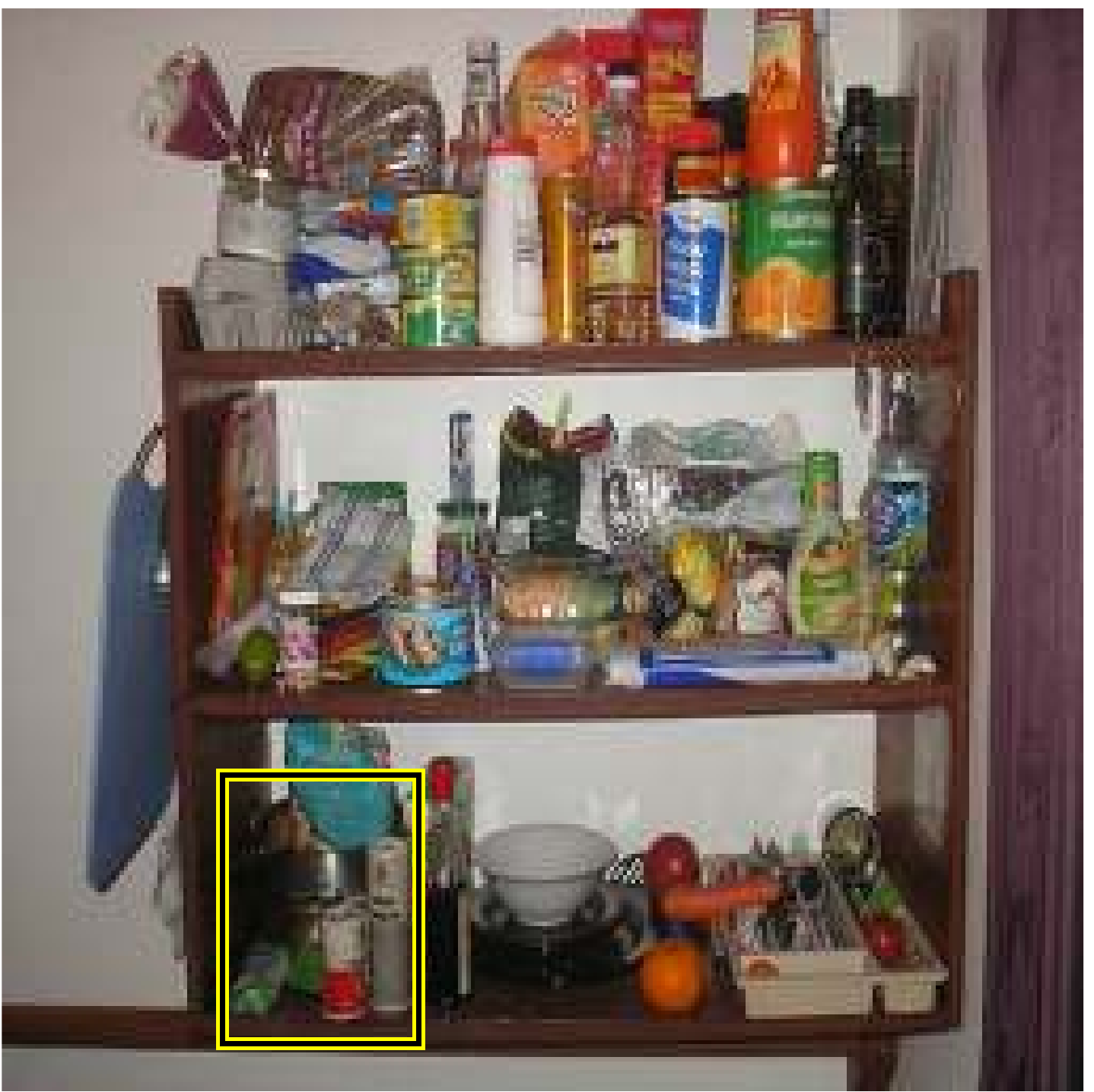} &
				\includegraphics[height=0.65in, width=0.85in]{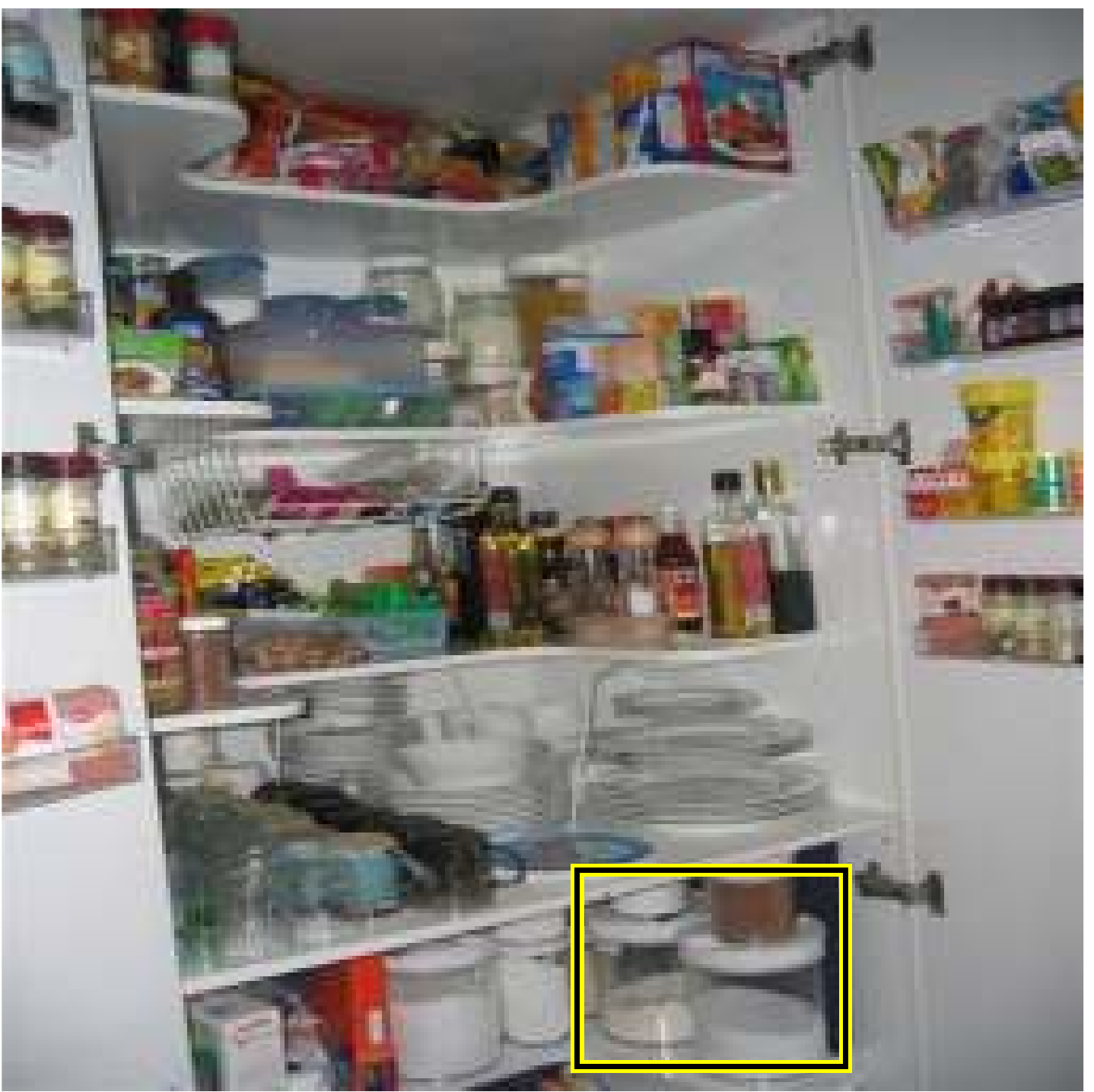} &
				\includegraphics[height=0.65in, width=0.85in]{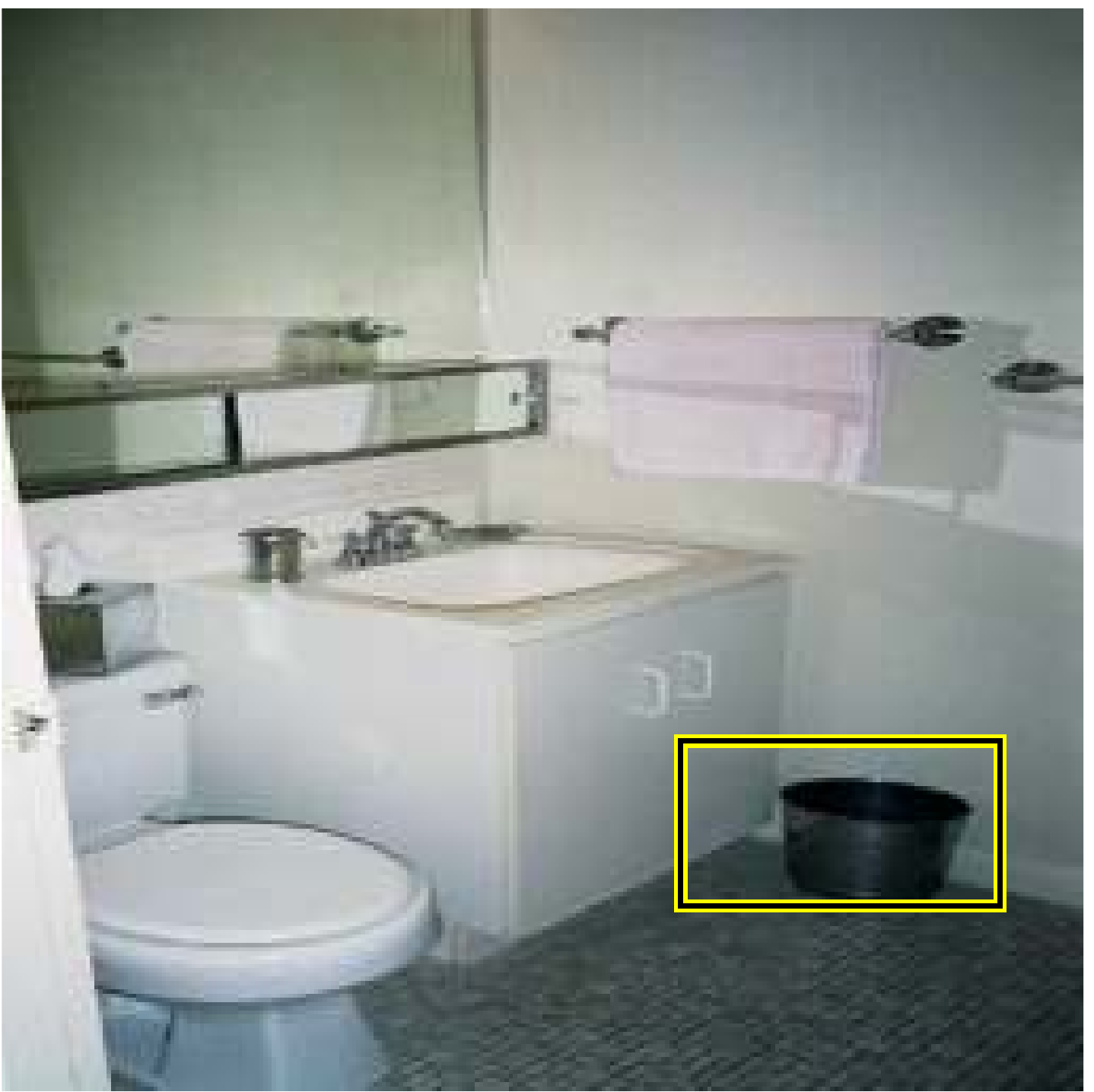} &
				\includegraphics[height=0.65in, width=0.85in]{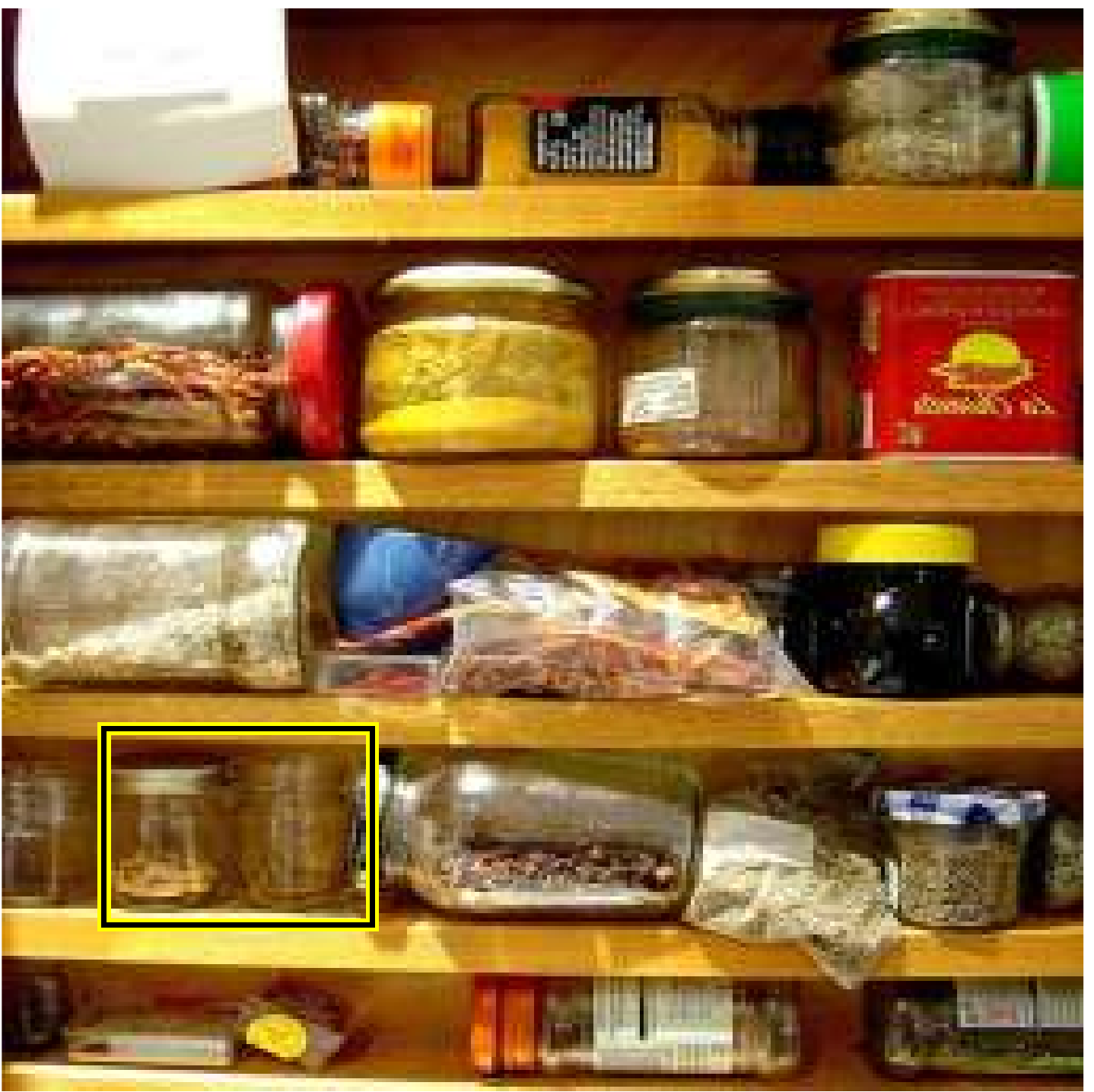} &
				\includegraphics[height=0.65in, width=0.85in]{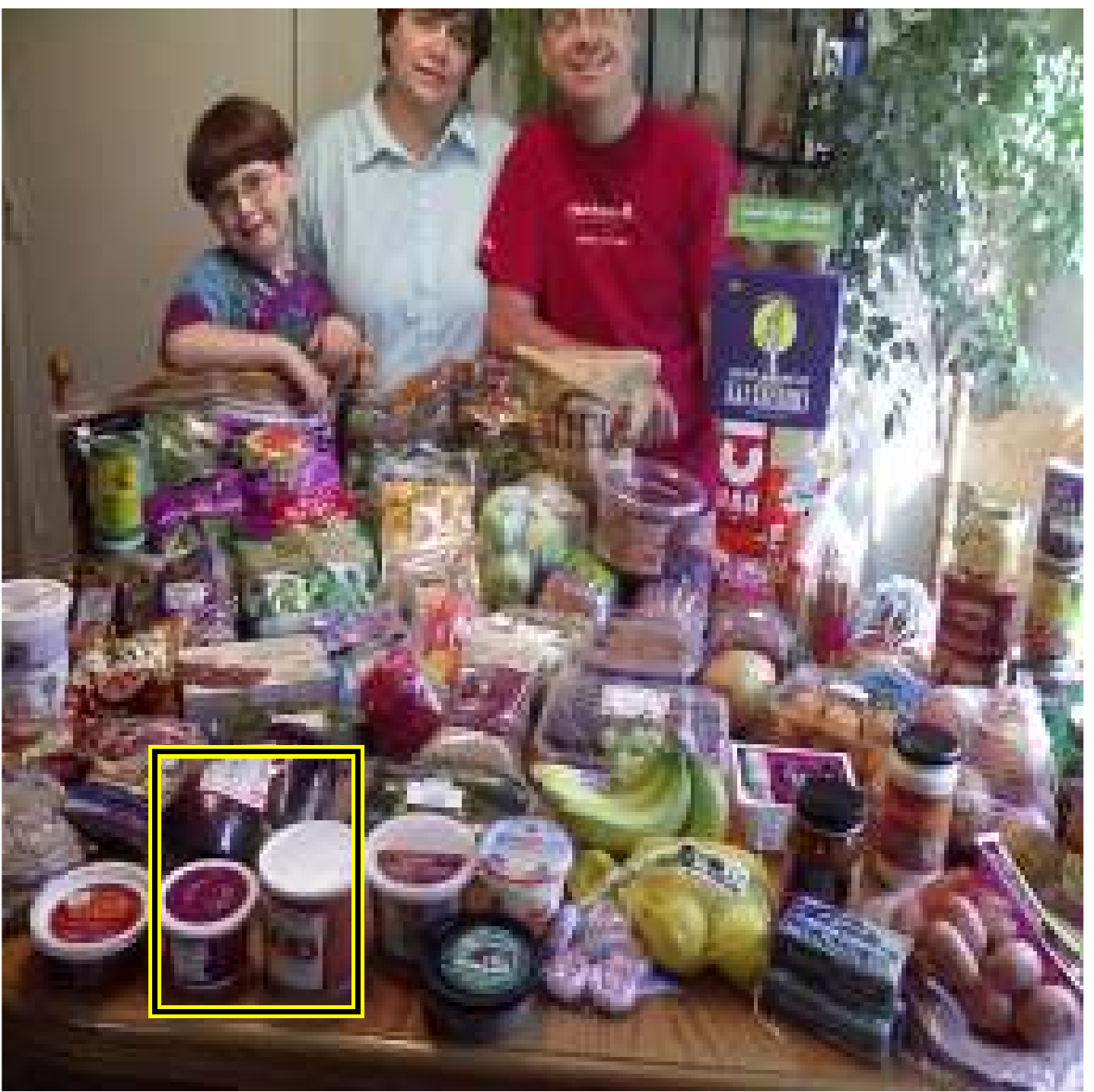} \\ [-0.05cm]
	\rotatebox{90}{\hspace{0.27cm}Part 31}$\;$ &
				\includegraphics[height=0.65in, width=0.85in]{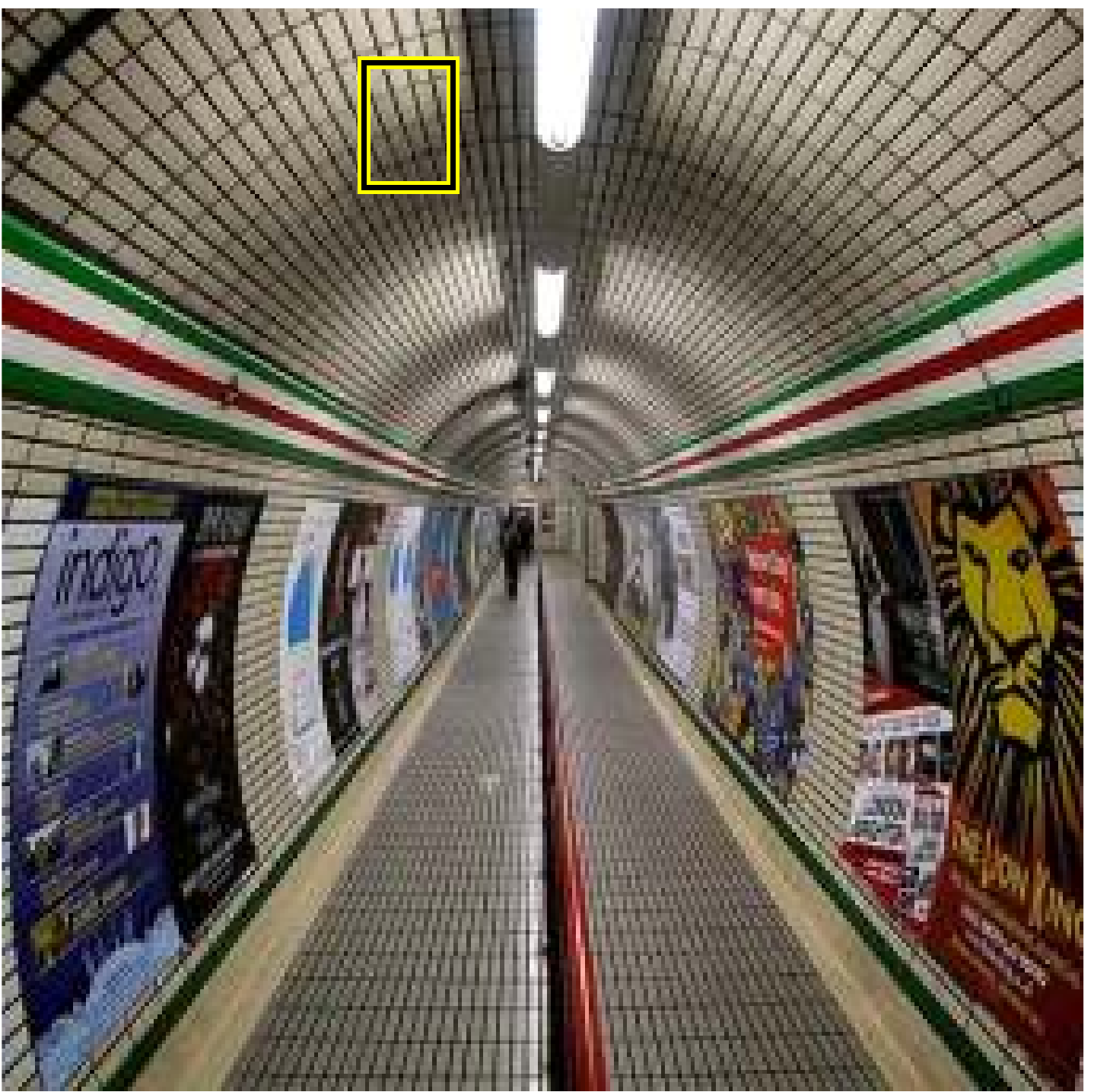} &
				\includegraphics[height=0.65in, width=0.85in]{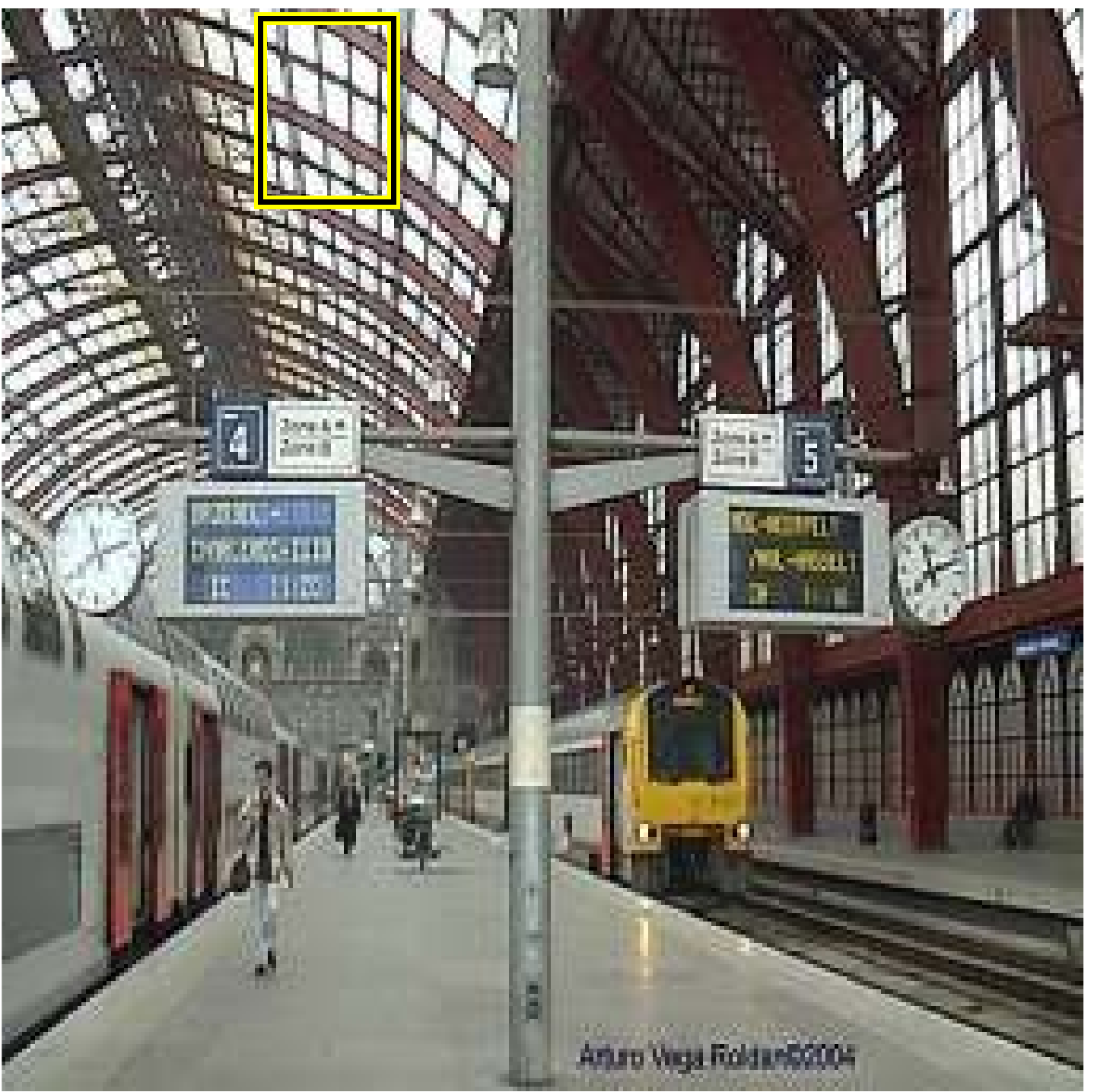} &
				\includegraphics[height=0.65in, width=0.85in]{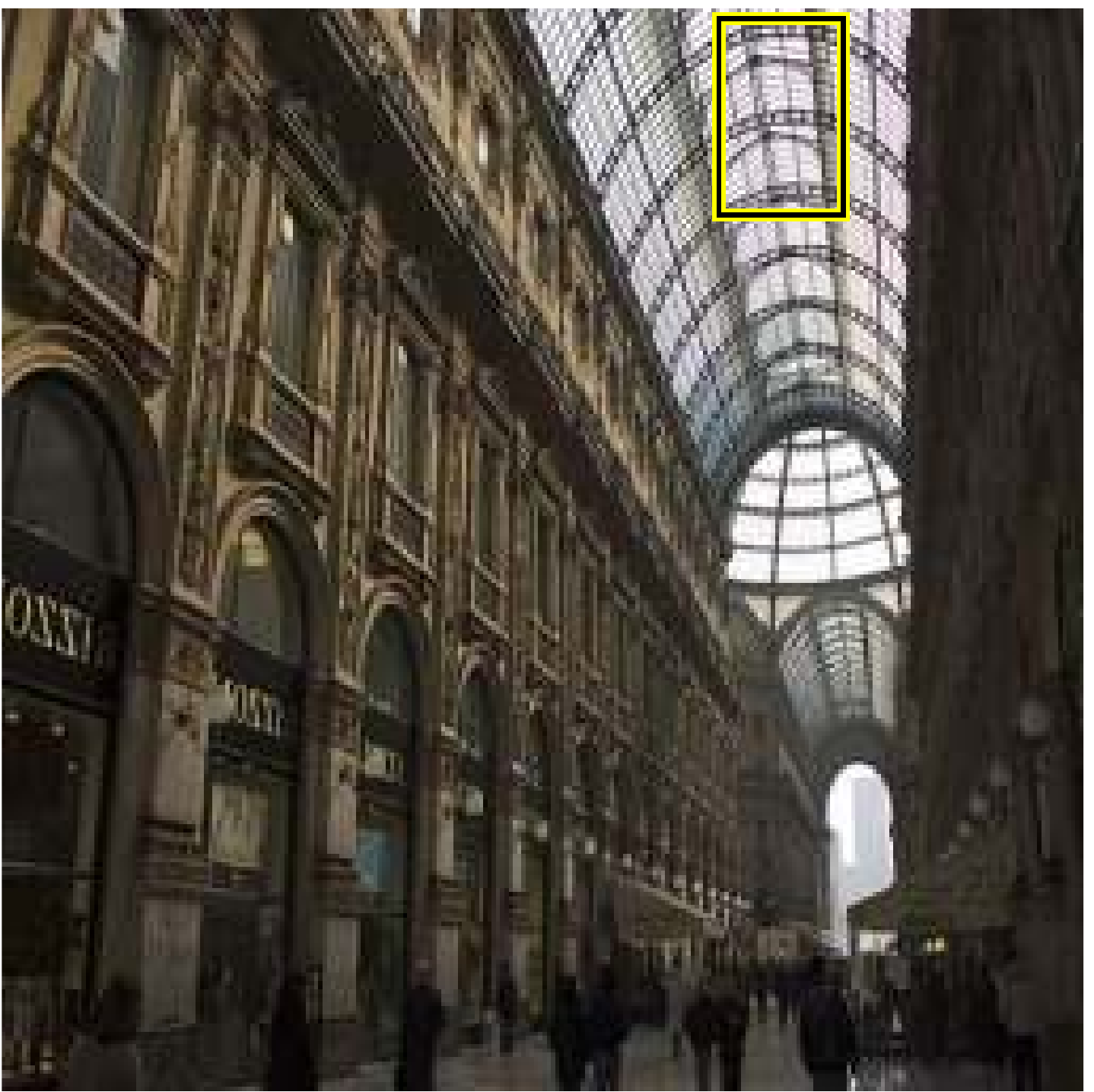} &
				\includegraphics[height=0.65in, width=0.85in]{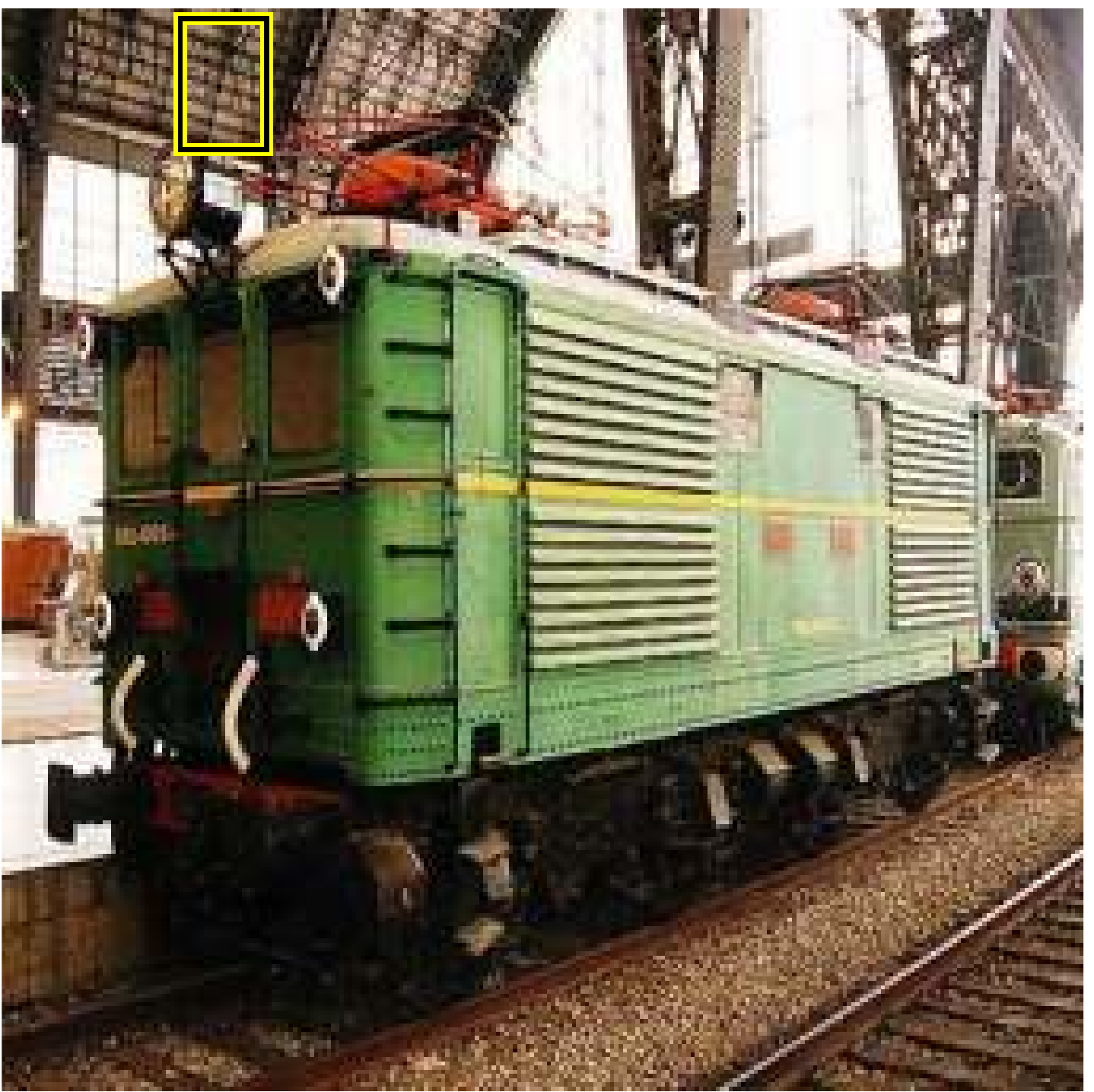} &
				\includegraphics[height=0.65in, width=0.85in]{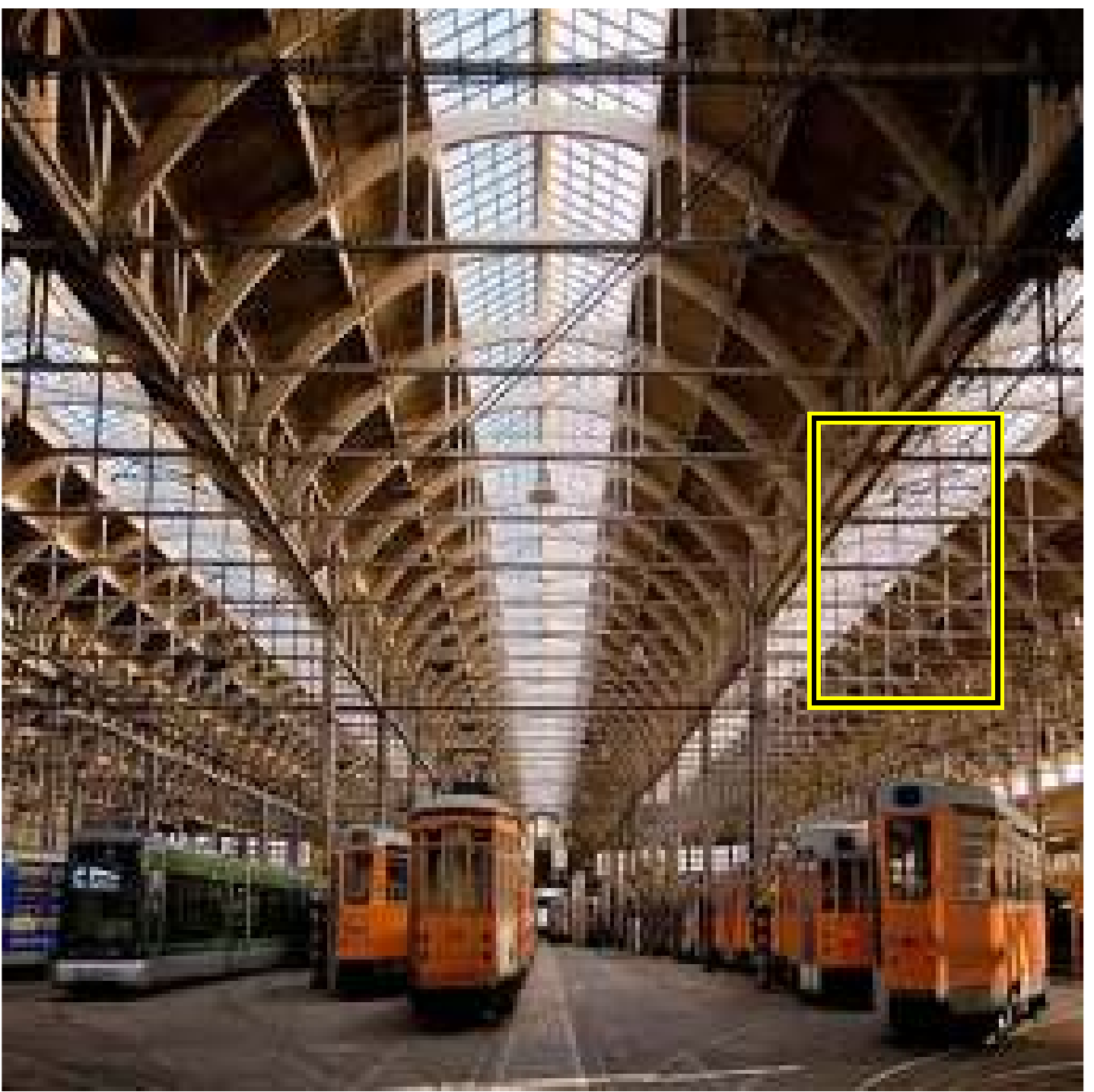} &
				\includegraphics[height=0.65in, width=0.85in]{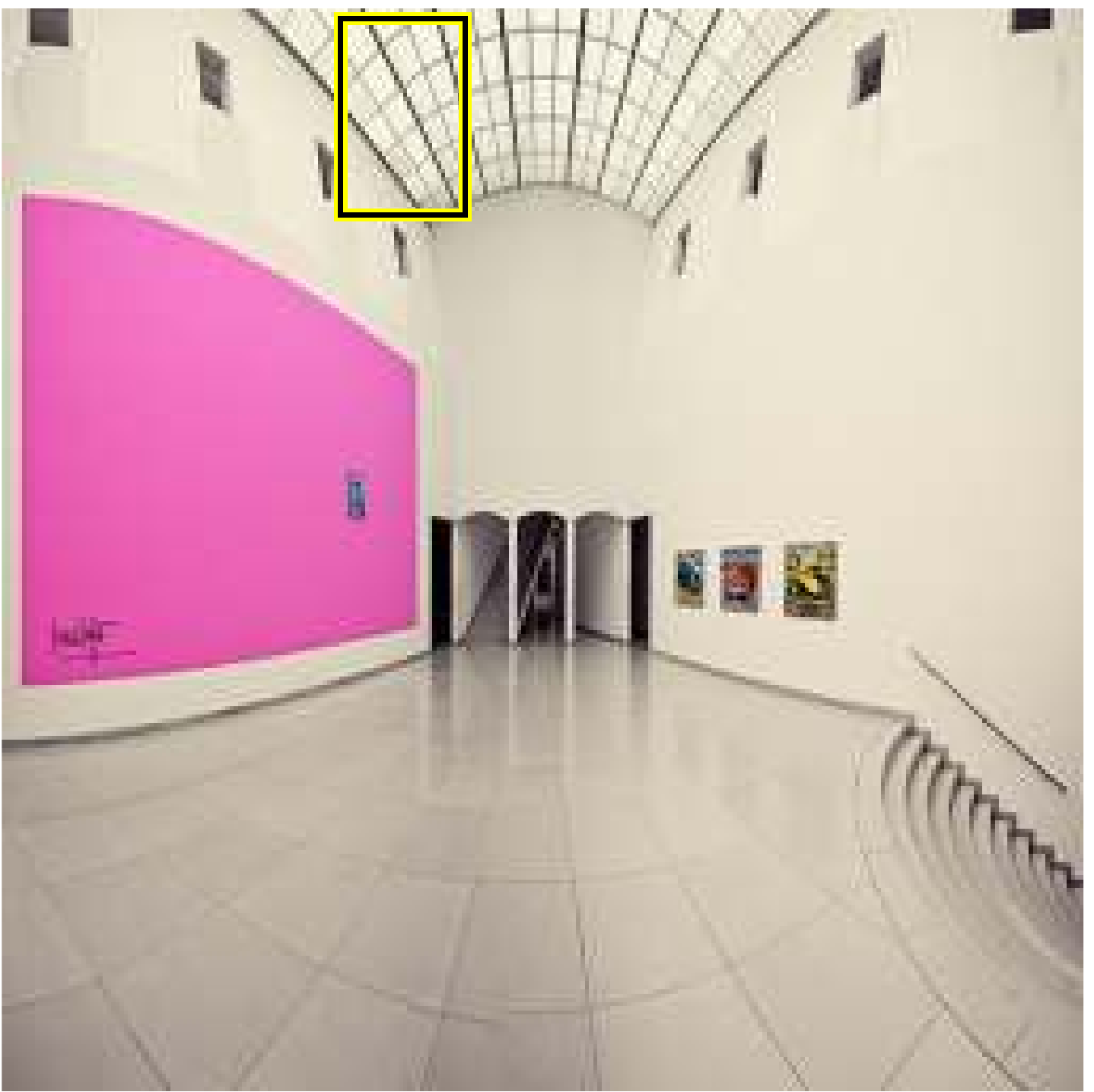} \\ [-0.05cm]
	\rotatebox{90}{\hspace{0.27cm}Part 33}$\;$ &
				\includegraphics[height=0.65in, width=0.85in]{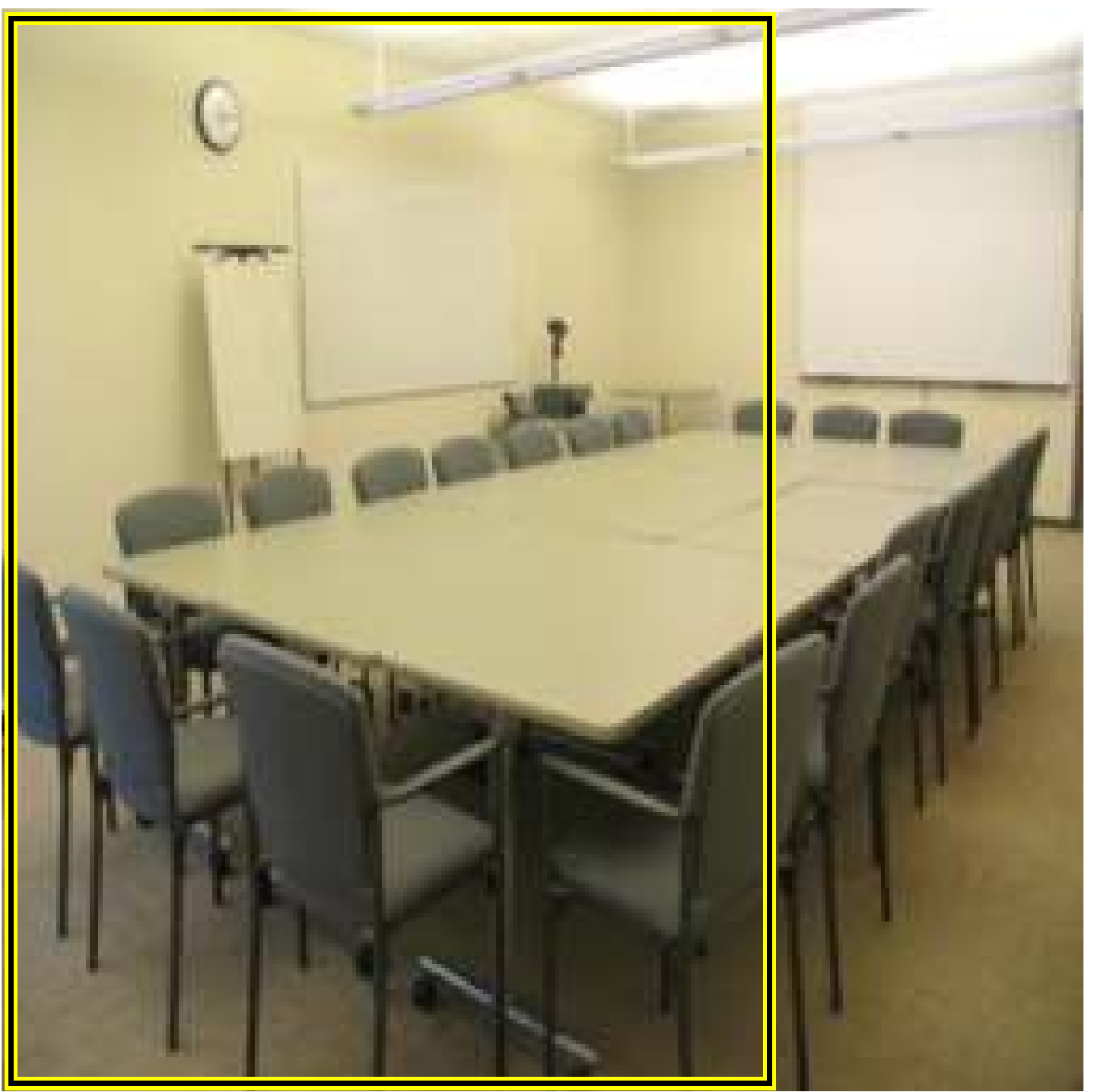} &
				\includegraphics[height=0.65in, width=0.85in]{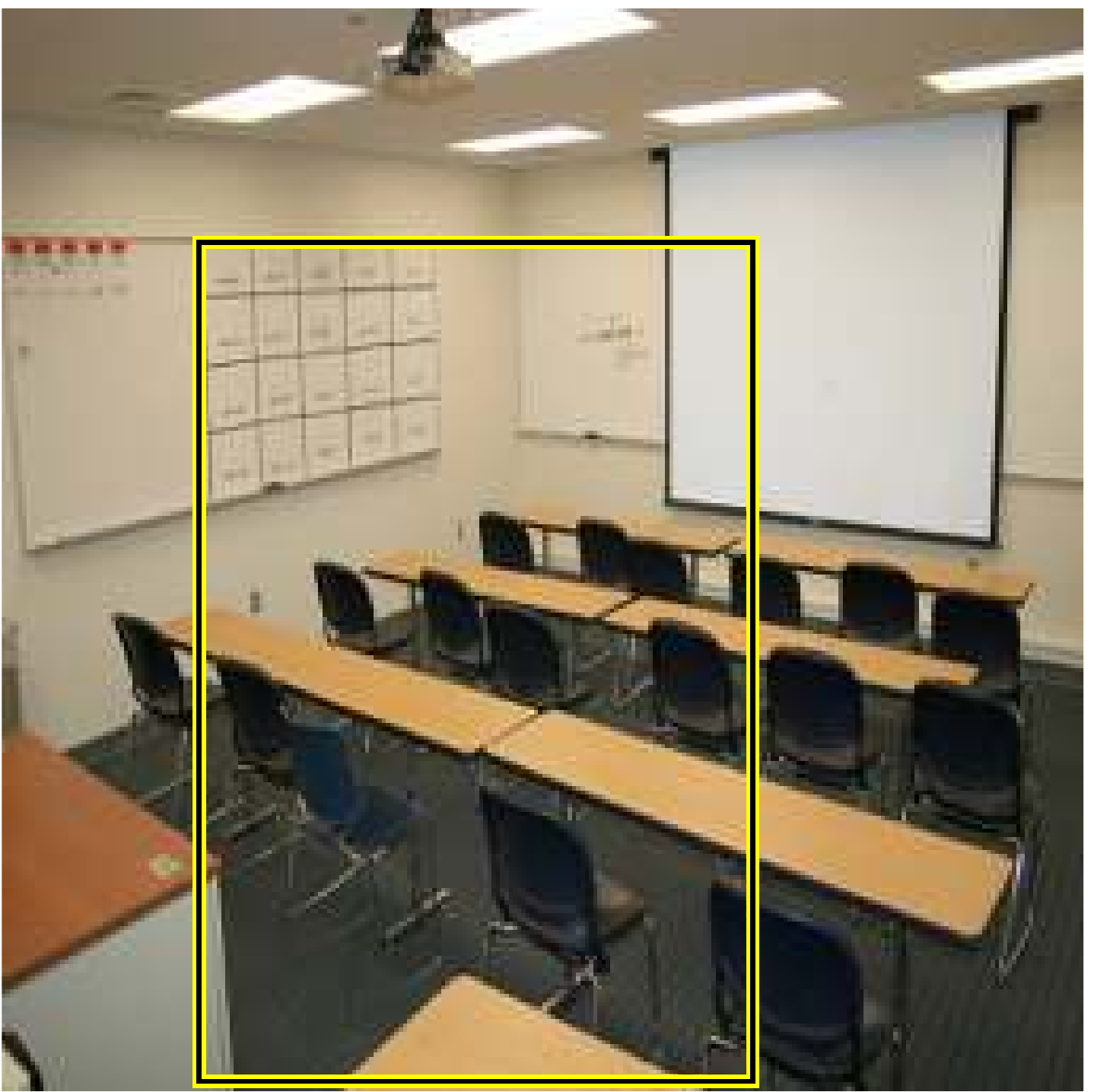} &
				\includegraphics[height=0.65in, width=0.85in]{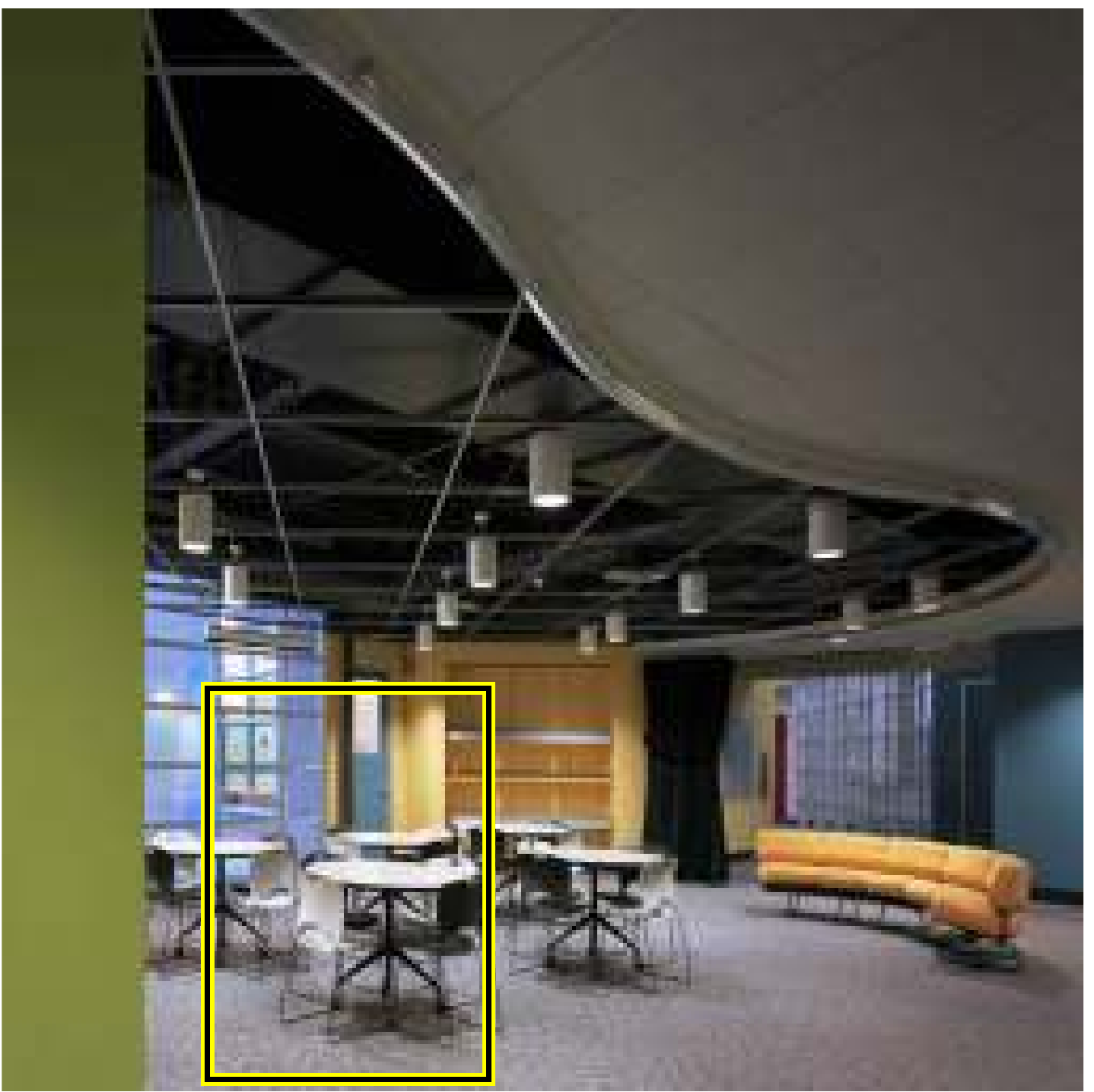} &
				\includegraphics[height=0.65in, width=0.85in]{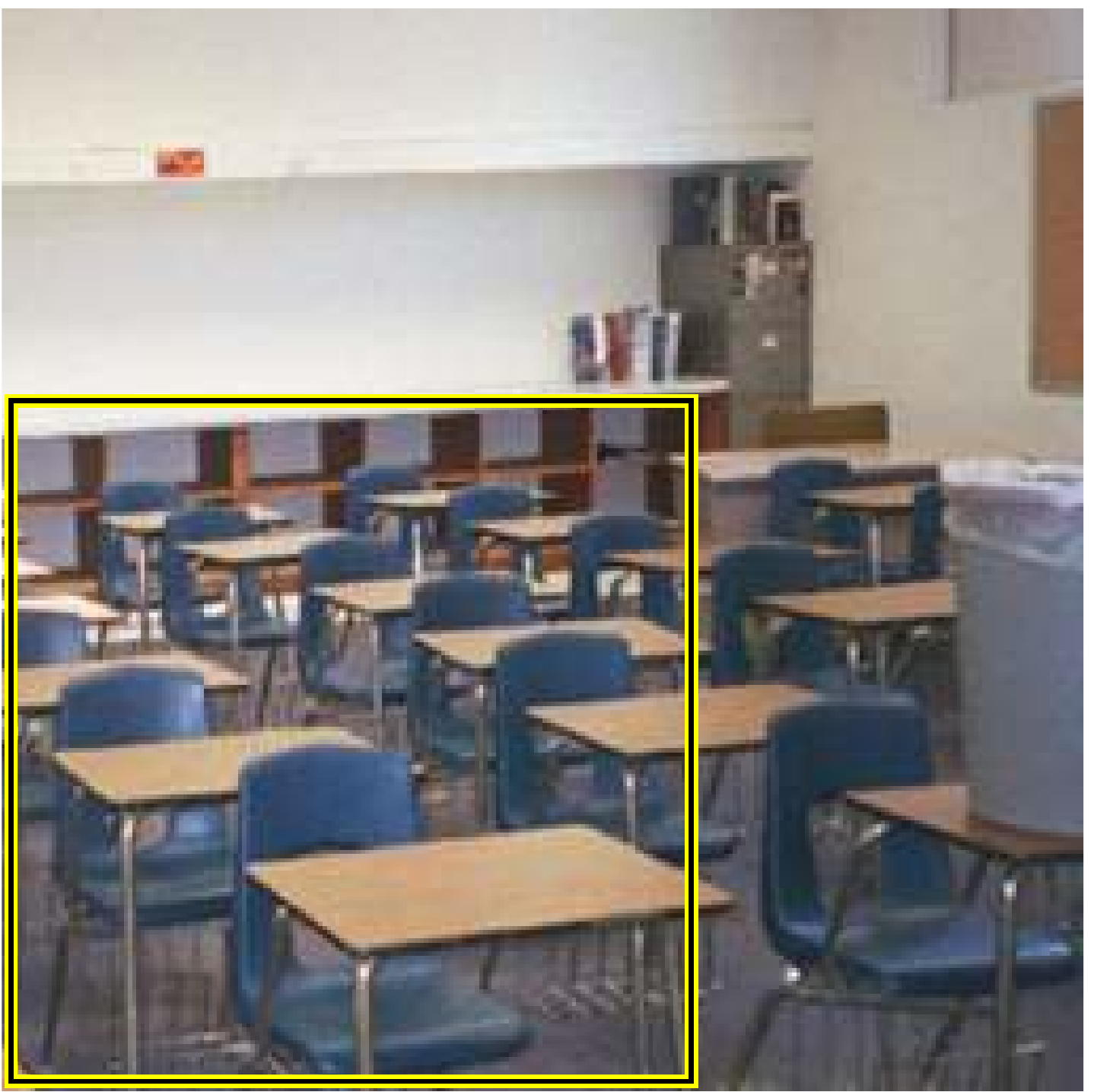} &
				\includegraphics[height=0.65in, width=0.85in]{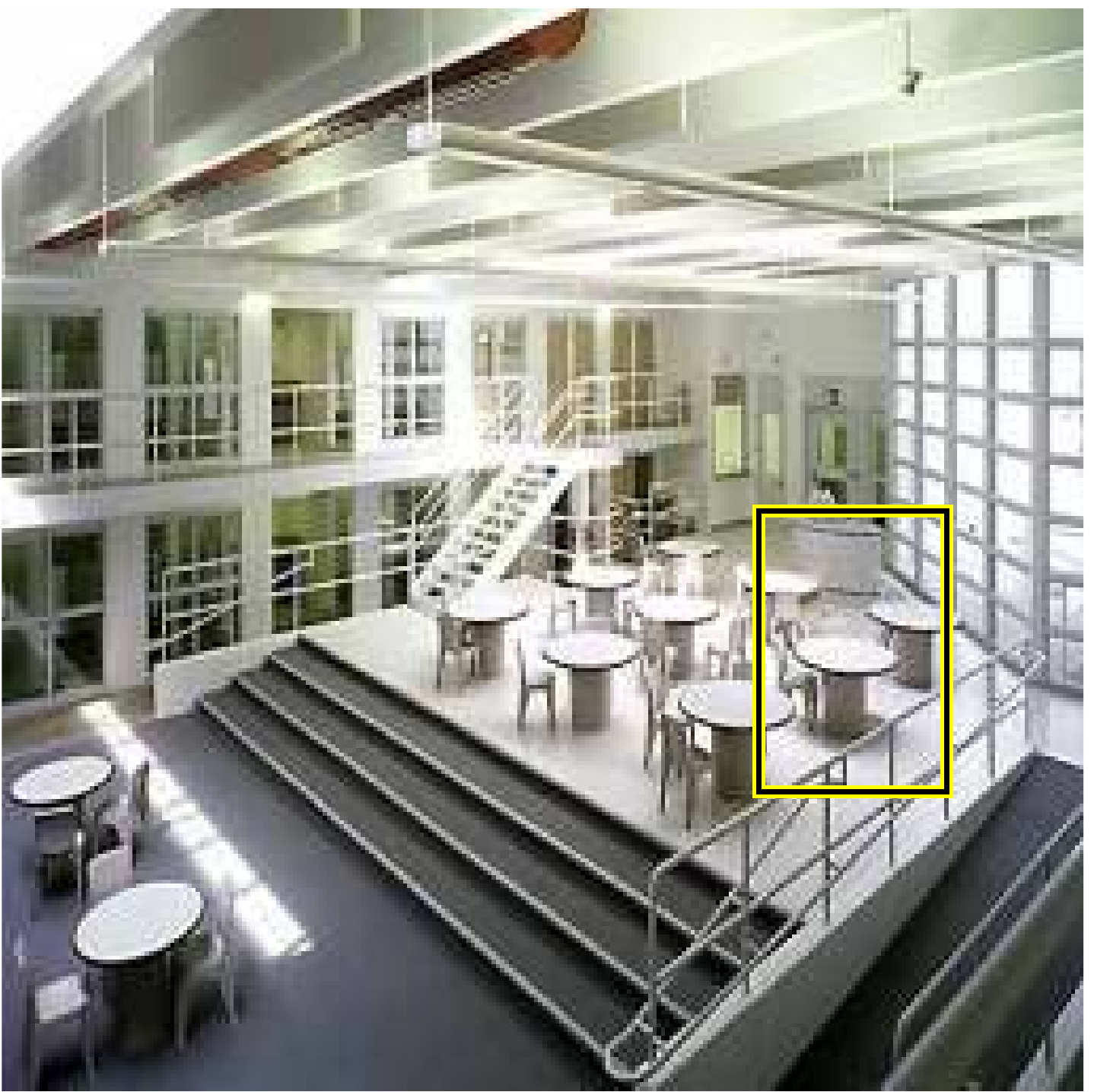} &
				\includegraphics[height=0.65in, width=0.85in]{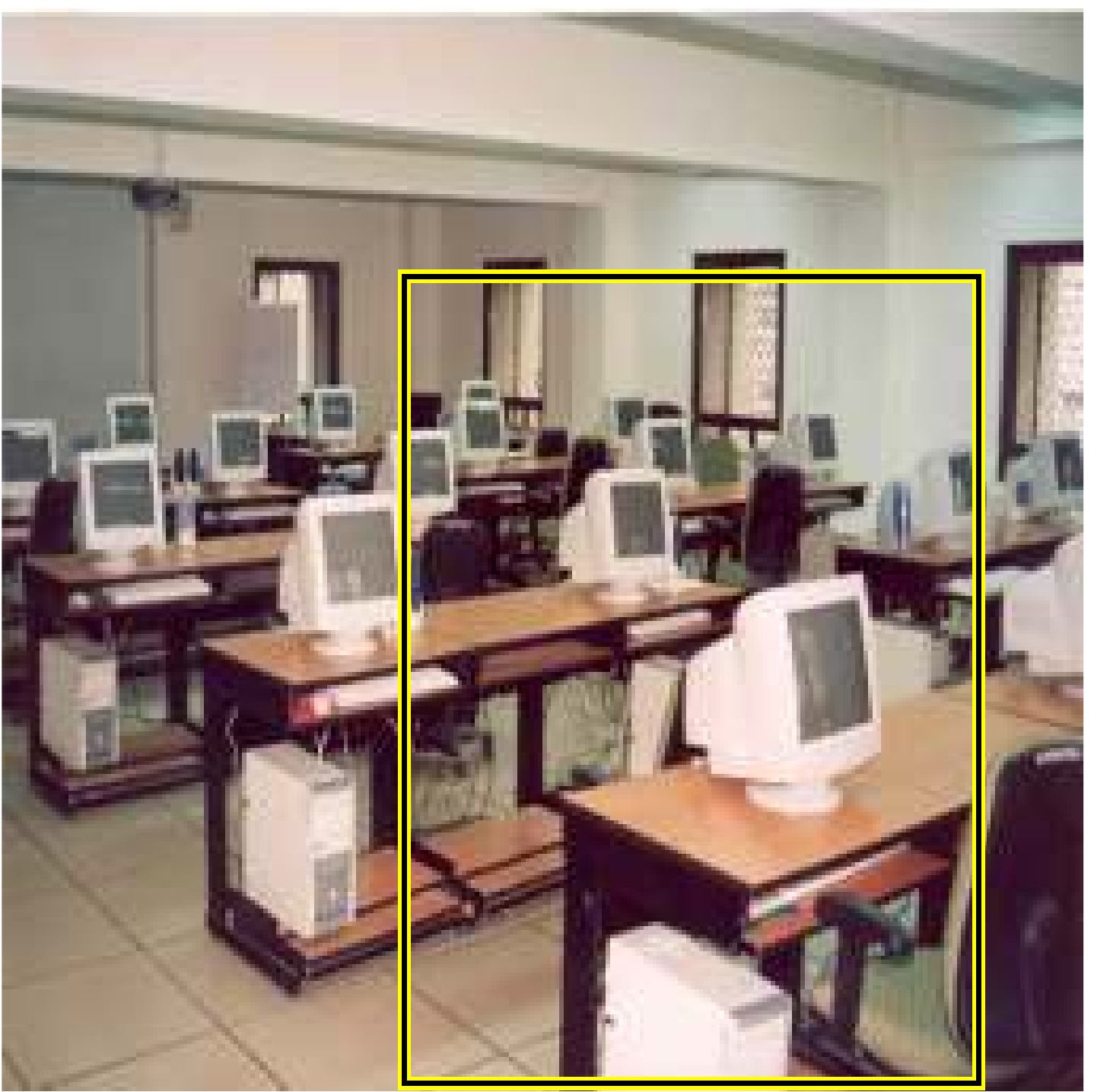} \\ [-0.05cm]
	\rotatebox{90}{\hspace{0.27cm}Part 41}$\;$ &
				\includegraphics[height=0.65in, width=0.85in]{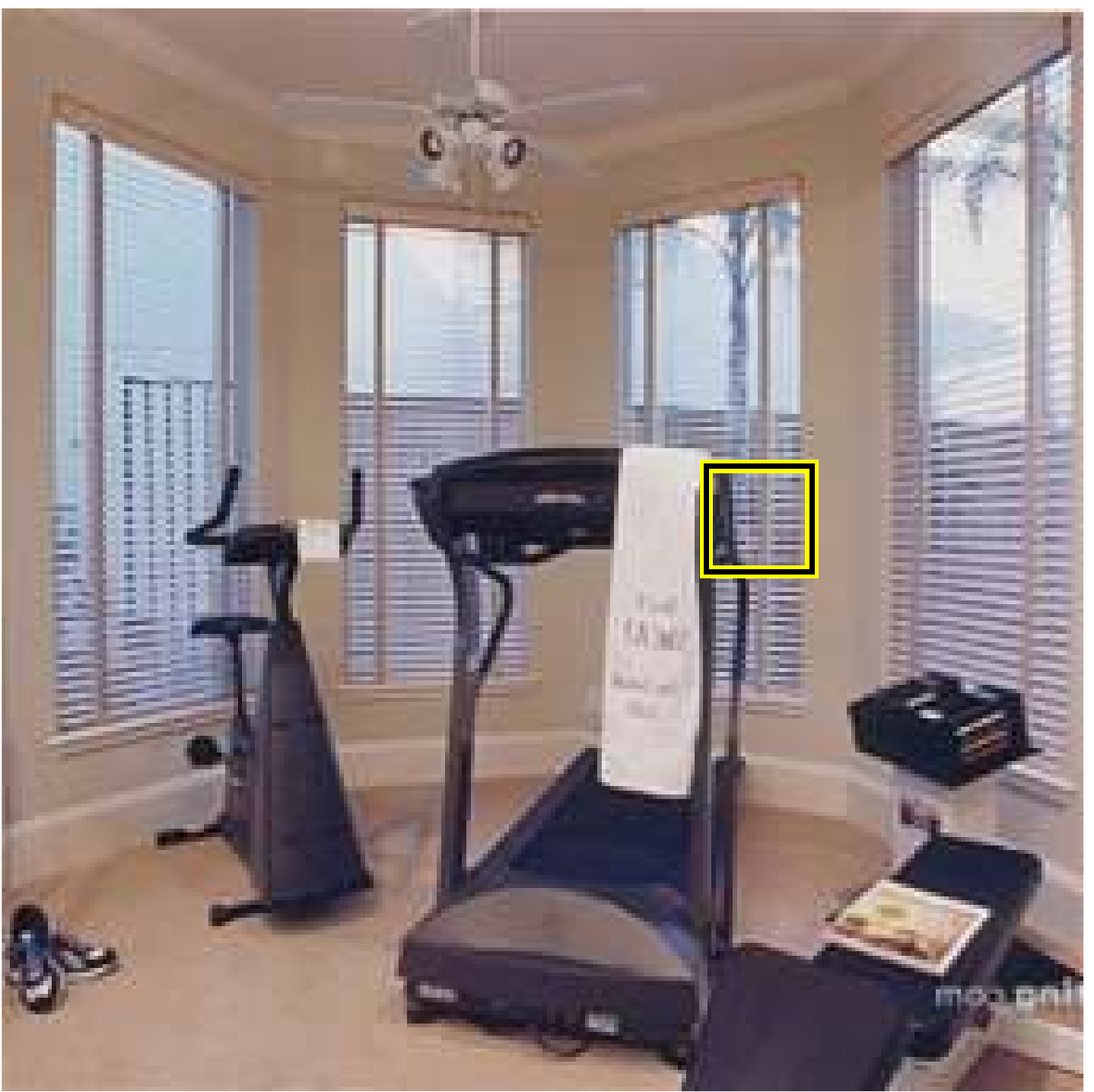} &
				\includegraphics[height=0.65in, width=0.85in]{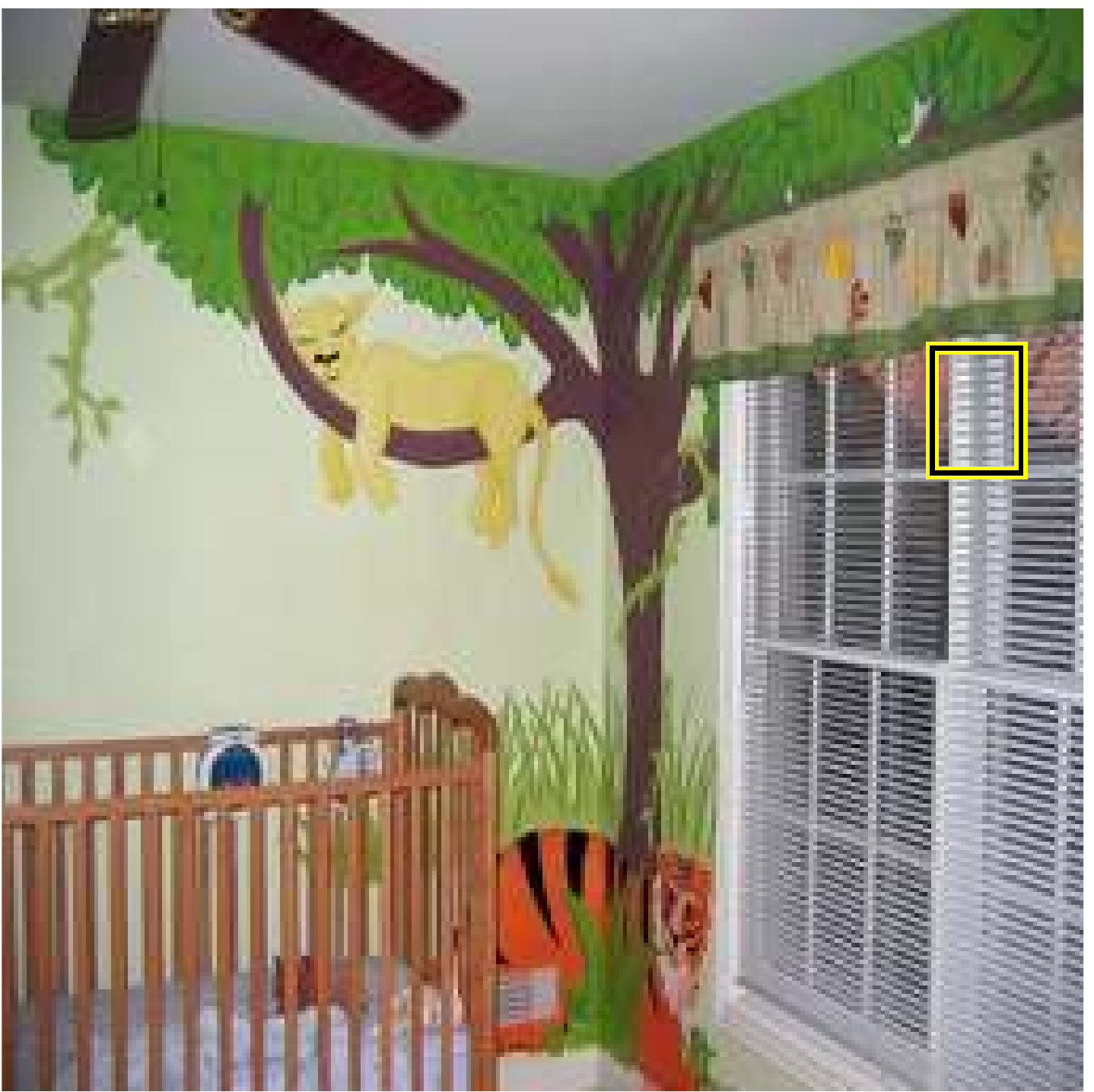} &
				\includegraphics[height=0.65in, width=0.85in]{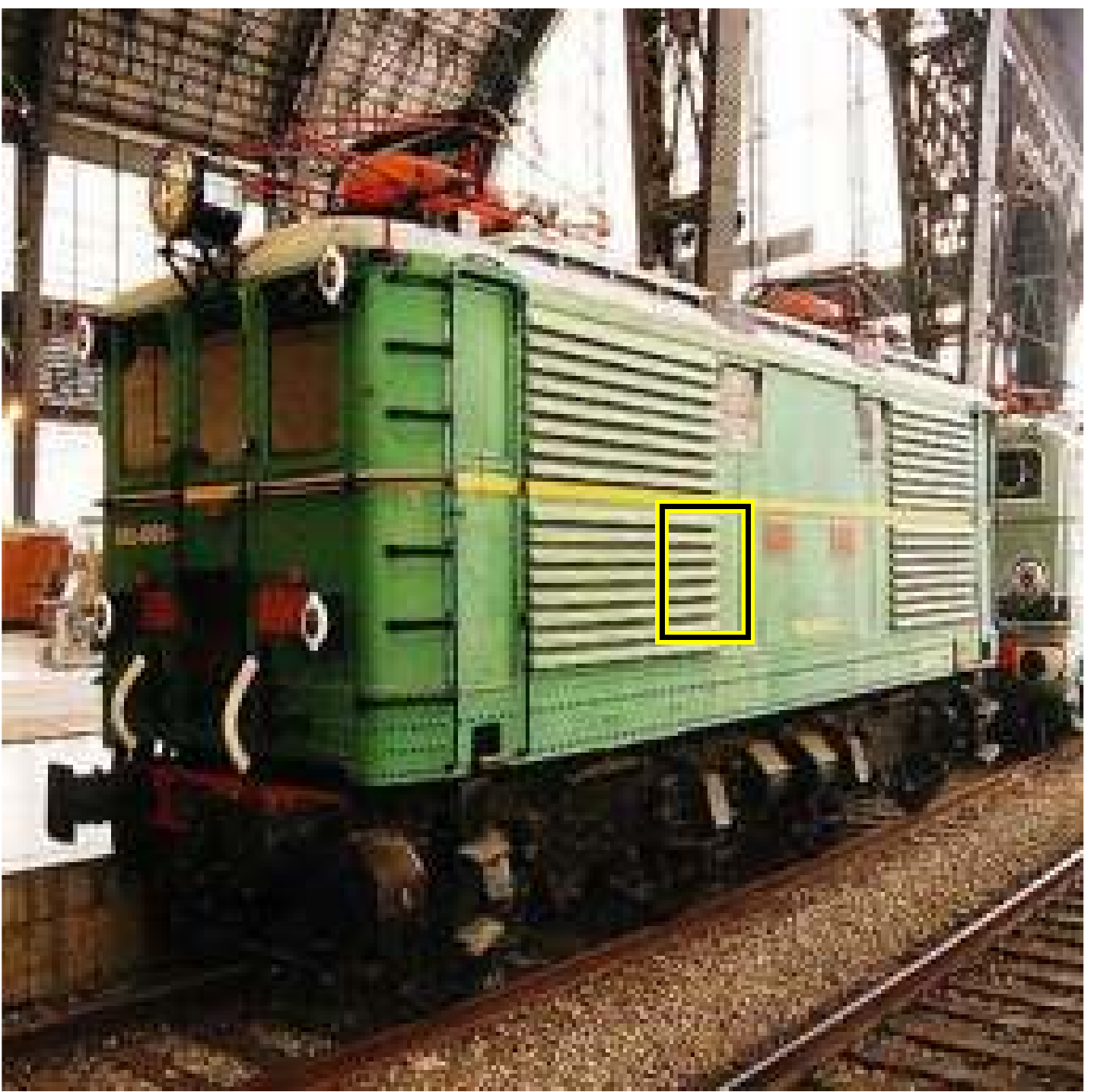} &
				\includegraphics[height=0.65in, width=0.85in]{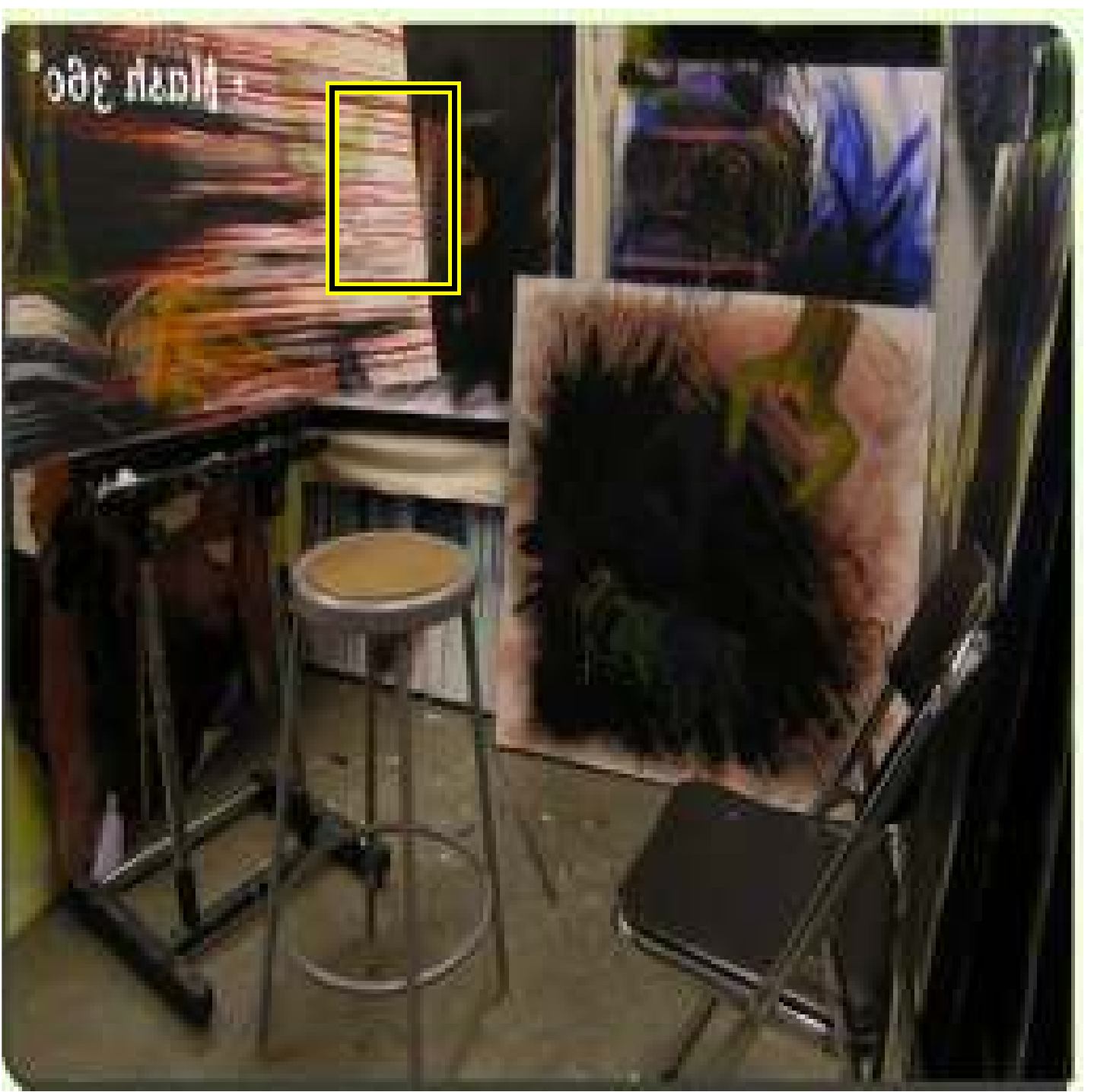} &
				\includegraphics[height=0.65in, width=0.85in]{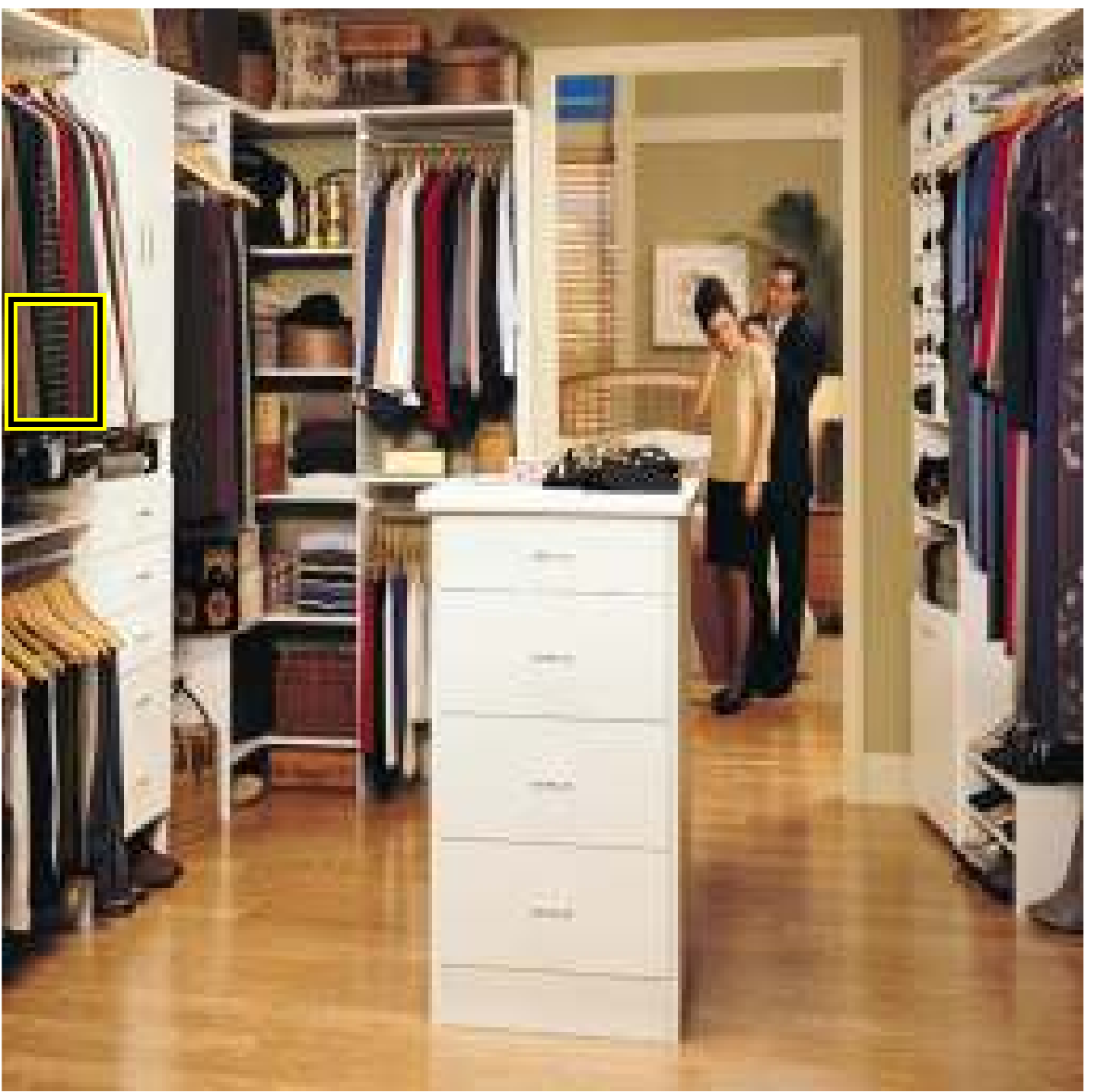} &
				\includegraphics[height=0.65in, width=0.85in]{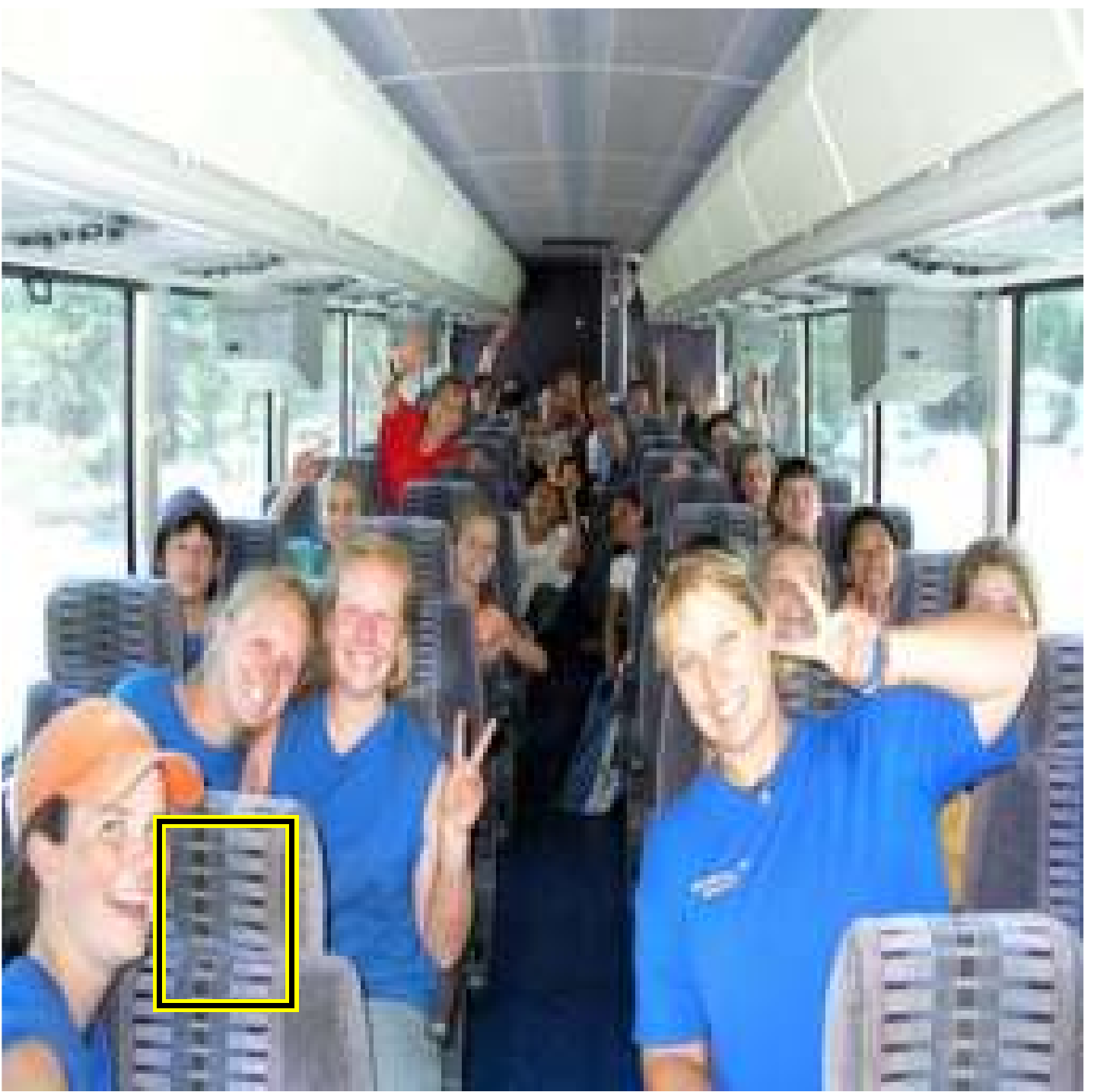} \\ [-0.05cm]
	\rotatebox{90}{\hspace{0.27cm}Part 48}$\;$ &
				\includegraphics[height=0.65in, width=0.85in]{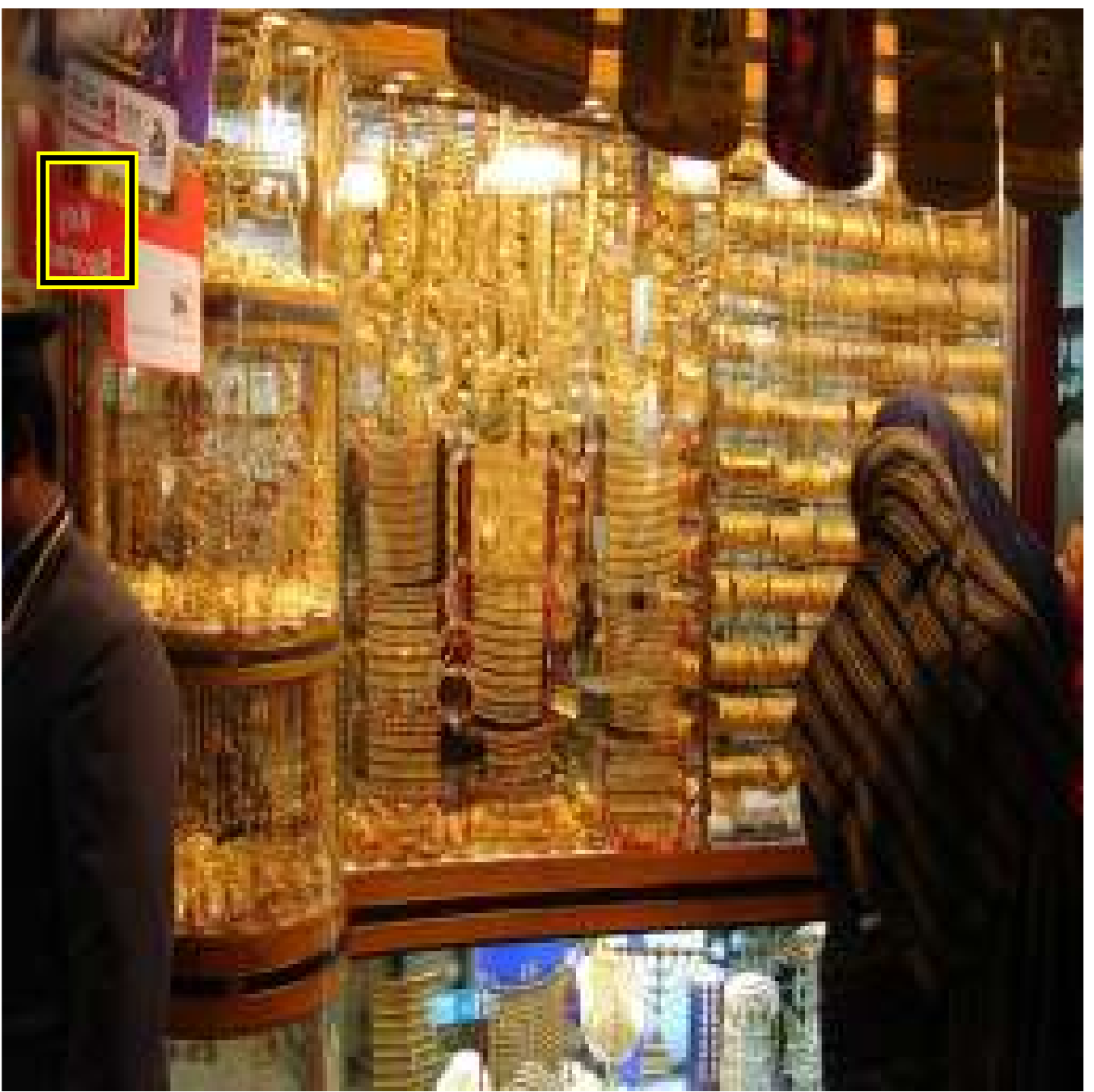} &
				\includegraphics[height=0.65in, width=0.85in]{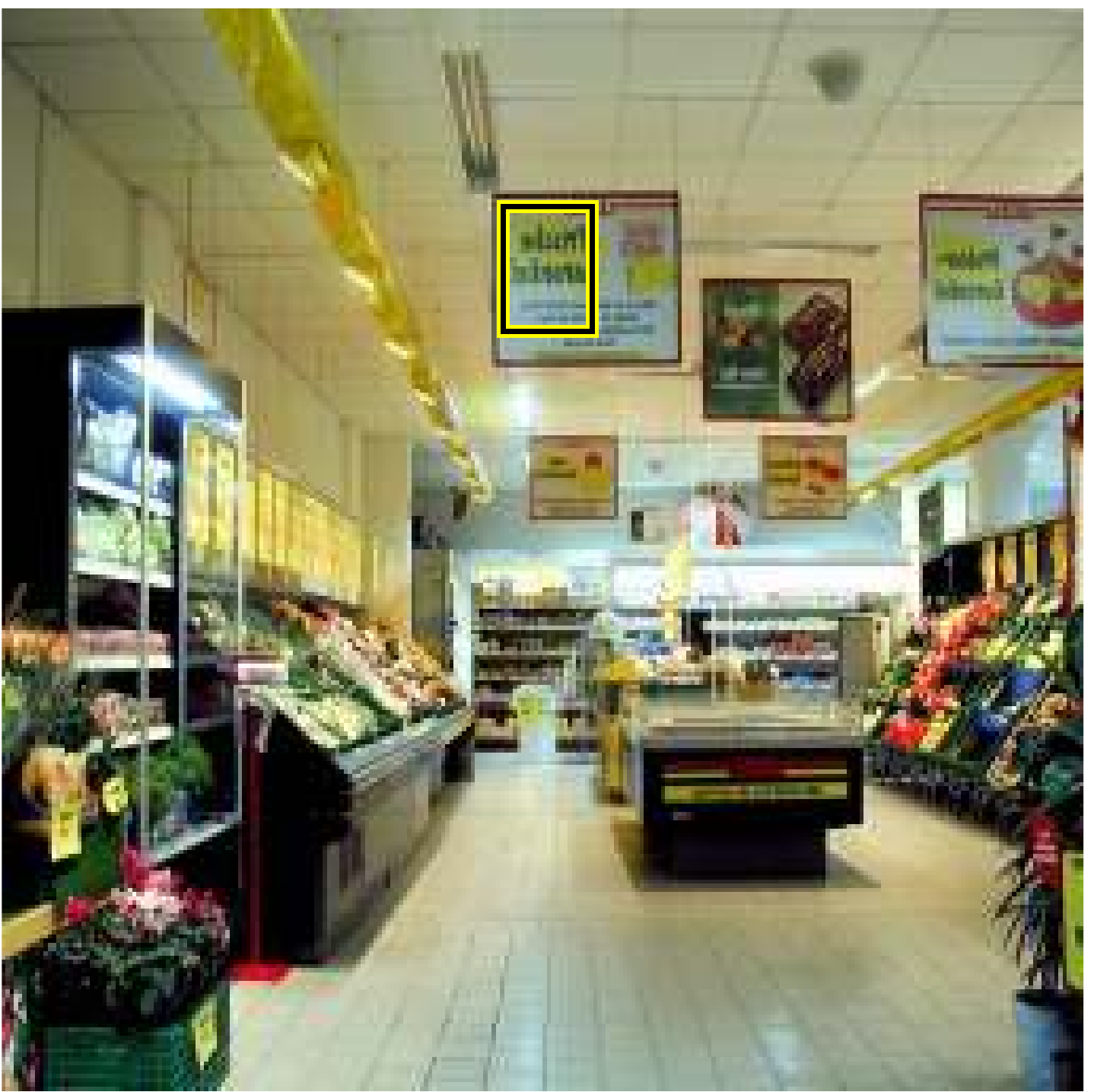} &
				\includegraphics[height=0.65in, width=0.85in]{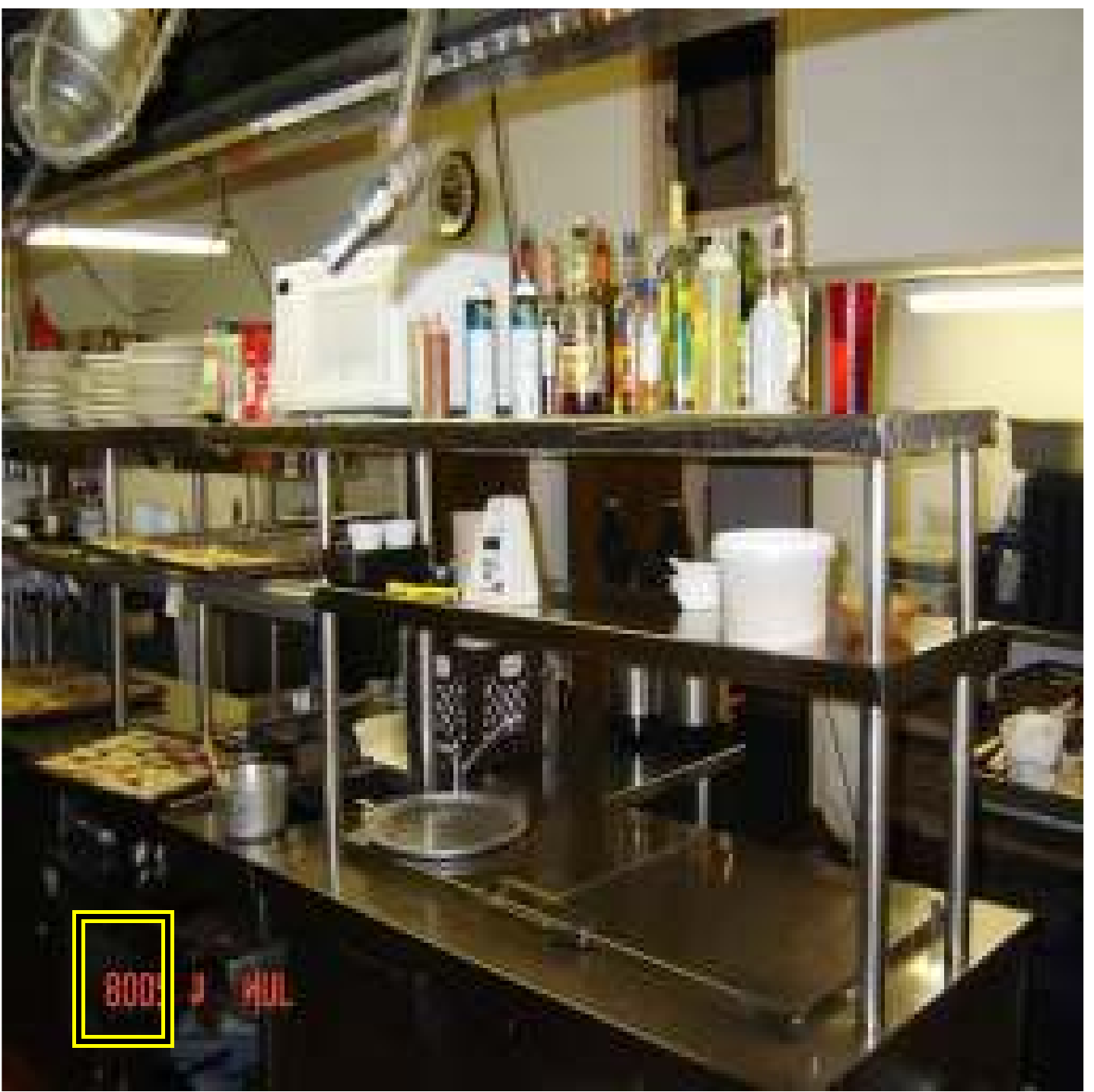} &
				\includegraphics[height=0.65in, width=0.85in]{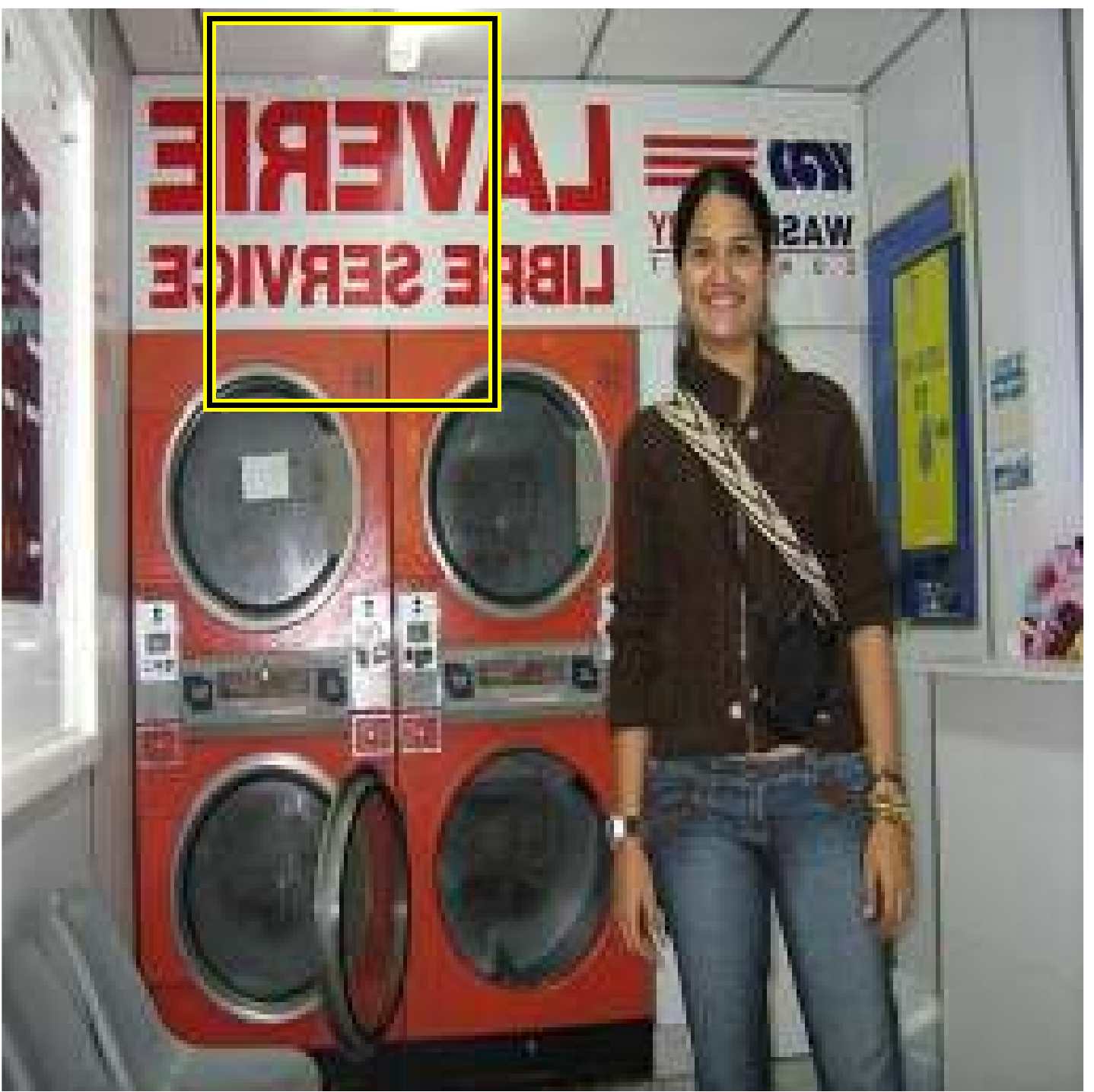} &
				\includegraphics[height=0.65in, width=0.85in]{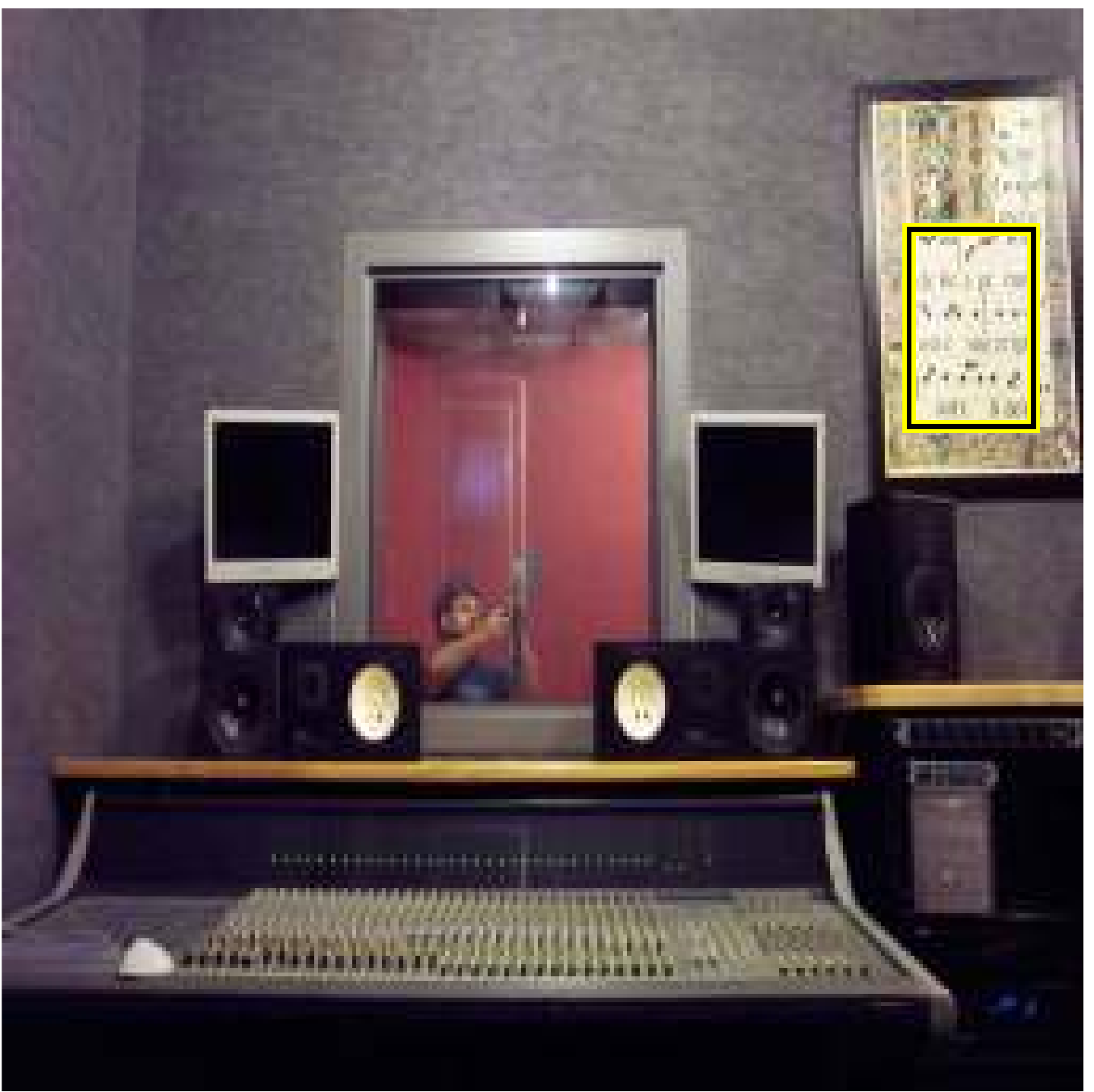} &
				\includegraphics[height=0.65in, width=0.85in]{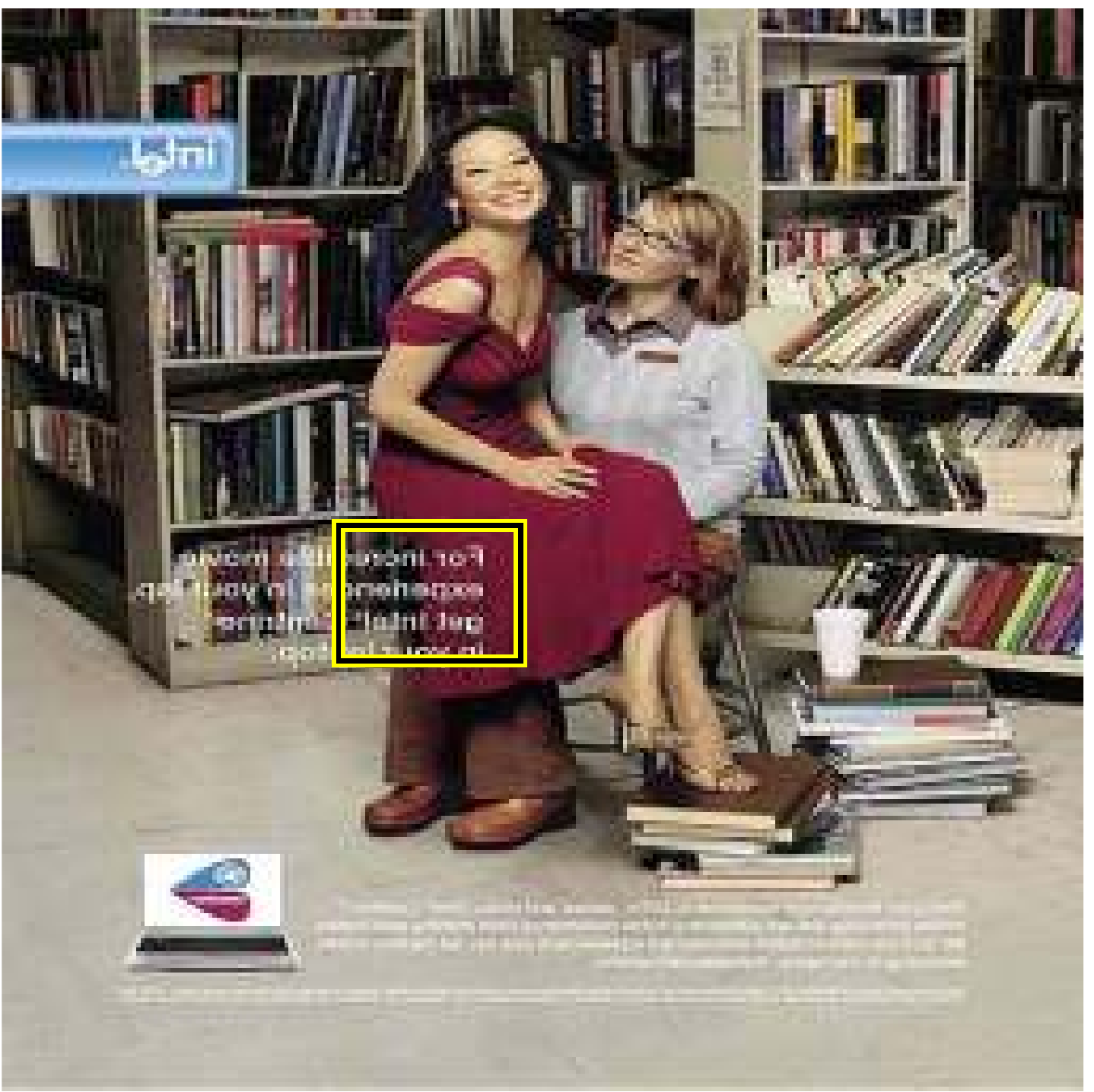} \\ [-0.05cm]
	\rotatebox{90}{\hspace{0.27cm}Part 51}$\;$ &
				\includegraphics[height=0.65in, width=0.85in]{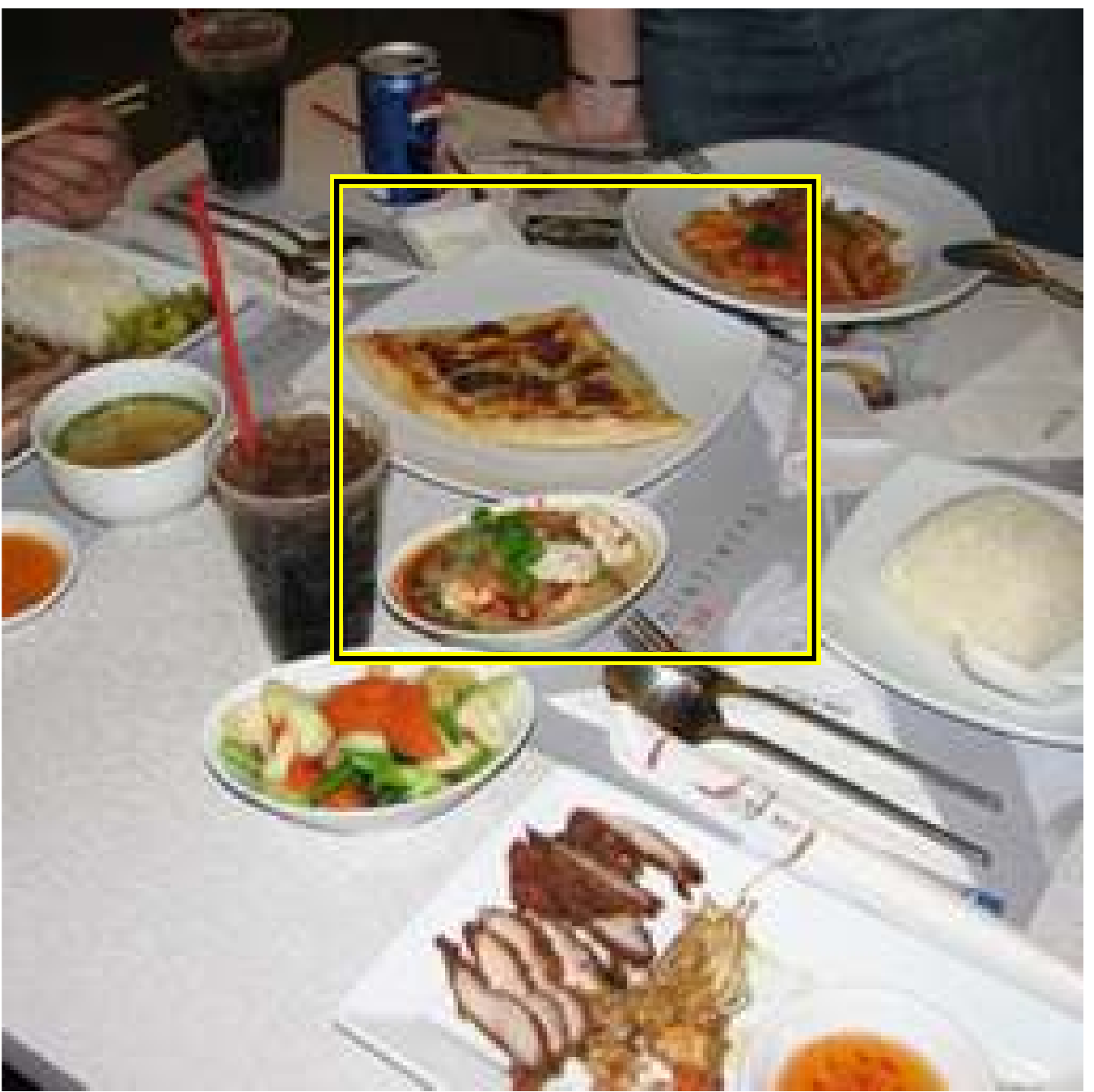} &
				\includegraphics[height=0.65in, width=0.85in]{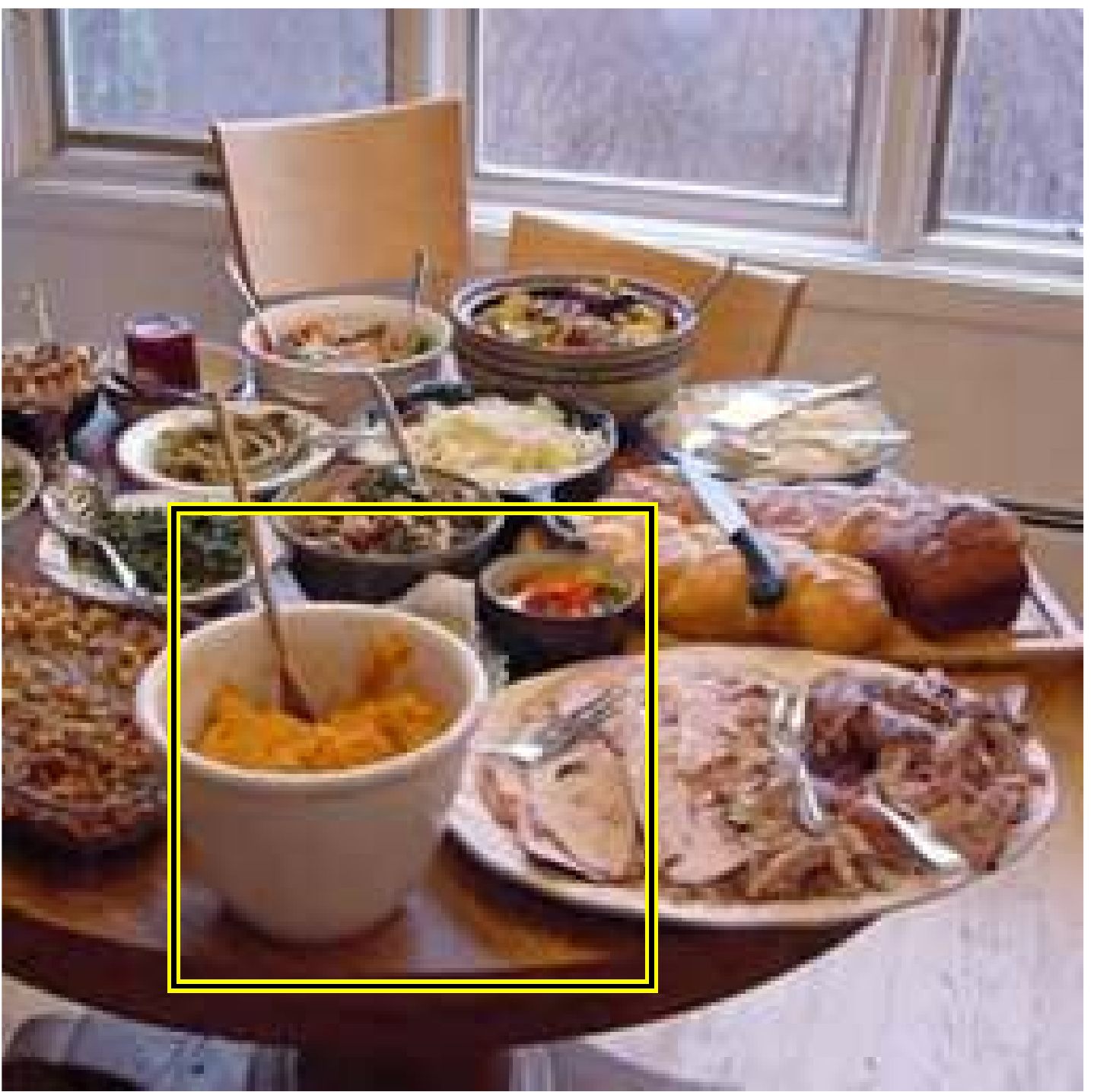} &
				\includegraphics[height=0.65in, width=0.85in]{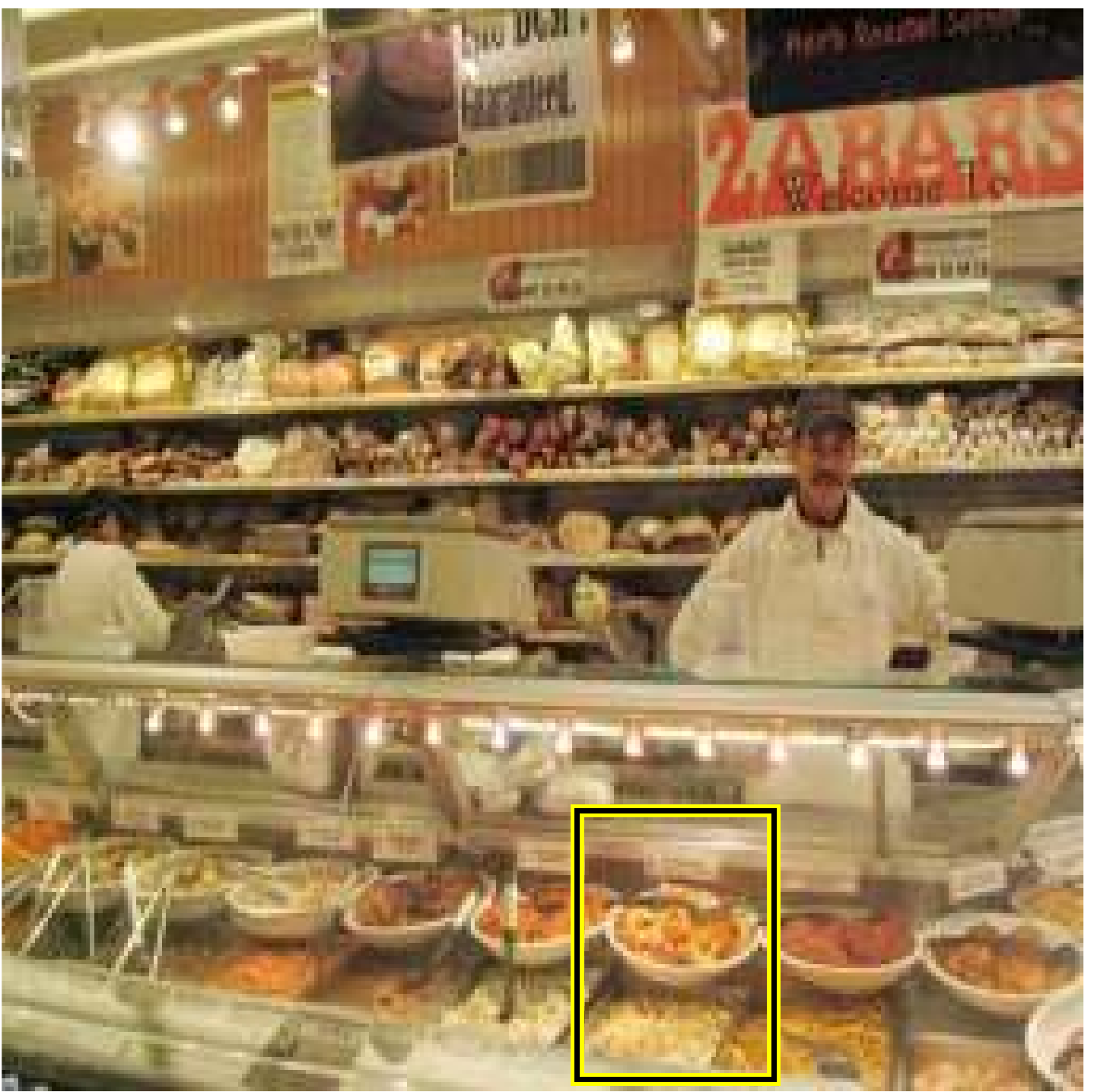} &
				\includegraphics[height=0.65in, width=0.85in]{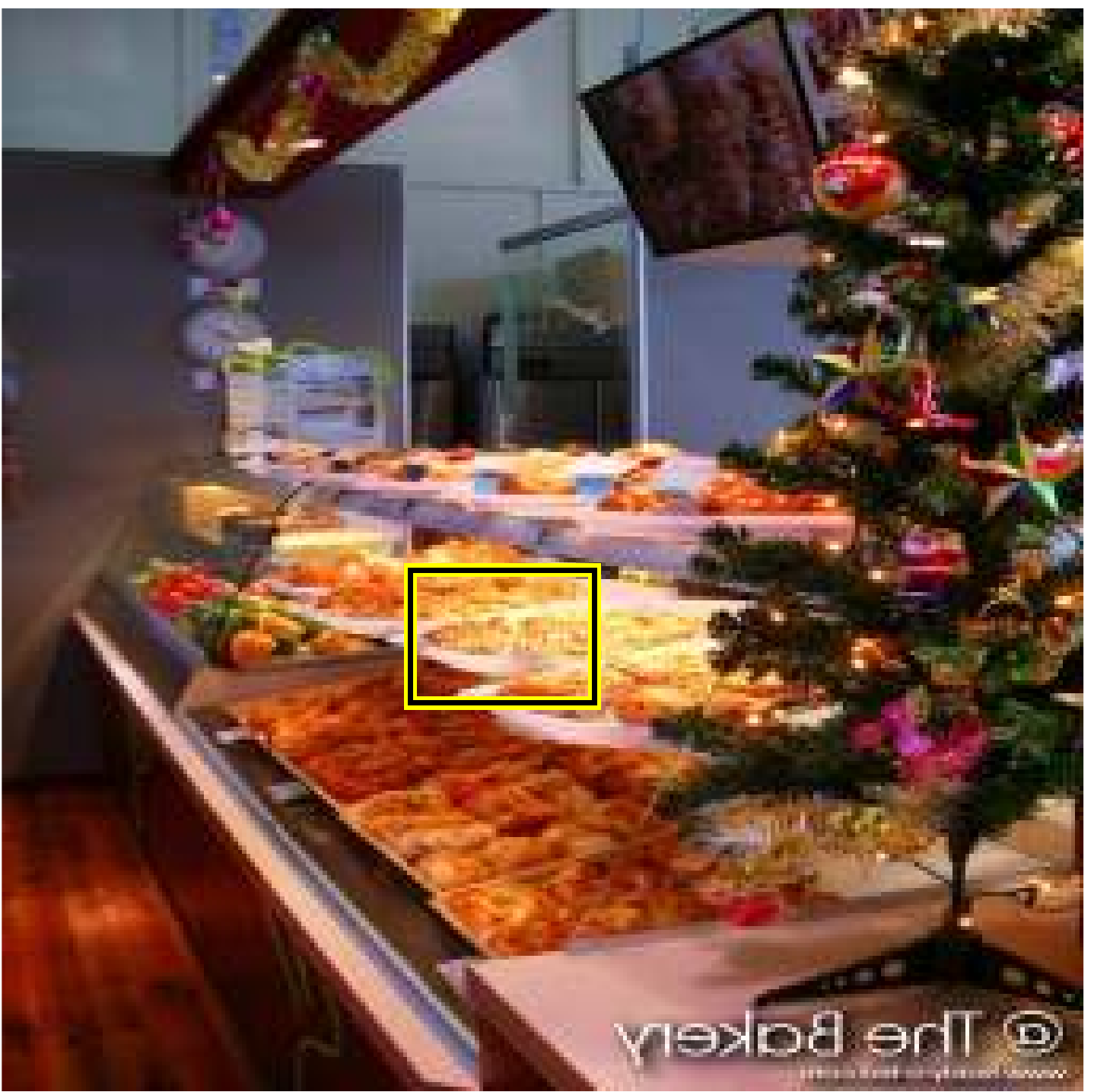} &
				\includegraphics[height=0.65in, width=0.85in]{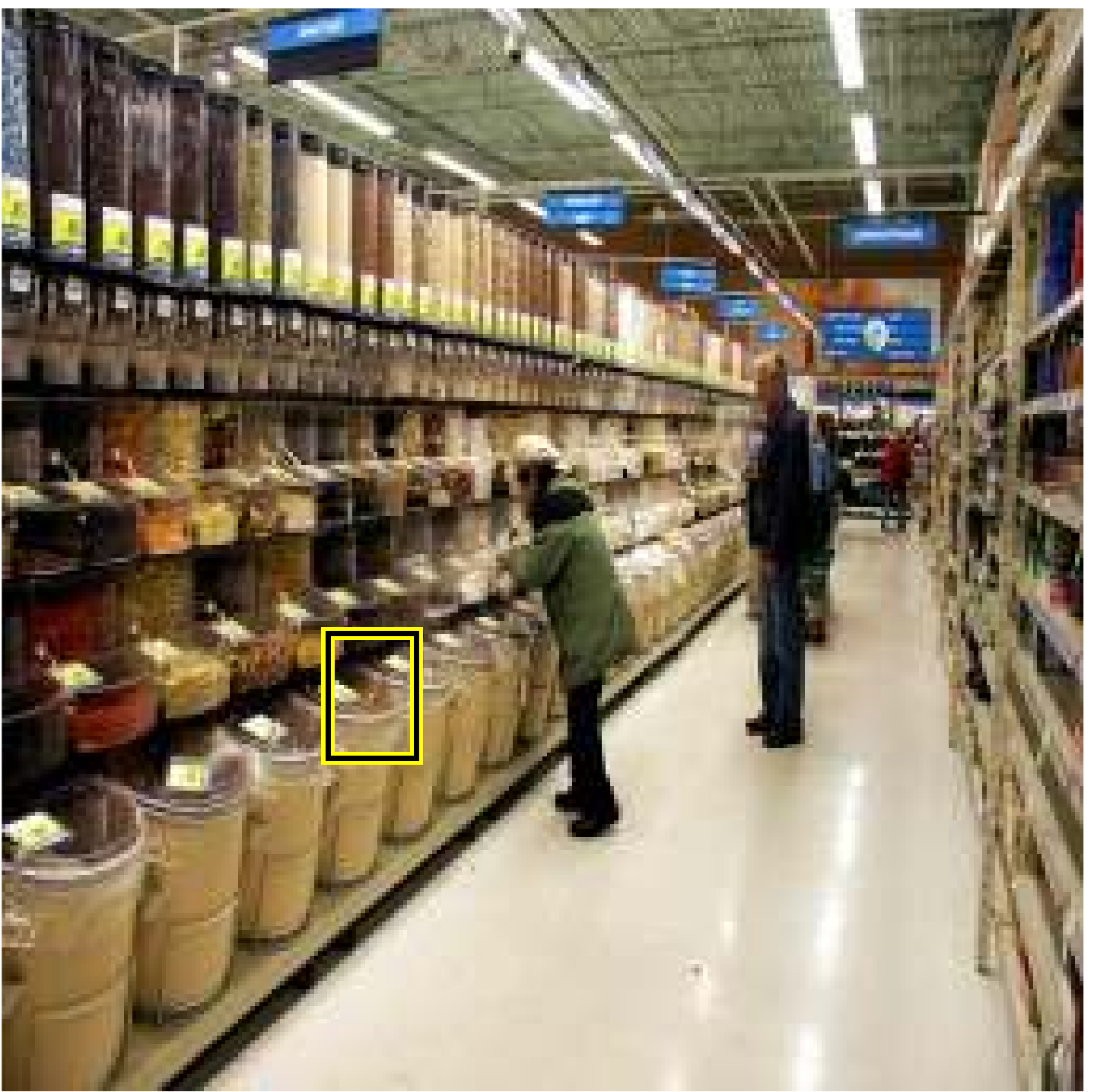} &
				\includegraphics[height=0.65in, width=0.85in]{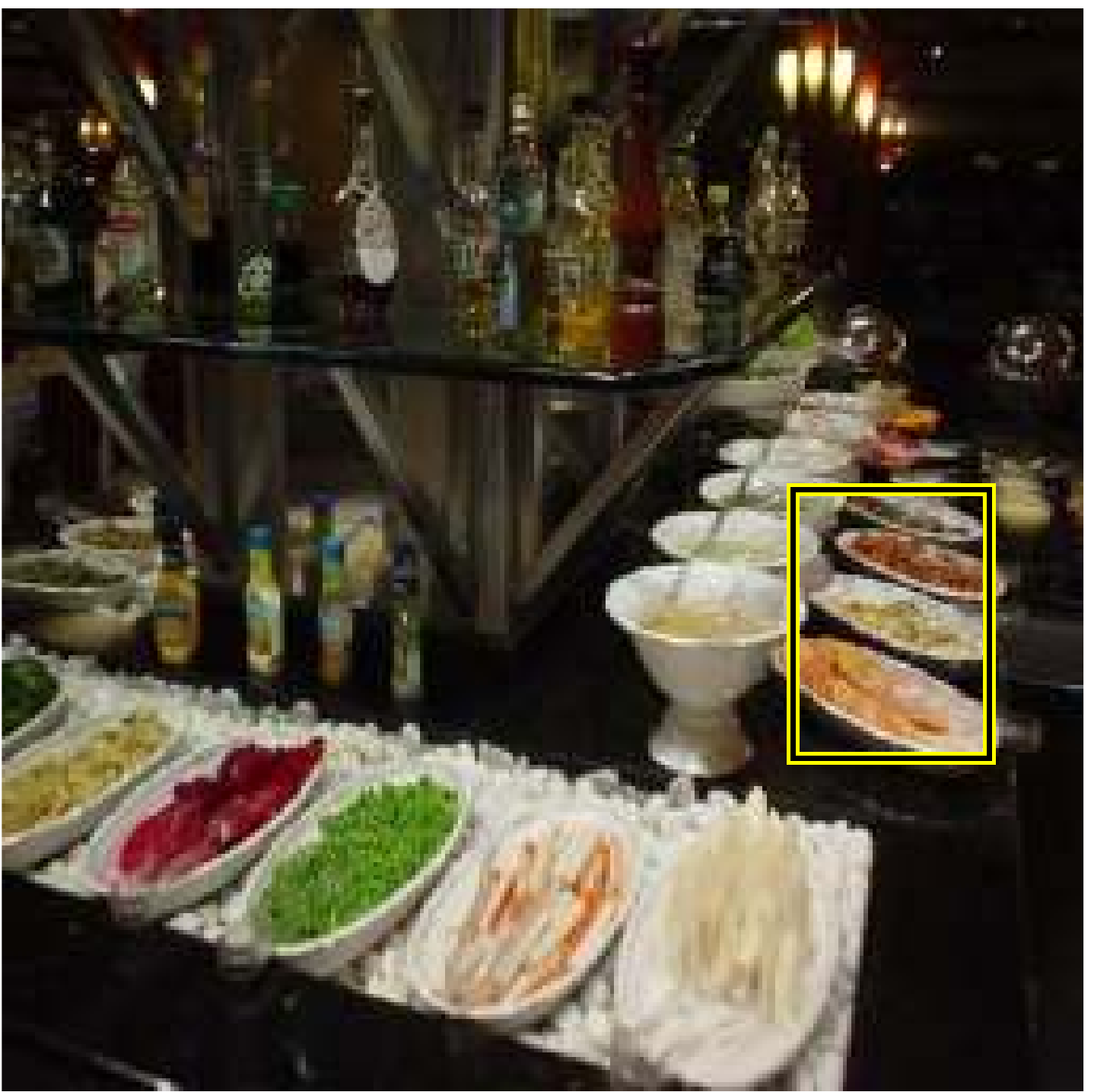}
	\end{tabular}
	\vspace{-1em}
	\caption{Top detections of parts on test images of the full dataset. The numbers in the first column match the part indices in Figure~\ref{fig:u_matrix_full}. Part detection is done in a multi-scale sliding window fashion and using a $\sq{256 \times 256}$ window. For visualization purposes images are stretched to have the same size.}
	\label{fig:top_scoring_patches_CNN_supplement}
	\vspace{-1em}
\end{figure*}

Figure~\ref{fig:top_scoring_patches_CNN_supplement} complements Figure~\ref{fig:top_scoring_patches_CNN} of the paper. The part indices in the first column match those of Figure~\ref{fig:u_matrix_full}. The rows show the highest scoring detections of a particular part on test images. 

Part 1 fires on \emph{clothing-rack}, part 22 appear to find \emph{container}, and part 33 detects \emph{table-top}. There are parts that capture low-level features such as the mesh pattern of part 31 and the high-frequency horizontal stripes of part 41. Also, there are parts that are selective for certain colors. For example, part 9 appears to respond to specific \emph{red} patterns (in particular \emph{fruits} and \emph{flowers}). Part 51 appears to fire on \emph{yellow-food} dishes. Part 48 is very well tuned to finding \emph{text}. 

According to the weight matrix (see Figure~\ref{fig:u_matrix_full} in the paper) Part 14 is highly weighted for \emph{nursery} and \emph{staircase} classes and it appears to detect a row of \emph{vertical-bars}. Part 21 is highly weighted for \emph{laundromat}, \emph{library}, and \emph{cloister} and it appears to respond strongly to \emph{arch}. Also note that part 21 is a strong negative part for \emph{bookstore} relative to \emph{library}. Presence of an \emph{arch}, in fact, is a very sensible differentiating pattern that could tell \emph{library} apart from \emph{bookstore}.

\section{Processing Time}
\label{sec:processing_time_supplement}
\textbf{Test time:}
the test procedure of our models involves three simple steps: 1) convolving part filters with the test image, 2) computing the part-based representation 3) finding the class with the highest classification score. Step 1 takes $O(mhd)$ time where $m$, $h$, and $d$ are the number of parts, latent locations, and dimensionality of the patch features. Step 2 takes $O(hR)$ time where $R$ is the number of pooling regions. Step 3 takes $O(nmR)$ time where $n$ is the number of classes. The bottleneck in test time is step 1 and 3 both of which depend on the number of parts $m$. So, a decrease in $m$ directly affects the test time. Note that both of these two steps are embarrassingly parallel processes.

\textbf{Training time:}
the training procedure involves two main steps: 1) learning part weights (line 3 in Algorithm~\ref{alg:np_optimization}) and 2) learning part filters (lines 4-7 in Algorithm~\ref{alg:np_optimization}). The first step is a standard multi-class SVM problem and is relatively fast to train. The bottleneck in training is the second step.

Learning part filters involves multiple nested loops: 1) joint training loop (lines 2-8 in Algorithm~\ref{alg:np_optimization}), 2) relabeling loop (lines 4-7 in Algorithm~\ref{alg:np_optimization}), 3) cache update loop (lines 4-9 in Algorithm~\ref{alg:np_cache}), and 4) the constraint generation loop of the QP solver (lines 3-10 in Algorithm~\ref{alg:qp_solver}). The number of iterations each loop takes depends on the training data and the hyper parameters of the model (i.e. $\lambda_w$ and $\lambda_u$).

We report the running time of our joint training algorithm separately for one experiment that uses HOG features and one that uses CNN features as the dimensionality of the features and the number of latent locations they consider is different. 

In our current implementation it takes 5 days to do joint training with 120 shared parts on the full MIT-indoor dataset on a 16-core machine using HOG features. It takes 2.5 days to do joint training with 372 parts on the full dataset on a 8 core machine using 60-dimensional PCA-reduced CNN features. Note that these time include learning all shared part filters and all 67 class-specific part weight vectors \emph{on a single machine}. In both experiments finding the most violated constraint (line 8 in Algorithm~\ref{alg:qp_solver}) takes more than half of the total running time. The second bottleneck for HOG features is growing the caches (line 6 in Algorithm~\ref{alg:np_cache}). This involves convolving the part filters (1152 dimensional HOG templates) with all training images (each containing $11000$ candidate locations). With the CNN features, however, the second bottleneck becomes the QP solver (line 7 in Algorithm~\ref{alg:np_cache}).  The QP solver that we use only uses a single core. In both cases the ratio of the time taken by the first bottleneck to the second one is 4 to 1.

The pipeline in previous methods such as~(\cite{juneja13,sun13,doersch13}) has several steps. For example, to discover parts, \cite{juneja13} applies multiple superpixel segmentations on the image to find initial seeds, trains exemplar LDA for each seed, enhances the candidate parts by harvesting similar patches in the dataset, and computes the entropy of the top-50 detections of each part over categories. They discard parts with high entropy as well as duplicates. Despite using several heuristics these methods are slow too. \cite{doersch13} do not comment on the processing time of their method in the paper but we know from personal correspondence that their code takes a long time to run. However, most of the steps in their method are independent; e.g. they start their method from multiple initial points to find the discriminative modes, they train 1-vs-all classifiers, etc. So, they distribute the processing load on a big cluster in order to run their experiments.

Our experimental results showed that we can obtain better performance than~\cite{juneja13} and~\cite{sun13} using a pool of randomly initialized parts (see Figure~\ref{fig:performance_HOG}). Note that creating a pool of random parts is very straightforward and fast. It only takes extracting features from random subwindow of a random image and applying a simple feature transformation on them (see Section~\ref{sec:init}).

\end{document}